\documentclass[anon]{colt2013} 

\usepackage{mathrsfs}
\usepackage{amscd}
\title[Mathematical Theory of Nonlinear Neural Networks In Very High Dimensions]{\textcolor{red}{Demystifying the Global Convergence Puzzle of Learning Over-parameterized ReLU Nets in Very High Dimensions}}

 \coltauthor{\textcolor{blue}{\Name{Peng He}} \Email{phe0328@163.com}\\
 \addr Shenyang 110042, Liaoning Province, China
 }

\begin{document}

\maketitle

\begin{abstract}
This theoretical paper is devoted to developing a rigorous theory for demystifying the global convergence phenomenon in a challenging scenario: learning over-parameterized Rectified Linear Unit (ReLU) nets for very high dimensional dataset under very mild assumptions. A major ingredient of our analysis is a fine-grained analysis of random activation matrices. The essential virtue of dissecting activation matrices is that it bridges the dynamics of optimization and angular distribution in high-dimensional data space. This angle-based detailed analysis leads to asymptotic characterizations of gradient norm and directional curvature of objective function at each gradient descent iteration, revealing that the empirical loss function enjoys nice geometrical properties in the overparameterized setting. Along the way, we significantly improve existing theoretical bounds on both over-parameterization condition and learning rate with very mild assumptions for learning very high dimensional data. Moreover, we uncover the role of the geometrical and spectral properties of the input data in determining desired over-parameterization size and global convergence rate. All these clues allow us to discover a novel geometric picture of nonconvex optimization in deep learning: angular distribution in high-dimensional data space $\mapsto$ spectrums of overparameterized activation matrices $\mapsto$ favorable geometrical properties of empirical loss landscape $\mapsto$  global convergence phenomenon. Furthremore, our theoretical results imply that gradient-based nonconvex optimization algorithms have much stronger statistical guarantees with much milder over-parameterization condition than exisiting theory states for learning very high dimensional data, which is rarely explored so far.
\end{abstract}

\begin{keywords}
Very High-Dimensional Learning, Deep Learning, Nonconvex Optimization, Random Matrix, Overparameterization, Global Convergence
\end{keywords}


\tableofcontents
\label{sec:Introduction}

\section{The Big Picture}

\subsection{The Global Convergence Puzzle}

Modern neural networks have achieved remarkable empirical sucess in a wide range of artifical intelligence applications including computer vision (\cite{he_2016} \cite{voulodimos_2018}), speech recognition (\cite{hinton_2012}), natural language processing (\cite{devlin_2018}), machine translation (\cite{wu_2016})) and game playing (\cite{silver_2017}). The success of deep learning stems from the surprising effectiveness of gradient-based optimization methods for overparmeterized networks, viz. the number of parameters exceeds the number of training samples. Theoretical studies show that learning neural networks are generally computationally infeasible (\cite{livni_2014}) and NP-hard in some cases (\cite{blum_1989}). Moreover, learning popular overparmeterized networks such as ReLU nets is a highly non-convex and non-smooth optimization problems. However, neural nets learned by first-order optimization methods such as (stochastic) gradient descent in conjunction with proper random intialization in the over-parameterized regime are often very effective in practice and reveal global convergence phenomenon (\cite{zhang_chi_2017}). This surprisingly positive empirical performances pose a rather puzzling situation which is far less understood theoretically and largely has remained a mystery. 

\subsection{Prior Art and Limitatons}
The global convergence puzzle raises tremendous interest in developing theoretical analysis of the training of modern neural networks. Due to the highly non-convex and non-smooth nature of modern neural networks, it is very challenging to understand the theoretical properties of learning networks. Recent research in nonconvex optimization shows that many iterative algorithms (e.g., gradient descent) are guaranteed to converge to a global minimum when proper random initialization (\cite{candes_2018} \cite{chi_2019} \cite{ma_2018_1}). With regard to learning overparametrized networks via gradient-based nonconvex optimization, in the last couple of years, there is a large body of literature on empirical evidence and theoretical guarantees for global convergence, which can be categorized in various aspects: trajectory analysis, landscape analysis and mean-field analysis and other research lines. (1) {\bf Trajectory Analysis}: For trajectory-based methods, a sequence of recent works (\cite{brutzkus_2017} \cite{li_2017} \cite{du_2017a} \cite{li_2018} \cite{du_2018} \cite{du_2018a} \cite{jacot_2018} \cite{soltanolkotabi_2017} \cite{wu_xiaoxia_2019} \cite{mannelli_2020}) tried to explain the global convergence of gradient-based algorithms. Many results were obtained based on additional assumptions such as (i) specific  sample distribution such as Gaussian input (\cite{brutzkus_2017} \cite{zhong_song_2017} \cite{li_2017}); (ii) specific data assumptions with additional hyper-parameters (\cite{song_2020}) and (iii) data labels are generated with a teacher (planted) neural network (\cite{brutzkus_2017} \cite{li_2017} \cite{zhang_2019}). Recently, global convergence guarantees of (stochastic) gradient descent have been established in polynomially many iterations when the network has polynomial width in terms of the sample size  (\cite{arora_2019} \cite{du_2018a} \cite{du_iclr} \cite{song_2020} \cite{oymak_2020} \cite{zhu_2018b} \cite{zou_2018} \cite{zou_2019} \cite{nguyen_2020}). In particular, convergence results on linear networks were obtained in (\cite{arora_2019a} \cite{bartlett_2018} \cite{hu_2020} \cite{wu_2019}). In addition, (\cite{zhang_yi_2020}) proved the convergence of adversarial training on polynomially wide two-layer rectified linear unit (ReLU) networks under a seperability assumption on the training data. However, their required over-parameterization sizes are still too large in practice. Some strong convergence results (\cite{nguyen_2021} \cite{noy_2021} \cite{song_2020} \cite{das_2019}) have been claimed based on additional training data or activation assumptions. However, these assumptions do not hold or are hard to verify in practical applications and thus are unrealistic. (2) {\bf Landscape Analysis}: A line of papers analyze geometric landscapes of objective functions (\cite{boob_2017} \cite{du_2017b} \cite{du_2018} \cite{ge_2018} \cite{nguyen_2017} \cite{safran_2016} \cite{safran_2017} \cite{soltanolkotabi_2018} \cite{venturi_2018} \cite{zhou_2017}) \cite{liu_2021}). In addition, the landscape of the squared loss surface with ReLU activations in the under-parameterized setting was investigated in (\cite{tian_2017}). However, the study on landscape analysis often makes assumptions that all local minima are global and all saddle points are strict (the existence of negative curvature) to guarantee gradient-based algorithm escape all saddle points and converge to a global minimum, but these assumptions are not satisfied for some non-linear neural networks or 3-layer linear neural networks. Some papers showed under some assumptions that for a wide non-linear fully connected neural network that there are no spurious valleys or almost all local minima are global minima. Some works also proved that the landscape of nonlinear networks exists spurious local minima when using random input data and labels generated with a teacher (planted) model. Another limitation of landscape analysis is that it is hard to obtain the convergence rate. (3) {\bf Mean-Field Analysis}: Another line of research utilizes mean-field approaches (\cite{chizat_2018a} \cite{mei_2018} \cite{mei_2019} \cite{nitanda_2017} \cite{rotskoff_2018} \cite{araujo_2019} \cite{sirignano_2020} \cite{lu_2020} \cite{chen_2020} \cite{ngugen_2020} \cite{fang_2020}) to prove the limiting behavior of neural networks and demonstrate the global convergence of gradient-based algorithm for training. Mean-field methods characterize the dynamics of the distribution of neurons and apply Wasserstein gradient flow to analyze gradient descent under the condition of infinite number of neurons. Although mean-field analysis allows neurons to move far away from initialization, their results often require the number of neurons to go to infinity, or exponential in related parameters. And it is not clear how to extend the results in mean-field analysis to explain gradient-based learning methods on finite-size neural networks.

In spite of quite a number of contributions understanding optimization, the results reported previously are not entirely satisfactory and the mathematical understanding of the orgin and mechanism of global convergence is still largely lacking. Compared with linear neural networks, the theory of nonlinear neural networks (which is the actual setting of interest), however, is still in its infancy. There have been attempts to extend the local minima are global property from linear to nonlinear networks, but recent results suggest that this property does not usually hold. Firstly, existing theoretical results show a significant gap between theoretical guarantees and practical performances, i.e., theoretical bounds on over-parameterization sizes and iteration complexities are larger than those used in practice by order of magnitude. Secondly, many previous works often rely on some assumptions which are hard to verify in practice. Thirdly, existing theories might not be sufficient to completely explain the success of learning modern neural networks by (stochastic gradient descent) in practice, due to lackness of a clear picture of whole training process and it is difficult to relate key quantities with the effects of the input data in training. To proceed, it is necessary to investigate these limitations in depths.
\subsection{Our Reflections and Motivations}
\textcolor{red}{\it Our Reflections on Neural Tangent Kernel (NTK) Theory.} 
Recently, significant progress was made by trajectory analysis based on NTK theory, which discovers the connection between deep learning and kernels and enables refined analysis of training overparameterized neural networks (\cite{belkin_2018a} \cite{belkin_2018b} \cite{jacot_2018} \cite{chizat_2018b} \cite{daniely_2016} \cite{daniely_2017} \cite{liang_2018_kernel} \cite{mianjy_2020} \cite{su_2019} \cite{ji_2019} \cite{huang_2020}). The seminal paper (\cite{jacot_2018}) analyzed the trajectory of gradient descent approach for smooth activation and infinitely width neural networks via a kernel called {\it neural tangent kernel (NTK)}. Neural tangent kernels are formed using the inner product between gradients of pairs of data points, which can be regarded as the reproducing kernels of the function space from the network structure. It is empirically observed that each row of parameter matrix in neural nets may take a small change during the optimization procedure in wide neural networks (\cite{jacot_2018} \cite{liu_2020}). Roughly speaking, if each neuron does not change much from its initial position, then the NTK is stable in the overparameterization regime, enabling (stochastic) gradient descent achieve zero training loss on neural networks. Along this way, the convergence and generalization of (stochastic) gradient descent methods for over-parameterized neural networks can be studied under the framework of the neural tangent kernel (NTK) (\cite{arora_2019} \cite{cao_2019} \cite{cao_2019b} \cite{chizat_2018b} \cite{du_2018a} \cite{du_iclr} \cite{huang_2020} \cite{jacot_2018} \cite{lee_2019} \cite{zhu_2018a} \cite{zhu_2018b} \cite{oymak_2020} \cite{song_2020}). 
However, we find there are some essential difficulties of existing NTK theory for explaining the success of overparameterized networks. Firstly, due to the non-smooth and non-convex properties of objective function, it is significantly challenging to track the dynamics of weight matrix in the parametric space. This leads to recent analysis of tracking the dynamics of each individual prediction in NTK theory via Gram kernel matrix (\cite{arora_2019} \cite{du_iclr} \cite{song_2020}) or Jacobian matrix (\cite{oymak_2020}). In essence, the tracking process in NTK theory are representd in a approximately linear system, which do not incorporate second-order (curvature) information. Secondly, they require that each neuron does not change much during training. To obtain desired over-parameterization condition, one should choice a very small allowed moving distance of each neuron. Thirdly, it is hard to connect parameters of NTK such as the mininum eigenvalue of Gram matrix with data-dependent properties in a explicit and concise manner. Additionally, an apparent gap between the test performance of neural networks learning by stochastic gradient descent and that of NTK method has been observed in recent experiments (\cite{arora_net_2019} \cite{li_2019}).
In summary, these limitations restrict the interpretability of NTK theory.

\textcolor{red}{\it Our Reflections on Overparameterization.} Recent empirical results in modern neural network learning imply the importance of overparameterization in training, whch movitates a rich body of works  on overparameterization for provable learning. A variety of theoretical papers in various contexts and architectures study how the training of neural networks benefits from over-parameterization, which is not fully explained in most existing theories. 
Compared with smooth activation functions, the convergence analysis on ReLU nets is more challenging, requiring significantly larger hidden units.
(\cite{arora_2019}) proved that the width $m=\Omega(n^7\text{poly}(1/\delta))$ is sufficient for learning. (\cite{du_iclr}) showed that the network width $m=\Omega(n^6\text{poly}(\log n,1/\delta))$ is sufficient for training, here $n$ is the number of input data and $\delta$ is the failure probability, and the randomness stems from the random initialization and learning algorithm, but not from the input data. Recently, (\cite{oymak_2020}) claimed that $m=\Omega(n^4)$ for provable learning with ReLU activation and $m=\Omega(n^2)$ with smooth activations. (\cite{song_2020}) achieved trainability improved the exponent of $n$ and the dependency of failure probability $\delta$, giving $m=\Omega(n^4\text{poly}(\log(n/\delta)))$ for learning in the same setting of (\cite{du_iclr})) and milder results under certain additional data-dependent assumptions. However, the exponent on $n$ in all of these results are still large, 
making the over-parameterization bound still intractable in practical applications. (\cite{ji_2019}) provided a poly-logarithmic width size for learning two-layer nets using logistic loss, relying upon the seperation margin of the limiting kernel. Quite recently, (\cite{kawaguchi_2020} \cite{kawaguchi_2021} \cite{nguyen_2021} \cite{noy_2021}) claimed improved over-parameterization conditions. Notwithstanding, these results rely on smooth activation units, different network architecture or specific data-dependent assumptions, which do not hold in practice or cannot be extended to ReLU activations.
More importantly, the role of the input data in determining the desired over-parameterization size and global convergence guarantee is quite unclear, which is also crucial to understand the convergence puzzle. 
The big gap of existing overparameterization conditions between practice and theory, with the indistinct connection to quantities concerning data samples, imply us the fact that current Gram matrix in NTK theory may be not the fundamental mathematical object governing the training process and practical overparameterization condition.
\textcolor{red}{\it In Search of ``DNA of Random Neural Networks''.} The reflections of existing NTK theory and over-parameterization prompt us to rethink the nature of global convergence phenomenon. To come to grips with the problem of global convergence puzzle which is of course the result of many intricate processes, we need to know what is the core mathematical object dominating the optimization. We now investigate the core mathematical object in NTK theory in more details. For trajectory-based methods, (\cite{arora_2019} \cite{du_iclr} \cite{oymak_2020} \cite{song_2020} \cite{tsuchida_2017} \cite{xie_2017}) depend on crucially on the spectral of the same Gram matrix with various notations: $\boldsymbol H^{\infty}$, $H^{cts}$ in (\cite{song_2020}) and $\boldsymbol\Sigma(\boldsymbol X)$ in (\cite{oymak_2020}) to characterize the convergence rate and required over-parameterization size. 
Intuitively, the geometrical properties of data set such as angles between two data samples would affect the convergence rate of GD. Nevertheless,  $\boldsymbol H^{\infty}$'s explicit expression makes it hard to grasp the effects of the input data in optimization from the spectrum of $\boldsymbol H^{\infty}$, implying that $\boldsymbol H^{\infty}$ may be not the basic mathematical object dominating in optimization. To understand the convergence behaviour of learning modern networks via (stochastic) gradient descent and the effects of the input data, it may be better to investigate more fundamental matrix other than $\boldsymbol H^{\infty}$. This motivates us to dissect random activation matrix $\boldsymbol\Psi$ containing activation pattern of weight vectors and data samples at intialization and during optimization, which is the central object of our theory. Roughly speaking, activation matrix $\boldsymbol\Psi$ has more concise form and, more importantly, more stability during training than $\boldsymbol H^{\infty}$. We believe that activation matrix is the more fundamental object in optimization process and regard activation matrices as \textcolor{blue}{\it ``DNA of random neural networks''}, governing the training process via information evolution in in activation matrices.

Our above reflections motivate us to ask the following questions which are crucial to demystify the global convergence puzzle:
\begin{enumerate}
\item{\it What are the fundamental mathematical objects and mechanism for understanding the convergence behavior of (stochastic) gradient descent on over-parameterized nets?} 
\item{\it How the over-parameterization acts in regard to the dynamics of optimization, and what is the minimal over-parameterization size for provable learning under mild assumptions?}

\item{\it What are the effects of the input data in determining the global convergence rate and required over-parameterization size?}
\end{enumerate}
Modern AI applications and statistical theory often deal with massive high dimensional datasets, where both the sample size $n$ and the dimension of data sample $d$ are very large. For many recent industrial AI-systems and models in modern scientific domains, the dimension $d$ is very large such that $d\geqslant n$ and can asymptotically increases with $n$. These \textcolor{blue}{\it very high dimensional learning problems} arsing from the real world in the 21-th century differ a lot than traditional machine learning tasks and have many challenges. Motivated by prior arts of modern neural network and the demand of learning in very high dimensional setting, in this paper we are especially interested in this very challenging scenario, viz. learning nonlinear neural networks via gradient descent for sufficiently large $n$ and $d\geqslant n$. 

\subsection{Our Contributions and Key Intuitions}
{\bf Main Contributions}:\\
We make progress on answering the above questions for learning over-parameterized two-layer ReLU networks via gradient descent using training samples of very high dimension, improving upon known state-of-the-art results and deepening theoretical understanding in several aspects. Our findings can be summarized as follows:
\begin{enumerate}
\item{\textcolor{blue}{\it A novel picture of learning ReLU nets: components, characteristics and mechanism}}. To understand what happens in learning overparameterized ReLU networks, we contract key components from various levels. Based on interactions of these components, we illustrate how training behaves and provide a novel explanation for global convergence guarantee in the over-parameterized regime.

Firstly, we understand the optimization mechanism from top to bottom as three levels: molecular level (high-level); atom level (mid-level); electron level (low-level), and the key components in each level as:
$$
\textcolor{blue}{
\text{Molecular Level Components: Gradient and Directional Curvature of Loss}}
$$
$$
\textcolor{red}{
\text{Atom Level Component: Overparameterized Random Activation Matrix}}
$$
$$
\textcolor{purple}{
\text{Electron Level Component: Angular Distribution in High-Dimensional Space}}
$$
Secondly, we find the key mathematical objects characterizing these components as:
$$
\textcolor{blue}{
\text{Molecular Level Characteristic: (Frobenius) Norm and Quadratic Form}}
$$
$$
\textcolor{red}{
\text{Atom Level Characteristic: Extreme Singular Values of Activation Matrices}}
$$
$$
\textcolor{purple}{
\text{Electron Level Characteristic: Extreme Angle $\oplus$ Eigenvalue of Angle-based Matrix}}
$$
With these components and characteristics in place, our picture of the underlying mechanism of the global convergence phenomenon can be summarized from bottom to up (from electron level to molecular level) as the following geometric picture:
\begin{align}\nonumber
&\text{Over-parameterization with Random Weight Initialization}\\\nonumber &\oplus\text{ Angular Information in Extreme Angle and Eigenvalue of Angle-based Matrix}\\\nonumber
\mapsto&\text{ The Smallest Singular Value of Initial Activation Matrix Is Large Enough}\\\nonumber
&\oplus\text{ Small Perturbation of Activation Matrix During Optimization}\\\nonumber
\mapsto&\text{ The Smallest Singular Values of Activation Matrix Remains Is Large Enough}\\\nonumber
\mapsto&\text{ Benign Properies of Gradient and Directional Curvature of Loss Function}\\\nonumber
&\text{ (Nice Geometrical Properies of Empirical Loss Function)}\\\nonumber
\mapsto&\text{ Global Convergence Guarantees of Gradient Descent}
\end{align}
Explicitly, by a full characterization of activation matrices via high-dimensional angular information at initialization and during optimization, we find that the over-parameterization condition enables the smallest singular value of initial activation matrix large enough and small pertuation of activation matrix at each GD iteration. This makes the smallest singular value of activation matrix remain sufficiently large during the whole training. Furthermore, the gradient and directional curvature of loss function can be efficiently bounded by means of spectral properties of activation matrices, leading to nice geometrical properties of empirical loss function and the final global convergence guarantee of gradient descent.

\item{\textcolor{blue}{\it Milder required over-parameterization condition and effects of over-parameterization}}. We improve the over-parameterization condition required for provable convergence using very  mild assumptions. Compared with previous results, our over-parameterization size clarifies its close and explicit relation with geometrical and spectral properties of training samples. Moreover, we reveal how the over-parameterization size affects the trainability of two-layer ReLU neural nets, viz. over-parameterization improves the extreme singular value bound of random activation matrix, which further has nice geometrical effects on the objective function via increasing gradient norm and decreasing directional curvature along gradient direction and higher order remainder term.
\item{\textcolor{blue}{\it A novel data-dependent global convergence guarantee and effects of input data}}. We analyze the dynamics of empirical quadratic loss and prove the convergence to a global minimum at a linear rate under very mild data assumptions, providing the data-dependent convergence rate and iteration complexity using practical learning rate. Our results also uncover the connection between global convergence rate and geometrical/spectral properties of training data, paving the way for understanding how the intrinsic structure of the input data affect the convergence guarantee. 
\end{enumerate}

{\bf Key Intuitions}:
\begin{enumerate}
\item \textcolor{blue}{\it Tracking Objective Function Rather Than Tracking Network Function.} Our first insight is that state-of-the-art convergence and over-parameterization results obtained from existing NTK theory rely on tracking network functions (individual predictions) approximated with the linearized form of the first order Taylor expansion, which are further governed by the smallest eigenvalue of a Gram matrix from activation kernel. We observe that difference with network functions, the dynamics of objective function for ReLU networks can be characterized in a nonlinear form via gradient, directional curvature and higher order remainder term, all of which can be effeciently controlled by extreme singular values or norms of activation matrix at per GD iteration. Therefore, if we have sharp spectral and norm estimates of activation matrices, we can obtain a tighter characterization of training process than that in NTK theory. This drives us to treat active matrix at each GD iteration with special care.

\item \textcolor{blue}{\it Controlling Active Matrix in a Larger Weight Neighbhood with Less Neurons.} Our second insight is that to control extreme singular values and norms of active matrix tightly, we need proper over-parameterization condition and allowed moving distance of weight vectors during GD. Furthermore, most of the activation patterns $\mathbb{I}\{\boldsymbol w_r^\top \boldsymbol x_i\geqslant 0\}$  do not change during training, together with more compact form of activation marix, implying activation matrix has more stability than Gram matrix from activation kernel in NTK theory. This stronger stable property further makes the spectrums and norms of random activation matrices be more tightly controlled with \textcolor{red}{milder over-parameterization size}  and \textcolor{red}{larger allowed moving distance of weight vectors} during GD than allowed moving distances choiced in existing NTK theory. This milder required over-parameterization size and larger allowed moving distance of weight vectors for controlling active matrix are demonstrated are crucial to improve the over-parameterization condition and final global convergence rate. This leads to the question of how to dissect active matrix at a fundamental level.

\item \textcolor{blue}{\it Dissecting Active Matrix in a More Geometric-Taste Manner.} Our third insight is to achieve the foregoing goal, we should carefully decompose active matrix from spatial and stochastic structures. At a basic level, properties of activation matrix are determined by the \textcolor{red}{angular distribution} arising from \textcolor{red}{high-dimensional geometry} of training data. Compared with the analysis of Gram matrices in existing NTK theory at the spectral level, our analysis of activation matrices at initialization and pertubations during optimization performs a more elaborate investigation of random matrices in terms of quantities concerning angle information. More explictily, by making full use of angle information in activation matrix, we can precisely \textcolor{red}{evaluate key expectations as high-dimensional integrals}, giving rise to tight estimates by further applying concerntration inequalities and precise characterization of activation matrix. 

Combining all these insights together, we can prove in a more geometric-taste investigation at a very basic level that an activation matrix does not change much during the optimization if the network is sufficient wide with random initialization. Furthermore, we can obtain the sharp lower bound of the smallest singular value of initial activation matrix and upper bound of the perturbation of activation matrix during training. As for what we expected, these bounds are the key ingredients to enable the gradient and directional curvature of objective function remain effeciently bounded and the final global convergence rate of GD. In addition, our geometric-taste characterization makes it easier to uncover the effect of the input data during optimization. 
\end{enumerate}
\subsection{Main Mathematical Weapons}
We apply powerful methods in high-dimensional probability (\cite{vershy_2018}) and non-asymptotic random matrix theory (\cite{ahlswede_2002} \cite{rudelson_1999}) including concentration inequalities such as Chebyshev's inequality, Hoeffding inequality, Bernstein inequality and recent developed matrix concentration inequalities such as matrix Bernstein inequality (\cite{tropp_2015} \cite{y_chen_2021}). Matrix trace and eigenvalue inequalities (\cite{horn_2013} \cite{garcia_2017}) such as Weyl' inequality (\cite{weyl_1912}), Ger\v{s}gorin disc theorem (\cite{gersgorin_1931}) and Schur's theorem (\cite{schur_1911}) are also frequently applied in our analysis. Some fundamental mathematical methods including analysis, algebra and probability can be found in classic and modern math books (\cite{moivre_1733} \cite{euler_1748} \cite{euler_1755} \cite{cauchy_1821} \cite{jordan_1894} \cite{goursat_1902} \cite{whittaker_1902} \cite{hardy_1908} \cite{poincare_1912} \cite{titchmarsh_1932} \cite{fisher_1941} \cite{aitken_1944} \cite{cramer_1946} \cite{doob_1953} \cite{feller_1957} \cite{bellman_1960} \cite{rudin_1964} \cite{gantmacher_1966} \cite{parthasarathy_1967} \cite{chung_1968} \cite{ross_1976} \cite{billingsley_1979} \cite{ross_1983} 
\cite{hua_1984} \cite{sun_1987} \cite{stewart_1990} \cite{ledoux_1991} \cite{lang_1993} \cite{billingsley_1999} \cite{ledoux_2001} \cite{varga_2004} \cite{lin_2010} \cite{tao_2012} \cite{golub_2012} \cite{horn_2013} \cite{foucart_2013} \cite{garcia_2017}).
\subsection{Roadmap of the Paper}

The remainder of this paper is organized as follows. We first start with formulating the problem and introducing notations in Section \ref{section_problem_for}. Then we formally state our main theoretical results in Section \ref{section_theory}, followed by immediate consequences, comparisons and discussions. Section \ref{section_convergence} is devoted to giving a proof sketch and elaborating more on the detailed proof of learning over-parameterized networks. At the end we conclude our work with reflections in Section \ref{section_conclusion}. We hope our reflections could shed light on new AI research.


\section{Problem Formulation}\label{section_problem_for}

\subsection{Network Architecture and Model Setup}\label{section_problem}
We consider a two-layer fully-connected neural network with $m$ neurons in the hidden layer and input dimension $d$, which has the network prediction function as
\begin{equation}\nonumber
f(\boldsymbol W,\boldsymbol a,\boldsymbol x)=\frac{1}{\sqrt{m}}\sum_{r=1}^ma_r\phi(\boldsymbol{w_r^\top x}),
\end{equation}
where $\boldsymbol x\in\mathbb R^d$ is the input vector, $\boldsymbol W=(\boldsymbol w_1,\cdots,\boldsymbol w_m)\in\mathbb R^{d\times m}$ denotes the hidden weight matrix and $\boldsymbol a=(a_1,\cdots,a_m)^\top\in\mathbb R^m$ denotes the output weight vector. Since ReLU activation (\cite{xu_2015}) is extensively used in modern networks due to its dominating empirical performance in optimization, we focus on networks with ReLU activation function $\phi(z)=\max\{z,0\}$, so $\phi'(z)=1$ if $z\geqslant0$ and $\phi'(z)=0$ otherwise. 

To train such a neural network, we access a data set consisting of $n$ feature/label pairs by $S=\{(\boldsymbol{x_i},y_i)\}_{i=1}^n$ drawn i.i.d. from an underlying distribution $\mathcal D$ over $\mathbb R^d\times\mathbb R$. We use $\boldsymbol X=(\boldsymbol x_1,\cdots,\boldsymbol x_n)\in\mathbb{R}^{d\times n}$ to denote the data matrix and assume thta no two input data vectors $\boldsymbol x_i$ and $\boldsymbol x_j$ for $i\not=j$ are co-aligned. This is the mildest data-dependent assumption, since no two input vectors are parallel for most real world datasets. In fact, it is impossible for any learning method to achieve zero training error if two parallel data points with different output labels. Without loss of generality, we can assume that $\|\boldsymbol{x_i}\|_2=1$ and $|y_i|\leqslant1$ for any $i\in[n]$, which is widely used in the literature.

We focus here for the sake of brevity on the quadratic loss, aiming to minimize the objective function
\begin{equation}\label{loss_func}
\mathcal L(\boldsymbol W)=\frac{1}{2}\sum_{i=1}^n(f(\boldsymbol W,\boldsymbol a,\boldsymbol x_i)-y_i)^2.
\end{equation}
If all the neurons are active ($\boldsymbol a_r\not=0$), then it is evident that their amplitudes of $\boldsymbol a_r$ can be easily absorbed into the corresponding layer weights ($\{\boldsymbol w_r\}$). Therefore, the task of optimizating over ($\boldsymbol W,\boldsymbol a$) is locally equivalent to optimization over $\boldsymbol W$ only. Therefore, we optimize $\boldsymbol W$ and treat $\boldsymbol a\in\mathbb R^m$ as fixed vector with  $\boldsymbol a_r$ initialized uniformly at random from $\{-1,+1\}$ for any $r\in[m]$. Without loss of generality, we use a common weight intialization scheme used in practice, viz. the intial weight vectors $\boldsymbol w_r(0)$ are i.i.d. randomly generated from ${\cal N}(\boldsymbol 0,\boldsymbol I_d)$ for each $r\in[m]$. 

The square loss is minimized by gradient descent and the weight matrix is updated as
\begin{equation}\label{grad_func}
\boldsymbol W(k+1)=\boldsymbol W(k)-\eta\frac{\partial\mathcal L(\boldsymbol W(k))}{\partial\boldsymbol W(k)},
\end{equation}
where $\eta>0$ is the step size (a.k.a. learning rate). Let $f_i=f(\boldsymbol W,\boldsymbol a,\boldsymbol x_i)$ and $\boldsymbol f=(f_1,\cdots,f_n)^\top\in\mathbb R^n$, then the gradient with respect to each weight vector is of the form\footnote{Here we define ``gradient'' using this expression as ReLU is not differentiable at 0. This is used in practice and widely adopted in the literature.}:
\begin{equation}\label{grad_wr_func}
\frac{\partial\mathcal L(\boldsymbol W)}{\partial\boldsymbol w_r}=\frac{1}{\sqrt{m}}\sum_{i=1}^n(f_i-y_i)a_r\mathbb{I}_{r,i}\boldsymbol x_i\in\mathbb{R}^{d},
\end{equation}
where $\mathbb{I}_{r,i}=\mathbb{I}\{\boldsymbol w_r^\top \boldsymbol x_i\geqslant 0\}$ with $\mathbb{I}$ denoting the indicator mapping. 

For $i,j\in[n]$, let $\theta_{ij}=\angle(\boldsymbol x_i,\boldsymbol x_j)$ be the angle between $\boldsymbol x_i$ and $\boldsymbol x_j$. We assume for any two different training samples $\boldsymbol x_i$ and $\boldsymbol x_j$, there exists a constant value $\widehat{\theta}>0$ such that $\min\{\theta_{ij},\pi-\theta_{ij}\}\geqslant\widehat{\theta}$. It is a quite mild assumption on training data, which is equivalent with the minimal distance assumption widely used in many previous work (\cite{zhu_2018b} \cite{zou_2019}) and is essential to guarantee global convergence for learning neural networks.

We further define a symmetric matrix
\begin{align}\nonumber
\boldsymbol\mho=\mathbb E_{\boldsymbol w\sim{\cal N}(0,\boldsymbol I)}[\phi'(\boldsymbol X^\top\boldsymbol w)\phi'(\boldsymbol w^\top\boldsymbol X)]\in\mathbb{R}^{n\times n}.
\end{align}
It is evident that
\begin{align}\nonumber
\boldsymbol\mho=\left(  \begin{array}{ccc}
   \mathbb E \left\{\mathbb I_{r,1}(0)\right\} & \cdots &     \mathbb E \left\{\mathbb I_{r,1}(0)\mathbb I_{r,n}(0)\right\}  \\
   \vdots & \ddots & \vdots  \\
  \mathbb E \left\{\mathbb I_{r,n}(0)\mathbb I_{r,1}(0)\right\}  & \cdots &   \mathbb E \left\{\mathbb I_{r,n}(0)\right\} 
                         \end{array}
              \right),\,\forall r\in[m].
\end{align}
It can be proved in Claim 1 in Section \ref{subsection_activation_matrices} that 
$$\mathbb E \left\{\mathbb I_{r,i}(0)\mathbb I_{r,j}(0)\right\}=\frac{\pi-\arccos\langle\boldsymbol x_i, \boldsymbol x_j\rangle}{2\pi}=\frac{\pi-\theta_{ij}}{2\pi}\quad\forall i,j\in[n],\forall r\in[m].$$ Therefore, $\boldsymbol\mho$ is an angle-based matrix characterizing angles between input data vectors, viz.
\begin{align}\label{matrix_mho_definition}
\boldsymbol\mho=&\left(  \begin{array}{ccc}
  1/2  & \cdots &    (\pi-\theta_{1n})/(2\pi)   \\
   \vdots & \ddots & \vdots  \\
 (\pi-\theta_{n1})/(2\pi)  & \cdots &   1/2  
                         \end{array}
              \right).
\end{align}
$\boldsymbol\mho$ characterizes the angular distribution from an overall perspective, which implies us to define $\lambda^*=\lambda_{\text{min}}(\mho)$.
To characterizes the angular distribution of the input data from an extreme perspective, we define the following angular parameters:
\begin{align}\label{theta_min_definition}
\theta_{\text{min}}&=\min_{1\leqslant i\not=j\leqslant n}\theta_{ij}=\min_{1\leqslant i\not=j\leqslant n}\angle(\boldsymbol x_i,\boldsymbol x_j)=\min_{1\leqslant i\not=j\leqslant n}\{\arccos\langle\boldsymbol x_i, \boldsymbol x_j\rangle\}\leqslant2\pi/3,\\\label{theta_max_definition}
\theta_{\text{max}}&=\max_{1\leqslant i\not=j\leqslant n}\theta_{ij}=\max_{1\leqslant i\not=j\leqslant n}\angle(\boldsymbol x_i,\boldsymbol x_j)=\max_{1\leqslant i\not=j\leqslant n}\{\arccos\langle\boldsymbol x_i, \boldsymbol x_j\rangle\},\\\label{theta_star_definition}
\theta^*&=\min\{\theta_{\text{min}},\pi-\theta_{\text{max}}\}.
\end{align}
The intevals of $\theta_{\text{min}},\theta_{\text{max}}$ and $\theta^*$ follows from Assumption 2. Intuitively, $\theta^*$ indicates the level of difficulty to distinguish data samples. When $\theta^*$ is large, all data points are well-seperated, whereas a small value of $\theta^*$ means there exist two input samples that are close to each other.
We will find that these geometric parameters $\theta^*,\theta_{\text{min}}$ and $\theta_{\text{max}}$ play a crucial role in convergence. The sharp upper bound of $\theta_{\text{min}}$ is $2\pi/3$, which is proved in (\ref{theta_star_upper_bound}) in step 3 of Section \ref{subsection_min_sigma_Psi0} and comes from Claim 2 in step 3 of Section \ref{subsection_min_sigma_Psi0}.

It is very convenient for subsequent analysis to introduce a matrix $\boldsymbol\Psi=\phi'(\boldsymbol X^\top\boldsymbol W)\in\mathbb{R}^{n\times m}$, which we call \textcolor{blue}{\it ''activation matrix''} and plays a crucial role in this paper. Here $\phi'$ is applied to each entry of the matrix $\boldsymbol X^\top\boldsymbol W$ and thus it is evident that each entry of matrix $\boldsymbol\Psi$ is $\phi'(\boldsymbol w_r^\top \boldsymbol x_i)=\mathbb{I}\{\boldsymbol w_r^\top \boldsymbol x_i\geqslant 0\})=\mathbb{I}_{i,r}$, giving rise to
\begin{align}\label{Psi_matrix}
\boldsymbol \Psi
=\left(  \begin{array}{ccc}
  \mathbb{I}_{1,1}  & \cdots &      \mathbb{I}_{m,1}   \\
   \vdots & \ddots & \vdots  \\
   \mathbb{I}_{1,n}  & \cdots &      \mathbb{I}_{m,n}    
                         \end{array}
              \right)
              =(\boldsymbol \psi_1,\cdots,\boldsymbol \psi_m)
              =\left( \begin{array}{c}
              \widetilde{\boldsymbol\psi_1} \\
              \vdots \\
              \widetilde{\boldsymbol\psi_n} 
        \end{array}
 \right)
\in\mathbb{R}^{n\times m},
\end{align}
where $\boldsymbol\psi_r(1\leqslant r\leqslant m)$ and $\widetilde{\boldsymbol\psi_i}(1\leqslant i\leqslant n)$ denote each column and row of $\boldsymbol \Psi$, respectively. $\mathbb{I}_{i,r}$ is called \textcolor{blue}{activation pattern} in the literature (\cite{du_iclr}) and hence we can call $\boldsymbol\Psi$ as activation matrix. $\boldsymbol\Psi$ is closely related with the matrix $\boldsymbol\mho$. Notice that $\boldsymbol\Psi$ is a time-dependent matrix in training process, which can be written as $\boldsymbol\Psi(k)$ at $k$-th iteration.

In summary, given the input data, we first have a data matrix $\boldsymbol X$, then the angular distribution from $\boldsymbol X^\top\boldsymbol X$ can be effeciently represented as a symmetric matrix $\boldsymbol\mho$ and an extreme angle $\theta^*$, which will be proved to be crucial to characterize the spectrums of active matrix $\boldsymbol\Psi$. Thus, the \textcolor{blue}{\it ``evolution route''} from input data to activation matrix is 
\begin{align}\nonumber
\textcolor{blue}{
\underbrace{\boldsymbol X\rightarrow\boldsymbol X^\top\boldsymbol X=(\cos\theta_{ij})_{n\times n}}_{\text{\textcolor{red}{Data Represtation}}}\mapsto\underbrace{\{\theta_{ij}\}_{1\leqslant i,j\leqslant n}\rightarrow\boldsymbol\mho\rightarrow\{\lambda^*=\lambda_{\text{min}}(\mho),\theta^*\}}_{\text{\textcolor{red}{Represtation of Angular Distribution}}}\mapsto\underbrace{\boldsymbol\Psi=\phi'(\boldsymbol X^\top\boldsymbol W)}_{\text{\textcolor{red}{Activation Matrix}}}.
}
\end{align}
In addition, we define the residure vector $\boldsymbol r\in\mathbb R^n$ with the $i$th entry given by
$r_i=f_i-y_i$ for $i\in[n]$. Thus, the matrix $\boldsymbol D_{\boldsymbol r}=\text{diag}(r_1,\cdots,r_n)$ is the diagonal matrix formed by the residure vector $\boldsymbol r=(r_1,\cdots,r_n)$. 

Finally, for the convenience of readers, we list the parameters and matrices throughout our presentation in the following Table 1.
\\\\
\begin{tabular}{|c|c|c|c|c|}
\hline
Pa. & Definition & Choice/Interval & Place\\
\hline
$d$ & The Dimension of Data Sample & $d\geqslant n$ (Very High Dimension) & Section \ref{section_problem} \\
\hline
$m$ & The Width of Neural Network & $\Omega(n^3\log^{10\hbar+3}(n/\delta)/\sin\theta^*)$ & Theorem \ref{theorem_GD_convergence} \\
\hline 
$\theta^*$ & Angular Parameter of Dataset & $\min\{\theta_{\text{min}},\pi-\theta_{\text{max}}\}$  & (\ref{theta_star_definition}) in Section \ref{section_problem} \\
\hline
$\lambda^*$ & $\lambda_{\text{min}}(\mho)$  & $[\lambda_{\text{min}}(\boldsymbol X^\top\boldsymbol X)/4,1/2]$ &  Section \ref{section_problem} \\
\hline
$\eta$ & Step Size (Learning Rate) & $\lambda^*/(2\kappa n^{3/2})$ & Theorem \ref{theorem_GD_convergence} \\
\hline
$\kappa$ & The Condition Number of $\boldsymbol X$ &  $\|\boldsymbol X\|_2/\sigma_{\min}(\boldsymbol X)$  & Theorem \ref{theorem_GD_convergence} \\
\hline 
$R$ &  Max. Allowed Change of Weight & $\sqrt{\pi\lambda^*\sin\theta^*/(2+\varepsilon)n} (\forall\varepsilon>0)$ & Lemma \ref{lemma_Psi_k_lower_bound} \\
\hline 
$\boldsymbol X$ & Data Matrix  & $(\boldsymbol x_1,\cdots,\boldsymbol x_n)\in\mathbb{R}^{d\times n}$  & Section \ref{section_problem} \\
\hline 
$\boldsymbol W$ & Weight Matrix  & $(\boldsymbol w_1,\cdots,\boldsymbol w_m)\in\mathbb R^{d\times m}$  & Section \ref{section_problem} \\
\hline 
$\boldsymbol \Psi$ & Activation Matrix  & $\phi'(\boldsymbol X^\top\boldsymbol W)$  & (\ref{Psi_matrix}) in Section \ref{section_problem} \\
\hline 
$\boldsymbol \mho$ & Symmetric Angle-Based Matrix &  $\mathbb E_{\boldsymbol w\sim{\cal N}(0,\boldsymbol I)}[\phi'(\boldsymbol X^\top\boldsymbol w)\phi'(\boldsymbol w^\top\boldsymbol X)]$ & (\ref{matrix_mho_definition}) in Section \ref{section_problem} \\
\hline 
\end{tabular}
\\\\ Table 1: Table of main parameters and matrices in this paper. Pa. stands for parameters, $n$ is the number of input data samples and $\delta$ is the faillure probability.
\subsection{Notations}

We let $[n]=\{1,2,\cdots,n\}$ and $|S|$ be the cardinality of a set $S$. $\mathbb{I}\{E\}$ or $\mathbb{I}_{E}$ denotes the indicator function for an event $E$. Throughout this paper, we make use of boldface letters to denote vectors and matrices. For a matrix $\boldsymbol A$, let $\boldsymbol A_{i,\cdot}$ and $\boldsymbol A_{\cdot,j}$ represent the $i$-th row and the $j$-th colmmn of $\boldsymbol A$, respectively. $(\boldsymbol A)_{i,j}$ or $\boldsymbol A_{i,j}$ denotes the $(i,j)$ entry of matrix $\boldsymbol A$, and $(\boldsymbol v)_i$ denotes the $i$th entry of vector $\boldsymbol v$, respectively. $\sigma_i(\boldsymbol A)$(resp. $\lambda_i(\boldsymbol A)$) stands for the $i$-th largest singular value (resp. eigenvalue) of $\boldsymbol A$. In particular, $\sigma_{\text{max}}(\boldsymbol A)$(resp. $\lambda_{\text{max}}(\boldsymbol A)$) represents the largest singular value (resp. eigenvalue) of $\boldsymbol A$, while $\sigma_{\text{min}}(\boldsymbol A)$(resp. $\lambda_{\text{min}}(\boldsymbol A)$) represents the smallest singular value (resp. eigenvalue) of $\boldsymbol A$. For matrix norms, $\|\boldsymbol A\|_2$(resp. $\|\boldsymbol A\|_\text{F}$) represent its spectral (resp. Frobenius) norm. For any matrices $\boldsymbol A, \boldsymbol B$ of compatible dimensions, $\boldsymbol A\circ\boldsymbol B$ stands for Hadamard (entrywise) product and $\langle\boldsymbol{A,B}\rangle$ denotes the inner product of $\boldsymbol{A}$ and $\boldsymbol{B}$. For any vector $\boldsymbol v$, denote by $\boldsymbol v_i$ the $i$-th element of $\boldsymbol v$ and $\boldsymbol D_{\boldsymbol v}$ (also written as $\text{diag}(\boldsymbol v)$) the diagonal matrix with elements $(\boldsymbol D_{\boldsymbol v})_{ii}$ as $\boldsymbol v_i$. For ReLU activation function $\phi(\cdot)=\max\{\cdot,0\}$, $\phi(\boldsymbol A)$ and $\phi'(\boldsymbol A)$ are termwise operations on matrix, viz. $(\phi(\boldsymbol A))_{i,j}=\phi(\boldsymbol A_{i,j})$ and $(\phi'(\boldsymbol A))_{i,j}=\phi'(\boldsymbol A_{i,j})$. $\boldsymbol 1_n=(1,\cdots,1)\top\in\mathbb R^n$.

In addition, we use the following asymptotic notations: for any real-valued functions $f(n)$ and $g(n)$, $f(n)=\mathcal O(g(n))$ or $f(n)\lesssim g(n)$ means that $\limsup_{n\to\infty}|f(n)/g(n)|\leqslant c$ or equivalently, $|f(n)|\leqslant cg(n)$ for some universal constant $c>0$ when $n$ is sufficiently large; $f(n)=\Omega(g(n))$ or $f(n)\gtrsim g(n)$ is equivalent to $g(n)\lesssim f(n)$; $f(n)=\Theta(g(n))$ or $f(n)\asymp g(n)$ means both $f(n)\lesssim g(n)$ and $f(n)\gtrsim g(n)$ hold true. In addition, we use $\widetilde{\mathcal O}(\cdot)$, $\widetilde{\Omega}(\cdot)$ and $\widetilde{\Theta}(\cdot)$ to hide logarithmic factors. Finally, we denote the normal distribution by $\mathcal N(\boldsymbol\mu,\boldsymbol\Sigma)$.
\section{Theoretical Guarantees and Implications}\label{section_theory}
In this section, we present our main results and findings. We demonstrate that gradient descent using random initialization can converge to the global minimum of the training loss, enjoying statistical accuracy and computational efficiency with milder over-parameterization condition and learning rate than previous results. Additionally, our results depend on very mild data assumption and do not assume the existence of any planted (true) model. 
Our first theorem concerns the loss decrease at per iteration of gradient descent.
\begin{theorem}\label{theorem_GD_lossfunction_update}
(\textcolor{blue}{Asymptotic Behavior of Training Loss Dynamics}) Suppose a two-layer overparameterized ReLU network trained by gradient descent under the scheme in Section \ref{section_problem} for data dimension $d\geqslant n$ and $n\leqslant m$. As long as $\sigma_{\text{min}}(\boldsymbol X)=\Omega(\log^{-\hbar}n)$ for certain $\hbar>0$ and the width satisfies $m=\Omega(n^3\log^{1+\varpi}(n/\delta))$ for any $\varpi>1$, then for $n$ tending to infinity, $k\in\mathbb Z^+$ and any $\delta\in(0,1)$, with probability exceeding $1-\delta$ over the random initialization, the gradient descent iterate obeys
    \begin{equation}\nonumber
    \mathcal L(\boldsymbol W(k+1))<\bigg\{1-\eta\left[2\lambda^*+o(1)\right]\sigma_{\min}^2(\boldsymbol X)
+\eta^2\left[2n^{3/2}+n/2+\mathcal O(n^{1/2})\right]
\|\boldsymbol X\|_2^2\bigg\}\mathcal L(\boldsymbol W(k)),
    \end{equation}
where $\lambda^*=\lambda_{\text{min}}(\mho)$ and the matrix $\mho$ is defined in (\ref{matrix_mho_definition}) in Section \ref{section_problem} and $\eta$ is the step size (learning rate).
\end{theorem}
{\bf Remark}: The parameter $\hbar$ is in honor of great Max Planck. Theorem \ref{theorem_GD_lossfunction_update} is a backbone to ensure the probable trainablitity of GD, stated in Theorem \ref{theorem_GD_convergence}. Theorem \ref{theorem_GD_lossfunction_update} provides a concise picture of the contraction of loss function at each GD itertion, where the coefficients are the forms of asymptotic expansions. Unlike most previous results on loss decreasing in training, we characterize the main ingredients of contraction rate as asymptotic expansions, which give the order of magnitude of the remainder and make the behavior of GD with regard to the number of training data more clearly and hence will be enlightening in today's big data era. The proof appears in Section \ref{section_convergence}. The central part of the proof of Theorem \ref{theorem_GD_lossfunction_update} is precise characterization of ojective function via the lower bound of gradient norm, upper bound of directional curvature and higher order remainder term. All these geometric quantities can be efficiently controlled via spectrums and norms of activation matrices in the overparameterized regime. This reveals that over-parameterization makes the landscape of objective function more favorable to optimization, explaining the global convergence puzzle for over-parameterized neural networks. Moreover, Theorem \ref{theorem_GD_lossfunction_update} indicates the crucial role of angle-based matrix $\mho$ in optimization, since the smallest eigenvalue $\lambda^*$ of matrix $\mho$ appears naturally in analyzing the lower bound of the smallest singualr value of activation matrix $\boldsymbol\Psi(k)$ in the overparameterized setting (see step 1 of Lemma \ref{lemma_Psi_k_lower_bound} in Section \ref{subsection_pieceing_together}).
In short, we find that the underlying mechanism of convergence process can be summarized as
$$\textcolor{blue}{\text{Angle-based Matrix }\xrightarrow{\text{Over-Para.}}\text{Activation Matrix } \rightarrow\text{Loss Geometry}\rightarrow\text{Loss Dynamics}}$$
Or in more explicitly manner:
$$
\textcolor{blue}{
\lambda^*=\lambda_{\text{min}}(\boldsymbol\mho)\xrightarrow{\text{Over-Para.}} \sigma_{\min}(\boldsymbol\Psi(k))\oplus\|\boldsymbol\Psi(k)\|\rightarrow\nabla \mathcal L(\boldsymbol W(k))\oplus\nabla^2\mathcal L(\boldsymbol W(k))\rightarrow d\mathcal L(\boldsymbol W(k)}
$$
Theorem \ref{theorem_GD_lossfunction_update} also implies the effect of the input data in optimization via spectrums of data matrix. Since $\hbar$ is only required to be positive, the condition $\sigma_{\text{min}}(\boldsymbol X)=\Omega(\log^{-\hbar}n)$ is mild in practice.
To make this more sense, we first explore the lower bound of $\sigma_{\text{min}}^2(\boldsymbol X)=\lambda_{\text{min}}(\boldsymbol X^\top\boldsymbol X)$ when sampling training data in very high dimensional data space $\mathbb R^d(d\geqslant n)$ from various aspects. Firstly, it is evident that $\lambda_{\text{min}}(\boldsymbol X^\top\boldsymbol X)\geqslant1-\sum_{j\not=i}|\cos\theta_{ij}|\in(0,1]$ for certain $i\in[n]$ via \textcolor{red}{Ger\v{s}gorin disc theorem (\cite{gersgorin_1931})}. For example, in very high dimensional space $\mathbb R^d$, data points i.i.d. sampling from many distributions such as i.i.d. isotropic random vectors tend to be almost orthogonal to each other(\cite{vershy_2018}), which is one of extraordinary phenomena in high dimensional space and could become more pronounced in very high-dimensional space. This ``almost orthogonal'' phenomenon in very high-dimensional space enables the term $1-\sum_{j\not=i}|\cos\theta_{ij}|$ be strictly positive in high probability in this case.
Roughly speaking, $\lambda_{\text{min}}(\boldsymbol X^\top\boldsymbol X)$ is strictly positive for many scenatrio and we could expect larger $\lambda_{\text{min}}(\boldsymbol X^\top\boldsymbol X)$ with larger angular parameter $\theta^*$ which means better seperation of dataset. In fact, as discussed in Theorem \ref{theorem_GD_convergence}, $\sigma_{\text{min}}(\boldsymbol X)$ could be expected to increase significantly for some data distributions and sufficiently large dimension $d$, giving rise to $\sigma_{\text{min}}(\boldsymbol X)=\Omega(\log^{-\hbar}n)$ for sufficiently large $\hbar$, leading to faster convergence rate. We shall discuss this problem in details in the remark of Theorem \ref{theorem_GD_convergence}.

We turn to state Theorem \ref{theorem_GD_convergence}, guranteeing a linear-rate convergence to a globel minimum when the network is over-parameterized with random initialization.
\begin{theorem}\label{theorem_GD_convergence}
(\textcolor{blue}{Convergence Guarantee of Gradient Descent}) Suppose a two-layer ReLU network in the over-parameterized setting ($n\leqslant m$) trained by gradient descent under the scheme in Section \ref{section_problem} for data dimension $d\geqslant n$. As long as $\sigma_{\text{min}}(\boldsymbol X)=\Omega(\log^{-\hbar}n)$ for certain $\hbar>0$, the step size  $\eta=\lambda^*/(2\kappa^2 n^{3/2})$ and the width of neural network $m$ satisfies $$m=\Omega\left(\frac{n^3\log^3(n/\delta)}{(\lambda^*)^3\sin\theta^*\sigma_{\min}^4(\boldsymbol X)}\right)\quad\text{or a conciser condition } m=\Omega\left(\frac{n^3\log^{10\hbar+3}(n/\delta)}{\sin\theta^*}\right),
$$
where $\lambda^*=\lambda_{\text{min}}(\mho)$ and the matrix $\mho$ is definied in (\ref{matrix_mho_definition}) in Section \ref{section_problem}, and $\kappa=\|\boldsymbol X\|_2/\sigma_{\min}(\boldsymbol X)$ denotes the condition number of data matrix $\boldsymbol X$, then for $n$ tending to infinity, $k\in\mathbb Z^+$ and any $\delta\in(0,1)$, with probability exceeding $1-\delta$ over the random initialization, the two-layer ReLU neural network $f(\boldsymbol W,\boldsymbol a,\boldsymbol x)$ trained by gradient descent obeys
    \begin{equation}\label{final_data_convergence_rate}
\mathcal L(\boldsymbol W(k))<\left[1-\eta(1+o(1))\lambda^*\sigma_{\min}^2(\boldsymbol X)\right]^k\mathcal L(\boldsymbol W(0)).
    \end{equation}       
Or equivalently,
    \begin{equation}\label{final_data_convergence_rate_2}
    \mathcal L(\boldsymbol W(k))<\left(1-\frac{(1+o(1))(\lambda^*)^2\sigma_{\min}^4(\boldsymbol X)}{2n^{3/2}\|\boldsymbol X\|_2^2}\right)^k\mathcal L(\boldsymbol W(0)).
    \end{equation}
Furthermore, for any $\epsilon>0$ and sufficiently large $n$, with probability exceeding $1-\delta$ over the random initialization, after 
\begin{equation}
T=\mathcal O\left(\frac{n^{3/2}\|\boldsymbol X\|_2^2\log(1/\epsilon)}{(\lambda^*)^2\sigma_{\text{min}}^4(\boldsymbol X)}\right)
\end{equation}
iterations, the learned network has empirical quadratic loss upper bounded as $\epsilon$.
\end{theorem}
{\bf Remark}: The proof of Theorem \ref{theorem_GD_convergence} appears in Section \ref{section_convergence}. Theorem \ref{theorem_GD_convergence} reveals that as long as the network width is sufficiently large, then randomly initialized gradient descent converges to zero training loss at a linear rate. A comparison with state-of-the-art results concerning global trainability of nonlinear overparameterized neural networks equipped with quadratic loss can be found in Table 2 and Table 3. We improve the desired over-parameterization condition and iteration complexity under milder data assumption over previous results. 
\\\\
\begin{tabular}{|c|c|c|c|c|}
\hline
Reference & Over-para. Condition & Activation \\
\hline
\cite{zou_2018} & $\Omega(n^{26}L^{38})$ & ReLU\\
\hline
\cite{zhu_2018b} & $\widetilde{\Omega}(kn^{24}L^{12}/\Delta^8)$ & 
Semi-Smooth(ReLU) \\
\hline
\cite{zou_2019} & $\widetilde\Omega(kn^8L^{12}/\Delta^4)$ & 
ReLU \\
\hline
\cite{oymak_2020} & $\Omega(n^9\|\boldsymbol X\|_2^6)/\Delta^4)$(ReLU)  & Analytic/ReLU \\
\hline
\cite{du_2018a} & $\widetilde\Omega(2^{\mathcal O(L)}n^4/\lambda_{\text{min}}^4(\boldsymbol K^{(L)}))$ & Lipschitz Smooth\\
\hline
\cite{du_iclr} & $\Omega(\lambda_0^{-4}n^6\delta^{-3})$ & ReLU\\
\hline
\cite{wu_xiaoxia_2019} & $\Omega(\lambda_0^{-4}n^6\delta^{-3})$ & ReLU\\
\hline
\cite{arora_2019} & $\Omega(\lambda_0^{-4}\kappa_0^{-2}n^7\text{poly}(1/\delta))$ & ReLU\\
\hline
\cite{song_2020} & $\Omega(\lambda_0^{-4}n^4\log^3(n/\delta))$ &  ReLU \\
\hline
\cite{liao_2021} & $ \Theta(\lambda_0^{-4}n^5K/\delta)$ & ReLU \\
\hline
\textcolor{red}{Our Result (Theorem \ref{theorem_GD_convergence})} & $\Omega(n^3\log^{10\hbar+3}(n/\delta)/\sin\theta^*)$ & ReLU \\
\hline
\end{tabular}
\\\\ Table 2: Comparison of leading works on overparameterizaion conditions and activation functions required to ensure the global convergence of gradient descent for learning nonlinear neural networks. $L$ is the number of hidden layers of neural networks. $k$ in (\cite{zhu_2018b} \cite{zou_2019}) is the dimension of the output. $\Delta$ in (\cite{zhu_2018b} \cite{zou_2019} \cite{oymak_2020}) is the minimum distance between two training data points.
$\lambda_0$ in (\cite{du_iclr} \cite{arora_2019} \cite{song_2020}) is the smallest eigenvalue of NTK kernel $\boldsymbol H^{\infty}(H^{cts})$. $\boldsymbol X$ in (\cite{oymak_2020}) are minimum seperation parameter and data matrix, respectively. 
$\boldsymbol K^{(L)}$ in \cite{du_2018a} is the Gram matrix for $L$-hidden-layer neural network. $\kappa_0=\mathcal O(\epsilon\delta/\sqrt{n})\in(0,1]$ in (\cite{arora_2019}) controls the magnitude of random initialization as $\boldsymbol w_r(0)\sim\mathcal N(\boldsymbol 0,\kappa_0^2\boldsymbol I)$. (\cite{liao_2021}) adopted a different network structure using multiple multiple randomly masked subnetworks with the step size $\mathcal O(\lambda_0/n^2)$, and $K$ in (\cite{liao_2021}) is a fixed number of iteration.
\\\\
\begin{tabular}{|c|c|c|c|c|}
\hline
Reference & Iteration Complexity & Step Size \\
\hline
\cite{zou_2018} & $ \mathcal O(n^8L^9)$ & $1/(n^{29}L^{47})$ \\
\hline
\cite{zhu_2018b} & $ \mathcal O(n^6L^2\log(1/\epsilon)/\Delta^2)$ & 
$1/(n^{28}L^{14}\log^5m)$ \\
\hline
\cite{zou_2019} & $ \mathcal O(n^2L^2\log(1/\epsilon)/\Delta)$ & $1/(n^8L^{14})$ \\
\hline
\cite{oymak_2020} & $ \mathcal O(n^2\|\boldsymbol X\|_2^2\log(1/\epsilon)/\Delta^2)$ & $n\bar{\eta}/(3\|\boldsymbol y\|_2^2\|\boldsymbol X\|_2^2)$ \\
\hline
\cite{du_2018a} & $ \mathcal O(2^{\mathcal O(L)}n^2\log(1/\epsilon)/\lambda_{\text{min}}^2(\boldsymbol K^{(L)}))$ & $\lambda_0/(2^{\mathcal O(L)}n^2)$\\
\hline
\cite{du_iclr} & $ \mathcal O(\lambda_0^{-2}n^2\log(1/\epsilon))$ & $\lambda_0/n^2$\\
\hline
\cite{wu_xiaoxia_2019} & $ \mathcal O(\lambda_0^{-2}n\log(1/\epsilon))$ &
$1/\|\boldsymbol H^{\infty}\|_2$ \\
\hline
\cite{song_2020} & $ \mathcal O(\lambda_0^{-2}n^2\log(1/\epsilon))$ & $\lambda_0/n^2$ \\
\hline
\textcolor{red}{Our Result (Theorem \ref{theorem_GD_convergence})} & $ \mathcal O((\lambda^*)^{-2}n^{3/2}\|\boldsymbol X\|_2^2\log(1/\epsilon)/\sigma_{\text{min}}^4(\boldsymbol X))$ & $\lambda^*/(2\kappa^2 n^{3/2})$ \\
\hline
\end{tabular}  
\\\\ Table 3: Perspective on iteration complexity and step size (a.k.a. learning rate) required to ensure the empirical loss upper bounded by $\epsilon$. Our results clarify the effects of spectrums of angle-based matrix $\mho$ and data matrix $\boldsymbol X$ in determining the iteration complexity and step size in a clear manner.\\

We would like to enumerate and clarify certain immediate yet important consequences implied by Theorem \ref{theorem_GD_convergence} as follows:
\begin{enumerate}
\item{\textcolor{blue}{\it The mildest over-parameterization size without additional assumptions}}. Our over-parameterization size desired plays a key role in guaranteeing the global linear convergence of gradient descent, depending upon $n$, $\delta$ and data-dependent quantities.

Regarding the dependence of $n$ and the failure probability $\delta$, Theorem \ref{theorem_GD_convergence} takes a step towards closing the gap between theory and practice, demonstrating that gradient descent succeeds under the condition that the network width exceeds the order of $n^3\log^3(n/\delta)$. Compared with previous works on the trainability of overparameterized networks (\cite{arora_2019} \cite{du_2018a} \cite{du_iclr} \cite{oymak_2020} \cite{li_2018} \cite{song_2020} \cite{wu_xiaoxia_2019} \cite{zou_2019} \cite{zhu_2018b} \cite{liao_2021}), our over-parameterization condition improves previous results by decreasing the exponent on $n$. 
We notice that cubic or quadratic bounds on the width for two-layer ReLU nets have been claimed in (\cite{song_2020} \cite{nguyen_2021} \cite{das_2019}). However, their results require more additional assumptions on training data or activation function. Assumption 1.2 and Assumption 1.3 in (\cite{song_2020} introduce additional parameters $\alpha$ and $\theta$ which are hard to be verified. In the case of $\alpha=n$ and $\theta=\sqrt{n}$, the overparameterization size $\Omega(\lambda_0^{-4}n^4\log^3(n/\delta))$ (\cite{song_2020}) in the general case is not be improved. Additionally, i.i.d. sub-gaussian data vector assumption and complex weight condition at initialization are needed in (\cite{nguyen_2021}) and results in (\cite{das_2019}) relies on additional activation assumptions. In contrast, our overparameterization condition holds without any additional data assumptions or activation assumptions. Very recently, (\cite{noy_2021} \cite{gao_2021}) established tight over-parameterization condition $\Omega(n^2)$. However, the result in (\cite{noy_2021}) was obtained using a different network structure with various initialization magnitude than our setting. \cite{gao_2021} adopted deep implicit network which has infinite layers and different network structure from ours, in addition, \cite{gao_2021} introduce an auxiliary parameter $0<\gamma<\min\{1/2,1/4c\}$ and other restrictive assumptions which should be verified in practice. (\cite{kawaguchi_2020}) provided a $\widetilde\Omega(n)$ width upper bound under the assumption of real analytic activation units including sigmoid and hyperbolic tangents, which differs our choice of ReLU activation unit. Due to non-smooth property of ReLU unit, it is reasonable to expect a significantly higher amounts of overparameterization than smooth activation functions. Therefore, we achieve the smallest exponent on $n$ in the over-parameterization condition under mild assumptions for learning very high dimensional data.

Besides the milder depependence of required over-parameterization size on $n$, our theory also demonstrates the connection between over-parameterization size and $\lambda^*$, $\theta^*$ and $\sigma_{\min}(\boldsymbol X)$ in a clear manner, revealing the role of the geometrical (angular) and spectral properties of training data in required network width, respectively.

Firstly, it can be seen from (\ref{matrix_mho_definition}) in Section \ref{section_problem} that $\lambda^*=\lambda_{\min}(\mho)$ characterizes the angular distribution of dataset. (\cite{arora_2019} \cite{du_iclr} \cite{tsuchida_2017} \cite{xie_2017}) adopt the Gram matrix $\boldsymbol H^{\infty}\in\mathbb R^{n\times n}$ induced by the ReLU activation function
(\cite{oymak_2020}) uses 
$\boldsymbol\Sigma(\boldsymbol X)$,
which is the same matrix of $\boldsymbol H^{\infty}$. $\lambda_0=\lambda_{\min}(\boldsymbol H^{\infty})$ is used for characterzing over-parameterization size in (\cite{arora_2019} \cite{du_iclr} \cite{tsuchida_2017} \cite{xie_2017}).
It can be seen from the defintions of $\boldsymbol\mho$ and $\boldsymbol H^{\infty}$ that $\lambda^*$ is more fundamental quantity than $\lambda_0$ to characterize the input data property. More importantly, it is more tractable to use $\lambda^*$ to characterize spectral and geometrical properties of dataset, enabling more fine-grained analysis based on angular distribution of the input data. 

Secondly, the angle parameter $\theta^*$ indicates the the level of seperation between data points. It makes intutive sense that $\theta^*$ would have an obvious affect on desired over-parameterization size and smaller $\theta^*$ would give rise to larger required over-parameterization size. Our result demonstrates this intuitive idea by a factor of $(\sin\theta^*)^{-1}$. 
(\cite{oymak_2020}) adopted $\Delta$ as a minimum seperation parameter of dataset, which requires $\min(\|\boldsymbol x_i-\boldsymbol x_j\|_2,\|\boldsymbol x_i+\boldsymbol x_j\|_2)\geqslant\Delta$ for any pair of points $\boldsymbol x_i$ and $\boldsymbol x_j$ (Due to $\delta$ denotes the failure probability in this paper, we use the notation $\Delta$ instead of $\delta$ in (\cite{oymak_2020}) for avoiding confusion). It is evident from elementary trigonometrical calculation that $\Delta$ in (\cite{oymak_2020}) are connected with our angular parameter $\theta^*$ via $\Delta=\sqrt{2(1-|\cos\theta^*|)}$, giving rise to $\sin\theta^*=\Delta\sqrt{1-0.25\Delta^2}=\Delta(1-0.125\Delta^2+\mathcal O(\Delta^4))=\Theta(\Delta)$. For further comparison, we rewrite previous results and ours in terms of the same parameters including $\Delta$. Notice that $\lambda_0$ in (\cite{du_iclr} \cite{wu_xiaoxia_2019} \cite{arora_2019} \cite{song_2020} \cite{liao_2021}) relates the minimum separation parameter $\Delta$. Applying the estimate in $\lambda_0=\Omega(\Delta/n^2)$ (\cite{oymak_2020}) in Table 2. It follows that our over-parameterization size significantly outperforms all of previous results for learning high dimensional data.

Thirdly, due to $\lambda^*\geqslant\lambda_{\text{min}}(\boldsymbol X^\top\boldsymbol X)/4$ from Lemma \ref{lemma_Psi0}, we could expect larger $\lambda^*$ for larger $\lambda_{\min}(\boldsymbol X^\top\boldsymbol X)$. Owing to the complexity of $\lambda_{\min}(\boldsymbol X^\top\boldsymbol X)$, we address the dicussion of $\lambda_{\min}(\boldsymbol X^\top\boldsymbol X)$ in the part of dicussing the effect of dataset upon optimization, where we show that we can expect larger $\lambda_{\min}(\boldsymbol X^\top\boldsymbol X)$ in high probability for learning very high dimension. In conclusion, we therefore see that for sufficiently large dimension $d$, $\lambda_{\min}(\boldsymbol X^\top\boldsymbol X)$ and $\lambda^*$ can be expected to be larger in high probability, giving rise to smaller required over-parameterization size $m$ in high probability. Furthermore, we reveal that smaller required amount of neurons $m$ is requireed for sufficiently large data dimension $d$, implying the advantage of very high dimensional learning. 

The conciser overparameterization condition of $m=\Omega(n^3\log^{10\hbar+3}(n/\delta)/\sin\theta^*)$ reveals a different  manner. Firstly, $\theta^*$ indicates the level of seperation between the input data from an extreme perspective. Secondly, intuitively, given a data distribution over $\mathbb R^d$ with $d\geqslant n$, when the number of samples $n$ increase, $\sigma_{\min}(\boldsymbol X)$ would significantly increase in high probability, which will be dicussed later. Clearly, $\hbar$ characterizes the decay rate of $\sigma_{\min}(\boldsymbol X)$ as $n$ increase and larger $\hbar$ corresponds to more difficult seperation dataset. Hence, $\hbar$ characterizes the level of seperation between data points from an overall perspective in term of spectrum decay rate of the training samples with respect to the number of samples. Therefore, the condition $m=\Omega(n^3\log^{10\hbar+3}(n/\delta)/\sin\theta^*)$ uncovers the instrinsic and subtle connection between desired network width and the input data by quantitatively representing the geometrical and spectral properties of the input data.

\item{\textcolor{blue}{\it The effects of high-dimensional geometry and spectrum of dataset upon optimization}}.
From the result (\ref{final_data_convergence_rate}) in Theorem \ref{theorem_GD_convergence}, we see that we can expect faster global convergence rate when $\sigma_{\min}(\boldsymbol X)$ is larger which often occurs when input data samples are better seperated, demonstrating the effect of the spectral property of data matrix on convergence. The other form of Theorem \ref{theorem_GD_convergence} represented by result (\ref{final_data_convergence_rate_2}) reveals the close connection between the convergence rate of GD and the sample number, spectrums of activation matrix and data matrix in a clear manner. Spectrums of activation matrix and data matrix are closely related, which are in essence determined by high-dimensional geometry of dataset. Therefore, our results uncover that high-dimensional geometry and spectrum of data set play a crucial role in optimiazation. More importantly, Theorem \ref{theorem_GD_convergence} allows us to understand and grasp the amazing convergence phenomena via dissecting $\lambda^*, \sigma_{\min}(\boldsymbol X)$ and $\|\boldsymbol X\|_2$ in (\ref{final_data_convergence_rate_2}) for various training dataset in very high dimensional regime, which are rarely understood till now. 

Firstly, in essence, $\lambda^*=\lambda_{\min}(\mho)$ reflects the angular distribution of dataset. Due to $\lambda_{\text{min}}(\boldsymbol X^\top\boldsymbol X)/4\leqslant\lambda^*\leqslant1/2$, $\lambda^*$ is well controlled by $\lambda_{\text{min}}(\boldsymbol X^\top\boldsymbol X)$. Therefore, we focus on the effect of $\lambda_{\text{min}}(\boldsymbol X^\top\boldsymbol X)$ and $\lambda_{\text{max}}(\boldsymbol X^\top\boldsymbol X)=\|\boldsymbol X\|_2(d\geqslant n)$ on training. It is evident that the data set with better seperation property, i.e., data points are well-separated, can be expected to be a larger value of $\lambda_{\min}(\boldsymbol X^\top\boldsymbol X)$. In fact, there is a subtle connection of spectrum of data matrix and geometrical property of data set. By virture of \textcolor{red}{Ger\v{s}gorin disc theorem (\cite{gersgorin_1931})}, there exists $i\in[n]$ such that $\lambda_{\min}(\boldsymbol X^\top\boldsymbol X)$ lies in one disc centered in $(1,0)$ with radius $\sum_{j\not=i}|\cos\theta_{ij}|$, giving rise to $\lambda_{\min}(\boldsymbol X^\top\boldsymbol X)\geqslant1-\sum_{j\not=i}|\cos\theta_{ij}|$. In very high dimensional space $\mathbb R^d(d\geqslant n)$, it is reasonable that the random variable $\sum_{j\not=i}|\cos\theta_{ij}|$ would concentrate around its expectation in overhelming probability due to concentration of measure phenomena in high dimensional space. 
For instance, in very high dimensional space $\mathbb R^d$, data points i.i.d. sampling from many distributions such as i.i.d. isotropic random vectors tend to be almost orthogonal to each other (\cite{vershy_2018}), which is one of extraordinary phenomena of probability in high dimensional space and could become more pronounced in very high-dimensional space. This ``almost orthogonal'' phenomenon in very high-dimensional space enables the term $1-\sum_{j\not=i}|\cos\theta_{ij}|$ be lower bounded from a positive number in high probability. We must emphasis here that it is extremely complex to characterize precise lower bound of $1-\sum_{j\not=i}|\cos\theta_{ij}|$. However, we can get an intutive feel for some scenarios. For example, we can obtain asymptotic equivalence $1-\sum_{j\not=i}|\cos\theta_{ij}|\approx1-n/\sqrt{d}$ under certain data conditions such as isotropic random vectors as $d$ is sufficiently large. Therefore, for applications satisfying $d\geqslant n^2$, which already appear in modern scientific domains and are going to become more common in the future, implying that that larger $d$ would increase the possibility of larger positive number $\lambda_{\min}(\boldsymbol X^\top\boldsymbol X)$ which ensures large diversity in the data set in the very high-dimensional regime.

We proceed to analyze the upper bound of $\|\boldsymbol X\|_2^2=\lambda_{\max}(\boldsymbol X^\top\boldsymbol X)$ for very high dimensional dataset. Let $\lambda_1\geqslant\cdots\geqslant\lambda_n$ are eigenvalues of $\boldsymbol X^\top\boldsymbol X$ and denote the mean and variance $\{\lambda_i\}_{i=1}^n$ by $M$ and $V$, respectively, viz. $M=\sum_{i=1}^n\lambda_i/n=\text{tr}(\boldsymbol X^\top\boldsymbol X)/n$ and $V=\sum_{i=1}^n(\lambda_i-M)^2/n$. We have that $M-\sqrt{n-1}V\leqslant\lambda_i\leqslant M+\sqrt{n-1}V$ for every $i\in[n]$ (\cite{wang_2006_book}). A routine algebraic computation gives rise to $V=[\sum_{i=1}^n\lambda_i^2/n-2\text{tr}(\boldsymbol X^\top\boldsymbol X)M+nM^2]/n=\{\sum_{i=1}^n\lambda_i^2/n-M[2\text{tr}(\boldsymbol X^\top\boldsymbol X)-nM]\}/n=\{\text{tr}[(\boldsymbol X^\top\boldsymbol X)^2]-[\text{tr}(\boldsymbol X^\top\boldsymbol X)]^2/n\}/n=\text{tr}[(\boldsymbol X^\top\boldsymbol X)^2]/n-M^2$. Taking account of $\|\boldsymbol x_i\|_2=1$ and $\langle\boldsymbol x_i,\boldsymbol x_j\rangle=\cos\theta_{ij}$, it is clear that $M=1$ and the $i$-th diagonal entry of $(\boldsymbol X^\top\boldsymbol X)^2$ is $\|(\boldsymbol X^\top\boldsymbol X)_{i,\cdot}\|_2^2)$, yielding $\text{tr}[(\boldsymbol X^\top\boldsymbol X)^2]=\sum_{i=1}^n\|(\boldsymbol X^\top\boldsymbol X)_{i,\cdot}\|_2^2=\|\boldsymbol X^\top\boldsymbol X\|_{\text{F}}^2$. Hence $\lambda_{\max}(\boldsymbol X^\top\boldsymbol X)\leqslant M+\sqrt{n-1}V=1+\sqrt{n-1}(\|\boldsymbol X^\top\boldsymbol X\|_{\text{F}}^2/n-1)<1+\|\boldsymbol X^\top\boldsymbol X\|_{\text{F}}^2/\sqrt{n}-\sqrt{n-1}$. Thanks to $\|(\boldsymbol X^\top\boldsymbol X)_{i,\cdot}\|_2^2=\cos^2\theta_{i1}+\cos^2\theta_{i2}+\cdots+\cos^2\theta_{in}(i\in[n])$, we arrive at $\|\boldsymbol X^\top\boldsymbol X\|_{\text{F}}^2=n+\sum_{i=1}^n\sum_{j\not=i}\cos^2\theta_{ij}$ and thus $\lambda_{\max}(\boldsymbol X^\top\boldsymbol X)\leqslant1+(\sum_{i=1}^n\sum_{j\not=i}\cos^2\theta_{ij})/\sqrt{n}
+\sqrt{n}-\sqrt{n-1}=1+(\sum_{i=1}^n\sum_{j\not=i}\cos^2\theta_{ij})/\sqrt{n}
+\mathcal O(n^{-1/2})$. Henceforth, we find a concise connection between angular information of vectors in dataset and $\lambda_{\max}(\boldsymbol X^\top\boldsymbol X)$. In very high dimensional space $\mathbb R^d(d\geqslant n)$, the random variable $\sum_{i=1}^n\sum_{j\not=i}\cos^2\theta_{ij}$ can be analyzed via powerful Talagrand concentration inequality, giving its concentration around its expectation in overwhelming probability due to concentration of measure phenomena in high dimensional space. Based on geometrical initution that the expectation of angles between vectors would be concentrated around $\pi/2$ in very high dimensional space, we can expect $\lambda_{\max}(\boldsymbol X^\top\boldsymbol X)$ can be efficiently upper bounded in overhelming probability in very high dimension regime. For instance, i.i.d. random vectors sampling certain distribution tend to be almost orthogonal (e.g. $\mathbb E\cos^2\theta_{ij}\sim1/d$, here the expectation is taken over training samples drawing i.i.d. from an underlying distribution) to each other (\cite{vershy_2018}), in this case we get immediately $\lambda_{\max}(\boldsymbol X^\top\boldsymbol X)\leqslant1+\mathcal O(n^{3/2}/d+n^{-1/2})$ in high probability, where randomness arises from drawing $n$ i.i.d. training samples from the underlying distribution over $\mathbb R^d$. In very high dimensional setting $d=n^{1+\varepsilon}$ for $\varepsilon>0$, we obtain $\lambda_{\max}(\boldsymbol X^\top\boldsymbol X)\leqslant1+\mathcal O(n^{1/2-\varepsilon}+n^{-1/2})$ and results for three typical dimensions (1) $d=n$: $\lambda_{\max}(\boldsymbol X^\top\boldsymbol X)\leqslant1+\mathcal O(n^{1/2})$; (2) $d=n^{3/2}$: $\lambda_{\max}(\boldsymbol X^\top\boldsymbol X)\leqslant1+\mathcal O(1)$; (3) $d\geqslant n^2$: $\lambda_{\max}(\boldsymbol X^\top\boldsymbol X)\leqslant1+\mathcal O(n^{-1/2})$. Again, we discover the effect of very high dimension $d$ on the upper bound of $\|\boldsymbol X\|_2$ that would be expected to be efficiently controlled in the very high dimensional regime.

Similarly, we have $\lambda_{\text{min}}(\boldsymbol X^\top\boldsymbol X)\geqslant M-\sqrt{n-1}V=1-\sqrt{n-1}(\|\boldsymbol X^\top\boldsymbol X\|_{\text{F}}^2/n-1)>1-(\sum_{i=1}^n\sum_{j\not=i}\cos^2\theta_{ij})/\sqrt{n}\approx
1-n^{3/2}\mathbb E\cos^2\theta_{ij}$ in high probability. It follows that for some data distribution, $\lambda_{\text{min}}(\boldsymbol X^\top\boldsymbol X)\geqslant1-n^{3/2}/d=\Omega(1)$ for sufficiently large $d>n^{3/2}$. Clearly, $\sigma_{\text{min}}(\boldsymbol X)=\Omega(\log^{-\hbar}n)$ holds for certain $\hbar>0$ in this case. The key insight is that the order of $\sigma_{\text{min}}(\boldsymbol X)$ would significantly increase when the data dimension is sufficiently large. Again, we find the role of large $d$ in increasing $\sigma_{\text{min}}(\boldsymbol X)$ and the condition $\sigma_{\text{min}}(\boldsymbol X)=\Omega(\log^{-\hbar}n)$ holds for certain $\hbar>0$ is mild in very high dimensional data. Thanks to $\lambda^*\geqslant\lambda_{\text{min}}(\boldsymbol X^\top\boldsymbol X)/4$, we can expect larger $\lambda^*$ for larger $\lambda_{\min}(\boldsymbol X^\top\boldsymbol X)$. Therefore, from our convergence result (\ref{final_data_convergence_rate_2}), it can be seen that for learning very high dimensional training data, larger $d$ would decreaes the upper bound of $\|\boldsymbol X\|_2$ and increase the lower bound of $\sigma_{\text{min}}(\boldsymbol X), \lambda^*$ in high probability, enabling faster global convergence rate in high probability. In other words, we find very high dimension $d$ would significantly improve the extreme spectral bounds of data matrix, leading to much milder overparameterization condition and faster convergence rate. This is what Theorem \ref{theorem_GD_convergence} tells us the role of very high data  dimension in optimiaztion.

\item{\textcolor{blue}{\it The data-dependent global convergence guarantee with the most practical learning rate}}. We provide the data-dependent convergence rate represented in (\ref{final_data_convergence_rate}) and (\ref{final_data_convergence_rate_2}), revealing the effects of angular distribution and spectrums of data matrix in an explicit way. We next examine our iteration complexity result for achieving at most $\epsilon$ training loss in Theorem \ref{theorem_GD_convergence}. Compared with previous results, our result of iteration complexity clearly reveals the role of spectrums of the input data in determining training speed. From the foregoing analysis of the lower bound of $\sigma_{\text{min}}(\boldsymbol X)$ and the upper bound of $\|\boldsymbol X\|_2$, we find that the iteration complexity would be expected $\Omega((\lambda^*)^{-2}n^{2-\tau}\log(1/\epsilon))(\tau\in(0,1])$ in high probability for sufficiently large dimension $d$, implying the dominating role of $\lambda^*$ in determining the iteration complexity in such scenario.
Again, we can expect faster global convergence rate of GD and smaller iteration complexity when learning dataset of sufficient larger dimension.
Furthermore, we infer that the condition number of $\boldsymbol X^\top\boldsymbol X$, i.e., $\lambda_{\max}(\boldsymbol X^\top\boldsymbol X)/\lambda_{\min}(\boldsymbol X^\top\boldsymbol X)$ may be effeciently upper bounded by $\mathcal O(n^{1/2})$ and even $o(n^{1/2})$ for sufficiently large dimension $d$ in high probability. Hence in this case the step size $\eta=\lambda^*/(2\kappa^2 n^{3/2})=\Omega(\lambda^*/n^2)$ and even $\eta=\Omega(\lambda^*/n^{2-\tau'})(\tau'\in(0,1])$ for sufficiently large dimension $d$. With the aid of $\lambda^*\geqslant\sigma_{\min}^2(\boldsymbol X)/4$, $\sigma_{\text{min}}(\boldsymbol X)=\Omega(\log^{-\hbar}n)$ for certain $\hbar>0$, we find that our step size $\eta=\Omega(1/(n^{2-\tau'}\log^{\hbar}n))=\Omega(1/n^{2-\tau''})(\tau''\in(0,1])$ for convergence guarantee is the most practical result than previous step sizes in the literature in terms of the order of the number of input data samples $n$, which are often $\Omega(\Delta/n^4)$ (due to $\lambda_0/n^2$ and the estimate in $\lambda_0=\Omega(\Delta/n^2)$ (\cite{oymak_2020})) or even more smaller being too small for practical applications. In short, we provide the data-dependent convergence rate and iteration complexity using the most practical step size and very mild assumptions on the training data, indicating the advantage of very high data dimension in enabling faster convergence, smaller iterations and larger learning rate.

In summary, the above analysis reveals the power of our results (\ref{final_data_convergence_rate}) and (\ref{final_data_convergence_rate_2}) in analyzing the global convergence phenomenon via examining spectrums of angle-based matrix $\boldsymbol\mho$ and data matrix $\boldsymbol X$. Our results also imply the crucial role of higher data dimension $d$ in determining smaller desired over-parameterization size and more practical step size (learning rate). These findings uncovers the possibility of discovering more exciting phenomena in very high dimensional learning, which is rarely explored so far. Since the angular distribution of the input data plays a crucial role in determining $\lambda^*$, $\sigma_{\min}(\boldsymbol X)$ and $\|\boldsymbol X\|_2$, we therefore infer that the angular distribution arising from high-dimensional geometry of dataset is the most fundamental element in determining the convergenec rate of gradient-based optimization. 
It should be emphasized that the angular distribution in high-dimensional space is extremely complex and subtle, partly due to the complex angular correlation in high-dimensional geometry. \textcolor{red}{In essence, the remarkable global convergence phenomena in highly nonconvex optimizations originates from the wonderful chemical reaction of high-dimensional geometry of training data and overparameterized random matrix}. We hope our results would shed light on the effects of high-dimensional geometry of training data under very mild and realistic data conditions in very high dimensional learning scenario. \textcolor{red}{In the highest-level viewpoint, our findings of the global convergence mechanism can be summarized as}
$$\textcolor{blue}{\text{High-Dimensional Geometry}\rightarrow\text{Spectrums of Random Matrices}\rightarrow\text{Global Convergence}}$$
\end{enumerate}
Finally, we compare our theory with NTK theory in the following Table 4.
\\\\
\begin{tabular}{|c|c|c|c|c|}
\hline
 & \textcolor{red}{Our Theory} & \textcolor{red}{NTK Theory} \\
\hline
\textcolor{blue}{The Object of Tracking} & Objective Function &  Network Function \\
\hline
\textcolor{blue}{Tracking System} & Nonlinear System & Approximate Linear System  \\
\hline
\textcolor{blue}{Core Matrix} & Activation Matrix $\boldsymbol\Psi$ & Gram Kernel Matrix $\boldsymbol H^{\infty}$\\
\hline
\textcolor{blue}{Key Quantities} & $\lambda^*,\theta^*$ & $\lambda_0=\lambda_{\text{min}}(\boldsymbol H^{\infty})$\\
\hline
\textcolor{blue}{Convergence Mechanism} & $\boldsymbol\mho\mapsto\boldsymbol\Psi(0)\approx\boldsymbol\Psi(k)\mapsto\mathcal L(\boldsymbol W)$ & $\boldsymbol H^{\infty}\approx\boldsymbol H(0)\approx\boldsymbol H(k)\mapsto\mathcal L(\boldsymbol W)$ \\
\hline
\textcolor{blue}{Angular Distribution} & Angle-Based Matrix $\boldsymbol\mho,\theta^*$ & Not Explicitly Stated \\
\hline
\textcolor{blue}{Effects of Over-para.}  & Enable $\sigma_{\text{min}}(\boldsymbol\Psi(k))\uparrow$ &  Enable $\lambda_{\text{min}}(\boldsymbol H(k))\uparrow$ \\
\hline
\textcolor{blue}{Effects of The Input Data}  & Over-Para. Size/Convergence & Not Explicitly Stated \\
\hline
\textcolor{blue}{Max. Allowed Weight Change} & $\Theta(\sqrt{\lambda^*/n})$ & $\Theta(\lambda_0/n)$ \\
\hline
\textcolor{blue}{Over-para. Condition (ReLU)} & $\Omega(n^3\log^{10\hbar+3}(n/\delta)/\sin\theta^*)$ & $ \Omega(\lambda_0^{-4}n^4\log^3(n/\delta))$ \\
\hline
\textcolor{blue}{Typical Step Size (ReLU)} & $\lambda^*/(2\kappa^2 n^{3/2})$ & $\lambda_0/n^2$ \\
\hline
\end{tabular}
\\\\ Table 4: Core ideas, ingredients and implications in our theory and NTK theory. 
\section{Convergence Analysis: In Search of the Origin}\label{section_convergence}
\subsection{High-Level Proof Strategy and Roadmap}

We summarize high-level proof strategyand sketch for proving Theorem \ref{theorem_GD_lossfunction_update} and Theorem \ref{theorem_GD_convergence}. Firstly, proving Theorem \ref{theorem_GD_lossfunction_update} is based on a very careful characterization of the trajectory of objective function during optimization, which is the most complex and challenging part of our analysis and can be divided into the following four steps.

{\textcolor{blue}{{\bf Step 1: Geometric Analysis of Loss: Gradient and Directional Curvature}}.
With the Taylor expansion of empirical loss function, we reduce the problem to mainly analyze the gradient's Frobenium norm $\|\nabla \mathcal L(\boldsymbol W)\|_{\text{F}}$ and the directional curvature as a quadratic form $\text{vec}(\nabla \mathcal L(\boldsymbol W(k)))^\top\nabla^2\mathcal L(\boldsymbol W(k))\text{vec}(\nabla \mathcal L(\boldsymbol W(k)))$. Then our analysis of gradient gives a Polyak-\L ojasiewics-like condition (\cite{polyak_1963} \cite{lojasiewicz_1963}) that $\|\nabla \mathcal L(\boldsymbol W)\|_{\text{F}}$ can be lower bounded with the smallest singular value of activation matrix $\sigma_{\min}(\boldsymbol\Psi(k))$ and the smallest eigenvalue of data matrix $\lambda_{\min}(\boldsymbol X^\top\boldsymbol X)$, while the analysis of the directional curvature demonstrates that it can be also upper bounded with quantities concerning activation matrix $\boldsymbol\Psi(k)$ and data matrix $\boldsymbol X$. Intuitively, taking advantage of concentration phenomena in high-dimensional probability, we can infer that with the random initialization and sufficiently large $m$, $\boldsymbol\Psi(k)$ is close to $\boldsymbol\Psi(0)$ in some sense during the whole optimization process. Henceforth, the key point of the whole analysis boils down to characterizing the properties of initialization matrix $\boldsymbol\Psi(0)$ and the perturbation of random matrices $\boldsymbol\Psi(k)$ to initialization matrix $\boldsymbol\Psi(0)$, which constitute the most complex and challenging ingredients of our proof. 

{\textcolor{blue}{{\bf Step 2: Exploring Initial Random Activation Matrix}}.
To characterize random activation matrix at initialization, we make use of concentration inequalities to obtain useful norm and singular value estimates in high probability. The most difficulty in this part is to get a tight lower bound of the smallest singular value of intial activation matrix. To achieve this, we apply the technique of decomposing the matrix from two aspects: (1) diagonal part and off-diagonal parts; (2) expectation and deviation parts. To tackle with complex matrix which entry has the form of complex inner product of vectors, we employ recently developed matrix Bernstein inequality with carefully estimates of related  expectations to arrive at a lower bound of the least singular value of activation matrix at initialization.

{\textcolor{blue}{{\bf Step 3: Evaluating Small Perturbation of Activation Matrix}}. The underlying intuition is that most of the elements of $\boldsymbol\Psi(k)-\boldsymbol\Psi(0)$ do not change during gradient descent training. Further, many key quantities of $\boldsymbol\Psi(0)$ and $\boldsymbol\Psi(k)-\boldsymbol\Psi(0)$ are i.i.d. random variables or random matrices with small variances and can be therefore controlled by various concentration inequalities. And evaluations of expectations in concentration inequalities are very challenging but also quite interesting. The most challenge of this part is to exactly evaluate related high-dimensional integrl rising from the expectation.

{\textcolor{blue}{{\bf Step 4: Connecting Everything to The Dynamics of Loss Function}}. Combining all the above results, we obtain the decrease of loss function and demonstrate that the loss function of GD process with random initialization do converge to zero. The most challenges in this part lie in dissecting the singular values of activation matrix $\boldsymbol \Psi(k)$ and many detailed asymptotic analysis of immediate expressions, requiring carefully choices of $R$ and over-parameterization size $m$.

In a nutshell, our road map of optimization analysis is
$$\textcolor{blue}{\boldsymbol \Psi(0)\oplus\|\boldsymbol \Psi(k)-\boldsymbol \Psi(0)\|\longrightarrow\boldsymbol \Psi(k)\longrightarrow\|\nabla \mathcal L(\boldsymbol W(k))\|_{\text{F}}\&\nabla^2\mathcal L(\boldsymbol W(k))\longrightarrow d\mathcal L(\boldsymbol W(k))}$$
The most important \textcolor{red}{``\it{chemical reaction equations}''} in various stages are
\begin{align}\nonumber
\textcolor{blue}{
\text{Step 1}: \text{Loss Decrease}
=-\eta\cdot\text{Gradient Norm}^2
+\frac{\eta^2}{2}\text{Directional Curvature}+\text{Higher-Order Term}}
\end{align}
\begin{align}\nonumber
\textcolor{blue}{
\text{Step 2}:\text{Over-Para.}\xrightarrow[m\uparrow]{\text{\textcolor{red}{Matrix Bernstein+Weyl}}} \sigma_{\min}(\boldsymbol\Psi(0))=\sigma_{\min}(\phi'(\boldsymbol X^\top\boldsymbol W(0)))\uparrow}
\end{align}
\begin{align}\nonumber
\textcolor{blue}{
\text{Step 3}:\text{Concentration $\oplus$ High-Dimensional Geometry }\xrightarrow[]{\text{\textcolor{red}{Bernstein+Ger\v{s}gorin+Weyl}}}
\|\boldsymbol \Psi(k)-\boldsymbol \Psi(0)\|_2}
\end{align}
\begin{align}\nonumber
\textcolor{blue}{
\text{Step 2+Step 3}:\sigma_{\min}(\boldsymbol\Psi(0))\uparrow-\|\boldsymbol \Psi(k)-\boldsymbol \Psi(0)\|_2\xrightarrow[m\uparrow]{\text{\textcolor{red}{Asymptotic Analysis}}}\sigma_{\min}(\boldsymbol\Psi(k))\uparrow}
\end{align}
\begin{align}\nonumber
\textcolor{blue}{
\text{Step 4.1}:\sigma_{\min}(\boldsymbol\Psi(k))\uparrow\mapsto
\|\nabla \mathcal L(\boldsymbol W(k))\|_{\text{F}}^2>[2\lambda^*+o(1)]\sigma_{\min}^2(\boldsymbol X)\mathcal L(\boldsymbol W(k))}
\end{align}
\begin{align}\nonumber
\textcolor{blue}{
\text{Step 4.2}:\text{vec}(\nabla \mathcal L(\boldsymbol W(k)))^\top\nabla^2\mathcal L(\boldsymbol W(k))\text{vec}(\nabla \mathcal L(\boldsymbol W(k)))<[4n^{3/2}+\mathcal O(n^{1/2})]\|\boldsymbol X\|_2^2\mathcal L(\boldsymbol W(k))}
\end{align}

With Theorem \ref{theorem_GD_lossfunction_update} in place, we can establish Theorem \ref{theorem_GD_convergence} as follows. The core idea is to that if $\boldsymbol w$ stays in a $R$-ball during training, then we can upper bound the actual moving distance of weight vectors during GD in high probability based on Theorem \ref{theorem_GD_lossfunction_update}. Then the over-parameterization condition can be obtained by requiring actual moving distance of weight vectors less than $R$. The iteration complexity is an immediate consequence of Theorem \ref{theorem_GD_lossfunction_update}.
%
\subsection{The Interplay between Gradient and Directional Curvature}
We start our analysis by discussing the differentiability of the objective function $\mathcal L(\boldsymbol W)$. By virture of the gradient expression (\ref{grad_wr_func}), the ojective function $\mathcal L(\boldsymbol W)$ is continuously differentiable with respect to $\boldsymbol W=(\boldsymbol w_1,\cdots,\boldsymbol w_m)\in\mathbb R^{d\times m}$ if $\angle(\boldsymbol w_r,\boldsymbol x_i)\not=\pi/2$ for every $r\in[m]$ and $i\in[n]$. It is evident that the Lebesgue measure of $\{\boldsymbol W=(\boldsymbol w_1,\cdots,\boldsymbol w_m)\in\mathbb R^{d\times m}:\exists r\in[m] \text{ such that }\angle(\boldsymbol w_r,\boldsymbol x_i)=\pi/2\text{ for certain }i\in[n]\}$ is zero. Then it can be seen from (\ref{grad_wr_func}) that the gradient $\nabla \mathcal L(\boldsymbol W)$ is continuously differentiable almost surely (almost everywhere). Therefore, the objective function $\mathcal L(\boldsymbol W)$ is twice continuously differentiable almost surely, viz. the Hessian $\nabla^2\mathcal L(\boldsymbol W)$ is well defined and continuous with respect to $\boldsymbol W$ almost surely. This allows us to arrive at the second order Taylor expansion of the objective function $\mathcal L(\boldsymbol W(k+1))$ in a neighborhood of $\boldsymbol W(k)(k\geqslant0)$ (\cite{lang_1993} \cite{rudin_1964}), viz. there exists $\theta\in(0,1)$ such that
\begin{align}\nonumber
&\mathcal L(\boldsymbol W(k+1))\\\nonumber
=&\mathcal L(\boldsymbol W(k))+\langle\nabla \mathcal L(\boldsymbol W(k)),\boldsymbol W(k+1)-\boldsymbol W(k)\rangle\\\nonumber
&+\frac{1}{2}\text{vec}(\boldsymbol W(k+1)-\boldsymbol W(k))^\top\nabla^2\mathcal L[\boldsymbol W(k)+\theta(\boldsymbol W(k+1)-\boldsymbol W(k))]\text{vec}(\boldsymbol W(k+1)-\boldsymbol W(k)),
\end{align}
Since each entry of the matrix $\nabla^2\mathcal L[\boldsymbol W(k)+\theta(\boldsymbol W(k+1)-\boldsymbol W(k))]$ is a continous function of $\boldsymbol W(k)$ almost surely, we have
\begin{align}\nonumber
\nabla^2\mathcal L[\boldsymbol W(k)+\theta(\boldsymbol W(k+1)-\boldsymbol W(k))]=\nabla^2\mathcal L(\boldsymbol W(k))+\widetilde{\nabla^2\mathcal L}[\boldsymbol W(k)+\theta(\boldsymbol W(k+1)-\boldsymbol W(k))],
\end{align}
where each entry of the matrix $\widetilde{\nabla^2\mathcal L}[\boldsymbol W(k)+\theta(\boldsymbol W(k+1)-\boldsymbol W(k))]$ is an infinitesimal if $\|\boldsymbol W(k+1)-\boldsymbol W(k)\|_{\text{F}}$ tends to zero, giving rise to
 \begin{align}\nonumber
&\mathcal L(\boldsymbol W(k+1))\\\nonumber
=&\mathcal L(\boldsymbol W(k))+\langle\nabla \mathcal L(\boldsymbol W(k)),\boldsymbol W(k+1)-\boldsymbol W(k)\rangle\\\nonumber
&+\frac{1}{2}\text{vec}(\boldsymbol W(k+1)-\boldsymbol W(k))^\top\nabla^2\mathcal L(\boldsymbol W(k))\text{vec}(\boldsymbol W(k+1)-\boldsymbol W(k))+\alpha(\boldsymbol W(k+1)-\boldsymbol W(k)).
\end{align}
where $\alpha(\boldsymbol W(k+1)-\boldsymbol W(k))$ is the Peano's remainder of Taylor expansion. It is easy to see that $\alpha(\boldsymbol W(k+1)-\boldsymbol W(k))=o(\|\boldsymbol W(k+1)-\boldsymbol W(k)\|_{\text{F}}^2)=o(\eta^2\|\nabla \mathcal L(\boldsymbol W(k))\|_{\text{F}}^2)$ if $\|\boldsymbol W(k+1)-\boldsymbol W(k)\|_{\text{F}}$ tends to zero, yielding
\begin{align}\nonumber
&\mathcal L(\boldsymbol W(k+1))\\\nonumber
=&\mathcal L(\boldsymbol W(k))-\eta\|\nabla \mathcal L(\boldsymbol W(k))\|_{\text{F}}^2\\\label{loss_function_Taylor}
&+\frac{\eta^2}{2}\text{vec}(\nabla \mathcal L(\boldsymbol W(k)))^\top\nabla^2\mathcal L(\boldsymbol W(k))\text{vec}(\nabla \mathcal L(\boldsymbol W(k)))+o(\eta^2\|\nabla \mathcal L(\boldsymbol W(k))\|_{\text{F}}^2).
\end{align}
Henceforth, to analyze the dynamics of gradient descent, it is crucial to characterize the gradient and directional curvature of the objective function, or equivalently, local geometrical properties of the objective function. 
\subsection{Gradient Lower Bound: Connection with Activation and Data Matrices}
We first characterizes the gradient information via properties of activation matrix, data matrix and loss function. 
\begin{lemma}(\textcolor{blue}{Gradient Lower Bound})\label{lemma_gradient_lower_bound}
In the over-parameterized setting $n\leqslant m$, the Frobenius norm of the gradient of the objective function $\mathcal L(\boldsymbol W(k))$ satisfies
\begin{align}\nonumber
\|\nabla \mathcal L(\boldsymbol W(k))\|_{\text{F}}^2\geqslant&
\frac{2}{m}\sigma_{\min}^2(\boldsymbol\Psi(k))\lambda_{\min}(\boldsymbol X^\top\boldsymbol X)\mathcal L(\boldsymbol W(k)).
\end{align}
\end{lemma}
{\bf Remark}: Lemma \ref{lemma_gradient_lower_bound} is analogue to Polyak-\L ojasiewicz condition (\cite{polyak_1963} \cite{lojasiewicz_1963}) except that we do not have a constant on the RHS. As Lemma \ref{lemma_Psi_k_lower_bound} in step 1 of Section \ref{subsection_pieceing_together} states, in the over-parameterization setting, sufficiently large $m$ ensures that the smallest singular value of activation matrix $\boldsymbol\Psi(k)$ is bounded away from zero during optimization. Hence, the loss function of sufficiently wide two-layer ReLU networks satisfies Polyak-\L ojasiewicz condition (\cite{polyak_1963} \cite{lojasiewicz_1963}) around the initialization point, guaranteeing the global convergence of GD. From the geometric aspect, Lemma \ref{lemma_gradient_lower_bound} indicates that the loss function will become zero if the gradient is zero and thus there are no spurious stationary points on the landscape of loss function.\\
\textbf{Proof}: 
We observe that it is of great convenience to represent $\nabla \mathcal L(\boldsymbol W(k))$ in a compact form of a product of matrices. Hiding the dependence on $k$ for notation simplicity, we can explicitly calculate the gradient matrix as
\begin{align}\nonumber
\nabla \mathcal L(\boldsymbol W)
=&\left( \frac{\partial L(\boldsymbol W)}{\partial\boldsymbol w_1},\cdots,
\frac{\partial L(\boldsymbol W)}{\partial\boldsymbol w_m} \right)\\\nonumber
=&\frac{1}{\sqrt{m}}\left( a_1\sum_{i=1}^n(f_i-y_i)\mathbb{I}_{1,i}\boldsymbol x_i,\cdots, a_m\sum_{i=1}^n(f_i-y_i)\mathbb{I}_{m,i}\boldsymbol x_i \right)\\\nonumber
=&\frac{1}{\sqrt{m}}(\boldsymbol x_1,\cdots,\boldsymbol x_n)\left(  \begin{array}{ccc}
  f_1-y_1&  &     \\
    & \ddots &   \\
    &  &      f_n-y_n 
          \end{array}
              \right)
              \left(  \begin{array}{cccc}
  \mathbb{I}_{1,1}a_1 & \mathbb{I}_{2,1}a_2 & \cdots &  \mathbb{I}_{m,1}a_m  \\
  \mathbb{I}_{1,2}a_1 & \mathbb{I}_{2,2}a_2 & \cdots &  \mathbb{I}_{m,2}a_m  \\
    & \cdots & \cdots &  \\
  \mathbb{I}_{1,n}a_1 & \mathbb{I}_{2,n}a_2 & \cdots &  \mathbb{I}_{m,n}a_m 
                      \end{array}
              \right)\\\nonumber
=&\frac{1}{\sqrt{m}}\boldsymbol X\left(  \begin{array}{ccc}
  f_1-y_1&  &     \\
    & \ddots &   \\
    &  &      f_n-y_n 
          \end{array}
              \right)
              \left(  \begin{array}{cccc}
  \mathbb{I}_{1,1}  & \mathbb{I}_{2,1} & \cdots &  \mathbb{I}_{m,1} \\
  \mathbb{I}_{1,2}  & \mathbb{I}_{2,2} & \cdots &  \mathbb{I}_{m,2} \\
    & \cdots &  \\
  \mathbb{I}_{1,n}  & \mathbb{I}_{2,n} & \cdots &  \mathbb{I}_{m,n}
                      \end{array}
              \right)
              \left(  \begin{array}{ccc}
  a_1&  &     \\
    & \ddots &   \\
    &  &      a_m 
          \end{array}
              \right)\\\nonumber
=&\frac{1}{\sqrt{m}}\boldsymbol X\boldsymbol D_{\boldsymbol r}\boldsymbol\Psi\boldsymbol D_{\boldsymbol a}.
\end{align}
Henceforth, we arrive at
\begin{align}\nonumber
&m\|\nabla \mathcal L(\boldsymbol W)\|_{\text{F}}^2
=\|\boldsymbol X\boldsymbol D_{\boldsymbol r}\boldsymbol\Psi\boldsymbol D_{\boldsymbol a}\|_{\text{F}}^2
=\text{tr}(\boldsymbol D_{\boldsymbol a}^\top\boldsymbol\Psi^\top\boldsymbol D_{\boldsymbol r}^\top\boldsymbol X^\top\boldsymbol X\boldsymbol D_{\boldsymbol r}\boldsymbol\Psi\boldsymbol D_{\boldsymbol a})\\\nonumber
\stackrel{\text{(i)}}{=}&\text{tr}(\boldsymbol D_{\boldsymbol r}^\top\boldsymbol X^\top\boldsymbol X\boldsymbol D_{\boldsymbol r}\boldsymbol\Psi\boldsymbol D_{\boldsymbol a}\cdot\boldsymbol D_{\boldsymbol a}^\top\boldsymbol\Psi^\top)
=\text{tr}(\boldsymbol D_{\boldsymbol r}^\top\boldsymbol X^\top\boldsymbol X\boldsymbol D_{\boldsymbol r}\boldsymbol\Psi\boldsymbol\Psi^\top)
\stackrel{\text{(ii)}}{\geqslant}\lambda_{\min}(\boldsymbol\Psi\boldsymbol\Psi^\top)\text{tr}(\boldsymbol D_{\boldsymbol r}^\top\boldsymbol X^\top\boldsymbol X\boldsymbol D_{\boldsymbol r})\\\nonumber
\stackrel{\text{(iii)}}{=}&\lambda_{\min}(\boldsymbol\Psi\boldsymbol\Psi^\top)\text{tr}(\boldsymbol X^\top\boldsymbol X\boldsymbol D_{\boldsymbol r}\boldsymbol D_{\boldsymbol r}^\top)
=\lambda_{\min}(\boldsymbol\Psi\boldsymbol\Psi^\top)\text{tr}(\boldsymbol X^\top\boldsymbol X\boldsymbol D_{\boldsymbol r}^2)
\stackrel{\text{(iv)}}{\geqslant}\lambda_{\min}(\boldsymbol\Psi\boldsymbol\Psi^\top)\lambda_{\min}(\boldsymbol X^\top\boldsymbol X)\text{tr}(\boldsymbol D_{\boldsymbol r}^2)\\\nonumber
\stackrel{\text{(v)}}{=}&2\sigma_{\min}^2(\boldsymbol\Psi)\lambda_{\min}(\boldsymbol X^\top\boldsymbol X)\mathcal L(\boldsymbol W),
\end{align}
where the first line follows from $\|\boldsymbol A\|_{\text{F}}^2=\text{tr}(\boldsymbol A^\top\boldsymbol A)$, (i) and (iii) arise from the famous {\textcolor{red}{trace identity} $\text{tr}(\boldsymbol A\boldsymbol B)=\text{tr}(\boldsymbol B\boldsymbol A)$, and the inequalities (ii) and (iv) come from the {\textcolor{red}{trace inequality} $\text{tr}(\boldsymbol A\boldsymbol B)\geqslant\lambda_{\min}(\boldsymbol A)\text{tr}(\boldsymbol B)$ if $\boldsymbol A$ and $\boldsymbol B$ are positive semidefinite matrices. The last line (v) is due to that since $\boldsymbol\Psi\in\mathbb R^{n\times m}$ and $n\leqslant m$ in the over-parameterized setting, we have $\lambda_{\min}(\boldsymbol\Psi\boldsymbol\Psi^\top)=\sigma_{\min}^2(\boldsymbol\Psi)$. We also use the simple facts that and the obvious facts that $\text{tr}(\boldsymbol D_{\boldsymbol r}^2)=2\mathcal L(\boldsymbol W)$ and $\boldsymbol D_{\boldsymbol a}\boldsymbol D_{\boldsymbol a}^\top=\boldsymbol I_n$ due to $a_i\in\{-1,+1\}$.
\subsection{Directional Curvature Upper Bound: Decompositon and Estimations}
We are in a position to characterize the directional curvature along $k$-th gradient direction in terms of activation matrix, data matrix and loss function. 
\begin{lemma}(\textcolor{blue}{Directional Curvature Upper Bound})\label{lemma_Hessian_upper_bound}
The directional curvature along $k$-th gradient direction satisfies
\begin{align}\nonumber
&\text{vec}(\nabla \mathcal L(\boldsymbol W(k)))^\top\nabla^2\mathcal L(\boldsymbol W(k))\text{vec}(\nabla \mathcal L(\boldsymbol W(k)))\\\nonumber
\leqslant&\frac{2}{m^2}\left[\|\boldsymbol\Psi(k)\|_{\text{F}}^2+\sum_{1\leqslant r\not=s\leqslant m}\sqrt{\|\boldsymbol\psi_r(k)\|_2\boldsymbol\psi_s(k)\|_2\langle\boldsymbol\psi_r(k),\boldsymbol\psi_s(k)\rangle}
\right]\sqrt{\sum_{1\leqslant i,j\leqslant n}\langle \boldsymbol x_i,\boldsymbol x_j\rangle^4}\cdot\mathcal L(\boldsymbol W(k))
\end{align}
\end{lemma}
{\bf Remark}: Lemma \ref{lemma_Hessian_upper_bound} reveals that although the exact expression of directional curvature can be very complex, it can be still upper bounded by norms concerning activation matrix, angular information of dataset and loss function. The key insight here is that all the norms and inner product in our upper bound of directional curvature can be efficiently controlled with the aid of estimates of initial activation matrix and perturbation bound of activation matrix by applying powerful matrix eigenvalue inequalities, (matrix) concentration inequalities and integral evaluation. \\
\textbf{Proof}: 
We first calculate the Hessian expression of the objective function $\mathcal L(\boldsymbol W)$ explicitly. Again, we hide the dependence on $k$ for notation simplicity. Recall that for each $r\in[m]$, the partial gradient of $\boldsymbol W$ with respect to $\boldsymbol w_r$ is of the form
$$
\frac{\partial\mathcal L(\boldsymbol W)}{\partial\boldsymbol w_r}=\frac{a_r}{\sqrt{m}}\sum_{i=1}^n(f_i-y_i)\mathbb{I}_{r,i}\boldsymbol x_i=\frac{a_r}{\sqrt{m}}\sum_{i=1}^nr_i\mathbb{I}_{r,i}\boldsymbol x_i.
$$
Hence, the Hessian of $\mathcal L(\boldsymbol W)$, viz. $\nabla^2\mathcal L(\boldsymbol W)$ is a $md\times md$ matrix which can be regarded as a $m\times m$ block matrix where each entry is a block of size $d\times d$. For each $r\in[m]$, the $r$-th diagonal block of the Hessian $\nabla^2\mathcal L(\boldsymbol W)$ is
$$
\frac{\partial^2\mathcal L(\boldsymbol W)}{\partial\boldsymbol w_r^2}
=\frac{a_r}{\sqrt{m}}\sum_{i=1}^n\frac{\partial f_i}{\partial\boldsymbol w_r}\mathbb{I}_{r,i}\boldsymbol x_i^\top
=\frac{a_r}{\sqrt{m}}\sum_{i=1}^n\frac{a_r}{\sqrt{m}}\mathbb{I}_{r,i}\boldsymbol x_i  \mathbb{I}_{r,i}\boldsymbol x_i^\top
=\frac{a_r^2}{m}\sum_{i=1}^n\mathbb{I}_{r,i}^2\boldsymbol x_i\boldsymbol x_i^\top,
$$
and for each $r,s\in[m]$ and $r\not=s$, the $(r,s)$-th off-diagonal block of the Hessian $\nabla^2\mathcal L(\boldsymbol W)$ is
$$\frac{\partial^2\mathcal L(\boldsymbol W)}{\partial\boldsymbol w_r\partial\boldsymbol w_s}
=\frac{a_r}{\sqrt{m}}\sum_{i=1}^n\frac{\partial f_i}{\partial\boldsymbol w_s}\mathbb{I}_{r,i}\boldsymbol x_i^\top
=\frac{a_r}{\sqrt{m}}\sum_{i=1}^n\frac{a_s}{\sqrt{m}}\mathbb{I}_{s,i}\boldsymbol x_i  \mathbb{I}_{r,i}\boldsymbol x_i^\top
=\frac{a_ra_s}{m}\sum_{i=1}^n\mathbb{I}_{r,i}\mathbb{I}_{s,i}\boldsymbol x_i\boldsymbol x_i^\top.$$
To proceed, it is convenient to rewrite the Hessian in the form of matrix products. Firstly, for each $r\in[m]$, we see that
\begin{align}\nonumber
&\frac{\partial^2\mathcal L(\boldsymbol W)}{\partial\boldsymbol w_r^2}\\\nonumber
=&\frac{a_r^2}{m}(\mathbb{I}_{r,1}\boldsymbol x_1,\cdots,\mathbb{I}_{r,n}\boldsymbol x_n)(\mathbb{I}_{r,1}\boldsymbol x_1,\cdots,\mathbb{I}_{r,n}\boldsymbol x_n)^\top\\\nonumber
=&\frac{a_r^2}{m}(\boldsymbol x_1,\cdots,\boldsymbol x_n)\left(  \begin{array}{ccc}
  \mathbb{I}_{r,1} &  &     \\
    & \ddots &   \\
    &  &      \mathbb{I}_{r,n}   
          \end{array}
              \right)
              \left(  \begin{array}{ccc}
  \mathbb{I}_{r,1} &  &     \\
    & \ddots &   \\
    &  &      \mathbb{I}_{r,n}   
          \end{array}
              \right)
              \left(  \begin{array}{c}
                        \boldsymbol x_1^\top \\
                        \vdots \\
                        \boldsymbol x_n^\top
                      \end{array}
              \right)
\\\label{H_rr}
=&\frac{1}{m}\boldsymbol X\text{diag}(\boldsymbol\psi_r)\boldsymbol X^\top=\frac{1}{m}\boldsymbol X\boldsymbol D_{\boldsymbol\psi_r}\boldsymbol X^\top.
\end{align}
Here, we use the fact that $a_i\in\{-1,+1\}$ and $\mathbb{I}_{r,i}^2=\mathbb{I}_{r,i}$. For each $r,s\in[m]$ and $r\not=s$, 
\begin{align}\nonumber
&\frac{\partial^2\mathcal L(\boldsymbol W)}{\partial\boldsymbol w_r\partial\boldsymbol w_s}\\\nonumber
=&\frac{a_ra_s}{m}(\mathbb{I}_{r,1}\boldsymbol x_1,\cdots,\mathbb{I}_{r,n}\boldsymbol x_n)(\mathbb{I}_{s,1}\boldsymbol x_1,\cdots,\mathbb{I}_{s,n}\boldsymbol x_n)^\top\\\nonumber
=&\frac{a_ra_s}{m}(\boldsymbol x_1,\cdots,\boldsymbol x_n)\left(  \begin{array}{ccc}
  \mathbb{I}_{r,1} &  &     \\
    & \ddots &   \\
    &  &      \mathbb{I}_{r,n}   
          \end{array}
              \right)
              \left(  \begin{array}{ccc}
  \mathbb{I}_{s,1} &  &     \\
    & \ddots &   \\
    &  &      \mathbb{I}_{s,n}   
          \end{array}
              \right)
              \left(  \begin{array}{c}
                        \boldsymbol x_1^\top \\
                        \vdots \\
                        \boldsymbol x_n^\top
                      \end{array}
              \right)
\\\label{H_rs}
=&\frac{a_ra_s}{m}\boldsymbol X\text{diag}(\boldsymbol\psi_r)\text{diag}(\boldsymbol\psi_s)\boldsymbol X^\top=\frac{a_ra_s}{m}\boldsymbol X\boldsymbol D_{\boldsymbol\psi_r}\boldsymbol D_{\boldsymbol\psi_s}\boldsymbol X^\top.
\end{align} 
For any matrix $\boldsymbol\Xi=(\boldsymbol\xi_1,\cdots,\boldsymbol\xi_m)\in\mathbb R^{d\times m}$, the key quantity in our analysis is the Hessian quadratic form $\text{vec}(\boldsymbol\Xi)^\top\nabla^2\mathcal L(\boldsymbol W)\text{vec}(\boldsymbol\Xi)$, which is the directional-curvature form for $\mathcal L(\boldsymbol W)$ or the \textcolor{blue}{\it directional cuvature} of $\mathcal L(\boldsymbol W)$ along the matrix $\boldsymbol\Xi$ of the same dimension as $\boldsymbol W$. We decompose the directional-curvature form into the diagonal part and off-diagonal part as
\begin{align}\nonumber
\text{vec}(\boldsymbol\Xi)^\top\nabla^2\mathcal L(\boldsymbol W)\text{vec}(\boldsymbol\Xi)
=\underbrace{\sum_{1\leqslant r\leqslant m}\boldsymbol\xi_r^\top\frac{\partial^2\mathcal L(\boldsymbol W)}{\partial\boldsymbol w_r^2}\boldsymbol\xi_r}_{\text{diagonal part}}
+\underbrace{\sum_{1\leqslant r\not=s\leqslant m}\boldsymbol\xi_r^\top\frac{\partial^2\mathcal L(\boldsymbol W)}{\partial\boldsymbol w_r\partial\boldsymbol w_s}\boldsymbol\xi_s}_{\text{off-diagonal part}}.
\end{align}
Substituting each column vector of the gradient matrix $\nabla\mathcal L(\boldsymbol W)$ for $\boldsymbol\xi_r$ and making use of (\ref{H_rr}) and (\ref{H_rs}), we arrive at
\begin{align}\nonumber
&\text{vec}(\nabla \mathcal L(\boldsymbol W))^\top\nabla^2\mathcal L(\boldsymbol W)\text{vec}(\nabla \mathcal L(\boldsymbol W)\\\nonumber
=&\sum_{r=1}^m \left[ \frac{\partial\mathcal L(\boldsymbol W)}{\partial\boldsymbol w_r} \right]^\top\frac{\partial^2\mathcal L(\boldsymbol W)}{\partial\boldsymbol w_r^2}\left[ \frac{\partial\mathcal L(\boldsymbol W)}{\partial\boldsymbol w_r} \right] 
+\sum_{1\leqslant r\not=s\leqslant m}\left[ \frac{\partial\mathcal L(\boldsymbol W)}{\partial\boldsymbol w_r} \right]^\top\frac{\partial^2\mathcal L(\boldsymbol W)}{\partial\boldsymbol w_r\partial\boldsymbol w_s}\left[ \frac{\partial L(\boldsymbol W)}{\partial\boldsymbol w_s} \right]\\\label{dir_curvature}
=&\underbrace{\sum_{r=1}^m\left[ \frac{\partial L(\boldsymbol W)}{\partial\boldsymbol w_r} \right]^\top\frac{1}{m}\boldsymbol X\boldsymbol D_{\boldsymbol\psi_r}\boldsymbol X^\top\left[ \frac{\partial L(\boldsymbol W)}{\partial\boldsymbol w_r} \right]}_{\heartsuit_1}
+\underbrace{\sum_{r\not=s}\left[ \frac{\partial L(\boldsymbol W)}{\partial\boldsymbol w_r} \right]^\top  \frac{a_ra_s}{m}\boldsymbol X\boldsymbol D_{\boldsymbol\psi_r}\boldsymbol D_{\boldsymbol\psi_s}\boldsymbol X^\top \left[ \frac{\partial L(\boldsymbol W)}{\partial\boldsymbol w_s} \right]}_{\heartsuit_2}.
\end{align}
In the sequel, we shall estimate the upper bounds of the diagonal and off-diagonal parts separately in (\ref{dir_curvature}) using careful algebraic calculations with trace identity. We use $\boldsymbol\xi_r$ as $\partial\mathcal L(\boldsymbol W)/\partial\boldsymbol w_r$ for notation simplicity and $\text{diag}(\boldsymbol\psi_r)$ as $\boldsymbol D_{\boldsymbol\psi_r}$ for identifying the diagonal structure of the matrix more clearly, when needed.\\\\
{\textcolor{blue}{{\bf Step 1: Bounding the diagonal term $\heartsuit_1$}}\\
We first control the diagonal part of the directional-curvature form $\heartsuit_1$ via a series of algebraic manipulations as
\begin{align}\nonumber
\heartsuit_1=
&\sum_{r=1}^m\left[ \frac{\partial L(\boldsymbol W)}{\partial\boldsymbol w_r} \right]^\top\frac{1}{m}\boldsymbol X\boldsymbol D_{\boldsymbol\psi_r}\boldsymbol X^\top\left[ \frac{\partial L(\boldsymbol W)}{\partial\boldsymbol w_r} \right]
=\frac{1}{m}\sum_{r=1}^m\boldsymbol\xi_r^\top\boldsymbol X\text{diag}(\boldsymbol\psi_r)\boldsymbol X^\top\boldsymbol\xi_r\\\nonumber
=&\frac{1}{m}\sum_{r=1}^m\boldsymbol\xi_r^\top\boldsymbol X\text{diag}(\boldsymbol\psi_r)^\top\text{diag}(\boldsymbol\psi_r)\boldsymbol X^\top\boldsymbol\xi_r=\frac{1}{m}\sum_{r=1}^m[\text{diag}(\boldsymbol\psi_r)\boldsymbol X^\top\boldsymbol\xi_r]^\top[\text{diag}(\boldsymbol\psi_r)\boldsymbol X^\top\boldsymbol\xi_r]\\\nonumber
=&\frac{1}{m}\sum_{r=1}^m\left\|\text{diag}(\boldsymbol\psi_r)\boldsymbol X^\top\boldsymbol\xi_r\right\|_2^2
=\frac{1}{m}\sum_{r=1}^m\left\|\text{diag}(\boldsymbol\psi_r)((\boldsymbol X^\top\boldsymbol\xi_r)_1,\cdots,(\boldsymbol X^\top\boldsymbol\xi_r)_n)^\top  \right\|_2^2\\\nonumber
=&\frac{1}{m}\sum_{r=1}^m\sum_{i=1}^n  (\boldsymbol\psi_r)_i^2 \langle\boldsymbol\xi_r,\boldsymbol x_i\rangle^2
=\frac{1}{m}\sum_{r=1}^m\sum_{i=1}^n  (\boldsymbol\psi_r)_i \langle\boldsymbol\xi_r,\boldsymbol x_i\rangle^2\\\nonumber
=&\frac{1}{m}\sum_{r=1}^m\sum_{i=1}^n  (\boldsymbol\psi_r)_i \Big<\frac{a_r}{\sqrt{m}}\sum_{j=1}^nr_j\mathbb{I}_{r,j}\boldsymbol x_j,\boldsymbol x_i\Big>^2\\\nonumber
=&\frac{1}{m^2}\sum_{r=1}^m\sum_{i=1}^n\left[(\boldsymbol\psi_r)_i \Big(\sum_{j=1}^n r_j\mathbb{I}_{r,j}\langle \boldsymbol x_i,\boldsymbol x_j\rangle\Big)^2\right]\\\nonumber
\stackrel{(\text{i})}{\leqslant}&\frac{1}{m^2}\sum_{r=1}^m\sum_{i=1}^n\left[(\boldsymbol\psi_r)_i \Big(\sum_{j=1}^nr_j^2\Big)\Big(\sum_{j=1}^n\mathbb{I}_{r,j}^2\langle \boldsymbol x_i,\boldsymbol x_j\rangle^2\Big)\right]\\\nonumber
=&\frac{2}{m^2}\sum_{r=1}^m\sum_{i=1}^n\left[(\boldsymbol\psi_r)_i\Big( \sum_{j=1}^n\mathbb{I}_{r,j}\langle \boldsymbol x_i,\boldsymbol x_j\rangle^2\Big)\right]\mathcal L(\boldsymbol W)\\\nonumber
\stackrel{(\text{ii})}{\leqslant}&\frac{2}{m^2}\sum_{r=1}^m\sqrt{\left[\sum_{i=1}^n(\boldsymbol\psi_r)_i^2\right]\cdot \left[\sum_{i=1}^n \Big( \sum_{j=1}^n\mathbb{I}_{r,j}\langle \boldsymbol x_i,\boldsymbol x_j\rangle^2\Big)^2\right]} \mathcal L(\boldsymbol W)\\\nonumber
\stackrel{(\text{iii})}{\leqslant}&\frac{2}{m^2}\sum_{r=1}^m\sqrt{\|\boldsymbol\psi_r\|_2^2\cdot \left[\sum_{i=1}^n \Big( \sum_{j=1}^n\mathbb{I}_{r,j}^2\Big)\Big(\sum_{j=1}^n\langle \boldsymbol x_i,\boldsymbol x_j\rangle^4\Big)\right]} \mathcal L(\boldsymbol W)\\\nonumber
=&\frac{2}{m^2}\sum_{r=1}^m\sqrt{\|\boldsymbol\psi_r\|_2^2\cdot \|\boldsymbol\psi_r\|_2^2\sum_{i,j=1}^n \langle\boldsymbol x_i,\boldsymbol x_j\rangle^4}\cdot \mathcal L(\boldsymbol W)\\\nonumber
=&\frac{2}{m^2}\sum_{r=1}^m\|\boldsymbol\psi_r\|_2^2\sqrt{ \sum_{i,j=1}^n \langle\boldsymbol x_i,\boldsymbol x_j\rangle^4}\cdot\mathcal L(\boldsymbol W)
=\frac{2}{m^2}\|\boldsymbol\Psi\|_{\text{F}}^2\sqrt{ \sum_{i,j=1}^n \langle\boldsymbol x_i,\boldsymbol x_j\rangle^4}\cdot\mathcal L(\boldsymbol W),
\end{align}
where (i),(ii) and (iii) are due to {\textcolor{red}{Cauchy-Schwartz inequality}.\\\\
{\textcolor{blue}{{\bf Step 2: bounding the off-diagonal term $\heartsuit_2$}}\\
We are now in a position to control the off-diagonal part of the directional-curvature form $\heartsuit_2$ with straightforward computations as 
\begin{align}\nonumber
\heartsuit_2=&\sum_{r\not=s}\left[ \frac{\partial L(\boldsymbol W)}{\partial\boldsymbol w_r} \right]^\top  \frac{a_ra_s}{m}\boldsymbol X\boldsymbol D_{\boldsymbol\psi_r}\boldsymbol D_{\boldsymbol\psi_s}\boldsymbol X^\top \left[ \frac{\partial L(\boldsymbol W)}{\partial\boldsymbol w_s} \right]
=\sum_{r\not=s}\boldsymbol\xi_r^\top  \frac{a_ra_s}{m}\boldsymbol X\boldsymbol D_{\boldsymbol\psi_r}\boldsymbol D_{\boldsymbol\psi_s}\boldsymbol X^\top \boldsymbol\xi_s\\\nonumber
=&\frac{1}{m}\sum_{r\not=s}\langle a_r\boldsymbol D_{\boldsymbol\psi_r}\boldsymbol X^\top\boldsymbol\xi_r, a_s\boldsymbol D_{\boldsymbol\psi_s}\boldsymbol X^\top\boldsymbol\xi_s\rangle \\\nonumber
=&\frac{1}{m}\sum_{r\not=s} a_r a_s\text{tr}\left(\boldsymbol\xi_r^\top  \boldsymbol X\boldsymbol D_{\boldsymbol\psi_r}\boldsymbol D_{\boldsymbol\psi_s}\boldsymbol X^\top \boldsymbol\xi_s\right)
\stackrel{\text{(i)}}{=}\frac{1}{m}\sum_{r\not=s} a_r a_s\text{tr}\left(\boldsymbol D_{\boldsymbol\psi_r}\boldsymbol D_{\boldsymbol\psi_s}\boldsymbol X^\top \boldsymbol\xi_s\boldsymbol\xi_r^\top  \boldsymbol X\right)\\\nonumber
=&\frac{1}{m}\sum_{r\not=s} a_r a_s\sum_{i=1}^n\left(\boldsymbol D_{\boldsymbol\psi_r}\boldsymbol D_{\boldsymbol\psi_s}\boldsymbol X^\top \boldsymbol\xi_s\boldsymbol\xi_r^\top  \boldsymbol X\right)_{ii}
=\frac{1}{m}\sum_{r\not=s} a_r a_s\sum_{i=1}^n\left(\boldsymbol D_{\boldsymbol\psi_r}\boldsymbol D_{\boldsymbol\psi_s}\right)_{ii}\left(\boldsymbol X^\top \boldsymbol\xi_s\boldsymbol\xi_r^\top  \boldsymbol X\right)_{ii}\\\nonumber
=&\frac{1}{m}\sum_{r\not=s} a_r a_s\sum_{i=1}^n(\boldsymbol\psi_r)_i(\boldsymbol\psi_s)_i  \left(\boldsymbol X^\top \boldsymbol\xi_s\right)_{i,1}\left(\boldsymbol\xi_r^\top  \boldsymbol X\right)_{1,i}\\\nonumber
=&\frac{1}{m}\sum_{r\not=s} a_r a_s\sum_{i=1}^n (\boldsymbol\psi_r)_i(\boldsymbol\psi_s)_i \langle\boldsymbol\xi_r,\boldsymbol x_i\rangle\langle\boldsymbol\xi_s,\boldsymbol x_i\rangle\\\nonumber
=&\frac{1}{m}\sum_{r\not=s} a_r a_s\sum_{i=1}^n (\boldsymbol\psi_r)_i(\boldsymbol\psi_s)_i \Big<\frac{a_r}{\sqrt{m}}\sum_{j=1}^nr_j\mathbb{I}_{r,j}\boldsymbol x_j,\boldsymbol x_i\Big>\Big<\frac{a_s}{\sqrt{m}}\sum_{j=1}^nr_j\mathbb{I}_{s,j}\boldsymbol x_j,\boldsymbol x_i\Big>\\\nonumber
=&\frac{1}{m^2}\sum_{r\not=s}\sum_{i=1}^n (\boldsymbol\psi_r)_i(\boldsymbol\psi_s)_i \Big(\sum_{j=1}^nr_j\mathbb{I}_{r,j}\langle\boldsymbol x_j,\boldsymbol x_i\rangle\Big)\Big(\sum_{j=1}^nr_j\mathbb{I}_{s,j}\langle\boldsymbol x_j,\boldsymbol x_i\rangle\Big)\\\nonumber
\stackrel{\text{(ii)}}{\leqslant}&\frac{1}{m^2}\sum_{r\not=s}\sum_{i=1}^n (\boldsymbol\psi_r)_i(\boldsymbol\psi_s)_i \sqrt{\Big(\sum_{j=1}^nr_j^2\Big)\Big( \sum_{j=1}^n\mathbb{I}_{r,j}^2\langle\boldsymbol x_j,\boldsymbol x_i\rangle^2\Big)}\sqrt{\Big(\sum_{j=1}^nr_j^2\Big)\Big(\sum_{j=1}^n\mathbb{I}_{s,j}^2\langle\boldsymbol x_j,\boldsymbol x_i\rangle^2\Big)}\\\nonumber
=&\frac{2}{m^2}\sum_{r\not=s}\left[\sum_{i=1}^n (\boldsymbol\psi_r)_i(\boldsymbol\psi_s)_i \cdot \sqrt{\Big( \sum_{j=1}^n\mathbb{I}_{r,j}\langle\boldsymbol x_j,\boldsymbol x_i\rangle^2\Big)\Big( \sum_{j=1}^n\mathbb{I}_{s,j}\langle\boldsymbol x_j,\boldsymbol x_i\rangle^2\Big)}\right]\cdot\mathcal L(\boldsymbol W)\\\nonumber
\stackrel{\text{(iii)}}{\leqslant}&\frac{2}{m^2}\sum_{r\not=s}\sqrt{\left[\sum_{i=1}^n (\boldsymbol\psi_r)_i^2(\boldsymbol\psi_s)_i^2\right] \left[\sum_{i=1}^n
\Big( \sum_{j=1}^n\mathbb{I}_{r,j}\langle\boldsymbol x_j,\boldsymbol x_i\rangle^2\Big)\Big( \sum_{j=1}^n\mathbb{I}_{s,j}\langle\boldsymbol x_j,\boldsymbol x_i\rangle^2\Big)\right] }\cdot\mathcal L(\boldsymbol W)\\\nonumber
\stackrel{\text{(iv)}}{\leqslant}&\frac{2}{m^2}\sum_{r\not=s}\sqrt{\left[\sum_{i=1}^n (\boldsymbol\psi_r)_i^2(\boldsymbol\psi_s)_i^2\right] \left[\sum_{i=1}^n
\sqrt{\Big( \sum_{j=1}^n\mathbb{I}_{r,j}^2\Big)\Big(\sum_{j=1}^n\langle \boldsymbol x_i,\boldsymbol x_j\rangle^4\Big)} \sqrt{\Big( \sum_{j=1}^n\mathbb{I}_{s,j}^2\Big)\Big(\sum_{j=1}^n\langle \boldsymbol x_i,\boldsymbol x_j\rangle^4\Big)} \right] }\cdot\mathcal L(\boldsymbol W)\\\nonumber
=&\frac{2}{m^2}\sum_{r\not=s}\sqrt{\langle\boldsymbol\psi_r,\boldsymbol\psi_s\rangle \|\boldsymbol\psi_r\|_2\|\boldsymbol\psi_s\|_2}\sqrt{\sum_{i,j=1}^n\langle \boldsymbol x_i,\boldsymbol x_j\rangle^4 }\cdot\mathcal L(\boldsymbol W).
\end{align}
Here (i) holds due to $\text{tr}(\boldsymbol A\boldsymbol B)=\text{tr}(\boldsymbol B\boldsymbol A)$, whereas (ii),(iii) and (iv) arise from {\textcolor{red}{Cauchy-Schwartz inequality}.\\
To proceed, we will need the following auxiliary lemma upper bounding the main ingredient of right-hand-side of the bound in Lemma \ref{lemma_Hessian_upper_bound}.
\begin{lemma}\label{lemma_Phi_fact_1}
\begin{align}\nonumber
&\sum_{1\leqslant r\not=s\leqslant m}\sqrt{\langle\boldsymbol\psi_r(k),\boldsymbol\psi_s(k)\rangle \|\boldsymbol\psi_r(k)\|_2\|\boldsymbol\psi_s(k)\|_2}\\\nonumber
<&m\|\boldsymbol\Psi(k)\|_{\text{F}}\left[\sum_{i=1}^n\|\widetilde{\boldsymbol\psi_i}(k)\|_2^4+\sum_{1\leqslant i\not=j\leqslant n}\langle\widetilde{\boldsymbol\psi_i}(k),\widetilde{\boldsymbol\psi_j}(k)\rangle^2-\frac{\|\boldsymbol\Psi(k)\|_{\text{F}}^4}{m}\right]^{1/4}.
\end{align}
\end{lemma}
{\bf Remark}: Since that the entries of $\boldsymbol\psi_r(0)$ are not mutually independent, while the entries of $\widetilde{\boldsymbol\psi_i}(0)$ are mutually independent due to the statistically indendency of $\boldsymbol w_1(0),\boldsymbol w_2(0),\cdots,\boldsymbol w_m(0)$, it is more convenient to obtain proper estimates of expressions concerning $\widetilde{\boldsymbol\psi_i}(0)$ in high probability than expressions concerning $\boldsymbol\psi_r(0)$. Lemma \ref{lemma_Phi_fact_1} establishes an intrinsic relationship of colmun and row vectors of activation matrix $\boldsymbol\Psi$. With the assistance of Lemma \ref{lemma_Phi_fact_1}, we can convert the upper bound of the expression of $\boldsymbol\psi_r(0)$ to upper bound of the expression of $\widetilde{\boldsymbol\psi_i}(0)$. Again, we hide the dependence on $k$ for notation simplicity.\\
\textbf{Proof}: Invoking {\textcolor{red}{Cauchy-Schwartz} three times and performing simple calculations yields
\begin{align}\nonumber
&\sum_{1\leqslant r\not=s\leqslant m}\sqrt{\langle\boldsymbol\psi_r,\boldsymbol\psi_s\rangle \|\boldsymbol\psi_r\|_2\|\boldsymbol\psi_s\|_2}\\\nonumber
\leqslant&\sqrt{m(m-1)\sum_{r\not=s}\langle\boldsymbol\psi_r,\boldsymbol\psi_s\rangle \|\boldsymbol\psi_r\|_2\|\boldsymbol\psi_s\|_2}\\\nonumber
\leqslant&\sqrt{m(m-1)\sqrt{\sum_{r\not=s}\langle\boldsymbol\psi_r,\boldsymbol\psi_s\rangle^2\sum_{r\not=s}\|\boldsymbol\psi_r\|_2^2\|\boldsymbol\psi_s\|_2^2}}\\\nonumber
=&\sqrt{m(m-1)\sqrt{\sum_{r\not=s}\langle\boldsymbol\psi_r,\boldsymbol\psi_s\rangle^2 \left[ \Big(\sum_{r=1}^m\|\boldsymbol\psi_r\|_2^2\Big)^2-\sum_{r=1}^m\|\boldsymbol\psi_r\|_2^4\right] } }\\\nonumber
\leqslant&\sqrt{m(m-1)\sqrt{\sum_{r\not=s}\langle\boldsymbol\psi_r,\boldsymbol\psi_s\rangle^2 \left[ \|\boldsymbol\Psi\|_{\text{F}}^4-\frac{1}{m}\left(\sum_{r=1}^m\|\boldsymbol\psi_r\|_2^2\right)^2\right] } }\\\nonumber
=&\sqrt{m(m-1)\sqrt{\sum_{r\not=s}\langle\boldsymbol\psi_r,\boldsymbol\psi_s\rangle^2 \left[ \|\boldsymbol\Psi\|_{\text{F}}^4-\frac{1}{m}\|\boldsymbol\Psi\|_{\text{F}}^4\right] } }\\\label{lemma_Phi_fact_1_result_1}
=&\sqrt{ (m-1)\sqrt{m(m-1)}\|\boldsymbol\Psi\|_{\text{F}}^2\sqrt{\sum_{r\not=s}\langle\boldsymbol\psi_r,\boldsymbol\psi_s\rangle^2} }<m\|\boldsymbol\Psi\|_{\text{F}}\left[\sum_{r\not=s}\langle\boldsymbol\psi_r,\boldsymbol\psi_s\rangle^2\right]^{1/4}.
\end{align}
Since $\boldsymbol\Psi\boldsymbol\Psi^\top\in\mathbb R^{n\times n}$ and $\boldsymbol\Psi^\top\boldsymbol\Psi\in\mathbb R^{m\times m}$ share the same non-zero eigenvalues with the same algebraic multiplicities, and a singular value of a symmetric matrix is the abosulte value of the corresponding eigenvalue, we have
\begin{align}\label{lemma_Phi_fact_1_result_2}
\|\boldsymbol\Psi\boldsymbol\Psi^\top\|_{\text{F}}^2=\sum_{l=1}^n\sigma_{l}^2(\boldsymbol\Psi\boldsymbol\Psi^\top)
=\sum_{l=1}^n\lambda_{l}^2(\boldsymbol\Psi\boldsymbol\Psi^\top)
=\sum_{l'=1}^m\lambda_{l'}^2(\boldsymbol\Psi^\top\boldsymbol\Psi)
=\sum_{l'=1}^m\sigma_{l'}^2(\boldsymbol\Psi^\top\boldsymbol\Psi)
=\|\boldsymbol\Psi^\top\boldsymbol\Psi\|_{\text{F}}^2.
\end{align}
Remind that
$$\boldsymbol \Psi=(\boldsymbol \psi_1,\cdots,\boldsymbol \psi_m)
=\left( \begin{array}{c}
              \widetilde{\boldsymbol\psi_1} \\
              \vdots \\
              \widetilde{\boldsymbol\psi_n} 
        \end{array}
 \right)\in\mathbb R^{n\times m},$$
we can express
$$\|\boldsymbol\Psi^\top\boldsymbol\Psi\|_{\text{F}}^2
=\|\left( \begin{array}{c}
              \boldsymbol\psi_1^\top \\
              \vdots \\
              \boldsymbol\psi_m^\top 
        \end{array}
 \right)(\boldsymbol \psi_1,\cdots,\boldsymbol \psi_m)
\|_{\text{F}}^2
=\sum_{r=1}^m\|\boldsymbol\psi_r\|_2^4
+\sum_{1\leqslant r\not=s\leqslant m}\langle\boldsymbol\psi_r,\boldsymbol\psi_s\rangle^2,$$
$$\|\boldsymbol\Psi\boldsymbol\Psi^\top\|_{\text{F}}^2
=\|\left( \begin{array}{c}
              \widetilde{\boldsymbol\psi_1} \\
              \vdots \\
              \widetilde{\boldsymbol\psi_n} 
        \end{array}
 \right)(\widetilde{\boldsymbol \psi_1}^\top,\cdots,\widetilde{\boldsymbol \psi_n}^\top)
\|_{\text{F}}^2=
\sum_{i=1}^n\|\widetilde{\boldsymbol\psi_i}\|_2^4
+\sum_{1\leqslant i\not=j\leqslant n}\langle\widetilde{\boldsymbol\psi_i},\widetilde{\boldsymbol\psi_j}\rangle^2.$$
These taken collectively with (\ref{lemma_Phi_fact_1_result_2}) implies that
$$\sum_{1\leqslant r\not=s\leqslant m}\langle\boldsymbol\psi_r,\boldsymbol\psi_s\rangle^2
=\sum_{i=1}^n\|\widetilde{\boldsymbol\psi_i}\|_2^4
+\sum_{1\leqslant i\not=j\leqslant n}\langle\widetilde{\boldsymbol\psi_i},\widetilde{\boldsymbol\psi_j}\rangle^2-\sum_{r=1}^m\|\boldsymbol\psi_r\|_2^4.$$
Applying \textcolor{red}{power mean inequality}, we see that
$$\left[\frac{\sum_{r=1}^m\|\boldsymbol\psi_r\|_2^4}{m}\right]^{1/4}
\geqslant\left[\frac{\sum_{r=1}^m\|\boldsymbol\psi_r\|_2^2}{m}\right]^{1/2}=\left[\frac{\|\boldsymbol\Psi\|_{\text{F}}^2}{m}\right]^{1/2}\quad\text{gives rise to}\quad\sum_{r=1}^m\|\boldsymbol\psi_r\|_2^4\geqslant\frac{\|\boldsymbol\Psi\|_{\text{F}}^4}{m}.$$
Hence, we can upper bound
$$\sum_{1\leqslant r\not=s\leqslant m}\langle\boldsymbol\psi_r,\boldsymbol\psi_s\rangle^2
\leqslant\sum_{i=1}^n\|\widetilde{\boldsymbol\psi_i}\|_2^4
+\sum_{1\leqslant i\not=j\leqslant n}\langle\widetilde{\boldsymbol\psi_i},\widetilde{\boldsymbol\psi_j}\rangle^2-\frac{\|\boldsymbol\Psi\|_{\text{F}}^4}{m}.$$
Together with (\ref{lemma_Phi_fact_1_result_1}), we arrive at the desired estimate in Lemma \ref{lemma_Phi_fact_1}.
\subsection{Characterizing Activation Matrices: Initial Matrix And Its Perturbation}\label{subsection_activation_matrices}
Before proceeding to analyze the dynamics of the loss function, we prove at this stage a number of important results regarding the intial activation matrix and the change of activation matrices that will be useful to us later on. The main tools of our proof are {\textcolor{red}{concentration inequalities} and {\textcolor{red}{matrix Bernstein inequality} in high-dimensional probability. Firstly, we define the function $\wp(z)$ for concise expressions.
$$\wp(z)=\mathbb P(|\xi|\leqslant z)\quad\text{for}\quad \xi\sim\mathcal N(0,1),$$
Notice that $\wp(z)$ often stands for elliptic function introduced by Weierstrass in 1895 (\cite{whittaker_1902}), however, \textcolor{red}{$\wp(R)$ in this paper has no connection with elliptic function}.
Let $R$ is the maximal allowed movement of each weight vector $\boldsymbol w_r(r\in[m])$ during training, i.e., for each $r\in[m]$, $\|\boldsymbol w_r(k)-\boldsymbol w_r(0)\|_2\leqslant R$ at $k$-th iteration in GD. By anti-concentration inequality of Gaussian, it is evident that 
\begin{align}\label{wp_R_definition}
\wp(R)=\int_{-R}^R\frac{1}{\sqrt{2\pi}}e^{-x^2/2}dx<\frac{2R}{\sqrt{2\pi}}<R,
\end{align}
In practice, recent empirical results (\cite{jacot_2018}, \cite{liu_2020}) showed that $|(\boldsymbol w_r(k))_i-(\boldsymbol w_r(0))_i|=\mathcal O(1/\sqrt{m})$, meaning that $R$ is a small number in (0,1/2), thus
\begin{align}\nonumber
\wp(R)&=\int_{-R}^R\frac{1}{\sqrt{2\pi}}e^{-x^2/2}dx>\frac{2}{\sqrt{2\pi}}\int_0^R\left(1-\frac{x^2}{2}\right)dx=\sqrt{\frac{2}{\pi}}\left(R-\frac{R^3}{3}\right)\\\nonumber
&=\sqrt{\frac{2}{\pi}}R\left(1-\frac{R^2}{3}\right)>\frac{11}{12}\sqrt{\frac{2}{\pi}}R>0.731R>\frac{7}{10}R.
\end{align}
Therefore, we have 
\begin{align}\label{wp_R_bound}
0<\frac{7}{10}R<\wp(R)<R<\frac{1}{2}.
\end{align}
\begin{lemma}\label{lemma_Psi0}
(\textcolor{blue}{Properties of Initial Activation Matrix}) For any $\delta\in(0,1)$, with probability exceeding $1-\delta$ over the random initialization, the following holds: 
$$\|\widetilde{\boldsymbol\psi_i}(0)\|_2^2\leqslant\frac{m}{2}+\sqrt{\frac{m}{2}\log\frac{1}{\delta}},\quad\forall i\in[n].$$
With probability exceeding $1-\delta$ over the random initialization, the following holds: 
$$\|\widetilde{\boldsymbol\psi_i}(0)\|_2^2\geqslant\frac{m}{2}-\sqrt{\frac{m}{2}\log\frac{1}{\delta}},\quad\forall i\in[n].$$
With probability exceeding $1-n\delta$ over the random initialization, the following holds:  
    \begin{align}\nonumber
      \|\boldsymbol\Psi(0)\|_{\text{F}}^2&\leqslant\frac{mn}{2}+n\sqrt{\frac{m}{2}\log\frac{1}{\delta}}.
\end{align}      
With probability exceeding $1-n\delta$ over the random initialization, the following holds:  
    \begin{align}\nonumber
      \|\boldsymbol\Psi(0)\|_{\text{F}}^2\geqslant\frac{mn}{2}-n\sqrt{\frac{m}{2}\log\frac{1}{\delta}}.
\end{align}      
With probability exceeding $1-2\delta$ over the random initialization, the following holds:
\begin{align}\nonumber
      \langle\widetilde{\boldsymbol\psi_i}(0),\widetilde{\boldsymbol\psi_j}(0)\rangle&\leqslant\frac{m}{2}+\sqrt{\frac{m}{2}\log\frac{1}{\delta}},\quad(i\neq j).
\end{align}
If $n\leqslant m$, i.e., in the over-parameterized setting, then with probability exceeding $1-(mn-m+n+2)\delta$ over the random initialization,
\begin{align}\nonumber
\sigma_{\min}^2(\boldsymbol \Psi_0)>\lambda^*m-n\sqrt{\left[2-\frac{\theta^*(\theta^*+3\pi)}{2\pi^2}\right]m\log(2n/\delta)}+\mathcal O\left(\sqrt{m\log(1/\delta)}+n\log(2n/\delta)/\sqrt{\delta}\right),
\end{align}
where $\lambda^*=\lambda_{\min}(\boldsymbol\mho)$ defined in (\ref{matrix_mho_definition}) in Section \ref{section_problem} and $\theta^*$ is defined in (\ref{theta_star_definition}) in Section \ref{section_problem}. 
\end{lemma}
{\bf Remark}: The lower bound of $\sigma_{\min}^2(\boldsymbol \Psi_0)$ implies the crucial role of the number of neurons and the geometric parameters characterizing the data set upon activation matrix. Sepecially, the lower bound of $\sigma_{\min}^2(\boldsymbol \Psi_0)$ can be expected to be larger for the larger $m,\lambda^*$ and $\theta^*$, implying better convergence abilities, as demonstrated in the sequel, and leading to our data-dependent theoretical guarantees. Other bounds reveal the concentration of measure phenomena in high dimensional space.\\ 
\textbf{Proof}: For each $i\in[n]$, $\|\widetilde{\boldsymbol\psi_i}(0)\|_2^2=\sum_{r=1}^m\mathbb I_{r,i}(0)$ is a sum of $m$ i.i.d. random variables bounded in $[0,1]$ with expectation $m/2$. Thus applying \textcolor{red}{Hoeffding inequality}, we have for any $t>0$,
\begin{align}\nonumber
\mathbb P\left[\|\widetilde{\boldsymbol\psi_i}(0)\|_2^2-\frac{m}{2}\geqslant t\right]\leqslant\exp\left(-\frac{2t^2}{m}\right).
\end{align}
Setting $\delta=\exp(-2t^2/m)$, we reach that with probability exceeding $1-\delta$,
\begin{align}\nonumber
\|\widetilde{\boldsymbol\psi_i}(0)\|_2^2\leqslant\frac{m}{2}+\sqrt{\frac{m}{2}\log\frac{1}{\delta}}.
\end{align}
By a union bound over all $i\in[n]$, we have that with probability exceeding $1-n\delta$,  
\begin{align}\nonumber
   \|\boldsymbol\Psi(0)\|_{\text{F}}^2=\sum_{i=1}^n\|\widetilde{\boldsymbol\psi_i}(0)\|_2^2\leqslant\frac{mn}{2}+n\sqrt{\frac{m}{2}\log\frac{1}{\delta}}.
\end{align}
In a similar manner, we see that with probability exceeding $1-\delta$,
$$\|\widetilde{\boldsymbol\psi_i}(0)\|_2^2\geqslant\frac{m}{2}-\sqrt{\frac{m}{2}\log\frac{1}{\delta}}.$$
Again, take a union bound over all $i\in[n]$, we have that with probability exceeding $1-n\delta$,  
\begin{align}\nonumber
   \|\boldsymbol\Psi(0)\|_{\text{F}}^2=\sum_{i=1}^n\|\widetilde{\boldsymbol\psi_i}(0)\|_2^2\geqslant\frac{mn}{2}-n\sqrt{\frac{m}{2}\log\frac{1}{\delta}}.
\end{align}
For $i\not=j\in[n]$, take a union bound over $(i,j)$ gives that with probability exceeding $1-2\delta$,
\begin{align}\nonumber
\langle\widetilde{\boldsymbol\psi_i}(0),\widetilde{\boldsymbol\psi_j}(0)\rangle\leqslant\|\widetilde{\boldsymbol\psi_i}(0)\|_2
\|\widetilde{\boldsymbol\psi_j}(0)\|_2\leqslant\frac{\|\widetilde{\boldsymbol\psi_i}(0)\|_2^2+\|\widetilde{\boldsymbol\psi_j}(0)\|_2^2}{2}\leqslant\frac{m}{2}+\sqrt{\frac{m}{2}\log\frac{1}{\delta}}.
\end{align}
To obtain a sharp lower bound on $\sigma_{\min}^2(\boldsymbol \Psi_0)$, we need much more delicate mathematical analysis, thus we leave it to the Section \ref{subsection_min_sigma_Psi0}.
\begin{lemma}\label{lemma_Psik_Psi0}(\textcolor{blue}{The Perturbation of Activation Matrix}) For any weight vector $\boldsymbol w_r\in\mathbb R^d$ $(r\in[m])$ satisfies $\|\boldsymbol w_r(k)-\boldsymbol w_r(0)\|_2\leqslant R$ for any $k\in\mathbb Z^+$, we have that for any $\delta\in(0,1)$, with probability exceeding $1-\delta$ over the random initialization, the following holds: 
$$\|\widetilde{\boldsymbol\psi_i}(k)-\widetilde{\boldsymbol\psi_i}(0)\|_2^2\leqslant\wp(R)m+\frac{2}{3}\log\frac{1}{\delta}+\sqrt{2\wp(R)[1-\wp(R)]m\log\frac{1}{\delta}},\quad\forall i\in[n].$$
With probability exceeding $1-n\delta$ over the random initialization, the following holds:  
    \begin{align}\nonumber
      \|\boldsymbol\Psi(k)-\boldsymbol\Psi(0)\|_{\text{F}}^2\leqslant\wp(R)mn+\frac{2}{3}n\log\frac{1}{\delta}+n\sqrt{2\wp(R)[1-\wp(R)]m\log\frac{1}{\delta}}.
    \end{align}
With probability exceeding $1-n\delta$ over the random initialization, the following holds:  
    \begin{align}\nonumber
      \|\boldsymbol\Psi(k)-\boldsymbol\Psi(0)\|_{\text{F}}^2\geqslant\wp(R)mn-\frac{2}{3}n\log\frac{1}{\delta}-n\sqrt{2\wp(R)[1-\wp(R)]m\log\frac{1}{\delta}}.
    \end{align}
    With probability exceeding $1-(3n-2)\delta$ over the random initialization, the following holds:
    \begin{align}\nonumber
    &\|\boldsymbol\Psi(k)-\boldsymbol\Psi(0)\|_2^2<\frac{2\csc\theta^*}{\pi}R^2mn+Rm+\frac{2n}{3}\log(1/\delta)+\mathcal O\left((\sqrt{R}n+1)\sqrt{Rm\log(1/\delta)}+R^4mn\right).
    \end{align}
Here $\wp(R)\in(0,1)$ is defined in (\ref{wp_R_definition}) in Section \ref{subsection_activation_matrices}. $\lambda^*=\lambda_{\min}(\boldsymbol\mho)$ defined in (\ref{matrix_mho_definition}) in Section \ref{section_problem} and $\theta^*$ is defined in (\ref{theta_star_definition}) in Section \ref{section_problem}.
\end{lemma}
{\bf Remark}: Recall that $\boldsymbol\Psi(k)=\phi'(\boldsymbol X^\top\boldsymbol W(k))$, then the upper bounds of $\|\widetilde{\boldsymbol\psi_i}(k)-\widetilde{\boldsymbol\psi_i}(0)\|_2$ and $\|\boldsymbol\Psi(k)-\boldsymbol\Psi(0)\|_2$ implies that activation matrix $\boldsymbol\Psi(k)$ is stable with respect to weight matrix $\boldsymbol W(k)$. More explicitly, if $\boldsymbol w$ stays in a $R$-ball during training, then we are able to upper bound the spectral norm of $\|\boldsymbol\Psi(k)-\boldsymbol\Psi(0)\|_2$. Other bounds reveal the concentration of measure phenomena in high dimensional space.\\
\textbf{Proof}: We define the event
$$A_{i,r}=\{\exists\boldsymbol w\in\mathcal B(\boldsymbol w_r(0),R):\mathbb I\{\left<\boldsymbol w,\boldsymbol x_i\right>\geqslant 0\}\not=\mathbb I\{\left<\boldsymbol w_r(0),\boldsymbol x_i\right>\geqslant 0\}\},\quad r\in[m],i\in[n].$$
In other words, for sample $\boldsymbol x_i$, the event $A_{i,r}$ denotes whether the activation pattern of neuron $r$, viz. activation pattern $\mathbb I\{\left<\boldsymbol w_r,\boldsymbol x_i\right>\}$ can change through training if the weight vector change is bounded in the $R$-ball  $\mathcal B(\boldsymbol w_r(0),R)=\{\boldsymbol w\in\mathbb R^d:\|\boldsymbol w-\boldsymbol w_r(0)\|_2\leqslant R\}$ centered at initialization point $\boldsymbol w_r(0)$. Hence, the event $A_{i,r}$ happens if and only if $|\boldsymbol w_r(0)^\top\boldsymbol x_i|\leqslant R$. For each $i\in[n]$, we devide all $m$ neurons into two disjoint sets:
$$S_i=\{r\in[m]:\mathbb I\{A_{i,r}\}=0\}\quad\text{and}\quad
\bar{S}_i=\{r\in[m]:\mathbb I\{A_{i,r}\}=1\}.$$
It is easy to see that 
$$\mathbb I\{\mathbb I_{r,i}(k)\not=\mathbb I_{r,i}(0)\}\leqslant\mathbb I\{A_{i,r}\}+\mathbb I\{\|\boldsymbol w_r(k)-\boldsymbol w_r(0)\|_2\leqslant R\},\,\forall i\in[n],r\in[m],\forall k\geqslant0.$$
Hence $\|\boldsymbol w_r(k)-\boldsymbol w_r(0)\|_2\leqslant R$ for any $k\in\mathbb Z^{+}$, then $r\in S_i$ means that $\mathbb I_{r,i}(k)=\mathbb I_{r,i}(0)$ for any $k\in\mathbb Z^{+}$. In other words, all neurons in $S_i$ will not change activation pattern on data-point $\boldsymbol x_i$ during gradient descent training. Therefore,
\begin{align}\nonumber
\|\widetilde{\boldsymbol\psi_i}(k)-\widetilde{\boldsymbol\psi_i}(0)\|_2^2\leqslant|\bar{S}_i|=\sum_{r=1}^m\mathbb I_{r\in\bar{S}_i}=\sum_{r=1}^m\mathbb E\mathbb I_{r\in\bar{S}_i}+\sum_{r=1}^m\left(\mathbb I_{r\in\bar{S}_i}-\mathbb E\mathbb I_{r\in\bar{S}_i}\right).
\end{align}
$$\mathbb E\mathbb I_{r\in\bar{S}_i}=\mathbb P(\mathbb I_{r\in\bar{S}_i}=1)=\mathbb P(\mathbb I\{A_{i,r}\}=1)=\mathbb P(A_{i,r})=\wp(R).$$
Since $\sum_{r=1}^m\left(\mathbb I_{r\in\bar{S}_i}-\mathbb E\mathbb I_{r\in\bar{S}_i}\right)$ is a sum of i.i.d. random variables with very small variances, we can obtain a tight upper bound of $\sum_{r=1}^m\left(\mathbb I_{r\in\bar{S}_i}-\mathbb E\mathbb I_{r\in\bar{S}_i}\right)$ via {\textcolor{red}{Bernstein inequality}.
Let
\begin{align}\nonumber
B=&\max_{1\leqslant r\leqslant m}|\mathbb I_{r\in\bar{S}_i}-\mathbb E\mathbb I_{r\in\bar{S}_i}|\leqslant1,\\\nonumber
V=&\sum_{r=1}^m\mathbb E[(\mathbb I_{r\in\bar{S}_i}-\mathbb E\mathbb I_{r\in\bar{S}_i})^2]=\sum_{r=1}^m\left\{\mathbb E[\mathbb I_{r\in\bar{S}_i}^2]
-[\mathbb E\mathbb I_{r\in\bar{S}_i}]^2\right\}\\\nonumber
=&\sum_{r=1}^m\left\{\mathbb E[\mathbb I_{r\in\bar{S}_i}]
-[\mathbb E\mathbb I_{r\in\bar{S}_i}]^2\right\}=\wp(R)[1-\wp(R)]m.
\end{align}
Applying {\textcolor{red}{Bernstein inequality}, we have for any $t>0$,
$$\mathbb P\left[\sum_{r=1}^m\left(\mathbb I_{r\in\bar{S}_i}-\mathbb E\mathbb I_{r\in\bar{S}_i}\right)\geqslant t\right]\leqslant\exp\left(-\frac{t^2/2}{V+Bt/3}\right).$$
Setting $\delta=\exp[-(t^2/2)/(V+Bt/3)]$, we have
$$\mathbb P\left[\sum_{r=1}^m\left(\mathbb I_{r\in\bar{S}_i}-\mathbb E\mathbb I_{r\in\bar{S}_i}\right)\geqslant \frac{B}{3}\log\frac{1}{\delta}\pm\sqrt{\frac{B^2}{9}\log^2\frac{1}{\delta}+2V\log\frac{1}{\delta}}\right]\leqslant\delta,$$
Since
$$\frac{B}{3}\log\frac{1}{\delta}\pm\sqrt{\frac{B^2}{9}\log^2\frac{1}{\delta}+2V\log\frac{1}{\delta}}\leqslant\frac{2B}{3}\log\frac{1}{\delta}+\sqrt{2V\log\frac{1}{\delta}},$$
we have
$$\mathbb P\left[\sum_{r=1}^m\left(\mathbb I_{r\in\bar{S}_i}-\mathbb E\mathbb I_{r\in\bar{S}_i}\right)\geqslant\frac{2B}{3}\log\frac{1}{\delta}+\sqrt{2V\log\frac{1}{\delta}}\right]\leqslant\delta,$$
Thus, with probability exceeding $1-\delta$,
\begin{align}\nonumber
&\sum_{r=1}^m\left(\mathbb I_{r\in\bar{S}_i}-\mathbb E\mathbb I_{r\in\bar{S}_i}\right)
\leqslant\frac{2B}{3}\log\frac{1}{\delta}+\sqrt{2V\log\frac{1}{\delta}}\leqslant\frac{2}{3}\log\frac{1}{\delta}+\sqrt{2\wp(R)[1-\wp(R)]m\log\frac{1}{\delta}}.
\end{align}
Henceforth, we arrive at with probability exceeding $1-\delta$ over the random initialization,
$$\|\widetilde{\boldsymbol\psi_i}(k)-\widetilde{\boldsymbol\psi_i}(0)\|_2^2\leqslant \wp(R)m+\frac{2}{3}\log\frac{1}{\delta}+\sqrt{2\wp(R)[1-\wp(R)]m\log\frac{1}{\delta}}.$$
By a union bound over all $i\in[n]$, we have that with probability exceeding $1-n\delta$,  
\begin{align}\nonumber
    \|\boldsymbol\Psi(k)-\boldsymbol\Psi(0)\|_{\text{F}}^2=\sum_{i=1}^n\|\widetilde{\boldsymbol\psi_i}(k)-\widetilde{\boldsymbol\psi_i}(0)\|_2^2\leqslant\wp(R)mn+\frac{2}{3}n\log\frac{1}{\delta}+n\sqrt{2\wp(R)[1-\wp(R)]m\log\frac{1}{\delta}}.
\end{align}
In a similar manner, we know that with probability exceeding $1-\delta$, we obtain
$$\|\widetilde{\boldsymbol\psi_i}(k)-\widetilde{\boldsymbol\psi_i}(0)\|_2^2\geqslant \wp(R)m-\frac{2}{3}\log\frac{1}{\delta}-\sqrt{2\wp(R)[1-\wp(R)]m\log\frac{1}{\delta}}.$$
Again, take a union bound over all $i\in[n]$, we have that with probability exceeding $1-n\delta$,  
\begin{align}\nonumber
    \|\boldsymbol\Psi(k)-\boldsymbol\Psi(0)\|_{\text{F}}^2=\sum_{i=1}^n\|\widetilde{\boldsymbol\psi_i}(k)-\widetilde{\boldsymbol\psi_i}(0)\|_2^2\geqslant\wp(R)mn-\frac{2}{3}n\log\frac{1}{\delta}-n\sqrt{2\wp(R)[1-\wp(R)]m\log\frac{1}{\delta}}.
\end{align}
To obtain a sharp upper bound on $\|\boldsymbol\Psi(k)-\boldsymbol\Psi(0)\|_2^2$, we need much more delicate mathematical analysis, thus we leave it to the Section \ref{subsection_spectral_norm_variation_Psi}.
\\This finishes the proof of Lemma \ref{lemma_Psik_Psi0}.
\subsection{Lower Bounding the Smallest Singular Value of Initial Activation Matrix}\label{subsection_min_sigma_Psi0}
{\textcolor{blue}{{\bf Step 1: Decomposing the Initial Activation Matrix}}\\
Regarding $\sigma_{\min}^2(\boldsymbol \Psi(0))$,
it can be seen from (\ref{Psi_matrix}) that
\begin{align}\nonumber
&\sigma_{\min}^2(\boldsymbol \Psi(0))\\\nonumber
\stackrel{\text{(i)}}{=}&\lambda_{\min}(\boldsymbol\Psi(0)\boldsymbol\Psi(0)^\top)\\\nonumber
=&\lambda_{\min}\left(\left( \begin{array}{c}
              \widetilde{\boldsymbol\psi_1}(0) \\
              \vdots \\
              \widetilde{\boldsymbol\psi_n}(0) 
        \end{array}
 \right)(\widetilde{\boldsymbol \psi_1}(0)^\top,\cdots,\widetilde{\boldsymbol \psi_n}(0)^\top)\right)\\\nonumber
=&\lambda_{\min}\left(\left(  \begin{array}{ccc}
  \|\widetilde{\boldsymbol\psi_1}(0)\|_2^2  & \cdots &      \langle\widetilde{\boldsymbol\psi_1}(0),\widetilde{\boldsymbol\psi_n}(0)\rangle   \\
   \vdots & \ddots & \vdots  \\
   \langle\widetilde{\boldsymbol\psi_n}(0),\widetilde{\boldsymbol\psi_1}(0)\rangle  & \cdots &     \|\widetilde{\boldsymbol\psi_n}(0)\|_2^2    
                         \end{array}
              \right)\right)\\\nonumber
              =&\lambda_{\min}\left(\left(  \begin{array}{ccc}
  \|\widetilde{\boldsymbol\psi_1}(0)\|_2^2 & \cdots &    0  \\
   \vdots & \ddots & \vdots  \\
   0 & \cdots &     \|\widetilde{\boldsymbol\psi_n}(0)\|_2^2 
                         \end{array} \right)+\left(  \begin{array}{ccc}
  0  & \cdots &      \langle\widetilde{\boldsymbol\psi_1}(0),\widetilde{\boldsymbol\psi_n}(0)\rangle   \\
   \vdots & \ddots & \vdots  \\
   \langle\widetilde{\boldsymbol\psi_n}(0),\widetilde{\boldsymbol\psi_1}(0)\rangle  & \cdots &     0   
                         \end{array} \right)\right)\\\nonumber
=&\lambda_{\min}(\text{diag}(\|\widetilde{\boldsymbol\psi_1}(0)\|_2^2,\cdots,\|\widetilde{\boldsymbol\psi_n}(0)\|_2^2)+\boldsymbol\Upsilon)\\\nonumber
\stackrel{(\text{ii})}{\geqslant}&\lambda_{\min}(\text{diag}(\|\widetilde{\boldsymbol\psi_1}(0)\|_2^2,\cdots,\|\widetilde{\boldsymbol\psi_n}(0)\|_2^2))+\lambda_{\min}(\boldsymbol\Upsilon)\\\label{sigma_min_Phi0_lower_bound}
=&\min_{1\leqslant i\leqslant n}\|\widetilde{\boldsymbol\psi_i}(0)\|_2^2+\lambda_{\min}(\boldsymbol\Upsilon),
\end{align}
Here (i) is due to that $\boldsymbol\Psi(0)\in\mathbb R^{n\times m}$ and $n\leqslant m$ in the over-parameterized setting, leading to $\sigma_{\min}^2(\boldsymbol\Psi(0))=\lambda_{\min}(\boldsymbol\Psi(0)\boldsymbol\Psi(0)^\top)$. The $(i,j)$ off-diagonal entry of symmetric matrix $\boldsymbol\Upsilon$ is $\langle\widetilde{\boldsymbol\psi_i}(0),\widetilde{\boldsymbol\psi_j}(0)\rangle^2$, whereas all the diagonal entries of symmetric matrix $\boldsymbol\Upsilon$ are zero. And (ii) comes from the \textcolor{red}{trace inequality}: $\lambda_{\min}(\boldsymbol A+\boldsymbol B)\geqslant\lambda_{\min}(\boldsymbol A)+\lambda_{\min}(\boldsymbol B)$ for Hermitian matrices $\boldsymbol A$ and $\boldsymbol B$, which can be easily proved via \textcolor{red}{Courant-Fischer theorem} on the variational characterization of eigenvalues of Hermitian matrices.\\
For the first term in (\ref{sigma_min_Phi0_lower_bound}), invoking (\ref{lemma_Psi0}), followed a union bound over $i\in[n]$, we see that with probability exceeding $1-n\delta$,
\begin{align}\label{Phi_i0_min_lower_bound}
\min_{1\leqslant i\leqslant n}\|\widetilde{\boldsymbol\psi_i}(0)\|_2^2\geqslant \frac{m}{2}-\sqrt{\frac{m}{2}\log\frac{1}{\delta}}.
\end{align}
We are in need of a high probability lower bound on $\lambda_{\min}(\boldsymbol\Upsilon)$. Due to the highly dependence of entries of $\boldsymbol\Upsilon$, it is very challenging to obtain a tight lower bound of $\lambda_{\min}(\boldsymbol\Upsilon)$.\\
{\textcolor{blue}{{\bf Step 2: Decomposing the Matrix $\boldsymbol\Upsilon$ as Expectation and Deviation Parts}}\\
We can decompose $\boldsymbol\Upsilon$ into its expectation matrix $\boldsymbol\Upsilon^*$ and deviation matrix $\Delta_{\boldsymbol\Upsilon}$.
\begin{align}\nonumber
\boldsymbol\Upsilon&=\left(  \begin{array}{ccc}
  0  & \cdots &      \langle\widetilde{\boldsymbol\psi_1}(0),\widetilde{\boldsymbol\psi_n}(0)\rangle   \\
   \vdots & \ddots & \vdots  \\
   \langle\widetilde{\boldsymbol\psi_n}(0),\widetilde{\boldsymbol\psi_1}(0)\rangle  & \cdots &     0   
                         \end{array} \right)\\\nonumber
                        &=\left(  \begin{array}{ccc}
  0  & \cdots &     \mathbb E( \langle\widetilde{\boldsymbol\psi_1}(0),\widetilde{\boldsymbol\psi_n}(0)\rangle)   \\
   \vdots & \ddots & \vdots  \\
 \mathbb E(  \langle\widetilde{\boldsymbol\psi_n}(0),\widetilde{\boldsymbol\psi_1}(0)\rangle)  & \cdots &   0   
                         \end{array}
              \right)\\\nonumber
              &+\left(  \begin{array}{ccc}
  0  & \cdots &      \langle\widetilde{\boldsymbol\psi_1}(0),\widetilde{\boldsymbol\psi_n}(0)\rangle-\mathbb E( \langle\widetilde{\boldsymbol\psi_1}(0),\widetilde{\boldsymbol\psi_n}(0)\rangle)    \\
   \vdots & \ddots & \vdots  \\
   \langle\widetilde{\boldsymbol\psi_n}(0),\widetilde{\boldsymbol\psi_1}(0)\rangle-\mathbb E( \langle\widetilde{\boldsymbol\psi_n}(0),\widetilde{\boldsymbol\psi_1}(0)\rangle)   & \cdots &   0   
                         \end{array}
              \right)\\\nonumber
              &=\boldsymbol\Upsilon^*+\Delta_{\boldsymbol\Upsilon}.
\end{align}
Here the $(i,j)$ entry of matrix $\boldsymbol\Upsilon^*$ is $\mathbb E( \langle\widetilde{\boldsymbol\psi_i}(0),\widetilde{\boldsymbol\psi_j}(0)\rangle)$, where the $(i,j)$ entry of matrix $\Delta_{\boldsymbol\Upsilon}$ is $ \langle\widetilde{\boldsymbol\psi_i}(0),\widetilde{\boldsymbol\psi_j}(0)\rangle-\mathbb E( \langle\widetilde{\boldsymbol\psi_i}(0),\widetilde{\boldsymbol\psi_j}(0)\rangle)$.\\\\
{\textcolor{blue}{{\bf Step 3: Estimating the Smallest Eigenvalue of Matrix $\boldsymbol\Upsilon^*$}}\\
It is evident that for $1\leqslant i,j\leqslant n$,
\begin{align}\label{Upsilon_star_ij_entry}
&\mathbb E( \langle\widetilde{\boldsymbol\psi_i}(0),\widetilde{\boldsymbol\psi_j}(0)\rangle)=\mathbb E\left(\sum_{r=1}^m\mathbb I_{r,i}(0)\mathbb I_{r,j}(0)\right)=\sum_{r=1}^m\mathbb E \left\{\mathbb I_{r,i}(0)\mathbb I_{r,j}(0)\right\}.
\end{align}
To proceed, we establish the following two claims concerning expectations.\\
\textcolor{blue}{\textbf{Claim 1}}:
\begin{align}\label{E_ri0_rj0}
\forall i,j\in[n],r\in[m], \mathbb E \left\{\mathbb I_{r,i}(0)\mathbb I_{r,j}(0)\right\}=\frac{\pi-\arccos\langle\boldsymbol x_i, \boldsymbol x_j\rangle}{2\pi}=\frac{\pi-\theta_{ij}}{2\pi}\in\left(0,\frac{1}{2}\right],\\\nonumber
\text{where }\theta_{ij}=\angle(\boldsymbol x_i, \boldsymbol x_j)=\arccos\langle\boldsymbol x_i, \boldsymbol x_j\rangle\in[0,\pi).
\end{align}
\textcolor{blue}{\textbf{Proof of Claim 1}}: With the spherical coordinate form of $\boldsymbol w_r(0)=(w_{r,1}(0),\cdots,w_{r,d}(0)):$
\begin{align}\nonumber
\left\{ \begin{array}{lll}
  w_{r,1}(0) &= \rho\cos\theta_1\\
  w_{r,2}(0) &= \rho\sin\theta_1\cos\theta_2 \\
  w_{r,3}(0) &= \rho\sin\theta_1\sin\theta_2\cos\theta_3 \\
  &\cdots\cdots\cdots\cdots\cdots\cdots \\
  w_{r,d-1}(0) &= \rho\sin\theta_1\sin\theta_2\sin\theta_3\cdots\cos\theta_{d-1} \\
  w_{r,d}(0) &= \rho\sin\theta_1\sin\theta_2\sin\theta_3\cdots\sin\theta_{d-1}
        \end{array}
\right.
\end{align}
where $\rho\geqslant0,0\leqslant\theta_i\leqslant\pi(i=1,\cdots,d-2)$ and $0\leqslant\theta_{d-1}\leqslant2\pi$.\\
We have
\begin{align}\nonumber
&\mathbb E \left\{\mathbb I_{r,i}(0)\mathbb I_{r,j}(0)\right\}\\\nonumber
=&\mathbb P\{\mathbb I_{r,i}(0)=1\text{ and }\mathbb I_{r,j}(0)=1\}\\\nonumber
=&\mathbb P\big\{\mathbb{I}\{\boldsymbol w_r(0)^\top \boldsymbol x_i\geqslant 0\}=1\text{ and }\mathbb{I}\{\boldsymbol w_r(0)^\top \boldsymbol x_j\geqslant 0\}=1\big\} \\\nonumber  
=&\mathbb P\big\{\boldsymbol w_r(0)^\top \boldsymbol x_i\geqslant 0\text{ and }\boldsymbol w_r(0)^\top \boldsymbol x_j\geqslant 0\big\}\\\nonumber
=&\idotsint\limits_{\begin{subarray}{c}-\infty\\\boldsymbol w_r(0)\top\boldsymbol x_i\geqslant0\\\boldsymbol w_r(0)\top\boldsymbol x_j\geqslant0\end{subarray}}^{\begin{subarray}{c}+\infty\end{subarray}}\frac{1}{(\sqrt{2\pi})^n}\exp\left\{-\frac{1}{2}\|\boldsymbol w_r(0)\|_2^2\right\}dw_{r,1}(0)dw_{r,2}(0)\cdots dw_{r,d}(0)\\\nonumber
=&\frac{1}{(\sqrt{2\pi})^n}\int_0^{+\infty}\exp\left\{-\frac{1}{2}\rho^2 \right\}\rho d\rho\int_{\theta_{ij}}^{\pi}(2\pi)^{\frac{1}{2}(n-2)}d\theta
\\\nonumber
=&\frac{\pi-\arccos\langle\boldsymbol x_i, \boldsymbol x_j\rangle}{2\pi}=\frac{\pi-\theta_{ij}}{2\pi}\in\left(0,\frac{1}{2}\right].
\end{align}
Note that $\mathbb E \left\{\mathbb I_{r,i}(0)\mathbb I_{r,j}(0)\right\}=1/2$ if and only if $i=j$. Claim 1 is thus proved.\\
\textcolor{blue}{\textbf{Claim 2}}:
\begin{align}\label{E_ri0_rell0_rj0}
\text{For }i,\ell,j\in[n], \text{ we have }\mathbb E \left\{\mathbb I_{r,i}(0)\mathbb I_{r,\ell}(0)\mathbb I_{r,j}(0)\right\}=\frac{2\pi-(\theta_{i\ell}+\theta_{\ell j}+\theta_{ji})}{4\pi}\in\left[0,\frac{1}{2}\right],\\\nonumber
\text{where }\theta_{i\ell}=\angle(\boldsymbol x_i, \boldsymbol x_\ell),
\theta_{\ell j}=\angle(\boldsymbol x_\ell, \boldsymbol x_j),\theta_{ji}=\angle(\boldsymbol x_j, \boldsymbol x_i)\in[0,\pi).
\end{align}
\textcolor{blue}{\textbf{Proof of Claim 2}}: 
For $\boldsymbol w(0)=(w_1(0),w_2(0),\cdots,w_d(0))\in\mathbb R^d$ and $i,\ell,j\in[n]$, we have
\begin{align}\nonumber
&\mathbb E\left\{\mathbb I_{r,i}(0)\mathbb I_{r,\ell}(0)\mathbb I_{r,j}(0)\right\}\\\nonumber
=&\mathbb P\{\mathbb I_{r,i}(0)=1\text{ and }\mathbb I_{r,\ell}(0)=1\text{ and }\mathbb I_{r,j}(0)=1\}\\\nonumber
=&\mathbb P\big\{\mathbb{I}\{\boldsymbol w_r(0)^\top \boldsymbol x_i\geqslant 0\}=1\text{ and }\mathbb{I}\{\boldsymbol w_r(0)^\top \boldsymbol x_\ell\geqslant 0\}=1\text{ and }\mathbb{I}\{\boldsymbol w_r(0)^\top \boldsymbol x_j\geqslant 0\}=1\big\} \\\nonumber  
=&\mathbb P\big\{\boldsymbol w_r(0)^\top \boldsymbol x_i\geqslant 0\text{ and }\boldsymbol w_r(0)^\top \boldsymbol x_\ell\geqslant 0\text{ and }\boldsymbol w_r(0)^\top \boldsymbol x_j\geqslant 0\big\}\\\nonumber
=&\idotsint\limits_{\begin{subarray}{c}-\infty\\\boldsymbol w_r(0)\top\boldsymbol x_i\geqslant0\\\boldsymbol w_r(0)\top\boldsymbol x_\ell\geqslant0\\\boldsymbol w_r(0)\top\boldsymbol x_j\geqslant0\end{subarray}}^{\begin{subarray}{c}+\infty\end{subarray}}\frac{1}{(\sqrt{2\pi})^n}\exp\left\{-\frac{1}{2}\|\boldsymbol w_r(0)\|_2^2\right\}dw_{r,1}(0)dw_{r,2}(0)\cdots dw_{r,d}(0)\\\nonumber
\stackrel{\text{(i)}}=&\frac{2\pi-\arccos\langle\boldsymbol x_i, \boldsymbol x_\ell\rangle-\arccos\langle\boldsymbol x_\ell, \boldsymbol x_j\rangle-\arccos\langle\boldsymbol x_j, \boldsymbol x_i\rangle}{4\pi}=\frac{2\pi-(\theta_{i\ell}+\theta_{\ell j}+\theta_{ji})}{4\pi}\in\left[0,\frac{1}{2}\right].
\end{align}
Here (i) comes from \cite{hua_1984}. Similarity, $\mathbb E \left\{\mathbb I_{r,i}(0)\mathbb I_{r,\ell}(0)\mathbb I_{r,j}(0)\right\}=1/2$ if and only if $i=\ell=j$. Claim 2 is thus proved.\\
An immediate consequence of Claim 2 is $\theta_{i\ell}+\theta_{\ell j}+\theta_{ji}\leqslant2\pi$ from $\mathbb E\left\{\mathbb I_{r,i}(0)\mathbb I_{r,\ell}(0)\mathbb I_{r,j}(0)\right\}\geqslant0$, giving the sharper upper bound of $\theta_{\text{min}}$: for distinct $i,j,k$, 
\begin{align}\label{theta_star_upper_bound}
\theta_{\text{min}}\leqslant\min\{\theta_{i\ell},\theta_{\ell j},\theta_{ji}\}\leqslant\frac{\theta_{i\ell}+\theta_{\ell j}+\theta_{ji}}{3}\leqslant\frac{2\pi}{3}.
\end{align}
With Claim 1 in place, combining with (\ref{Upsilon_star_ij_entry}) and (\ref{E_ri0_rj0}), we obtain
\begin{align}\nonumber
&\mathbb E( \langle\widetilde{\boldsymbol\psi_i}(0),\widetilde{\boldsymbol\psi_j}(0)\rangle)
=\sum_{r=1}^m\frac{\pi-\theta_{ij}}{2\pi}
=\frac{m(\pi-\theta_{ij})}{2\pi}.
\end{align}
Recall the definition of $\boldsymbol\mho$ defined in Section \ref{section_problem}: for any $r\in[m]$,
\begin{align}\nonumber
\boldsymbol\mho=&\mathbb E_{\boldsymbol w\sim{\cal N}(0,\boldsymbol I)}[\phi'(\boldsymbol X^\top\boldsymbol w)\phi'(\boldsymbol w^\top\boldsymbol X)]
=\left(  \begin{array}{ccc}
   \mathbb E \left\{\mathbb I_{r,i}(0)\right\} & \cdots &     \mathbb E \left\{\mathbb I_{r,i}(0)\mathbb I_{r,j}(0)\right\}  \\
   \vdots & \ddots & \vdots  \\
  \mathbb E \left\{\mathbb I_{r,i}(0)\mathbb I_{r,j}(0)\right\}  & \cdots &   \mathbb E \left\{\mathbb I_{r,i}(0)\right\} 
                         \end{array}
              \right)
\\\nonumber
=&\left(  \begin{array}{ccc}
  1/2  & \cdots &    (\pi-\theta_{ij})/(2\pi)   \\
   \vdots & \ddots & \vdots  \\
 (\pi-\theta_{ij})/(2\pi)  & \cdots &   1/2  
                         \end{array}
              \right).
\end{align}
We can therefore express the following expectations using entries of $\boldsymbol\mho$ as:
\begin{align}\label{E_ri0_rj0_mho}
 \mathbb E \left\{\mathbb I_{r,i}(0)\mathbb I_{r,j}(0)\right\}&=\frac{\pi-\theta_{ij}}{2\pi}=\boldsymbol\mho_{ij},\\\label{E_ri0_rell0_rj0_mho}
 \mathbb E\left\{\mathbb I_{r,i}(0)\mathbb I_{r,\ell}(0)\mathbb I_{r,j}(0)\right\}&=\frac{2\pi-(\theta_{i\ell}+\theta_{\ell j}+\theta_{ji})}{4\pi}=\frac{2(\mho_{i\ell}+\mho_{\ell j}+\mho_{ji})-1}{4}.
\end{align}
Remind the geometric parameters characterizing the data set defined in Section \ref{section_problem}:
\begin{align}\nonumber
\theta_{\text{min}}&=\min_{1\leqslant i\not=j\leqslant n}\theta_{ij}=\min_{1\leqslant i\not=j\leqslant n}\angle(\boldsymbol x_i,\boldsymbol x_j)=\min_{1\leqslant i\not=j\leqslant n}\{\arccos\langle\boldsymbol x_i, \boldsymbol x_j\rangle\}\leqslant2\pi/3,\\\nonumber
\theta_{\text{max}}&=\max_{1\leqslant i\not=j\leqslant n}\theta_{ij}=\max_{1\leqslant i\not=j\leqslant n}\angle(\boldsymbol x_i,\boldsymbol x_j)=\max_{1\leqslant i\not=j\leqslant n}\{\arccos\langle\boldsymbol x_i, \boldsymbol x_j\rangle\},\\\nonumber
\theta^*&=\min\{\theta_{\text{min}},\pi-\theta_{\text{max}}\}.
\end{align}
which admits the following interval of the entry of matrix $\boldsymbol\mho$ as:
\begin{align}\label{mho_ij_bound}
\boldsymbol\mho_{ii}=\frac{1}{2},
\boldsymbol\mho_{ij}=\frac{\pi-\theta_{ij}}{2\pi}\in[\mho_{\text{min}},\mho_{\text{max}}]=\left[\frac{\pi-\theta_{\text{max}}}{2\pi},\frac{\pi-\theta_{\text{min}}}{2\pi}\right]\in\left(0,\frac{1}{2}\right),\forall i\not=j.
\end{align}
Notice that $\mho_{\text{max}}$ is the maximum value of off-diagonal entries of matrix $\boldsymbol\mho$, but not the maximum value of all the entries of matrix $\boldsymbol\mho$.
\\Recall the parameter $\lambda^*=\lambda_{\min}(\boldsymbol\mho)$ defined in (\ref{matrix_mho_definition}) in Section \ref{section_problem}, we have
\begin{align}\nonumber
&\lambda_{\min}(\boldsymbol\Upsilon^*)=\lambda_{\min}\left(\left(  \begin{array}{ccc}
  0  & \cdots &     \mathbb E( \langle\widetilde{\boldsymbol\psi_1}(0),\widetilde{\boldsymbol\psi_n}(0)\rangle)   \\
   \vdots & \ddots & \vdots  \\
 \mathbb E(  \langle\widetilde{\boldsymbol\psi_n}(0),\widetilde{\boldsymbol\psi_1}(0)\rangle)  & \cdots &   0   
                         \end{array}
              \right)\right)\\\nonumber
              =&
              \lambda_{\min}\left(\left(  \begin{array}{ccc}
  0  & \cdots &    m(\pi-\theta_{ij})/(2\pi)   \\
   \vdots & \ddots & \vdots  \\
 m(\pi-\theta_{ij})/(2\pi)  & \cdots &   0   
                         \end{array}
              \right)\right)\\\nonumber
                            =&
              \lambda_{\min}\left(m\left(\left(  \begin{array}{ccc}
  1/2  & \cdots &    (\pi-\theta_{ij})/(2\pi)   \\
   \vdots & \ddots & \vdots  \\
 (\pi-\theta_{ij})/(2\pi)  & \cdots &   1/2  
                         \end{array}
              \right)-\frac{1}{2}\boldsymbol I_n\right)\right)\\\nonumber
              =&
              \lambda_{\min}\left(m\left(\boldsymbol\mho-\frac{1}{2}\boldsymbol I_n\right)\right)=m\left(\lambda_{\min}(\boldsymbol\mho)-\frac{1}{2}\right)=m\left(\lambda^*-\frac{1}{2}\right).
\end{align}
Applying {\textcolor{red}{Weyl's inequality} (\cite{weyl_1912}) gives
\begin{align}\label{lambda_min_Upsilon_lower_bound_1}
\lambda_{\min}(\boldsymbol\Upsilon)\geqslant&
\lambda_{\min}(\boldsymbol\Upsilon^*)-\|\Delta_{\boldsymbol\Upsilon}\|_2=
 m\left(\lambda^*-\frac{1}{2}\right)-\|\Delta_{\boldsymbol\Upsilon}\|_2.
\end{align}
{\textcolor{blue}{{\bf Step 4: Estimating $\|\boldsymbol\Delta_{\Upsilon}\|_2$ via Matrix Bernstein Inequality}}\\
Since the $(i,j)(1\leqslant i\not=j\leqslant n)$ entry of matrix  $\boldsymbol\Delta_{\Upsilon}$ is $$\langle\widetilde{\boldsymbol\psi_i}(0),\widetilde{\boldsymbol\psi_j}(0)\rangle-\mathbb E( \langle\widetilde{\boldsymbol\psi_i}(0),\widetilde{\boldsymbol\psi_j}(0)\rangle)
=\sum_{r=1}^m \underbrace{\left[\mathbb I_{r,i}(0)\mathbb I_{r,j}(0)-\mathbb E \left\{\mathbb I_{r,i}(0)\mathbb I_{r,j}(0)\right\}\right]}_{\varrho_{ij}^{(r)}} \triangleq\sum_{r=1}^m\varrho_{ij}^{(r)},$$we have
\begin{align}\nonumber
\|\boldsymbol\Delta_{\Upsilon}\|_2
=\left\|\sum_{r=1}^m  \underbrace{\left(  \begin{array}{ccc}
  0  & \cdots &      \varrho_{ij}^{(r)}  \\
   \vdots & \ddots & \vdots  \\
   \varrho_{ij}^{(r)}  & \cdots &     0   
                         \end{array} \right)}_{\boldsymbol\Delta_{\Upsilon}^{(r)}}  \right\|_2
=\left\|\sum_{r=1}^m \boldsymbol\Delta_{\Upsilon}^{(r)}  \right\|_2
\end{align}
It remains to estimate $\|\sum_{r=1}^m\boldsymbol\Delta_{\Upsilon}^{(r)}\|_2$. Observe that $\sum_{r=1}^m\boldsymbol\Delta_{\Upsilon}^{(r)}$ is a summation of independent random matrices due to the independency of $\{\boldsymbol w_r(0)\}_{r=1}^m$, we can apply {\textcolor{red}{matrix Bernstein inequality} (\cite{tropp_2015}) in high-dimensional probability to obtain a sharp upper bound of $\|\sum_{r=1}^m\boldsymbol\Delta_{\Upsilon}^{(r)}\|_2$. Sepcifically, let
\begin{align}\nonumber
B&=\max_{1\leqslant r\leqslant m}\left\|\boldsymbol\Delta_{\Upsilon}^{(r)}\right\|_2,\\\nonumber
V&=\max\{\sum_{1\leqslant r\leqslant m}\|\mathbb E(\boldsymbol\Delta_{\Upsilon}^{(r)}
(\boldsymbol\Delta_{\Upsilon}^{(r)})^\top)\|_2,\sum_{1\leqslant r\leqslant m}\|\mathbb E((\boldsymbol\Delta_{\Upsilon}^{(r)}
)^\top\boldsymbol\Delta_{\Upsilon}^{(r)})\|_2\}=\sum_{1\leqslant r\leqslant m}\|\mathbb E(\boldsymbol\Delta_{\Upsilon}^{(r)})^2\|_2,
\end{align}
then {\textcolor{red}{matrix Bernstein inequality} yields that with probability exceeding $1-\delta$,
\begin{align}\nonumber
&\|\boldsymbol\Delta_{\Upsilon}\|_2=\left\|\sum_{r=1}^m\boldsymbol\Delta_{\Upsilon}^{(r)}\right\|_2
\leqslant\frac{2}{3}B\log\left(\frac{2n}{\delta}\right)+\sqrt{2V\log\left(\frac{2n}{\delta}\right)}.
\end{align}
In what follows, we shall upper bound these two terms $B$ and $V$ seperately.\\
{\textcolor{blue}{\bf Substep 4-1: Bounding $B=\max_{1\leqslant r\leqslant m}\left\|\boldsymbol\Delta_{\Upsilon}^{(r)}\right\|_2$}}\\
For $B$, with the aid of {\textcolor{red}{Chebyshev inequality}, one obtains that for any $t>0$,
\begin{align}\nonumber
&\mathbb P\{|\mathbb I_{r,i}(0)\mathbb I_{r,j}(0)-\mathbb E \left\{\mathbb I_{r,i}(0)\mathbb I_{r,j}(0)\right\}|\geqslant t\}\leqslant\frac{\text{var}[\mathbb I_{r,i}(0)\mathbb I_{r,j}(0)]}{t^2}\\\nonumber
=&\frac{\mathbb E[\mathbb I_{r,i}^2(0)\mathbb I_{r,j}^2(0)]
-[\mathbb E\mathbb I_{r,i}(0)\mathbb I_{r,j}(0)]^2}{t^2}
=\frac{\mathbb E[\mathbb I_{r,i}(0)\mathbb I_{r,j}(0)]
-[\mathbb E\mathbb I_{r,i}(0)\mathbb I_{r,j}(0)]^2}{t^2}
\stackrel{(\text{i})}{=}\frac{\boldsymbol\mho_{ij}-\boldsymbol\mho_{ij}^2}{t^2},
\end{align} 
where (i) comes from (\ref{E_ri0_rj0_mho}). 
\\Therefore, with probability exceeding $1-\delta$,
\begin{align}\nonumber
|\mathbb I_{r,i}(0)\mathbb I_{r,j}(0)-\mathbb E \left\{\mathbb I_{r,i}(0)\mathbb I_{r,j}(0)\right\}|\leqslant\sqrt{\frac{\boldsymbol\mho_{ij}-\boldsymbol\mho_{ij}^2}{\delta}}.
\end{align}
For a fixed $i\in[n]$, remind that
\begin{align}\nonumber
\sum_{1\leqslant j\not=i\leqslant n}|\varrho_{ij}^{(r)}|
=\sum_{1\leqslant j\not=i\leqslant n}\left|\mathbb I_{r,i}(0)\mathbb I_{r,j}(0)-\mathbb E \left\{\mathbb I_{r,i}(0)\mathbb I_{r,j}(0)\right\}\right|.
\end{align}
Thus, applying a union bound over $j\in[n]$ and $j\not=i$, we have with probability exceeding $1-(n-1)\delta$,
\begin{align}\nonumber
&\sum_{1\leqslant j\not=i\leqslant n}|\varrho_{ij}^{(r)}|\\\nonumber
=&\sum_{1\leqslant j\not=i\leqslant n}\left|\mathbb I_{r,i}(0)\mathbb I_{r,j}(0)-\mathbb E \left\{\mathbb I_{r,i}(0)\mathbb I_{r,j}(0)\right\}\right|\\\nonumber
\leqslant&\sum_{1\leqslant j\not=i\leqslant n}\sqrt{\frac{\boldsymbol\mho_{ij}-\boldsymbol\mho_{ij}^2}{\delta}}\leqslant(n-1)\max_{1\leqslant j\not=i\leqslant n}\sqrt{\frac{\boldsymbol\mho_{ij}-\boldsymbol\mho_{ij}^2}{\delta}}\\\nonumber
\stackrel{\text{(i)}}{=}&(n-1)\sqrt{\frac{\max_{1\leqslant j\not=i\leqslant n}\boldsymbol\mho_{ij}-(\max_{1\leqslant j\not=i\leqslant n}\boldsymbol\mho_{ij})^2}{\delta}}\\\nonumber
\stackrel{\text{(ii)}}{=}&(n-1)\sqrt{\delta^{-1}\left[ \frac{\pi-\theta_{\text{min}}}{2\pi}-\left(\frac{\pi-\theta_{\text{min}}}{2\pi}\right)^2 \right]}\\\nonumber
=&\frac{n-1}{2}\sqrt{\frac{\pi^2-\theta_{\text{min}}^2}{\pi^2\delta}}.
\end{align}
Here (i) due to $\boldsymbol\mho_{ij}\in(0,1/2)$ and (ii) comes from (\ref{mho_ij_bound}).\\
For a fixed $r\in[m]$, making use of {\textcolor{red}{Ger\v{s}gorin disc theorem} (\cite{gersgorin_1931}) for symmetric matrix $\boldsymbol\Delta_{\Upsilon}^{(r)}$, there exists $i\in[n]$, with probability exceeding $1-(n-1)\delta$,
\begin{align}\nonumber
\left\|\boldsymbol\Delta_{\Upsilon}^{(r)}\right\|_2=\lambda_{\text{max}}(\boldsymbol\Delta_{\Upsilon}^{(r)})  \leqslant\sum_{1\leqslant j\not=i\leqslant n}|\varrho_{ij}^{(r)}|\leqslant\frac{n-1}{2}\sqrt{\frac{\pi^2-\theta_{\text{min}}^2}{\pi^2\delta}}.
\end{align}
Again, by a union bound over $r\in[m]$, we arrive with probability exceeding $1-m(n-1)\delta$,
\begin{align}\label{Delta_Upsilon_B_upper_bound}
B=\max_{1\leqslant r\leqslant m}\left\|\boldsymbol\Delta_{\Upsilon}^{(r)}\right\|_2\leqslant\frac{n-1}{2}\sqrt{\frac{\pi^2-\theta_{\text{min}}^2}{\pi^2\delta}}
\leqslant\frac{n-1}{2}\sqrt{\frac{\pi^2-(\theta^*)^2}{\pi^2\delta}}.
\end{align}
{\textcolor{blue}{\bf Substep 4-2: Bounding $V=\sum_{1\leqslant r\leqslant m}\|\mathbb E(\boldsymbol\Delta_{\Upsilon}^{(r)})^2\|_2$}}\\
Turing to $V$, be careful that diagonal entries of $\boldsymbol\Delta_{\Upsilon}^{(r)}$ is zero, then straightforward calculation gives the $(i,j)$ entry of $(\boldsymbol\Delta_{\Upsilon}^{(r)})^2(1\leqslant i\not=j\leqslant n)$ is
\begin{align}\nonumber
&((\boldsymbol\Delta_{\Upsilon}^{(r)})^2)_{i,j}\\\nonumber
=&\sum_{\ell=1}^n(\boldsymbol\Delta_{\Upsilon}^{(r)})_{i\ell}(\boldsymbol\Delta_{\Upsilon}^{(r)})_{\ell j}\\\nonumber
=&\sum_{\ell=1}^n\varrho_{i\ell}^{(r)}\varrho_{\ell j}^{(r)}-\varrho_{ii}^{(r)}\varrho_{i j}^{(r)}-\varrho_{ij}^{(r)}\varrho_{jj}^{(r)}\\\nonumber
=&\sum_{\ell=1}^n\left[\mathbb I_{r,i}(0)\mathbb I_{r,\ell}(0)-\mathbb E \left\{\mathbb I_{r,i}(0)\mathbb I_{r,\ell}(0)\right\}\right]\left[\mathbb I_{r,\ell}(0)\mathbb I_{r,j}(0)-\mathbb E \left\{\mathbb I_{r,\ell}(0)\mathbb I_{r,j}(0)\right\}\right]
-(\varrho_{ii}^{(r)}+\varrho_{jj}^{(r)})\varrho_{ij}^{(r)}\\\nonumber
=&\sum_{\ell=1}^n\left[\mathbb I_{r,i}(0)\mathbb I_{r,\ell}(0)-\boldsymbol\mho_{i\ell}\right]\left[\mathbb I_{r,\ell}(0)\mathbb I_{r,j}(0)-\boldsymbol\mho_{\ell j}\right]-(\varrho_{ii}^{(r)}+\varrho_{jj}^{(r)})\varrho_{ij}^{(r)}.
\end{align}
With the aid of (\ref{E_ri0_rj0_mho}) and (\ref{E_ri0_rell0_rj0_mho}), for $1\leqslant i\not=j\leqslant n$, one has 
\begin{align}\nonumber
&\mathbb E((\boldsymbol\Delta_{\Upsilon}^{(r)})^2)_{i,j}\\\nonumber
=&\mathbb E\left\{\sum_{\ell=1}^n\left[\mathbb I_{r,i}(0)\mathbb I_{r,\ell}(0)-\boldsymbol\mho_{i\ell}\right]\left[\mathbb I_{r,\ell}(0)\mathbb I_{r,j}(0)-\boldsymbol\mho_{\ell j}\right]\right\}
-\mathbb E\left[ (\varrho_{ii}^{(r)}+\varrho_{jj}^{(r)})\varrho_{ij}^{(r)} \right]\\\nonumber
=&\mathbb E\left\{\sum_{\ell=1}^n\left[\mathbb I_{r,i}(0)\mathbb I_{r,\ell}(0)\mathbb I_{r,\ell}(0)\mathbb I_{r,j}(0)-\boldsymbol\mho_{\ell j}\mathbb I_{r,i}(0)\mathbb I_{r,\ell}(0)-\boldsymbol\mho_{i\ell}\mathbb I_{r,\ell}(0)\mathbb I_{r,j}(0)+\boldsymbol\mho_{i\ell}\boldsymbol\mho_{\ell j}\right]\right\}-\boldsymbol\mho_{ij}\\\nonumber
=&\sum_{\ell=1}^n\left\{\mathbb E\left[\mathbb I_{r,i}(0)\mathbb I_{r,\ell}(0)\mathbb I_{r,j}(0)\right]-\boldsymbol\mho_{\ell j}\boldsymbol\mho_{i\ell}  -\boldsymbol\mho_{i\ell}\boldsymbol\mho_{\ell j}+\boldsymbol\mho_{i\ell}\boldsymbol\mho_{\ell j}\right\}-\boldsymbol\mho_{ij}\\\label{E_Delta_Upsilon_2_ij}
=&\sum_{\ell=1}^n\left\{\frac{2(\mho_{i\ell}+\mho_{\ell j}+\mho_{ji})-1}{4}-\boldsymbol\mho_{i\ell}\boldsymbol\mho_{\ell j}\right\}-\boldsymbol\mho_{ij}.
\end{align}
In a similar manner, for $i\in[n]$, we have
\begin{align}\nonumber
&\mathbb E((\boldsymbol\Delta_{\Upsilon}^{(r)})^2)_{i,i}\\\nonumber
=&\sum_{\ell=1}^n\left\{\mathbb E\left[\mathbb I_{r,i}(0)\mathbb I_{r,\ell}(0)\mathbb I_{r,i}(0)\right]-\boldsymbol\mho_{i\ell}\boldsymbol\mho_{\ell i}\right\}-\mathbb E\left[ (\varrho_{ii}^{(r)})^2  \right] \\\nonumber
=&\sum_{\ell=1}^n\left\{\mathbb E\left[\mathbb I_{r,i}(0)\mathbb I_{r,\ell}(0)\right]-\boldsymbol\mho_{i\ell}\boldsymbol\mho_{\ell i}\right\}-\mathbb E\left[ \left(\mathbb I_{r,i}(0)-\frac{1}{2}\right)^2 \right] \\\label{E_Delta_Upsilon_2_ii}
=&\sum_{\ell=1}^n\left(\boldsymbol\mho_{i\ell}-\boldsymbol\mho_{i\ell}^2\right)
-\frac{1}{4}
=\sum_{j=1}^n\left(\boldsymbol\mho_{ij}-\boldsymbol\mho_{ij}^2\right)
-\frac{1}{4}.
\end{align}
Invoking {\textcolor{red}{Ger\v{s}gorin disc theorem} (\cite{gersgorin_1931}) for symmetric matrix $\mathbb E\left[(\boldsymbol\Delta_{\Upsilon}^{(r)})^2\right]$, by (\ref{E_Delta_Upsilon_2_ij}) and (\ref{E_Delta_Upsilon_2_ii}), we see that there exists certain $i\in[n]$ such that
\begin{align}\nonumber
&\left\|\mathbb E\left[(\boldsymbol\Delta_{\Upsilon}^{(r)})^2\right]\right\|_2=\lambda_{\text{max}}((\boldsymbol\Delta_{\Upsilon}^{(r)})^2)\leqslant\mathbb E((\boldsymbol\Delta_{\Upsilon}^{(r)})^2)_{i,i}+\sum_{1\leqslant j\not=i\leqslant n}\left|\mathbb E((\boldsymbol\Delta_{\Upsilon}^{(r)})^2)_{i,j}\right|\\\nonumber
=&\sum_{j=1}^n\left(\boldsymbol\mho_{ij}-\boldsymbol\mho_{ij}^2\right)
-\frac{1}{4}
+\sum_{1\leqslant j\not=i\leqslant n}\left|\sum_{\ell=1}^n\left\{\frac{2(\mho_{i\ell}+\mho_{\ell j}+\mho_{ji})-1}{4}-\boldsymbol\mho_{i\ell}\boldsymbol\mho_{\ell j}\right\}-\boldsymbol\mho_{ij}\right|\\\nonumber
=&(\boldsymbol\mho_{ii}-\boldsymbol\mho_{ii}^2)+\sum_{1\leqslant j\not=i\leqslant n}\left(\boldsymbol\mho_{ij}-\boldsymbol\mho_{ij}^2\right)
-\frac{1}{4}+\sum_{1\leqslant j\not=i\leqslant n}\left|\sum_{\ell=1}^n\left\{\frac{2(\mho_{i\ell}+\mho_{\ell j}+\mho_{ji})-1}{4}-\boldsymbol\mho_{i\ell}\boldsymbol\mho_{\ell j}\right\}-\boldsymbol\mho_{ij}\right|\\\nonumber
\stackrel{\text{(i)}}{=}&\sum_{1\leqslant j\not=i\leqslant n}\left(\boldsymbol\mho_{ij}-\boldsymbol\mho_{ij}^2\right)
+\sum_{1\leqslant j\not=i\leqslant n}\left|\frac{1}{2}\sum_{\ell=1}^n\left\{\mho_{i\ell}+\mho_{\ell j}-2\boldsymbol\mho_{i\ell}\boldsymbol\mho_{\ell j}\right\}+\left(\frac{n}{2}-1\right)\mho_{ij}-\frac{n}{4}\right|\\\nonumber
\stackrel{\text{(ii)}}{\leqslant}&\sum_{1\leqslant j\not=i\leqslant n}\left(\boldsymbol\mho_{ij}-\boldsymbol\mho_{ij}^2\right)
+\sum_{1\leqslant j\not=i\leqslant n}\left|\frac{1}{2}\sum_{\ell=1}^n(\mho_{i\ell}+\mho_{\ell j}-2\boldsymbol\mho_{i\ell}\boldsymbol\mho_{\ell j})+\left(\frac{n}{2}-1\right)\mho_{ij}\right|+\frac{n(n-1)}{4}\\\nonumber
\stackrel{\text{(iii)}}{=}&\frac{n^2-n}{4}+\sum_{1\leqslant j\not=i\leqslant n}\left(\boldsymbol\mho_{ij}-\boldsymbol\mho_{ij}^2\right)
+\frac{1}{2}\sum_{1\leqslant j\not=i\leqslant n}\left[\sum_{\ell=1}^n(\mho_{i\ell}+\mho_{\ell j}-2\boldsymbol\mho_{i\ell}\boldsymbol\mho_{\ell j})+(n-2)\mho_{ij}\right]\\\nonumber
=&\frac{n^2-n}{4}+\sum_{1\leqslant j\not=i\leqslant n}\left[\frac{n}{2}\mho_{ij}-\mho_{ij}^2\right]+\frac{1}{2}\sum_{1\leqslant j\not=i\leqslant n}\sum_{\ell=1}^n(\mho_{i\ell}+\mho_{\ell j})-\sum_{1\leqslant j\not=i\leqslant n}\sum_{\ell=1}^n\mho_{i\ell}\mho_{\ell j}\\\nonumber
\stackrel{\text{(iv)}}{\leqslant}&\frac{n^2-n}{4}+\sum_{1\leqslant j\not=i\leqslant n}\left[\frac{n}{2}\mho_{\text{max}}-\mho_{\text{max}}^2\right]+\frac{1}{2}n(n-1)\cdot2\mho_{\text{max}}-n(n-1)\mho_{\text{min}}^2\\\nonumber
=&\left(\frac{1}{4}+\frac{3}{2}\mho_{\text{max}}-\mho_{\text{min}}^2\right)n^2-\left(\frac{1}{4}+\frac{1}{2}\mho_{\text{max}}+\mho_{\text{max}}^2-\mho_{\text{min}}^2\right)n
+\left(\frac{1}{2}-\mho_{\text{max}}\right)^2
-\frac{1}{4}-\sum_{1\leqslant j\not=i\leqslant n}\mho_{ij}\\\nonumber
\stackrel{\text{(v)}}{=}&\left[\frac{1}{4}+\frac{3}{2}\cdot\frac{\pi-\theta_{\text{min}}}{2\pi}-\left(\frac{\pi-\theta_{\text{max}}}{2\pi}\right)^2\right]n^2\\\nonumber
&-\left[\frac{1}{4}+\frac{1}{2}\cdot\frac{\pi-\theta_{\text{min}}}{2\pi}+\left(\frac{\pi-\theta_{\text{min}}}{2\pi}\right)^2-\left(\frac{\pi-\theta_{\text{max}}}{2\pi}\right)^2\right]n
+\frac{\theta_{\text{min}}^2}{4\pi^2}-\frac{1}{4}-\sum_{1\leqslant j\not=i\leqslant n}\mho_{ij}\\\nonumber
=&\left[\frac{1}{4}+\frac{2\pi^2+(2\pi\theta_{\text{max}}-\theta_{\text{max}}^2)-3\pi\theta_{\text{min}}
}{4\pi^2}\right]n^2\\\nonumber
&-\left[\frac{1}{4}+\frac{\pi^2+(2\pi\theta_{\text{max}}-\theta_{\text{max}}^2)+(\theta_{\text{min}}^2
-3\pi\theta_{\text{min}})}{4\pi^2}\right]n+\frac{\theta_{\text{min}}^2}{4\pi^2}-\frac{1}{4}-\sum_{1\leqslant j\not=i\leqslant n}\mho_{ij}\\\nonumber
\stackrel{\text{(vi)}}{<}&\left[\frac{1}{4}+\frac{2\pi^2+[2\pi(\pi-\theta^*)-(\pi-\theta^*)^2]-3\pi\theta^*
}{4\pi^2}\right]n^2\\\nonumber
&-\left[\frac{1}{4}+\frac{\pi^2+(2\pi\theta^*-(\theta^*)^2)+[(\pi-\theta^*)^2
-3\pi(\pi-\theta^*)]}{4\pi^2}\right]n+\frac{\pi^2}{4\pi^2}-\frac{1}{4}-(n-1)\frac{\theta^*}{2\pi}\\\label{E_Upsilon_square_2_norm}
=&\left[1-\frac{\theta^*(\theta^*+3\pi)}{4\pi^2}\right]n^2-\frac{5\theta^*}{4\pi}n
\end{align}
where (i) and (v) use (\ref{mho_ij_bound}) and $\mho_{ji}=\mho_{ij}$; (ii) holds due to {\textcolor{red}{triangle inequality}; (iii) follows from the following fact due to $0<\mho_{ij}\leqslant1/2$ from (\ref{mho_ij_bound}):
\begin{align}\nonumber
\frac{1}{2}\sum_{\ell=1}^n(\mho_{i\ell}+\mho_{\ell j}-2\boldsymbol\mho_{i\ell}\boldsymbol\mho_{\ell j})+\left(\frac{n}{2}-1\right)\mho_{ij}
=\frac{1}{2}\sum_{\ell=1}^n\left[\mho_{i\ell}(1-2\boldsymbol\mho_{\ell j})+\mho_{\ell j}\right]+\left(\frac{n}{2}-1\right)\mho_{ij}\geqslant0;
\end{align}
(iv) and (v) use (\ref{mho_ij_bound}) with $\mho_{\text{max}}<1/2$; (vi) makes use of the fact that
\begin{align}\nonumber
0<\theta^*\leqslant\theta_{\text{min}}\leqslant\theta_{\text{max}}\leqslant\pi-\theta^*<\pi.
\end{align}
Therefore, (\ref{E_Upsilon_square_2_norm}) leads to
\begin{align}\label{Delta_Upsilon_V_upper_bound}
V= m\|\mathbb E(\boldsymbol\Delta_{\Upsilon}^{(r)})^2\|_2<
m\left[\left(1-\frac{\theta^*(\theta^*+3\pi)}{4\pi^2}\right)n^2-\frac{5\theta^*}{4\pi}n\right].
\end{align}
Invoking {\textcolor{red}{matrix Bernstein inequality} yields, with probability exceeding $1-2\delta$,
\begin{align}\nonumber
&\|\boldsymbol\Delta_{\Upsilon}\|_2=\left\|\sum_{r=1}^m\boldsymbol\Delta_{\Upsilon}^{(r)}\right\|_2
\leqslant\frac{2}{3}B\log\left(\frac{2n}{\delta}\right)+\sqrt{2V\log\left(\frac{2n}{\delta}\right)}.
\end{align}
With the above bounds (\ref{Delta_Upsilon_B_upper_bound}) and (\ref{Delta_Upsilon_V_upper_bound}) in place and a union bound, we have that with probability exceeding $1-m(n-1)\delta-2\delta=1-(mn-m+2)\delta$,
\begin{align}\label{Delta_Upsilon_spectral_upper_bound}
\|\boldsymbol\Delta_{\Upsilon}\|_2<\frac{n-1}{3}\sqrt{\frac{\pi^2-(\theta^*)^2}{\pi^2\delta}}\log\left(\frac{2n}{\delta}\right)+\sqrt{2m\left[\left(1-\frac{\theta^*(\theta^*+3\pi)}{4\pi^2}\right)n^2-\frac{5\theta^*}{4\pi}n\right]\log\left(\frac{2n}{\delta}\right)}.
\end{align}
{\textcolor{blue}{{\bf Step 5: Estimating the Smallest Eigenvalue of Matrix $\boldsymbol\Upsilon$ via Weyl's Inequality}}\\
Substituting (\ref{Delta_Upsilon_spectral_upper_bound}) into (\ref{lambda_min_Upsilon_lower_bound_1}) which applies \textcolor{red}{Weyl's inequality}, one has that with probability exceeding $1-(mn-m+2)\delta$,
\begin{align}\nonumber
&\lambda_{\min}(\boldsymbol\Upsilon)\\\nonumber
\geqslant&
m\left(\lambda^*-\frac{1}{2}\right)-\frac{n-1}{3}\sqrt{\frac{\pi^2-(\theta^*)^2}{\pi^2\delta}}\log\left(\frac{2n}{\delta}\right)-\sqrt{2m\left[\left(1-\frac{\theta^*(\theta^*+3\pi)}{4\pi^2}\right)n^2-\frac{5\theta^*}{4\pi}n\right]\log\left(\frac{2n}{\delta}\right)}\\\nonumber
>&
m\left(\lambda^*-\frac{1}{2}\right)-\frac{n}{3}\sqrt{\left[1-\left(\frac{\theta^*}{\pi}\right)^2\right]\frac{1}{\delta}}\log\left(\frac{2n}{\delta}\right)-\sqrt{2m\left[\left(1-\frac{\theta^*(\theta^*+3\pi)}{4\pi^2}\right)n^2\right]\log\left(\frac{2n}{\delta}\right)}\\\label{lambda_min_Upsilon_lower_bound_2}
\stackrel{\text{(i)}}{>}&
m\left(\lambda^*-\frac{1}{2}\right)-\frac{n}{3}\left[1-\frac{1}{2}\left(\frac{\theta^*}{\pi}\right)^2\right]\sqrt{\frac{1}{\delta}}\log\left(\frac{2n}{\delta}\right)-\sqrt{2m\left[\left(1-\frac{\theta^*(\theta^*+3\pi)}{4\pi^2}\right)n^2\right]\log\left(\frac{2n}{\delta}\right)},
\end{align}
where (i) results from the inequality $\sqrt{1-x^2}<1-x^2/2$ for $|x|<1$.\\\\
{\textcolor{blue}{{\bf Step 6: Estimating the Smallest Singular Value of the Initial Activation Matrix}}\\
Substituting the bounds (\ref{Phi_i0_min_lower_bound}) and (\ref{lambda_min_Upsilon_lower_bound_2}) into (\ref{sigma_min_Phi0_lower_bound}) and using a union bound gives, with probability exceeding $1-n\delta-(mn-m+2)\delta=1-(mn-m+n+2)\delta$,
\begin{align}\nonumber
&\sigma_{\min}^2(\boldsymbol \Psi_0)\\\nonumber
\geqslant&\min_{1\leqslant i\leqslant n}\|\widetilde{\boldsymbol\psi_i}(0)\|_2^2+\lambda_{\min}(\boldsymbol\Upsilon)\\\nonumber
>&\frac{m}{2}-\sqrt{\frac{m}{2}\log\frac{1}{\delta}}+m\left(\lambda^*-\frac{1}{2}\right)-\frac{n}{3}\left[1-\frac{1}{2}\left(\frac{\theta^*}{\pi}\right)^2\right]\sqrt{\frac{1}{\delta}}\log\left(\frac{2n}{\delta}\right)\\\nonumber
-&\sqrt{2m\left[\left(1-\frac{\theta^*(\theta^*+3\pi)}{4\pi^2}\right)n^2\right]\log\left(\frac{2n}{\delta}\right)}\\\nonumber
=&m\lambda^*-\sqrt{\frac{m}{2}\log\frac{1}{\delta}}-\frac{n}{3}\left[1-\frac{1}{2}\left(\frac{\theta^*}{\pi}\right)^2\right]\log\left(\frac{2n}{\delta}\right)-\sqrt{2m\left[\left(1-\frac{\theta^*(\theta^*+3\pi)}{4\pi^2}\right)n^2\right]\log\left(\frac{2n}{\delta}\right)}\\\nonumber
=&m\lambda^*-\sqrt{m}n\sqrt{2\left(1-\frac{\theta^*(\theta^*+3\pi)}{4\pi^2}\right)}\sqrt{\log\left(\frac{2n}{\delta}\right)}
-\sqrt{\frac{m}{2}\log\frac{1}{\delta}}-\frac{n}{3}\left[1-\frac{1}{2}\left(\frac{\theta^*}{\pi}\right)^2\right]\sqrt{\frac{1}{\delta}}\log\left(\frac{2n}{\delta}\right)\\\nonumber
=&m\lambda^*-\sqrt{m}n\sqrt{2\left(1-\frac{\theta^*(\theta^*+3\pi)}{4\pi^2}\right)}\sqrt{\log\left(\frac{2n}{\delta}\right)}-\left[\sqrt{\frac{m}{2}\log\frac{1}{\delta}}+\frac{n[2\pi^2-(\theta^*)^2]}{6\pi^2}\sqrt{\frac{1}{\delta}}\log\left(\frac{2n}{\delta}\right)\right]\\\label{min_sigma_Psi0_final_result}
=&\lambda^*m-n\sqrt{\left[2-\frac{\theta^*(\theta^*+3\pi)}{2\pi^2}\right]m\log(2n/\delta)}+\mathcal O\left(\sqrt{m\log(1/\delta)}+n\log(2n/\delta)/\sqrt{\delta}\right).
\end{align}
{\textcolor{blue}{{\bf Step 7: Exploring the Lower and Upper Bounds of $\lambda^*$}}\\
We now move to lower bound $\lambda^*$, revealing its connection to the spectral property of data matrix $\boldsymbol X$. We know from (\ref{matrix_mho_definition}) that the off-diagonal entry of matrix $\boldsymbol\mho$ can be expressed as:
\begin{align}\nonumber
&\boldsymbol\mho_{ij}=\frac{\pi-\arccos\langle\boldsymbol x_i, \boldsymbol x_j\rangle}{2\pi}=\frac{\pi-(\pi/2-\arcsin\langle\boldsymbol x_i, \boldsymbol x_j\rangle)}{2\pi}\\\nonumber
=&\frac{1}{4}+\frac{1}{2\pi}\arcsin\langle\boldsymbol x_i, \boldsymbol x_j\rangle=\frac{1}{4}+\frac{1}{2\pi}\arcsin(\boldsymbol x_i'\boldsymbol x_j)\\\nonumber
\stackrel{\text{(i)}}{=}&\frac{1}{4}+\frac{1}{2\pi}\int_0^{\boldsymbol x_i'\boldsymbol x_j}(\arcsin t)'dt=\frac{1}{4}+\frac{1}{2\pi}\int_0^{\boldsymbol x_i'\boldsymbol x_j}\frac{1}{\sqrt{1-t^2}}dt=\frac{1}{4}+\frac{1}{2\pi}\int_0^{\boldsymbol x_i'\boldsymbol x_j}(1-t^2)^{-1/2}dt\\\nonumber
\stackrel{\text{(ii)}}{=}&\frac{1}{4}+\frac{1}{2\pi}\int_0^{\boldsymbol x_i'\boldsymbol x_j}\left[1+{-1/2\choose 1}(-t^2)+{-1/2\choose 2}(-t^2)^2+{-1/2\choose 3}(-t^2)^3+\cdots\right]dt\\\nonumber
=&\frac{1}{4}+\frac{1}{2\pi}\int_0^{\boldsymbol x_i'\boldsymbol x_j}\left[1+\frac{1}{2\cdot1!}t^2+\frac{1\cdot3}{2^2\cdot2!}t^4+\frac{1\cdot3\cdot5}{2^3\cdot3!}t^6+\frac{1\cdot3\cdot5\cdot7}{2^4\cdot4!}t^8+\cdots\right]dt\\\nonumber
\stackrel{\text{(iii)}}{=}&\frac{1}{4}+\frac{1}{2\pi}\int_0^{\boldsymbol x_i'\boldsymbol x_j}\left[1+\frac{1!!}{2!!}t^2+\frac{3!!}{4!!}t^4+\frac{5!!}{6!!}t^6+\frac{7!!}{8!!}t^8+\cdots\right]dt\\\nonumber
=&\frac{1}{4}+\frac{1}{2\pi}\left[\boldsymbol x_i'\boldsymbol x_j+\frac{1!!}{2!!}\cdot\frac{(\boldsymbol x_i'\boldsymbol x_j)^3}{3}+\frac{3!!}{4!!}\cdot\frac{(\boldsymbol x_i'\boldsymbol x_j)^5}{5}+\frac{5!!}{6!!}\cdot\frac{(\boldsymbol x_i'\boldsymbol x_j)^7}{7}+\frac{7!!}{8!!}\cdot\frac{(\boldsymbol x_i'\boldsymbol x_j)^9}{9}+\cdots\right]\\\label{mho_ij_arcsin_series}
\stackrel{\text{(iv)}}{=}&\frac{1}{4}+\frac{1}{2\pi}\sum_{\nu=0}^{+\infty}\frac{(2\nu-1)!!}{(2\nu)!!}\cdot\frac{(\boldsymbol x_i'\boldsymbol x_j)^{2\nu+1}}{2\nu+1}.
\end{align}
Here we make use of the \textcolor{red}{fundamental theorem of calculus} in (i); (ii) follows from the \textcolor{red}{binomial theorem} (\cite{hardy_1908}) with $-1<\boldsymbol x_i'\boldsymbol x_j<1$ and $m=-1/2$:
\begin{align}\nonumber
&(1+x)^m=1+{m\choose1}x+{m\choose2}x^2+{m\choose3}x^3+\cdots\\\nonumber
=&1+mx+\frac{m(m-1)}{2!}x^2+\frac{m(m-1)(m-2)}{3!}x^3+\cdots,\text{ for }-1<x<1 \text{ and }m\text{ is rational}.
\end{align}
(iii) and (iv) are due to $n!!=n(n-2)(n-4)\cdots$ and $0!!=(-1)!!=1$, respectively.\\
The binomial series $(1+x)^m=\sum_{\nu=0}^{+\infty}{m\choose\nu}x^{\nu}$ is convergent for $x=1$ if and only if $m>-1$ (\cite{hardy_1908}), leading to
\begin{align}\label{mho_ii_arcsin_series}
\frac{1}{4}+\frac{1}{2\pi}\sum_{\nu=0}^{+\infty}\frac{(2\nu-1)!!}{(2\nu)!!}\cdot\frac{1}{2\nu+1}=\frac{1}{4}+\frac{1}{2\pi}\arcsin1=\frac{1}{2}=\boldsymbol\mho_{ii}\quad\forall i\in[n].
\end{align}
Taking the series expressions (\ref{mho_ij_arcsin_series}) and (\ref{mho_ii_arcsin_series}) with the fact of $(\boldsymbol X^\top\boldsymbol X)_{ij}=\boldsymbol x_i\top\boldsymbol x_j$ yields
\begin{align}\nonumber
\boldsymbol\mho=\frac{1}{4}\boldsymbol1_n\boldsymbol1_n^\top+\frac{1}{2\pi}\sum_{\nu=0}^{+\infty}\frac{(2\nu-1)!!}{(2\nu)!!}\cdot\frac{(\boldsymbol X^\top\boldsymbol X)^{\circ(2\nu+1)}}{2\nu+1},
\end{align}
here $(\boldsymbol X^\top\boldsymbol X)^{\circ(2\nu+1)}=\underbrace{(\boldsymbol X^\top\boldsymbol X)\circ\cdots\circ(\boldsymbol X^\top\boldsymbol X)}_{(2\nu+1)\text{ terms}}$ denotes the matrix Hadamard power.\\
We can utilize \textcolor{red}{trace inequality}: $\lambda_{\min}(\boldsymbol A+\boldsymbol B)\geqslant\lambda_{\min}(\boldsymbol A)+\lambda_{\min}(\boldsymbol B)$ for Hermitian matrices $\boldsymbol A$ and $\boldsymbol B$ to deduce that
\begin{align}\nonumber
\lambda^*=\lambda_{\text{min}}(\boldsymbol\mho)\geqslant\frac{1}{4}\lambda_{\min}(\boldsymbol1_n\boldsymbol1_n^\top)+\frac{1}{2\pi}\sum_{\nu=0}^{+\infty}\frac{(2\nu-1)!!}{(2\nu)!!}\cdot\frac{\lambda_{\text{min}}[(\boldsymbol X^\top\boldsymbol X)^{\circ(2\nu+1)}]}{2\nu+1}.
\end{align}
We can then control $\lambda^*$ as
\begin{align}\nonumber
&\lambda_{\text{min}}[(\boldsymbol X^\top\boldsymbol X)^{\circ i}]
\geqslant\lambda_{\text{min}}[(\boldsymbol X^\top\boldsymbol X)^{\circ (i-1)}]\min_{1\leqslant i\leqslant n}\boldsymbol (\boldsymbol X^\top\boldsymbol X)_{ii}=\lambda_{\text{min}}[(\boldsymbol X^\top\boldsymbol X)^{\circ (i-1)}]\min_{1\leqslant i\leqslant n}\|\boldsymbol x_i\|_2^2\\\nonumber
=&\lambda_{\text{min}}[(\boldsymbol X^\top\boldsymbol X)^{\circ (i-1)}]\geqslant\lambda_{\text{min}}[(\boldsymbol X^\top\boldsymbol X)^{\circ (i-2)}]\min_{1\leqslant i\leqslant n}\boldsymbol (\boldsymbol X^\top\boldsymbol X)_{ii}=\lambda_{\text{min}}[(\boldsymbol X^\top\boldsymbol X)^{\circ (i-2)}]\\\nonumber
\geqslant&\cdots\geqslant\lambda_{\text{min}}[(\boldsymbol X^\top\boldsymbol X)^{\circ (i-3)}]\geqslant\cdots\geqslant\lambda_{\text{min}}(\boldsymbol X^\top\boldsymbol X),\quad\forall i\in\mathbb Z^+,
\end{align}
where all the inequalities result from (1) Schur's theorem discovered in 1911 (\cite{schur_1911}):
$
\lambda_{\text{min}}(\boldsymbol A\circ\boldsymbol B)\geqslant\lambda_{\text{min}}(\boldsymbol A)\min_{1\leqslant i\leqslant n}\boldsymbol B_{ii}\text{ for two }n\times n\text{ positive semidefinite matrices }\boldsymbol A\text{ and }\boldsymbol B,
$
here $\boldsymbol A\circ\boldsymbol B$ stands for matrix Hadamard (entrywise) product. (2) \textcolor{red}{Schur's multiplication theorem} (\cite{wang_2006_book}): $\boldsymbol A\circ\boldsymbol B$ is positive semidefinite if both $\boldsymbol A$ and $\boldsymbol B$ are positive semidefinite. It is obvious that $\lambda_{\min}(\boldsymbol1_n\boldsymbol1_n^\top)$=0, as a consequence,
\begin{align}\nonumber
&\lambda^*\geqslant\frac{1}{2\pi}\lambda_{\text{min}}(\boldsymbol X^\top\boldsymbol X)\sum_{\nu=0}^{+\infty}\frac{(2\nu-1)!!}{(2\nu)!!}\cdot\frac{1}{2\nu+1}\\\label{lambda_star_lower_bound}
=&\frac{1}{2\pi}\lambda_{\text{min}}(\boldsymbol X^\top\boldsymbol X)\arcsin1=\frac{\lambda_{\text{min}}(\boldsymbol X^\top\boldsymbol X)}{4},
\end{align}
Recall that (\ref{matrix_mho_definition}), we arrive at
\begin{align}\label{lambda_star_upper_bound}
\lambda^*\leqslant\frac{\sum_{i=1}^n\lambda_i(\boldsymbol\mho)}{n}\leqslant\frac{\text{tr}(\boldsymbol\mho)}{n}=\frac{1}{2}.
\end{align}
which completes the last piece of proving Lemma \ref{lemma_Psi0}.
\subsection{Upper Bounding the Perturbation of Activation Matrix During Training}\label{subsection_spectral_norm_variation_Psi}
{\textcolor{blue}{{\bf Step 1: Decomposing Activation Variation Matrix}}\\
We start with decomposing the symmetric marix $(\boldsymbol\Psi(k)-\boldsymbol\Psi(0))(\boldsymbol\Psi(k)-\boldsymbol\Psi(0))^\top$
\begin{align}\nonumber
&\|\boldsymbol\Psi(k)-\boldsymbol\Psi(0)\|_2^2\\\nonumber
=&\|(\boldsymbol\Psi(k)-\boldsymbol\Psi(0))(\boldsymbol\Psi(k)-\boldsymbol\Psi(0))^\top\|_2\\\nonumber
=&\|\text{diag}(\|\widetilde{\boldsymbol\psi_1}(k)-\widetilde{\boldsymbol\psi_1}(0)\|_2^2,\cdots,\|\widetilde{\boldsymbol\psi_n}(k)-\widetilde{\boldsymbol\psi_n}(0)\|_2^2)+\boldsymbol N\|_2\\\nonumber
\leqslant&\|\text{diag}(\|\widetilde{\boldsymbol\psi_1}(k)-\widetilde{\boldsymbol\psi_1}(0)\|_2^2,\cdots,\|\widetilde{\boldsymbol\psi_n}(k)-\widetilde{\boldsymbol\psi_n}(0)\|_2^2)\|_2+\|\boldsymbol N\|_2 \\\nonumber
=&\max_{1\leqslant i\leqslant n}\|\widetilde{\boldsymbol\psi_i}(k)-\widetilde{\boldsymbol\psi_i}(0)\|_2^2+\|\boldsymbol N\|_2,
\end{align}
where the $(i,j)$ off-diagonal entry of symmetric matrix $\boldsymbol N$ is $\langle\widetilde{\boldsymbol\psi_i}(k)-\widetilde{\boldsymbol\psi_i}(0),\widetilde{\boldsymbol\psi_j}(k)-\widetilde{\boldsymbol\psi_j}(0)\rangle$, and all the diagonal entries of symmetric matrix $\boldsymbol N$ are zero. 
Therefore, to obtain a tight estimate $\|\boldsymbol\Psi(k)-\boldsymbol\Psi(0)\|_2^2$, it suffices to estimate $\max_{1\leqslant i\leqslant n}\|\widetilde{\boldsymbol\psi_i}(k)-\widetilde{\boldsymbol\psi_i}(0)\|_2^2$ and $\|\boldsymbol N\|_2$.\\
Firstly, applying Lemma \ref{lemma_Psik_Psi0} with a union bound over $i\in[n]$, we see that with probability exceeding $1-n\delta$,
$$\max_{1\leqslant i\leqslant n}\|\widetilde{\boldsymbol\psi_i}(k)-\widetilde{\boldsymbol\psi_i}(0)\|_2^2\leqslant \wp(R)m+\frac{2}{3}\log\frac{1}{\delta}+\sqrt{2\wp(R)[1-\wp(R)]m\log\frac{1}{\delta}}.$$
Secondly, we can obtain an upper bound of $\|\boldsymbol N\|_2$ via {\textcolor{red}{Ger\v{s}gorin disc theorem} (\cite{gersgorin_1931}) for symmetric matrix $\boldsymbol N$, which needs an upper bound of $\sum_{1\leqslant j\not=i\leqslant n}|\boldsymbol N_{ij}|$.
For a fixed $i\in[n]$, since
\begin{align}\nonumber
\sum_{1\leqslant j\not=i\leqslant n}|\boldsymbol N_{ij}|
=\sum_{1\leqslant j\not=i\leqslant n}\left|\langle\widetilde{\boldsymbol\psi_i}(k)-\widetilde{\boldsymbol\psi_i}(0),\widetilde{\boldsymbol\psi_j}(k)-\widetilde{\boldsymbol\psi_j}(0)\rangle\right|
\end{align}
Owing to the symmetry of $\langle\widetilde{\boldsymbol\psi_i}(k)-\widetilde{\boldsymbol\psi_i}(0),\widetilde{\boldsymbol\psi_j}(k)-\widetilde{\boldsymbol\psi_j}(0)\rangle$, we just need to analyze the upper bound of $\langle\widetilde{\boldsymbol\psi_i}(k)-\widetilde{\boldsymbol\psi_i}(0),\widetilde{\boldsymbol\psi_j}(k)-\widetilde{\boldsymbol\psi_j}(0)\rangle$.
\\\\
{\textcolor{blue}{{\bf Step 2: Upper Bounding the Inner Product $\langle\widetilde{\boldsymbol\psi_i}(k)-\widetilde{\boldsymbol\psi_i}(0),\widetilde{\boldsymbol\psi_j}(k)-\widetilde{\boldsymbol\psi_j}(0)\rangle$}}\\
{\textcolor{blue}{\bf Substep 2-1: Bounding the Summation via Bernstein Inequality}}\\
Remind that
$$A_{i,r}=\{|\boldsymbol w_r(0)^\top\boldsymbol x_i|\leqslant R\},\quad i\in[n],r\in[m].$$
$$S_i=\{r\in[m]:\mathbb I\{A_{i,r}\}=0\}\quad\text{and}\quad
\bar{S}_i=\{r\in[m]:\mathbb I\{A_{i,r}\}=1\},\quad i\in[n],$$
which are defined in the proof of Lemma \ref{lemma_Psik_Psi0} in Section \ref{subsection_activation_matrices}.\\
Since
$$\mathbb I\{\mathbb I_{r,i}(k)\not=\mathbb I_{r,i}(0)\}\leqslant\mathbb I\{A_{i,r}\}+\mathbb I\{\|\boldsymbol w_r(k)-\boldsymbol w_r(0)\|_2\leqslant R\},\,\forall i\in[n],r\in[m],\forall k\geqslant0.$$
we see that if $\|\boldsymbol w_r(k)-\boldsymbol w_r(0)\|_2\leqslant R$ for any $k\in\mathbb Z^{+}$, then $r\in S_i$ means $\mathbb I_{r,i}(k)=\mathbb I_{r,i}(0)$ for any $k\in\mathbb Z^{+}$ (In other words, all neurons $\boldsymbol w_r$ of $r\in S_i$ will not change activation pattern on data-point $\boldsymbol x_i$ during gradient descent training), thus $\mathbb I_{r\in\bar{S}_i}\mathbb I_{r\in\bar{S}_j}=0$ implies $r\in S_i$ or $r\in S_j$, and hence $[\mathbb I_{r,i}(k)-\mathbb I_{r,i}(0)][\mathbb I_{r,j}(k)-\mathbb I_{r,j}(0)]=0$.
Therefore, for $i\not=j\in[n]$, since $\widetilde{\boldsymbol\psi_i}(k)=\mathbb I_{r,i}(k)$, $\widetilde{\boldsymbol\psi_j}(k)=\mathbb I_{r,j}(k)$, we have
\begin{align}\nonumber
&\langle\widetilde{\boldsymbol\psi_i}(k)-\widetilde{\boldsymbol\psi_i}(0),\widetilde{\boldsymbol\psi_j}(k)-\widetilde{\boldsymbol\psi_j}(0)\rangle\\\nonumber
=&\sum_{r=1}^m[\mathbb I_{r,i}(k)-\mathbb I_{r,i}(0)][\mathbb I_{r,j}(k)-\mathbb I_{r,j}(0)]\\\label{inner_prod_variation_Psi_1}
\stackrel{\text{(i)}}\leqslant&\sum_{r=1}^m\mathbb I_{r\in\bar{S}_i}\mathbb I_{r\in\bar{S}_j}=\sum_{r=1}^m\underbrace{\mathbb I\{r\in\bar{S}_i\,\text{and}\,r\in\bar{S}_j\}}_{\varsigma_{ij}^{(r)}}\triangleq\sum_{r=1}^m\varsigma_{ij}^{(r)}.
\end{align}
where (i) is due to the fact that $[\mathbb I_{r,i}(k)-\mathbb I_{r,i}(0)][\mathbb I_{r,j}(k)-\mathbb I_{r,j}(0)]\leqslant\mathbb I_{r\in\bar{S}_i}\mathbb I_{r\in\bar{S}_j}$.\\
We decompose $\sum_{r=1}^m\varsigma_{ij}^{(r)}$ into its expectation and deviation parts:
\begin{align}\label{inner_prod_variation_Psi_2}
\sum_{r=1}^m\varsigma_{ij}^{(r)}=\sum_{r=1}^m\mathbb E\mathbb \varsigma_{ij}^{(r)}+\sum_{r=1}^m\left(\varsigma_{ij}^{(r)}-\mathbb E\mathbb \varsigma_{ij}^{(r)}\right).
\end{align}
Since $\sum_{r=1}^m\left(\varsigma_{ij}^{(r)}-\mathbb E\mathbb \varsigma_{ij}^{(r)}\right)$ is a sum of i.i.d. random variables with very small variances, we can obtain a tight upper bound of $\sum_{r=1}^m\left(\varsigma_{ij}^{(r)}-\mathbb E\mathbb \varsigma_{ij}^{(r)}\right)$ via Bernstein inequality.
Let
\begin{align}\nonumber
B=&\max_{1\leqslant r\leqslant m}|\varsigma_{ij}^{(r)}-\mathbb E\mathbb \varsigma_{ij}^{(r)}|\leqslant1,\\\nonumber
V=&\sum_{r=1}^m\mathbb E[(\varsigma_{ij}^{(r)}-\mathbb E\mathbb \varsigma_{ij}^{(r)})^2]=\sum_{r=1}^m\left\{\mathbb E[(\varsigma_{ij}^{(r)})^2]
-[\mathbb E\mathbb\varsigma_{ij}^{(r)}]^2\right\}\\\nonumber
=&\sum_{r=1}^m\left\{\mathbb E[\mathbb\varsigma_{ij}^{(r)}]
-[\mathbb E\mathbb\varsigma_{ij}^{(r)}]^2\right\}=\eth_{ij}(1-\eth_{ij})m.
\end{align}
Here we define $\eth_{ij}\triangleq\mathbb E[\mathbb\varsigma_{ij}^{(r)}]$ for $r\in[m]$ for notation simplicity with the fact that $\varsigma_{ij}^{(r)}(r\in[m])$ are i.i.d. random variables. Applying {\textcolor{red}{Bernstein inequality} over i.i.d. random variables $\varsigma_{ij}^{(r)}-\mathbb E\mathbb \varsigma_{ij}^{(r)}(r\in[m])$, we have for any $t>0$,
$$\mathbb P\left[\sum_{r=1}^m\left(\varsigma_{ij}^{(r)}-\mathbb E\mathbb \varsigma_{ij}^{(r)}\right)\geqslant t\right]\leqslant\exp\left(-\frac{t^2/2}{V+Bt/3}\right),$$
Setting $\delta=\exp[-(t^2/2)/(V+Bt/3)]$, we have
$$\mathbb P\left[\sum_{r=1}^m\left(\varsigma_{ij}^{(r)}-\mathbb E\mathbb \varsigma_{ij}^{(r)}\right)\geqslant \frac{B}{3}\log\frac{1}{\delta}\pm\sqrt{\frac{B^2}{9}\log^2\frac{1}{\delta}+2V\log\frac{1}{\delta}}\right]\leqslant\delta,$$
Since
$$\frac{B}{3}\log\frac{1}{\delta}\pm\sqrt{\frac{B^2}{9}\log^2\frac{1}{\delta}+2V\log\frac{1}{\delta}}\leqslant\frac{2B}{3}\log\frac{1}{\delta}+\sqrt{2V\log\frac{1}{\delta}},$$
we have
$$\mathbb P\left[\sum_{r=1}^m\left(\varsigma_{ij}^{(r)}-\mathbb E\mathbb \varsigma_{ij}^{(r)}\right)\geqslant\frac{2B}{3}\log\frac{1}{\delta}+\sqrt{2V\log\frac{1}{\delta}}\right]\leqslant\delta,$$
Thus, with probability exceeding $1-\delta$,
\begin{align}\label{inner_prod_variation_Psi_3}
&\sum_{r=1}^m\left(\varsigma_{ij}^{(r)}-\mathbb E\mathbb \varsigma_{ij}^{(r)}\right)
\leqslant\frac{2B}{3}\log\frac{1}{\delta}+\sqrt{2V\log\frac{1}{\delta}}\leqslant\frac{2}{3}\log\frac{1}{\delta}+\sqrt{2\eth_{ij}(1-\eth_{ij})m\log\frac{1}{\delta}}.
\end{align}
Henceforth, combining (\ref{inner_prod_variation_Psi_1}), (\ref{inner_prod_variation_Psi_2}) and (\ref{inner_prod_variation_Psi_3}), we arrive at with probability exceeding $1-\delta$ over the random initialization,
\begin{align}\label{inner_prod_variation_Psi_4}
\langle\widetilde{\boldsymbol\psi_i}(k)-\widetilde{\boldsymbol\psi_i}(0),\widetilde{\boldsymbol\psi_j}(k)-\widetilde{\boldsymbol\psi_j}(0)\rangle\leqslant m\eth_{ij}+\frac{2}{3}\log\frac{1}{\delta}+\sqrt{2\eth_{ij}(1-\eth_{ij})m\log\frac{1}{\delta}}.
\end{align}
{\textcolor{blue}{\bf Substep 2-2: Bounding the Expectation Term $\eth_{ij}$}}\\
We are in a position to upper bound the expectation term $\eth_{ij}$, which is a very funny process with manipulating lovely integrals, geometry and (inverse) trigonometrical functions. For any $r\in[m]$, we have
\begin{align}\nonumber
&\eth_{ij}
=\mathbb E[\mathbb\varsigma_{ij}^{(r)}]\\\nonumber
=&\mathbb P(\mathbb\varsigma_{ij}^{(r)}=1)=\mathbb P(\mathbb I\{r\in\bar{S}_i\,\text{and}\,r\in\bar{S}_j\}=1)\\\nonumber
=&\mathbb P(\text{The events }A_{ir}\,\text{and}\,A_{jr}\text{ happen simultaneously})\\\nonumber
=&\mathbb P(|\boldsymbol w_r(0)^\top\boldsymbol x_i|\leqslant R\quad\text{and}\quad|\boldsymbol w_r(0)^\top\boldsymbol x_j|\leqslant R).
\end{align}
Remind that the intial weight vector $\boldsymbol w_r(0)=(w_{r,1}(0),w_{r,2}(0),\cdots,w_{r,d}(0))\in\mathbb R^d$ is i.i.d. randomly generated from ${\cal N}(0,\boldsymbol I)$, thus
\begin{align}\nonumber
&\eth_{ij}=\idotsint\limits_{\begin{subarray}{c}-\infty\\|\boldsymbol w(0)\top\boldsymbol x_i|\leqslant R\\|\boldsymbol w(0)\top\boldsymbol x_j|\leqslant R\end{subarray}}^{\begin{subarray}{c}+\infty\end{subarray}}\frac{1}{(\sqrt{2\pi})^n}\exp\left\{-\frac{1}{2}\|\boldsymbol w_r(0)\|_2^2\right\}dw_{r,1}(0)dw_{r,2}(0)\cdots dw_{r,d}(0).
\end{align}
For $i\not=j\in[n]$, recall that the geometric parameter $\theta_{ij}=\angle(\boldsymbol x_i,\boldsymbol x_j)\in[\theta^*,\pi-\theta^*]$ defined in Section \ref{section_problem}. Intuitively, larger $\min\{\theta_{ij},\pi-\theta_{ij}\}$ which means the more seperation of $\boldsymbol x_i$ and $\boldsymbol x_j$ would give rise to smaller $\eth_{ij}$. In addition, smaller $R$ would also yield smaller $\eth_{ij}$. The key idea to evaluate the integral $\eth_{ij}$ is that we first reduce the high-dimensional integral to two-dimensional integral by means of an orthogonal transformation, then we dissect the range of integration by carefully analyzing various subsets of the solution set of trigonometrical inequalities. With the help of exact expression of the range of integration, we can further obtain the asymptotic expansion of final integral $\theta_{ij}$ by classical analysis method. It is evident that there is an orthogonal matrix $\boldsymbol\Gamma$ such that $\boldsymbol\Gamma\boldsymbol x_i=(1,0,0,\cdots,0)$ and $\boldsymbol\Gamma\boldsymbol x_j=(\cos\theta_{ij},\sin\theta_{ij},0,\cdots,0)$, then $\boldsymbol\nu=\boldsymbol\Gamma\boldsymbol w(0)=(\nu_1,\nu_2,\cdots,\nu_d)\in\mathbb R^d$ leads to
\begin{align}\nonumber
\eth_{ij}=&|\det(\boldsymbol\Gamma)|\idotsint\limits_{\begin{subarray}{c}-\infty\\|(\boldsymbol\Gamma\boldsymbol w(0))^\top\boldsymbol\Gamma\boldsymbol x_i|\leqslant R\\|(\boldsymbol\Gamma\boldsymbol w(0))^\top\boldsymbol\Gamma\boldsymbol x_j|\leqslant R\end{subarray}}^{\begin{subarray}{c}+\infty\end{subarray}}\frac{1}{(\sqrt{2\pi})^n}\exp\left\{-\frac{1}{2}\|\boldsymbol\Gamma\boldsymbol\nu\|_2^2\right\}d\nu_1d\nu_2\cdots d\nu_d\\\nonumber
=&\idotsint\limits_{\begin{subarray}{c}-\infty\\|\boldsymbol\nu\top\boldsymbol\Gamma\boldsymbol x_i|\leqslant R\\|\boldsymbol\nu\top\boldsymbol\Gamma\boldsymbol x_j|\leqslant R\end{subarray}}^{\begin{subarray}{c}+\infty\end{subarray}}\frac{1}{(\sqrt{2\pi})^n}\exp\left\{-\frac{1}{2}\|\boldsymbol\nu\|_2^2\right\}d\nu_1d\nu_2\cdots d\nu_d\\\nonumber
=&\idotsint\limits_{\begin{subarray}{c}-\infty\\|\nu_1|\leqslant R\\|(\cos\theta_{ij})\nu_1+(\sin\theta_{ij})\nu_2|\leqslant R\end{subarray}}^{\begin{subarray}{c}+\infty\end{subarray}}\frac{1}{(\sqrt{2\pi})^n}\exp\left\{-\frac{1}{2}(\nu_1^2+\nu_2^2+\cdots+\nu_d^2)\right\}d\nu_1d\nu_2\cdots d\nu_d\\\nonumber
=&\frac{1}{(\sqrt{2\pi})^n}\iint\limits_{\begin{subarray}{c}|\nu_1|\leqslant R\\|(\cos\theta_{ij})\nu_1+(\sin\theta_{ij})\nu_2|\leqslant R\end{subarray}}\exp\left\{-\frac{1}{2}(\nu_1^2+\nu_2^2)\right\}d\nu_1d\nu_2\prod_{\mu=3}^d\int_{-\infty}^{+\infty}\exp\left\{-\frac{\nu_\mu^2}{2}\right\}d\nu_\mu\\\nonumber
=&\frac{1}{2\pi}\iint\limits_{\begin{subarray}{c}|\nu_1|\leqslant R\\|(\cos\theta_{ij})\nu_1+(\sin\theta_{ij})\nu_2|\leqslant R\end{subarray}}\exp\left\{-\frac{1}{2}(\nu_1^2+\nu_2^2)\right\}d\nu_1d\nu_2
\end{align}
Using the polar coordinate form of $(\nu_1,\nu_2)=(\rho\sin\alpha,\rho\cos\alpha)$ with $\rho\geqslant0$ and $\alpha\in[0,2\pi)$, one finds that $|(\cos\theta_{ij})\nu_1+(\sin\theta_{ij})\nu_2|\leqslant R$ is equivalent to $\rho|\sin(\alpha+\theta_{ij})|\leqslant R$. If $\rho\leqslant R$, then $\rho|\sin\alpha|\leqslant\rho\leqslant R$ and $\rho|\sin(\alpha+\theta_{ij})|\leqslant\rho\leqslant R$ hold simultaneously for any $\alpha\in[0,2\pi)$, giving rise to
\begin{align}\nonumber
\eth_{ij}=&\frac{1}{2\pi}\iint\limits_{\begin{subarray}{c}\rho|\sin\alpha|\leqslant R\\\rho|\sin(\alpha+\theta_{ij})|\leqslant R\end{subarray}}\rho e^{-\rho^2/2}d\rho d\alpha\\\nonumber
\stackrel{\text{(i)}}=&\frac{1}{2\pi}\int_0^R\left(\int_0^{2\pi}d\alpha\right)\rho e^{-\rho^2/2}d\rho+\frac{1}{2\pi}\int_R^{+\infty}\left(\int_{\aleph(\theta_{ij})}d\alpha\right)\rho e^{-\rho^2/2}d\rho \\\label{eth_ij_1}
=&1-e^{-R^2/2}+\frac{1}{2\pi}\int_R^{+\infty}\left(\int_{\aleph(\theta_{ij})}d\alpha\right)\rho e^{-\rho^2/2}d\rho 
\end{align}
where $\aleph(\theta_{ij})$ in (i) denotes the range of integration with respect to $\alpha$. More explicitly, 
$$\aleph(\theta_{ij})=\{\alpha||\sin\alpha|<R/\rho\quad\text{and}\quad|\sin(\alpha+\theta_{ij})|<R/\rho,\quad\alpha\in[0,2\pi)\}.$$
Notice that we do not include the set $S_{\alpha0}=\{\alpha||\sin\alpha|=R/\rho\quad\text{or}\quad|\sin(\alpha+\theta_{ij})|=R/\rho,\quad\alpha\in[0,2\pi)\}$ into the set $\aleph(\theta_{ij})$, since $S_{\alpha0}$ is a set of Lebesgue measure zero in $\mathbb R$ and thus does no affect the value of the integral term in (\ref{eth_ij_1}). In addition, our definition of $\aleph(\theta_{ij})$ makes further analysis more convenient.\\
{\textcolor{blue}{\bf Subsubstep 2-2-1: Analyze the Integral Domain $\aleph(\theta_{ij})$}}\\
We now proceed to analyze the integral domain $\aleph(\theta_{ij})$ carefully, which plays a key role in evaluating the integral term in (\ref{eth_ij_1}).\\
In the case of $\rho>R$, let $\beta=\arcsin(R/\rho)\in(0,\pi/2)$ and solve the trigonometrical inequalities, we obtain the following structure of solution sets.\\
{\textcolor{red}{1.} The solution set of $|\sin\alpha|<R/\rho$ with $\alpha\in[0,2\pi)$ is
\begin{align}\nonumber
&\{\alpha||\sin\alpha|<R/\rho,\alpha\in[0,2\pi)\}=\{\alpha|-R/\rho<\sin\alpha<R/\rho,\alpha\in[0,2\pi)\}\\\nonumber
=&\{\alpha|2k\pi\leqslant\alpha<2k\pi+\beta,k\in\mathbb Z,\alpha\in[0,2\pi)\}\\\nonumber
\cup&\{\alpha|(2k+1)\pi-\beta<\alpha<(2k+1)\pi+\beta,k\in\mathbb Z,\alpha\in[0,2\pi)\}\\\nonumber
\cup&\{\alpha|(2k+2)\pi-\beta<\alpha\leqslant(2k+2)\pi,k\in\mathbb Z,\alpha\in[0,2\pi)\}\\\nonumber
=&\{\alpha|0\leqslant\alpha<\beta\}
\cup\{\alpha|\pi-\beta<\alpha<\pi+\beta\}\cup\{\alpha|2\pi-\beta<\alpha<2\pi\}\\\nonumber
=&[0,\beta)
\cup(\pi-\beta,\pi+\beta)\cup(2\pi-\beta,2\pi)=\mathfrak A_1\cup\mathfrak A_2\cup\mathfrak A_3.
\end{align}
Notice that the interval lengths of $\mathfrak A_1$, $\mathfrak A_2$ and $\mathfrak A_2$ are $\beta,2\beta$ and $\beta$, respectively.\\
In addition, the total interval length of $\{\alpha||\sin\alpha|<R/\rho,\alpha\in[0,2\pi)\}$ is $4\beta$.\\\\
{\textcolor{red}{2.} The solution set of $|\sin(\alpha+\theta_{ij})|<R/\rho$ with $\alpha\in[0,2\pi)$ is
\begin{align}\nonumber
&\{\alpha||\sin(\alpha+\theta_{ij})|<R/\rho,\alpha\in[0,2\pi)\}=\{\alpha|-R/\rho<\sin(\alpha+\theta_{ij})<R/\rho,\alpha\in[0,2\pi)\}\\\nonumber
=&\{\alpha|2k\pi\leqslant\alpha+\theta_{ij}<2k\pi+\beta,k\in\mathbb Z,\alpha\in[0,2\pi)\}\\\nonumber
\cup&\{\alpha|(2k+1)\pi-\beta<\alpha+\theta_{ij}<(2k+1)\pi+\beta,k\in\mathbb Z,\alpha\in[0,2\pi)\}\\\nonumber
\cup&\{\alpha|(2k+2)\pi-\beta<\alpha+\theta_{ij}\leqslant(2k+2)\pi,k\in\mathbb Z,\alpha\in[0,2\pi)\}=\mathfrak B_1\cup\mathfrak B_2\cup\mathfrak B_3.
\end{align}
{\textcolor{blue}{(2.1)}
\begin{align}\nonumber
&\mathfrak B_1=\{\alpha|2k\pi\leqslant\alpha+\theta_{ij}<2k\pi+\beta,k\in\mathbb Z,\alpha\in[0,2\pi)\}\\\nonumber
=&\{\alpha|2k\pi\leqslant\alpha+\theta_{ij}<2k\pi+\beta,k\leqslant-1,k\in\mathbb Z,\alpha\in[0,2\pi)\}\\\nonumber
\cup&\{\alpha|0\leqslant\alpha<\beta-\theta_{ij},\alpha\in[0,2\pi)\}\cup\{\alpha|2\pi-\theta_{ij}<\alpha<2\pi+\beta-\theta_{ij},\alpha\in[0,2\pi)\}\\\nonumber
\cup&\{\alpha|2k\pi\leqslant\alpha+\theta_{ij}<2k\pi+\beta,k\geqslant2,k\in\mathbb Z,\alpha\in[0,2\pi)\}\\\nonumber
=&\{\alpha|0\leqslant\alpha<\beta-\theta_{ij},\alpha\in[0,2\pi)\}\cup\{\alpha|2\pi-\theta_{ij}<\alpha<2\pi+\beta-\theta_{ij},\alpha\in[0,2\pi)\}.
\end{align}
If $\beta\leqslant\theta_{ij}$, then $\{\alpha|0\leqslant\alpha<\beta-\theta_{ij},\alpha\in[0,2\pi)\}=\emptyset$ and $\{\alpha|2\pi-\theta_{ij}<\alpha<2\pi+\beta-\theta_{ij},\alpha\in[0,2\pi)\}=(2\pi-\theta_{ij},2\pi+\beta-\theta_{ij})$. Thus $$\mathfrak B_1=(2\pi-\theta_{ij},2\pi+\beta-\theta_{ij})\quad\text{for}\quad\beta\leqslant\theta_{ij};$$
If $\beta>\theta_{ij}$, $\{\alpha|0\leqslant\alpha<\beta-\theta_{ij},\alpha\in[0,2\pi)\}=[0,\beta-\theta_{ij})$ and $\{\alpha|2\pi-\theta_{ij}<\alpha<2\pi+\beta-\theta_{ij},\alpha\in[0,2\pi)\}=(2\pi-\theta_{ij},2\pi)$. Thus 
$$\mathfrak B_1=[0,\beta-\theta_{ij})\cup(2\pi-\theta_{ij},2\pi)\quad\text{for}\quad\beta>\theta_{ij}.$$
Notice that the total interval length of $\mathfrak B_1$ is always $\beta$ in any cases.\\
{\textcolor{blue}{(2.2)}
\begin{align}\nonumber
&\mathfrak B_2=\{\alpha|(2k+1)\pi-\beta<\alpha+\theta_{ij}<(2k+1)\pi+\beta,k\in\mathbb Z,\alpha\in[0,2\pi)\}\\\nonumber
=&\{\alpha|(2k+1)\pi-\beta<\alpha+\theta_{ij}<(2k+1)\pi+\beta,k\leqslant-1,k\in\mathbb Z,\alpha\in[0,2\pi)\}\\\nonumber
\cup&\{\alpha|\pi-\theta_{ij}-\beta<\alpha<\pi-\theta_{ij}+\beta,\alpha\in[0,2\pi)\}
\cup\{\alpha|3\pi-\theta_{ij}-\beta<\alpha<3\pi-\theta_{ij}+\beta,\alpha\in[0,2\pi)\}\\\nonumber
\cup&\{\alpha|(2k+1)\pi-\beta<\alpha+\theta_{ij}<(2k+1)\pi+\beta,k\geqslant2,k\in\mathbb Z,\alpha\in[0,2\pi)\}\\\nonumber
=&\{\alpha|\pi-\theta_{ij}-\beta<\alpha<\pi-\theta_{ij}+\beta,\alpha\in[0,2\pi)\}
\cup\{\alpha|3\pi-\theta_{ij}-\beta<\alpha<3\pi-\theta_{ij}+\beta,\alpha\in[0,2\pi)\}.
\end{align}
If $\beta+\theta_{ij}\leqslant\pi$, then $\{\alpha|\pi-\theta_{ij}-\beta<\alpha<\pi-\theta_{ij}+\beta,\alpha\in[0,2\pi)\}=(\pi-\theta_{ij}-\beta,\pi-\theta_{ij}+\beta)$ and $\{\alpha|3\pi-\theta_{ij}-\beta<\alpha<3\pi-\theta_{ij}+\beta,\alpha\in[0,2\pi)\}=\emptyset$. Thus
$$\mathfrak B_2=(\pi-\theta_{ij}-\beta,\pi-\theta_{ij}+\beta)\quad\text{for}\quad\beta+\theta_{ij}\leqslant\pi;$$
If $\beta+\theta_{ij}>\pi$, then $\{\alpha|\pi-\theta_{ij}-\beta<\alpha<\pi-\theta_{ij}+\beta,\alpha\in[0,2\pi)\}=(0,\pi-\theta_{ij}+\beta)$ and $\{\alpha|3\pi-\theta_{ij}-\beta<\alpha<3\pi-\theta_{ij}+\beta,\alpha\in[0,2\pi)\}=(3\pi-\theta_{ij}-\beta,2\pi)$. Thus
$$\mathfrak B_2=(0,\pi-\theta_{ij}+\beta)\cup(3\pi-\theta_{ij}-\beta,2\pi)\quad\text{for}\quad\beta+\theta_{ij}>\pi.$$
Notice that the total interval length of $\mathfrak B_2$ is always $2\beta$ in any cases.\\
{\textcolor{blue}{(2.3)}
\begin{align}\nonumber
&\mathfrak B_3=\{\alpha|(2k+2)\pi-\beta<\alpha+\theta_{ij}\leqslant(2k+2)\pi,k\in\mathbb Z,\alpha\in[0,2\pi)\}\\\nonumber
=&\{\alpha|(2k+2)\pi-\beta<\alpha+\theta_{ij}\leqslant(2k+2)\pi,k\leqslant-1,k\in\mathbb Z,\alpha\in[0,2\pi)\}\\\nonumber
\cup&\{\alpha|2\pi-\beta-\theta_{ij}<\alpha\leqslant2\pi-\theta_{ij},\alpha\in[0,2\pi)\}\\\nonumber
\cup&\{\alpha|(2k+2)\pi-\beta<\alpha+\theta_{ij}\leqslant(2k+2)\pi,k\geqslant1,k\in\mathbb Z,\alpha\in[0,2\pi)\}\\\nonumber
=&\{\alpha|2\pi-\beta-\theta_{ij}<\alpha\leqslant2\pi-\theta_{ij},\alpha\in[0,2\pi)\}\\\nonumber
=&(2\pi-\beta-\theta_{ij},2\pi-\theta_{ij}].
\end{align}
Notice that the interval length of $\mathfrak B_3$ is $\beta$.\\
In addition, the total interval length of $\{\alpha||\sin\alpha+\theta_{ij}|<R/\rho,\alpha\in[0,2\pi)\}$ is $4\beta$.\\
To proceed, we analyze the concrete form of $\aleph(\theta_{ij})$ and its total length $|\aleph(\theta_{ij})|$ according to the following eight various cases:
\begin{align}\nonumber
\left\{\begin{array}{ll}
       \text{I}:R<\rho\leqslant\sqrt{2}R\text{ or }\pi/4\leqslant\beta<\pi/2\left\{ \begin{array}{ll}
                                       \text{I-a}:0<\theta_{ij}\leqslant\pi/2\left\{ \begin{array}{ll}
                                       \text{I-a-1}:\beta\leqslant\theta_{ij} \\
                                       \text{I-a-2}:\beta>\theta_{ij}
                                \end{array} \right. \\
                                       \text{I-b}:\pi/2<\theta_{ij}\leqslant\pi\left\{ \begin{array}{ll}
                                       \text{I-b-1}:\beta+\theta_{ij}\leqslant\pi \\
                                       \text{I-b-2}:\beta+\theta_{ij}>\pi
                                \end{array} \right.
                                \end{array} \right.\\
       \text{II}:\rho>\sqrt{2}R\text{ or }0<\beta<\pi/4\left\{ \begin{array}{ll}
                                       \text{II-a}:0<\theta_{ij}\leqslant\pi/2\left\{ \begin{array}{ll}
                                       \text{II-a-1}:\beta\leqslant\theta_{ij} \\
                                       \text{II-a-2}:\beta>\theta_{ij}
                                \end{array} \right. \\
                                       \text{II-b}:\pi/2<\theta_{ij}\leqslant\pi\left\{ \begin{array}{ll}
                                       \text{II-b-1}:\beta+\theta_{ij}\leqslant\pi \\
                                       \text{II-b-2}:\beta+\theta_{ij}>\pi
                                \end{array} \right.
                                \end{array} \right.
       \end{array} \right.
\end{align}
\\{\textcolor{red}{(I).} When $R<\rho\leqslant\sqrt{2}R$ or $\pi/4\leqslant\beta<\pi/2$.\\\\
{\textcolor{blue}{(I-a)} If $0<\theta_{ij}\leqslant\pi/2$, then $\beta+\theta_{ij}<\pi$.\\\\
{\textcolor{orange}{(I-a-1)} If $\beta\leqslant\theta_{ij}$, then
\begin{align}\nonumber
&\{\alpha||\sin(\alpha+\theta_{ij})|<R/\rho,\alpha\in[0,2\pi)\}=\mathfrak B_1\cup\mathfrak B_2\cup\mathfrak B_3\\\nonumber
=&(2\pi-\theta_{ij},2\pi+\beta-\theta_{ij})\cup(\pi-\theta_{ij}-\beta,\pi-\theta_{ij}+\beta)\cup(2\pi-\beta-\theta_{ij},2\pi-\theta_{ij}].
\end{align}
\begin{align}\nonumber
&\aleph(\theta_{ij})=\{\alpha||\sin\alpha|<R/\rho\quad\text{and}\quad|\sin(\alpha+\theta_{ij})|<R/\rho,\quad\alpha\in[0,2\pi)\}\\\nonumber
=&(\mathfrak A_1\cup\mathfrak A_2\cup\mathfrak A_3)\cap(\mathfrak B_1\cup\mathfrak B_2\cup\mathfrak B_3)\\\nonumber
=&[\mathfrak A_1\cap(\mathfrak B_1\cup\mathfrak B_2\cup\mathfrak B_3)]\cup[\mathfrak A_2\cap(\mathfrak B_1\cup\mathfrak B_2\cup\mathfrak B_3)]\cup[\mathfrak A_3\cap(\mathfrak B_1\cup\mathfrak B_2\cup\mathfrak B_3)]\\\nonumber
=&\{[0,\beta)\cap(\mathfrak B_1\cup\mathfrak B_2\cup\mathfrak B_3)\}
\cup\{(\pi-\beta,\pi+\beta)\cap(\mathfrak B_1\cup\mathfrak B_2\cup\mathfrak B_3)\}\cup\{(2\pi-\beta,2\pi)\cap(\mathfrak B_1\cup\mathfrak B_2\cup\mathfrak B_3)\}\\\nonumber
=&\{[0,\beta)\cap(\pi-\theta_{ij}-\beta,\pi-\theta_{ij}+\beta)\}\\\nonumber
&\cup\{[(\pi-\beta,\pi+\beta)\cap(\pi-\theta_{ij}-\beta,\pi-\theta_{ij}+\beta)]\cup[(\pi-\beta,\pi+\beta)\cap(2\pi-\beta-\theta_{ij},2\pi-\theta_{ij}]]\}\\\nonumber
&\cup\{(2\pi-\beta,2\pi)\cap(2\pi-\theta_{ij},2\pi+\beta-\theta_{ij})\}.
\end{align}
It is evident that
\begin{align}\nonumber
&|[0,\beta)\cap(\pi-\theta_{ij}-\beta,\pi-\theta_{ij}+\beta)|=\max\{2\beta+\theta_{ij}-\pi,0\},\\\nonumber
&|(\pi-\beta,\pi+\beta)\cap(\pi-\theta_{ij}-\beta,\pi-\theta_{ij}+\beta)|=2\beta-\theta_{ij},\\\nonumber
&|
(\pi-\beta,\pi+\beta)\cap(2\pi-\beta-\theta_{ij},2\pi-\theta_{ij}]|=\max\{2\beta+\theta_{ij}-\pi,0\},\\\nonumber
&|(2\pi-\beta,2\pi)\cap(2\pi-\theta_{ij},2\pi+\beta-\theta_{ij})|=2\beta-\theta_{ij}.
\end{align}
Therefore, the total interval length of $\aleph(\theta_{ij})$ in the case of (I-a-1) is
\begin{align}\label{aleph_I_a_1}
|\aleph(\theta_{ij})|=4\beta-2\theta_{ij}+2\max\{2\beta+\theta_{ij}-\pi,0\}.
\end{align}
{\textcolor{orange}{(I-a-2)} If $\beta>\theta_{ij}$, then
\begin{align}\nonumber
&\{\alpha||\sin(\alpha+\theta_{ij})|<R/\rho,\alpha\in[0,2\pi)\}=\mathfrak B_1\cup\mathfrak B_2\cup\mathfrak B_3\\\nonumber
=&[0,\beta-\theta_{ij})\cup(2\pi-\theta_{ij},2\pi)\cup(\pi-\theta_{ij}-\beta,\pi-\theta_{ij}+\beta)\cup(2\pi-\beta-\theta_{ij},2\pi-\theta_{ij}].
\end{align}
\begin{align}\nonumber
&\aleph(\theta_{ij})=\{\alpha||\sin\alpha|<R/\rho\quad\text{and}\quad|\sin(\alpha+\theta_{ij})|<R/\rho,\quad\alpha\in[0,2\pi)\}\\\nonumber
=&(\mathfrak A_1\cup\mathfrak A_2\cup\mathfrak A_3)\cap(\mathfrak B_1\cup\mathfrak B_2\cup\mathfrak B_3)\\\nonumber
=&[\mathfrak A_1\cap(\mathfrak B_1\cup\mathfrak B_2\cup\mathfrak B_3)]\cup[\mathfrak A_2\cap(\mathfrak B_1\cup\mathfrak B_2\cup\mathfrak B_3)]\cup[\mathfrak A_3\cap(\mathfrak B_1\cup\mathfrak B_2\cup\mathfrak B_3)]\\\nonumber
=&\{[0,\beta)\cap(\mathfrak B_1\cup\mathfrak B_2\cup\mathfrak B_3)\}
\cup\{(\pi-\beta,\pi+\beta)\cap(\mathfrak B_1\cup\mathfrak B_2\cup\mathfrak B_3)\}\cup\{(2\pi-\beta,2\pi)\cap(\mathfrak B_1\cup\mathfrak B_2\cup\mathfrak B_3)\}\\\nonumber
=&\{[[0,\beta)\cap(0,\beta-\theta_{ij})]\cup[[0,\beta)\cap(\pi-\theta_{ij}-\beta,\pi-\theta_{ij}+\beta)]\}\\\nonumber
&\cup\{[(\pi-\beta,\pi+\beta)\cap(\pi-\theta_{ij}-\beta,\pi-\theta_{ij}+\beta)]\cup[(\pi-\beta,\pi+\beta)\cap(2\pi-\beta-\theta_{ij},2\pi-\theta_{ij}]]\}\\\nonumber
&\cup\{[(2\pi-\beta,2\pi)\cap(2\pi-\theta_{ij},2\pi)\cup[(2\pi-\beta,2\pi)\cap(2\pi-\beta-\theta_{ij},2\pi-\theta_{ij}]]\}.
\end{align}
It is easy to see that
\begin{align}\nonumber
&|[0,\beta)\cap(0,\beta-\theta_{ij})|=\theta_{ij},\\\nonumber
&|[0,\beta)\cap(\pi-\theta_{ij}-\beta,\pi-\theta_{ij}+\beta)|=\max\{2\beta+\theta_{ij}-\pi,0\},\\\nonumber
&|(\pi-\beta,\pi+\beta)\cap(\pi-\theta_{ij}-\beta,\pi-\theta_{ij}+\beta)|=2\beta-\theta_{ij},\\\nonumber
&|(\pi-\beta,\pi+\beta)\cap(2\pi-\beta-\theta_{ij},2\pi-\theta_{ij}]|=\max\{2\beta+\theta_{ij}-\pi,0\},\\\nonumber
&|(2\pi-\beta,2\pi)\cap(2\pi-\theta_{ij},2\pi)|=\beta-\theta_{ij},\\\nonumber
&|(2\pi-\beta,2\pi)\cap(2\pi-\beta-\theta_{ij},2\pi-\theta_{ij}]|=\beta-\theta_{ij}.
\end{align}
Therefore, the total interval length of $\aleph(\theta_{ij})$ in the case of (I-a-2) is
\begin{align}\label{aleph_I_a_2}
|\aleph(\theta_{ij})|=4\beta-2\theta_{ij}+2\max\{2\beta+\theta_{ij}-\pi,0\}.
\end{align}
{\textcolor{blue}{(I-b)} If $\pi/2<\theta_{ij}<\pi$, then $\beta<\theta_{ij}$.\\\\
{\textcolor{orange}{(I-b-1)} If $\beta+\theta_{ij}\leqslant\pi$, then
\begin{align}\nonumber
&\{\alpha||\sin(\alpha+\theta_{ij})|<R/\rho,\alpha\in[0,2\pi)\}=\mathfrak B_1\cup\mathfrak B_2\cup\mathfrak B_3\\\nonumber
=&(2\pi-\theta_{ij},2\pi+\beta-\theta_{ij})\cup(\pi-\theta_{ij}-\beta,\pi-\theta_{ij}+\beta)\cup(2\pi-\beta-\theta_{ij},2\pi-\theta_{ij}].
\end{align}
\begin{align}\nonumber
&\aleph(\theta_{ij})=\{\alpha||\sin\alpha|<R/\rho\quad\text{and}\quad|\sin(\alpha+\theta_{ij})|<R/\rho,\quad\alpha\in[0,2\pi)\}\\\nonumber
=&(\mathfrak A_1\cup\mathfrak A_2\cup\mathfrak A_3)\cap(\mathfrak B_1\cup\mathfrak B_2\cup\mathfrak B_3)\\\nonumber
=&[\mathfrak A_1\cap(\mathfrak B_1\cup\mathfrak B_2\cup\mathfrak B_3)]\cup[\mathfrak A_2\cap(\mathfrak B_1\cup\mathfrak B_2\cup\mathfrak B_3)]\cup[\mathfrak A_3\cap(\mathfrak B_1\cup\mathfrak B_2\cup\mathfrak B_3)]\\\nonumber
=&\{[0,\beta)\cap(\mathfrak B_1\cup\mathfrak B_2\cup\mathfrak B_3)\}
\cup\{(\pi-\beta,\pi+\beta)\cap(\mathfrak B_1\cup\mathfrak B_2\cup\mathfrak B_3)\}\cup\{(2\pi-\beta,2\pi)\cap(\mathfrak B_1\cup\mathfrak B_2\cup\mathfrak B_3)\}\\\nonumber
=&\{[0,\beta)\cap(\pi-\theta_{ij}-\beta,\pi-\theta_{ij}+\beta)\}\\\nonumber
&\cup\{[(\pi-\beta,\pi+\beta)\cap(\pi-\theta_{ij}-\beta,\pi-\theta_{ij}+\beta)]\cup[(\pi-\beta,\pi+\beta)\cap(2\pi-\beta-\theta_{ij},2\pi-\theta_{ij}]]\}\\\nonumber
&\cup\{(2\pi-\beta,2\pi)\cap(2\pi-\theta_{ij},2\pi+\beta-\theta_{ij})\}.
\end{align}
It is easy to see that
\begin{align}\nonumber
&|[0,\beta)\cap(\pi-\theta_{ij}-\beta,\pi-\theta_{ij}+\beta)|=2\beta+\theta_{ij}-\pi,\\\nonumber
&|(\pi-\beta,\pi+\beta)\cap(\pi-\theta_{ij}-\beta,\pi-\theta_{ij}+\beta)|=\max\{2\beta-\theta_{ij},0\},\\\nonumber
&|(\pi-\beta,\pi+\beta)\cap(2\pi-\beta-\theta_{ij},2\pi-\theta_{ij}]|=2\beta+\theta_{ij}-\pi,\\\nonumber
&|(2\pi-\beta,2\pi)\cap(2\pi-\theta_{ij},2\pi+\beta-\theta_{ij})|=\max\{2\beta-\theta_{ij},0\}.
\end{align}
Therefore, the total interval length of $\aleph(\theta_{ij})$ in the case of (I-b-1) is
\begin{align}\label{aleph_I_b_1}
|\aleph(\theta_{ij})|=4\beta+2\theta_{ij}-2\pi+2\max\{2\beta-\theta_{ij},0\}.
\end{align}
{\textcolor{orange}{(I-b-2)} If $\beta+\theta_{ij}>\pi$, then
\begin{align}\nonumber
&\{\alpha||\sin(\alpha+\theta_{ij})|<R/\rho,\alpha\in[0,2\pi)\}=\mathfrak B_1\cup\mathfrak B_2\cup\mathfrak B_3\\\nonumber
=&(2\pi-\theta_{ij},2\pi+\beta-\theta_{ij})\cup(0,\pi-\theta_{ij}+\beta)\cup(3\pi-\theta_{ij}-\beta,2\pi)\cup(2\pi-\beta-\theta_{ij},2\pi-\theta_{ij}].
\end{align}
\begin{align}\nonumber
&\aleph(\theta_{ij})=\{\alpha||\sin\alpha|<R/\rho\quad\text{and}\quad|\sin(\alpha+\theta_{ij})|<R/\rho,\quad\alpha\in[0,2\pi)\}\\\nonumber
=&(\mathfrak A_1\cup\mathfrak A_2\cup\mathfrak A_3)\cap(\mathfrak B_1\cup\mathfrak B_2\cup\mathfrak B_3)\\\nonumber
=&[\mathfrak A_1\cap(\mathfrak B_1\cup\mathfrak B_2\cup\mathfrak B_3)]\cup[\mathfrak A_2\cap(\mathfrak B_1\cup\mathfrak B_2\cup\mathfrak B_3)]\cup[\mathfrak A_3\cap(\mathfrak B_1\cup\mathfrak B_2\cup\mathfrak B_3)]\\\nonumber
=&\{[0,\beta)\cap(\mathfrak B_1\cup\mathfrak B_2\cup\mathfrak B_3)\}
\cup\{(\pi-\beta,\pi+\beta)\cap(\mathfrak B_1\cup\mathfrak B_2\cup\mathfrak B_3)\}\cup\{(2\pi-\beta,2\pi)\cap(\mathfrak B_1\cup\mathfrak B_2\cup\mathfrak B_3)\}\\\nonumber
=&\{[0,\beta)\cap(0,\pi-\theta_{ij}+\beta)\}\\\nonumber
&\cup\{[(\pi-\beta,\pi+\beta)\cap(2\pi-\theta_{ij},2\pi+\beta-\theta_{ij})]\cup[(\pi-\beta,\pi+\beta)\cap(0,\pi-\theta_{ij}+\beta)]\\\nonumber
&\cup[(\pi-\beta,\pi+\beta)\cap(2\pi-\beta-\theta_{ij},2\pi-\theta_{ij}]]\}\\\nonumber
&\cup\{[(2\pi-\beta,2\pi)\cap(2\pi-\theta_{ij},2\pi+\beta-\theta_{ij})]\cup
[(2\pi-\beta,2\pi)\cap(3\pi-\theta_{ij}-\beta,2\pi)]\}.
\end{align}
It is easy to see that
\begin{align}\nonumber
&|[0,\beta)\cap(0,\pi-\theta_{ij}+\beta)|=\beta,\\\nonumber
&|(\pi-\beta,\pi+\beta)\cap(2\pi-\theta_{ij},2\pi+\beta-\theta_{ij})|=\beta+\theta_{ij}-\pi,\\\nonumber
&|(\pi-\beta,\pi+\beta)\cap(0,\pi-\theta_{ij}+\beta)|=\max\{2\beta-\theta_{ij},0\},\\\nonumber
&|(\pi-\beta,\pi+\beta)\cap(2\pi-\beta-\theta_{ij},2\pi-\theta_{ij}]|=\beta,\\\nonumber
&|(2\pi-\beta,2\pi)\cap(2\pi-\theta_{ij},2\pi+\beta-\theta_{ij})|=\max\{2\beta-\theta_{ij},0\},\\\nonumber
&|(2\pi-\beta,2\pi)\cap(3\pi-\theta_{ij}-\beta,2\pi)|=\beta+\theta_{ij}-\pi.
\end{align}
Therefore, the total interval length of $\aleph(\theta_{ij})$ in the case of (I-b-2) is
\begin{align}\label{aleph_I_b_2}
|\aleph(\theta_{ij})|=4\beta+2\theta_{ij}-2\pi+2\max\{2\beta-\theta_{ij},0\}.
\end{align}
\\{\textcolor{red}{(II).} When $\rho>\sqrt{2}R$ or $0<\beta<\pi/4$, we proceed by investigating two cases.\\\\
{\textcolor{blue}{(II-a)} If $0<\theta_{ij}\leqslant\pi/2$, then $\beta+\theta_{ij}<\pi$.\\\\
{\textcolor{orange}{(II-a-1)} If $\beta\leqslant\theta_{ij}$, then
\begin{align}\nonumber
&\{\alpha||\sin(\alpha+\theta_{ij})|<R/\rho,\alpha\in[0,2\pi)\}=\mathfrak B_1\cup\mathfrak B_2\cup\mathfrak B_3\\\nonumber
=&(2\pi-\theta_{ij},2\pi+\beta-\theta_{ij})\cup(\pi-\theta_{ij}-\beta,\pi-\theta_{ij}+\beta)\cup(2\pi-\beta-\theta_{ij},2\pi-\theta_{ij}].
\end{align}
\begin{align}\nonumber
&\aleph(\theta_{ij})=\{\alpha||\sin\alpha|<R/\rho\quad\text{and}\quad|\sin(\alpha+\theta_{ij})|<R/\rho,\quad\alpha\in[0,2\pi)\}\\\nonumber
=&(\mathfrak A_1\cup\mathfrak A_2\cup\mathfrak A_3)\cap(\mathfrak B_1\cup\mathfrak B_2\cup\mathfrak B_3)\\\nonumber
=&[\mathfrak A_1\cap(\mathfrak B_1\cup\mathfrak B_2\cup\mathfrak B_3)]\cup[\mathfrak A_2\cap(\mathfrak B_1\cup\mathfrak B_2\cup\mathfrak B_3)]\cup[\mathfrak A_3\cap(\mathfrak B_1\cup\mathfrak B_2\cup\mathfrak B_3)]\\\nonumber
=&\{[0,\beta)\cap(\mathfrak B_1\cup\mathfrak B_2\cup\mathfrak B_3)\}
\cup\{(\pi-\beta,\pi+\beta)\cap(\mathfrak B_1\cup\mathfrak B_2\cup\mathfrak B_3)\}\cup\{(2\pi-\beta,2\pi)\cap(\mathfrak B_1\cup\mathfrak B_2\cup\mathfrak B_3)\}\\\nonumber
=&\emptyset\cup\{(\pi-\beta,\pi+\beta)\cap(\pi-\theta_{ij}-\beta,\pi-\theta_{ij}+\beta)\}\cup\{(2\pi-\beta,2\pi)\cap(2\pi-\theta_{ij},2\pi+\beta-\theta_{ij})\}.
\end{align}
It is evident that
\begin{align}\nonumber
&|(\pi-\beta,\pi+\beta)\cap(\pi-\theta_{ij}-\beta,\pi-\theta_{ij}+\beta)|=\max\{2\beta-\theta_{ij},0\},\\\nonumber
&|(2\pi-\beta,2\pi)\cap(2\pi-\theta_{ij},2\pi+\beta-\theta_{ij})|=\max\{2\beta-\theta_{ij},0\}.
\end{align}
Therefore, the total interval length of $\aleph(\theta_{ij})$ in the case of (II-a-1) is
\begin{align}\label{aleph_II_a_1}
|\aleph(\theta_{ij})|=2\max\{2\beta-\theta_{ij},0\}.
\end{align}
{\textcolor{orange}{(II-a-2)} If $\beta>\theta_{ij}$, then
\begin{align}\nonumber
&\{\alpha||\sin(\alpha+\theta_{ij})|<R/\rho,\alpha\in[0,2\pi)\}=\mathfrak B_1\cup\mathfrak B_2\cup\mathfrak B_3\\\nonumber
=&[0,\beta-\theta_{ij})\cup(2\pi-\theta_{ij},2\pi)\cup(\pi-\theta_{ij}-\beta,\pi-\theta_{ij}+\beta)\cup(2\pi-\beta-\theta_{ij},2\pi-\theta_{ij}].
\end{align}
\begin{align}\nonumber
&\aleph(\theta_{ij})=\{\alpha||\sin\alpha|<R/\rho\quad\text{and}\quad|\sin(\alpha+\theta_{ij})|<R/\rho,\quad\alpha\in[0,2\pi)\}\\\nonumber
=&(\mathfrak A_1\cup\mathfrak A_2\cup\mathfrak A_3)\cap(\mathfrak B_1\cup\mathfrak B_2\cup\mathfrak B_3)\\\nonumber
=&[\mathfrak A_1\cap(\mathfrak B_1\cup\mathfrak B_2\cup\mathfrak B_3)]\cup[\mathfrak A_2\cap(\mathfrak B_1\cup\mathfrak B_2\cup\mathfrak B_3)]\cup[\mathfrak A_3\cap(\mathfrak B_1\cup\mathfrak B_2\cup\mathfrak B_3)]\\\nonumber
=&\{[0,\beta)\cap[0,\beta-\theta_{ij})\}\\\nonumber
&\cup\{(\pi-\beta,\pi+\beta)\cap(\pi-\theta_{ij}-\beta,\pi-\theta_{ij}+\beta)\}\\\nonumber
&\cup\{[(2\pi-\beta,2\pi)\cap(2\pi-\theta_{ij},2\pi)]\cup[(2\pi-\beta,2\pi)\cap(2\pi-\beta-\theta_{ij},2\pi-\theta_{ij}]]\}.
\end{align}
It is evident that
\begin{align}\nonumber
&|[0,\beta)\cap[0,\beta-\theta_{ij})|=\beta-\theta_{ij},\\\nonumber
&|(\pi-\beta,\pi+\beta)\cap(\pi-\theta_{ij}-\beta,\pi-\theta_{ij}+\beta)|=2\beta-\theta_{ij},\\\nonumber
&|(2\pi-\beta,2\pi)\cap(2\pi-\theta_{ij},2\pi)|=\theta_{ij},\\\nonumber
&|(2\pi-\beta,2\pi)\cap(2\pi-\beta-\theta_{ij},2\pi-\theta_{ij}]|
=\beta-\theta_{ij}.
\end{align}
Therefore, the total interval length of $\aleph(\theta_{ij})$ in the case of (II-a-2) is
\begin{align}\label{aleph_II_a_2}
|\aleph(\theta_{ij})|=4\beta-2\theta_{ij}=2\max\{2\beta-\theta_{ij},0\}\,\text{due to }\beta>\theta_{ij}.
\end{align}
{\textcolor{blue}{(II-b)} If $\pi/2<\theta_{ij}<\pi$, then $\beta<\theta_{ij}$.\\\\
{\textcolor{orange}{(II-b-1)} If $\beta+\theta_{ij}\leqslant\pi$, then
\begin{align}\nonumber
&\{\alpha||\sin(\alpha+\theta_{ij})|<R/\rho,\alpha\in[0,2\pi)\}=\mathfrak B_1\cup\mathfrak B_2\cup\mathfrak B_3\\\nonumber
=&(2\pi-\theta_{ij},2\pi+\beta-\theta_{ij})\cup(\pi-\theta_{ij}-\beta,\pi-\theta_{ij}+\beta)\cup(2\pi-\beta-\theta_{ij},2\pi-\theta_{ij}].
\end{align}
\begin{align}\nonumber
&\aleph(\theta_{ij})=\{\alpha||\sin\alpha|<R/\rho\quad\text{and}\quad|\sin(\alpha+\theta_{ij})|<R/\rho,\quad\alpha\in[0,2\pi)\}\\\nonumber
=&(\mathfrak A_1\cup\mathfrak A_2\cup\mathfrak A_3)\cap(\mathfrak B_1\cup\mathfrak B_2\cup\mathfrak B_3)\\\nonumber
=&[\mathfrak A_1\cap(\mathfrak B_1\cup\mathfrak B_2\cup\mathfrak B_3)]\cup[\mathfrak A_2\cap(\mathfrak B_1\cup\mathfrak B_2\cup\mathfrak B_3)]\cup[\mathfrak A_3\cap(\mathfrak B_1\cup\mathfrak B_2\cup\mathfrak B_3)]\\\nonumber
=&\{[0,\beta)\cap(\mathfrak B_1\cup\mathfrak B_2\cup\mathfrak B_3)\}
\cup\{(\pi-\beta,\pi+\beta)\cap(\mathfrak B_1\cup\mathfrak B_2\cup\mathfrak B_3)\}\cup\{(2\pi-\beta,2\pi)\cap(\mathfrak B_1\cup\mathfrak B_2\cup\mathfrak B_3)\}\\\nonumber
=&\{[0,\beta)\cap(\pi-\theta_{ij}-\beta,\pi-\theta_{ij}+\beta)\}\cup\{(\pi-\beta,\pi+\beta)\cap(2\pi-\theta_{ij}-\beta,2\pi-\theta_{ij})\}\cup\emptyset.
\end{align}
It is evident that
\begin{align}\nonumber
&|[0,\beta)\cap(\pi-\theta_{ij}-\beta,\pi-\theta_{ij}+\beta)|=\max\{2\beta+\theta_{ij}-\pi,0\},\\\nonumber
&|(\pi-\beta,\pi+\beta)\cap(2\pi-\theta_{ij}-\beta,2\pi-\theta_{ij})|=\max\{2\beta+\theta_{ij}-\pi,0\}.
\end{align}
Therefore, the total interval length of $\aleph(\theta_{ij})$ in the case of (II-b-1) is
\begin{align}\label{aleph_II_b_1}
|\aleph(\theta_{ij})|=2\max\{2\beta+\theta_{ij}-\pi,0\}.
\end{align}
{\textcolor{orange}{(II-b-2)} If $\beta+\theta_{ij}>\pi$, then
\begin{align}\nonumber
&\{\alpha||\sin(\alpha+\theta_{ij})|<R/\rho,\alpha\in[0,2\pi)\}=\mathfrak B_1\cup\mathfrak B_2\cup\mathfrak B_3\\\nonumber
=&(0,\pi-\theta_{ij}+\beta)\cup(2\pi-\theta_{ij},2\pi+\beta-\theta_{ij})\cup(2\pi-\beta-\theta_{ij},2\pi-\theta_{ij}]\cup(3\pi-\theta_{ij}-\beta,2\pi).
\end{align}
\begin{align}\nonumber
&\aleph(\theta_{ij})=\{\alpha||\sin\alpha|<R/\rho\quad\text{and}\quad|\sin(\alpha+\theta_{ij})|<R/\rho,\quad\alpha\in[0,2\pi)\}\\\nonumber
=&(\mathfrak A_1\cup\mathfrak A_2\cup\mathfrak A_3)\cap(\mathfrak B_1\cup\mathfrak B_2\cup\mathfrak B_3)\\\nonumber
=&[\mathfrak A_1\cap(\mathfrak B_1\cup\mathfrak B_2\cup\mathfrak B_3)]\cup[\mathfrak A_2\cap(\mathfrak B_1\cup\mathfrak B_2\cup\mathfrak B_3)]\cup[\mathfrak A_3\cap(\mathfrak B_1\cup\mathfrak B_2\cup\mathfrak B_3)]\\\nonumber
=&\{[0,\beta)\cap(0,\pi-\theta_{ij}+\beta)\}\\\nonumber
&\cup\{[(\pi-\beta,\pi+\beta)\cap(2\pi-\theta_{ij},2\pi+\beta-\theta_{ij})]\cup[(\pi-\beta,\pi+\beta)\cap(2\pi-\beta-\theta_{ij},2\pi-\theta_{ij}]]\}\\\nonumber
&\cup\{(2\pi-\beta,2\pi)\cap(3\pi-\theta_{ij}-\beta,2\pi)\}
\end{align}
It is evident that
\begin{align}\nonumber
&|[0,\beta)\cap(0,\pi-\theta_{ij}+\beta)|=\beta,\\\nonumber
&|(\pi-\beta,\pi+\beta)\cap(2\pi-\theta_{ij},2\pi+\beta-\theta_{ij})|=\beta+\theta_{ij}-\pi,\\\nonumber
&|(\pi-\beta,\pi+\beta)\cap(2\pi-\beta-\theta_{ij},2\pi-\theta_{ij}]|=\beta,\\\nonumber
&|(2\pi-\beta,2\pi)\cap(3\pi-\theta_{ij}-\beta,2\pi)|=\beta+\theta_{ij}-\pi.
\end{align}
Therefore, the total interval length of $\aleph(\theta_{ij})$ in the case of (II-b-2) is
\begin{align}\label{aleph_II_b_2}
|\aleph(\theta_{ij})|=4\beta+2\theta_{ij}-2\pi=2\max\{2\beta+\theta_{ij}-\pi,0\}\,\text{due to }\beta+\theta_{ij}>\pi.
\end{align}
Combining all the results of $|\aleph(\theta_{ij})|$ in all the cases (\ref{aleph_I_a_1}), (\ref{aleph_I_a_2}), (\ref{aleph_I_b_1}),  (\ref{aleph_I_b_2}), (\ref{aleph_II_a_1}), (\ref{aleph_II_a_2}), (\ref{aleph_II_b_1}) and (\ref{aleph_II_b_2}), we find that
\begin{align}\nonumber
\left\{\begin{array}{ll}
       \text{I}:\pi/4\leqslant\beta<\pi/2\left\{ \begin{array}{ll}
                                       \text{I-a}:0<\theta_{ij}\leqslant\pi/2:|\aleph(\theta_{ij})|=2(2\beta-\theta_{ij})+2\max\{2\beta+\theta_{ij}-\pi,0\} \\
                                       \text{I-b}:\pi/2<\theta_{ij}<\pi:|\aleph(\theta_{ij})|=2(2\beta+\theta_{ij}-\pi)+2\max\{2\beta-\theta_{ij},0\}
                                \end{array} \right.\\
       \text{II}:0<\beta<\pi/4\left\{ \begin{array}{ll}
                                       \text{II-a}:0<\theta_{ij}\leqslant\pi/2:|\aleph(\theta_{ij})|=2\max\{2\beta-\theta_{ij},0\} \\
                                       \text{II-b}:\pi/2<\theta_{ij}<\pi:|\aleph(\theta_{ij})|=2\max\{2\beta+\theta_{ij}-\pi,0\}
                                \end{array} \right.
       \end{array} \right.
\end{align}
On account of $2\beta\geqslant\pi/2\geqslant\theta_{ij}$ in the case of (I-a), $2\beta+\theta_{ij}>\pi$ in the case of (I-b),  $2\beta+\theta_{ij}<\pi$ in the case of (II-a) and $2\beta<\pi/2<\theta_{ij}$ in the case of (II-b), \textcolor{red}{we discover that $|\aleph(\theta_{ij})|$ in any case can be always represented in a suprisingly unified and beautiful form:
\begin{align}\label{aleph_concise_form}
|\aleph(\theta_{ij})|=2\left(\max\{2\beta-\theta_{ij},0\}+\max\{2\beta+\theta_{ij}-\pi,0\}\right).
\end{align}
In essence, (\ref{aleph_concise_form}) orgins from high-dimensional data geometry and the multiplier factor 2 in (\ref{aleph_concise_form}) is due to the geometric symmetry of the solution sets of trigonometrical inequalities $|\sin\alpha|<R/\rho$ and $|\sin(\alpha+\theta_{ij})|<R/\rho$ in $[0,2\pi)$}.\\
With (\ref{aleph_concise_form}) at hand, it is easy to express $|\aleph(\theta_{ij})|$ more explictly for various value of $\theta_{ij}$ as
\begin{align}\label{aleph_acute_theta}
\text{If }0<\theta_{ij}\leqslant\pi/2,\text{ then }
|\aleph(\theta_{ij})|=\left\{ \begin{array}{ll}
                                       8\beta-2\pi, &\quad  (\pi-\theta_{ij})/2<\beta<\pi/2, \\
                                       4\beta-2\theta_{ij},&\quad \theta_{ij}/2<\beta\leqslant(\pi-\theta_{ij})/2,\\
                                       0, &\quad  0<\beta\leqslant\theta_{ij}/2.
                                \end{array} \right.
\end{align}
\begin{align}\label{aleph_obtuse_theta}
\text{If }\pi/2<\theta_{ij}\leqslant\pi,\text{ then }
|\aleph(\theta_{ij})|=\left\{ \begin{array}{ll}
                                       8\beta-2\pi, &\quad  \theta_{ij}/2<\beta<\pi/2, \\
                                       4\beta+2\theta_{ij}-2\pi,&\quad (\pi-\theta_{ij})/2<\beta\leqslant\theta_{ij}/2,\\
                                       0, &\quad  0<\beta\leqslant(\pi-\theta_{ij})/2.
                                \end{array} \right.
\end{align}
Clearly, we can further integrate $|\aleph(\theta_{ij})|$ in (\ref{aleph_acute_theta}) and (\ref{aleph_obtuse_theta}) for any $\theta_{ij}\in(0,\pi)$ as one form:
\begin{align}\nonumber
|\aleph(\theta_{ij})|=\left\{ \begin{array}{ll}
                                       8\beta-2\pi, &\quad  \max\{\theta_{ij},\pi-\theta_{ij}\}/2<\beta<\pi/2, \\
                                       4\beta-2\min\{\theta_{ij},\pi-\theta_{ij}\},&\quad \min\{\theta_{ij},\pi-\theta_{ij}\}/2<\beta\leqslant\max\{\theta_{ij},\pi-\theta_{ij}\}/2,\\
                                       0, &\quad  0<\beta\leqslant\min\{\theta_{ij},\pi-\theta_{ij}\}/2.
                                \end{array} \right.
\end{align}
With the aid of the obvious fact that
$$\min\{\theta_{ij},\pi-\theta_{ij}\}=\frac{\pi}{2}-\left|\theta_{ij}-\frac{\pi}{2}\right|\text{ and }\min\{\theta_{ij},\pi-\theta_{ij}\}+\max\{\theta_{ij},\pi-\theta_{ij}\}=\pi,\,\forall\theta_{ij}\in(0,\pi),$$
we obtain the final expression of $|\aleph(\theta_{ij})|$ as
\begin{align}\label{aleph_final_form}
|\aleph(\theta_{ij})|=\left\{ \begin{array}{ll}
                                       8\beta-2\pi, &\quad  \pi/2-\widetilde{\theta_{ij}}<\beta<\pi/2, \\
                                       4\beta-\pi+\left|2\theta_{ij}-\pi\right|,&\quad \widetilde{\theta_{ij}}<\beta\leqslant\pi/2-\widetilde{\theta_{ij}},\\
                                       0, &\quad  0<\beta\leqslant\widetilde{\theta_{ij}}.
                                \end{array} \right.
\end{align}
Here we introduce an angle parameter \textcolor{blue}{$\widetilde{\theta_{ij}}=\min\{\theta_{ij},\pi-\theta_{ij}\}/2$} for notation simplicity. Remind that $0<\theta^*\leqslant\theta_{ij}\leqslant\pi-\theta^*<\pi$ from (\ref{theta_star_definition}), we have $\widetilde{\theta_{ij}}\in[\theta^*/2,\pi/4]$. \\
\textcolor{red}{We must emphasize that a very important property of $|\aleph(\theta_{ij})|$ unvealed by the above expressions is that $|\aleph(\theta_{ij})|$ vanishes when $\beta$ is sufficiently small or equivalently $\rho$ is sufficiently large for any pair of $\boldsymbol x_i$ and $\boldsymbol x_j$, which makes the integral in (\ref{eth_ij_1}) be upper controlled efficiently. Moreover, a tight upper bound of the integral in (\ref{eth_ij_1}) is crucial to control the inner product $\langle\widetilde{\boldsymbol\psi_i}(k)-\widetilde{\boldsymbol\psi_i}(0),\widetilde{\boldsymbol\psi_j}(k)-\widetilde{\boldsymbol\psi_j}(0)\rangle$ in (\ref{inner_prod_variation_Psi_4}), while a sharp upper bound of  $\langle\widetilde{\boldsymbol\psi_i}(k)-\widetilde{\boldsymbol\psi_i}(0),\widetilde{\boldsymbol\psi_j}(k)-\widetilde{\boldsymbol\psi_j}(0)\rangle$ plays a key role in upper bounding $\|\boldsymbol\Psi(k)-\boldsymbol\Psi(0)\|_2$ and hence leads to our final convergence guarantee. Therefore, the expression $|\aleph(\theta_{ij})|$ in (\ref{aleph_concise_form}) is one of the key immediate results in our paper, which is the reason why we analyze $\aleph(\theta_{ij})$ and $|\aleph(\theta_{ij})|$ in the above very detailed level.}\\\\
{\textcolor{blue}{\bf Subsubstep 2-2-2: Evaluating the Asymptotic Behavior of Integral $\eth_{ij}$}}\\
Recall that $\beta=\arcsin(R/\rho)\in(0,\pi/2)$ for $\rho>R$, we have 
\begin{align}\nonumber
\left\{ \begin{array}{ll}
       \pi/2-\widetilde{\theta_{ij}}<\beta<\pi/2\,\text{ if and only if }\,R<\rho<R\sec\widetilde{\theta_{ij}}, \\
       \widetilde{\theta_{ij}}<\beta\leqslant\pi/2-\widetilde{\theta_{ij}}\,\text{ if and only if }\,R\sec\widetilde{\theta_{ij}}\leqslant\rho<R\csc\widetilde{\theta_{ij}}, \\
       0<\beta\leqslant\widetilde{\theta_{ij}}\,\text{ if and only if }\,\rho\geqslant R\csc\widetilde{\theta_{ij}}.
        \end{array} \right.
\end{align}
Therefore, the length of integral domain $\aleph(\theta_{ij})$ in (\ref{aleph_final_form}) leads to the integral term in (\ref{eth_ij_1}) as
\begin{align}\nonumber
&\int_R^{+\infty}\left(\int_{\aleph(\theta_{ij})}d\alpha\right)\rho e^{-\rho^2/2}d\rho\\\label{sum_club_1_2}
=&\underbrace{\int_{R}^{R\sec\widetilde{\theta_{ij}}}(8\beta-2\pi)\rho e^{-\rho^2/2}d\rho}_{\textcolor{red}{\clubsuit_1}}+\underbrace{\int_{R\sec\widetilde{\theta_{ij}}}^{R\csc\widetilde{\theta_{ij}}}\left(4\beta-\pi+\left|2\theta_{ij}-\pi\right|\right)\rho e^{-\rho^2/2}d\rho}_{\textcolor{red}{\clubsuit_2}}.
\end{align}
As for \textcolor{red}{$\clubsuit_1$}, we see that
\begin{align}\nonumber
&\textcolor{red}{\clubsuit_1}\\\nonumber
=&8\int_{R}^{R\sec\widetilde{\theta_{ij}}}\arcsin\frac{R}{\rho}\cdot\rho e^{-\rho^2/2}d\rho-2\pi\int_{R}^{R\sec\widetilde{\theta_{ij}}}\rho e^{-\rho^2/2}d\rho\\\nonumber
=&8\int_{R}^{R\sec\widetilde{\theta_{ij}}}-\arcsin\frac{R}{\rho}d(e^{-\rho^2/2})-2\pi\left[\exp\left(-\frac{R^2}{2}\right)-\exp\left(-\frac{R^2\sec^2\widetilde{\theta_{ij}}}{2}\right)\right]\\\nonumber
\stackrel{\text{(i)}}=&4\pi\exp\left(-\frac{R^2}{2}\right)-8\arcsin(\cos\widetilde{\theta_{ij}})\exp\left(-\frac{R^2\sec^2\widetilde{\theta_{ij}}}{2}\right)\\\nonumber
&+8\int_{R}^{R\sec\widetilde{\theta_{ij}}}\left(\arcsin\frac{R}{\rho}\right)'\exp\left(-\frac{R^2}{2}\right)d\rho
-2\pi\left[\exp\left(-\frac{R^2}{2}\right)-\exp\left(-\frac{R^2\sec^2\widetilde{\theta_{ij}}}{2}\right)\right]\\\nonumber
=&4\pi\exp\left(-\frac{R^2}{2}\right)-8\left(\frac{\pi}{2}-\widetilde{\theta_{ij}}\right)\exp\left(-\frac{R^2\sec^2\widetilde{\theta_{ij}}}{2}\right)\\\nonumber
&-8R\int_{R}^{R\sec\widetilde{\theta_{ij}}}\frac{d\rho}{\rho\sqrt{\rho^2-R^2}\exp(\rho^2/2)}
-2\pi\left[\exp\left(-\frac{R^2}{2}\right)-\exp\left(-\frac{R^2\sec^2\widetilde{\theta_{ij}}}{2}\right)\right]\\\nonumber
=&2\pi\exp\left(-\frac{R^2}{2}\right)+(8\widetilde{\theta_{ij}}-2\pi)\exp\left(-\frac{R^2\sec^2\widetilde{\theta_{ij}}}{2}\right)-8R\int_{R}^{R\sec\widetilde{\theta_{ij}}}\frac{d\rho}{\rho\sqrt{\rho^2-R^2}\exp(\rho^2/2)},
\end{align}
where (i) is due to integration by part.\\
Expanding $\exp\{-\rho^2/2\}$ by power series and
inverting the order of integral and summation by uniform convergence, we obtain
\begin{align}\nonumber
&\int_{R}^{R\sec\widetilde{\theta_{ij}}}\frac{d\rho}{\rho\sqrt{\rho^2-R^2}\exp(\rho^2/2)}\\\nonumber
=&\int_{R}^{R\sec\widetilde{\theta_{ij}}}\sum_{k=0}^{+\infty}
\frac{(-1)^k\rho^{2k-1}}{2^k\cdot k!\sqrt{\rho^2-R^2}}d\rho=\sum_{k=0}^{+\infty}\int_{R}^{R\sec\widetilde{\theta_{ij}}}
\frac{(-1)^k\rho^{2k-1}}{2^k\cdot k!\sqrt{\rho^2-R^2}}d\rho\\\nonumber
=&\int_{R}^{R\sec\widetilde{\theta_{ij}}}\frac{d\rho}{\rho\sqrt{\rho^2-R^2}}-\int_{R}^{R\sec\widetilde{\theta_{ij}}}\frac{\rho d\rho}{2\sqrt{\rho^2-R^2}}+\int_{R}^{R\sec\widetilde{\theta_{ij}}}\frac{\mathcal O(\rho^4)d\rho}{\rho\sqrt{\rho^2-R^2}}\\\nonumber
=&\int_{R}^{R\sec\widetilde{\theta_{ij}}}\frac{d\rho}{\rho\sqrt{\rho^2-R^2}}-\int_{R}^{R\sec\widetilde{\theta_{ij}}}\frac{\rho d\rho}{2\sqrt{\rho^2-R^2}}+\mathcal O\left(\int_{R}^{R\sec\widetilde{\theta_{ij}}}\frac{\rho^3d\rho}{\sqrt{\rho^2-R^2}}\right).
\end{align}
Changing the integral variable as $\rho=R\sec\alpha$, then $d\rho=R\sin\alpha\sec^2\alpha d\alpha$ and we have
\begin{align}\label{0127_int_1}
\int_{R}^{R\sec\widetilde{\theta_{ij}}}\frac{d\rho}{\rho\sqrt{\rho^2-R^2}}
=\int_0^{\widetilde{\theta_{ij}}}\frac{R\sin\alpha\sec^2\alpha d\alpha}{R^2\sec\alpha\tan\alpha}=\int_0^{\widetilde{\theta_{ij}}}\frac{d\alpha}{R}=\frac{\widetilde{\theta_{ij}}}{R}.
\end{align}
Changing the integral variable as $t=\sqrt{\rho^2-R^2}$, we get
\begin{align}\label{0127_int_2}
\int_{R}^{R\sec\widetilde{\theta_{ij}}}\frac{\rho d\rho}{2\sqrt{\rho^2-R^2}}
=\int_0^{R\tan\widetilde{\theta_{ij}}}\frac{dt}{2}=\frac{R\tan\widetilde{\theta_{ij}}}{2}.
\end{align}
In addition,
\begin{align}\label{0127_int_3}
\left|\int_{R}^{R\sec\widetilde{\theta_{ij}}}\frac{\rho^3d\rho}{\sqrt{\rho^2-R^2}}\right|
\leqslant R^2\sec^2\widetilde{\theta_{ij}}\int_{R}^{R\sec\widetilde{\theta_{ij}}}\frac{\rho d\rho}{\sqrt{\rho^2-R^2}}=R^3\sec^2\widetilde{\theta_{ij}}\tan\widetilde{\theta_{ij}}.
\end{align}
Combining (\ref{0127_int_1}), (\ref{0127_int_2}) and (\ref{0127_int_3}) leads to the \textcolor{red}{asymptotic expansion of} $\textcolor{red}{\clubsuit_1}$ as
\begin{align}\nonumber
&\textcolor{red}{\clubsuit_1}\\\nonumber
=&2\pi\exp\left\{-\frac{R^2}{2}\right\}+(8\widetilde{\theta_{ij}}-2\pi)\exp\left\{-\frac{R^2\sec^2\widetilde{\theta_{ij}}}{2}\right\}\\\nonumber
&-8R\left[\frac{\widetilde{\theta_{ij}}}{R}-\frac{R\tan\widetilde{\theta_{ij}}}{2}+\mathcal O\left(R^3\sec^2\widetilde{\theta_{ij}}\tan\widetilde{\theta_{ij}}\right)\right]\\\nonumber
=&2\pi\left[\exp\left\{-\frac{R^2}{2}\right\}-\exp\left\{-\frac{R^2\sec^2\widetilde{\theta_{ij}}}{2}\right\}\right]-8\widetilde{\theta_{ij}}\left[1-\exp\left\{-\frac{R^2\sec^2\widetilde{\theta_{ij}}}{2}\right\}\right]\\\nonumber
&+4R^2\tan\widetilde{\theta_{ij}}+\mathcal O\left(R^4\sec^2\widetilde{\theta_{ij}}\tan\widetilde{\theta_{ij}}\right).
\end{align}
Regarding \textcolor{red}{$\clubsuit_2$}, similarly, we obtain
\begin{align}\nonumber
&\textcolor{red}{\clubsuit_2}\\\nonumber
=&4\int_{R\sec\widetilde{\theta_{ij}}}^{R\csc\widetilde{\theta_{ij}}}\arcsin\frac{R}{\rho}\cdot\rho e^{-\rho^2/2}d\rho+(\left|2\theta_{ij}-\pi\right|-\pi)\int_{R\sec\widetilde{\theta_{ij}}}^{R\csc\widetilde{\theta_{ij}}}\rho e^{-\rho^2/2}d\rho\\\nonumber
=&4\int_{R\sec\widetilde{\theta_{ij}}}^{R\csc\widetilde{\theta_{ij}}}-\arcsin\frac{R}{\rho}d(e^{-\rho^2/2})+(\left|2\theta_{ij}-\pi\right|-\pi)\left[\exp\left\{-\frac{R^2\sec^2\widetilde{\theta_{ij}}}{2}\right\}-\exp\left\{-\frac{R^2\csc^2\widetilde{\theta_{ij}}}{2}\right\}\right]\\\nonumber
=&4\arcsin(\cos\widetilde{\theta_{ij}})\exp\left\{-\frac{R^2\sec^2\widetilde{\theta_{ij}}}{2}\right\}-4\arcsin(\sin\widetilde{\theta_{ij}})\exp\left\{-\frac{R^2\csc^2\widetilde{\theta_{ij}}}{2}\right\}\\\nonumber
&+4\int_{R\sec\widetilde{\theta_{ij}}}^{R\csc\widetilde{\theta_{ij}}}\left[\arcsin\frac{R}{\rho}\right]'e^{-R^2/2}d\rho
+(\left|2\theta_{ij}-\pi\right|-\pi)\left[\exp\left\{-\frac{R^2\sec^2\widetilde{\theta_{ij}}}{2}\right\}-\exp\left\{-\frac{R^2\csc^2\widetilde{\theta_{ij}}}{2}\right\}\right]\\\nonumber
=&(2\pi-4\widetilde{\theta_{ij}})\exp\left\{-\frac{R^2\sec^2\widetilde{\theta_{ij}}}{2}\right\}-4\widetilde{\theta_{ij}}\exp\left\{-\frac{R^2\csc^2\widetilde{\theta_{ij}}}{2}\right\}\\\nonumber
&+4\int_{R\sec\widetilde{\theta_{ij}}}^{R\csc\widetilde{\theta_{ij}}}\left[\arcsin\frac{R}{\rho}\right]'e^{-R^2/2}d\rho
+(\left|2\theta_{ij}-\pi\right|-\pi)\left[\exp\left\{-\frac{R^2\sec^2\widetilde{\theta_{ij}}}{2}\right\}-\exp\left\{-\frac{R^2\csc^2\widetilde{\theta_{ij}}}{2}\right\}\right]\\\nonumber
=&(\pi-4\widetilde{\theta_{ij}}+\left|2\theta_{ij}-\pi\right|)\exp\left\{-\frac{R^2\sec^2\widetilde{\theta_{ij}}}{2}\right\}+(\pi-4\widetilde{\theta_{ij}}-\left|2\theta_{ij}-\pi\right|)\exp\left\{-\frac{R^2\csc^2\widetilde{\theta_{ij}}}{2}\right\}\\\nonumber
&-4R\int_{R\sec\widetilde{\theta_{ij}}}^{R\csc\widetilde{\theta_{ij}}}
\frac{d\rho}{\rho\sqrt{\rho^2-R^2}\exp(\rho^2/2)}
\end{align}
Proceeding in a similar manner, we find immediately
\begin{align}\nonumber
&\int_{R\sec\widetilde{\theta_{ij}}}^{R\csc\widetilde{\theta_{ij}}}\frac{d\rho}{\rho\sqrt{\rho^2-R^2}\exp(\rho^2/2)}\\\nonumber
=&\sum_{k=0}^{+\infty}\int_{R\sec\widetilde{\theta_{ij}}}^{R\csc\widetilde{\theta_{ij}}}
\frac{(-1)^k\rho^{2k-1}}{2^k\cdot k!\sqrt{\rho^2-R^2}}d\rho\\\nonumber
=&\int_{R\sec\widetilde{\theta_{ij}}}^{R\csc\widetilde{\theta_{ij}}}\frac{d\rho}{\rho\sqrt{\rho^2-R^2}}-\int_{R\sec\widetilde{\theta_{ij}}}^{R\csc\widetilde{\theta_{ij}}}\frac{\rho d\rho}{2\sqrt{\rho^2-R^2}}+\mathcal O\left(\int_{R\sec\widetilde{\theta_{ij}}}^{R\csc\widetilde{\theta_{ij}}}\frac{\rho^3d\rho}{\sqrt{\rho^2-R^2}}\right).
\end{align}
\begin{align}\label{0127_int_1_2}
\int_{R\sec\widetilde{\theta_{ij}}}^{R\csc\widetilde{\theta_{ij}}}\frac{d\rho}{\rho\sqrt{\rho^2-R^2}}
=\int_{\widetilde{\theta_{ij}}}^{\pi/2-\widetilde{\theta_{ij}}}\frac{d\alpha}{R}=\frac{\pi-4\widetilde{\theta_{ij}}}{2R}.
\end{align}
\begin{align}\nonumber
&\int_{R\sec\widetilde{\theta_{ij}}}^{R\csc\widetilde{\theta_{ij}}}\frac{\rho d\rho}{2\sqrt{\rho^2-R^2}}
=\int_{R\tan\widetilde{\theta_{ij}}}^{R\cot\widetilde{\theta_{ij}}}\frac{dt}{2}=\frac{R(\cot\widetilde{\theta_{ij}}-\tan\widetilde{\theta_{ij}})}{2}\\\label{0127_int_2_2}
=&\frac{R}{2}\left(\frac{\cos\widetilde{\theta_{ij}}}{\sin\widetilde{\theta_{ij}}}-\frac{\sin\widetilde{\theta_{ij}}}{\cos\widetilde{\theta_{ij}}}\right)=\frac{R}{2}\cdot\frac{\cos^2\widetilde{\theta_{ij}}-\sin^2\widetilde{\theta_{ij}}}{\sin\widetilde{\theta_{ij}}\cos\widetilde{\theta_{ij}}}=R\frac{\cos2\widetilde{\theta_{ij}}}{\sin2\widetilde{\theta_{ij}}}=R\cot2\widetilde{\theta_{ij}}.
\end{align}
In addition,
\begin{align}\label{0127_int_3_2}
\left|\int_{R\sec\widetilde{\theta_{ij}}}^{R\csc\widetilde{\theta_{ij}}}\frac{\rho^3d\rho}{\sqrt{\rho^2-R^2}}\right|
\leqslant R^2\csc^2\widetilde{\theta_{ij}}\int_{R\sec\widetilde{\theta_{ij}}}^{R\csc\widetilde{\theta_{ij}}}\frac{\rho d\rho}{\sqrt{\rho^2-R^2}}=2R^3\csc^2\widetilde{\theta_{ij}}\cot2\widetilde{\theta_{ij}}.
\end{align}
Combining (\ref{0127_int_1_2}), (\ref{0127_int_2_2}) and (\ref{0127_int_3_2}) leads to the \textcolor{red}{asymptotic expansion of} $\textcolor{red}{\clubsuit_2}$ as
\begin{align}\nonumber
&\textcolor{red}{\clubsuit_2}\\\nonumber
=&(\pi-4\widetilde{\theta_{ij}}+\left|2\theta_{ij}-\pi\right|)\exp\left\{-\frac{R^2\sec^2\widetilde{\theta_{ij}}}{2}\right\}+(\pi-4\widetilde{\theta_{ij}}-\left|2\theta_{ij}-\pi\right|)\exp\left\{-\frac{R^2\csc^2\widetilde{\theta_{ij}}}{2}\right\}\\\nonumber
&-4R\left[\frac{\pi-4\widetilde{\theta_{ij}}}{2R}-R\cot2\widetilde{\theta_{ij}}+\mathcal O\left(2R^3\csc^2\widetilde{\theta_{ij}}\cot2\widetilde{\theta_{ij}}\right)\right]\\\nonumber
=&(\pi-4\widetilde{\theta_{ij}})\left[\exp\left\{-\frac{R^2\sec^2\widetilde{\theta_{ij}}}{2}\right\}+\exp\left\{-\frac{R^2\csc^2\widetilde{\theta_{ij}}}{2}\right\}-2\right]\\\nonumber
&+\left|2\theta_{ij}-\pi\right|\left[\exp\left\{-\frac{R^2\sec^2\widetilde{\theta_{ij}}}{2}\right\}-\exp\left\{-\frac{R^2\csc^2\widetilde{\theta_{ij}}}{2}\right\}\right]\\\nonumber
&+4R^2\cot2\widetilde{\theta_{ij}}+\mathcal O\left(R^4\csc^2\widetilde{\theta_{ij}}\cot2\widetilde{\theta_{ij}}\right).
\end{align}
Now, substituting the asymptotic expansions of $\textcolor{red}{\clubsuit_1}$ and $\textcolor{red}{\clubsuit_2}$ into (\ref{sum_club_1_2}) and expanding the Maclaurin's series of exponential functions, we can evaluate the integral, say
\begin{align}\nonumber
&\int_R^{+\infty}\left(\int_{\aleph(\theta_{ij})}d\alpha\right)\rho e^{-\rho^2/2}d\rho\\\nonumber
=&2\pi\left[\exp\left\{-\frac{R^2}{2}\right\}-\exp\left\{-\frac{R^2\sec^2\widetilde{\theta_{ij}}}{2}\right\}\right]-8\widetilde{\theta_{ij}}\left[1-\exp\left\{-\frac{R^2\sec^2\widetilde{\theta_{ij}}}{2}\right\}\right]\\\nonumber
&+4R^2\tan\widetilde{\theta_{ij}}+\mathcal O\left(R^4\sec^2\widetilde{\theta_{ij}}\tan\widetilde{\theta_{ij}}\right)\\\nonumber
&+(\pi-4\widetilde{\theta_{ij}})\left[\exp\left\{-\frac{R^2\sec^2\widetilde{\theta_{ij}}}{2}\right\}+\exp\left\{-\frac{R^2\csc^2\widetilde{\theta_{ij}}}{2}\right\}-2\right]\\\nonumber
&+\left|2\theta_{ij}-\pi\right|\left[\exp\left\{-\frac{R^2\sec^2\widetilde{\theta_{ij}}}{2}\right\}-\exp\left\{-\frac{R^2\csc^2\widetilde{\theta_{ij}}}{2}\right\}\right]+4R^2\cot2\widetilde{\theta_{ij}}+\mathcal O\left(R^4\csc^2\widetilde{\theta_{ij}}\cot2\widetilde{\theta_{ij}}\right)\\\nonumber
=&\pi R^2\tan^2\widetilde{\theta_{ij}}+\mathcal O\left(R^4\right)-4R^2\widetilde{\theta_{ij}}\sec^2\widetilde{\theta_{ij}}
+\mathcal O\left(R^4\right)+4R^2\tan\widetilde{\theta_{ij}}+\mathcal O\left(R^4\right)\\\nonumber
&+\frac{4\widetilde{\theta_{ij}}-\pi}{2}R^2(\sec^2\widetilde{\theta_{ij}}+\csc^2\widetilde{\theta_{ij}})+\mathcal O\left(R^4\right)+\frac{\left|2\theta_{ij}-\pi\right|}{2}R^2(\csc^2\widetilde{\theta_{ij}}-\sec^2\widetilde{\theta_{ij}})+4R^2\cot2\widetilde{\theta_{ij}}
+\mathcal O\left(R^4\right)\\\nonumber
\stackrel{\text{(i)}}=&(\pi\tan^2\widetilde{\theta_{ij}}-4\widetilde{\theta_{ij}}\sec^2\widetilde{\theta_{ij}}  +4\tan\widetilde{\theta_{ij}}+\frac{4\widetilde{\theta_{ij}}-\pi}{2}(\sec^2\widetilde{\theta_{ij}}+\csc^2\widetilde{\theta_{ij}})
\\\nonumber
&+\frac{\pi-4\widetilde{\theta_{ij}}}{2}(\csc^2\widetilde{\theta_{ij}}-\sec^2\widetilde{\theta_{ij}})+4\cot2\widetilde{\theta_{ij}})R^2+\mathcal O\left(R^4\right)\\\nonumber
=&\left[\pi\tan^2\widetilde{\theta_{ij}}-4\widetilde{\theta_{ij}}\sec^2\widetilde{\theta_{ij}}  +4\tan\widetilde{\theta_{ij}}+(4\widetilde{\theta_{ij}}-\pi)\sec^2\widetilde{\theta_{ij}}
+4\cot2\widetilde{\theta_{ij}}\right]R^2+\mathcal O\left(R^4\right)\\\nonumber
=&(4\tan\widetilde{\theta_{ij}}
+4\cot2\widetilde{\theta_{ij}}-\pi)R^2+\mathcal O\left(R^4\right)\stackrel{\text{(ii)}}=(4\csc2\widetilde{\theta_{ij}}-\pi)R^2+\mathcal O\left(R^4\right),
\end{align}
where (i) is due to $|\theta_{ij}-\pi/2|=\pi/2-2\widetilde{\theta_{ij}}$ which follows from $\widetilde{\theta_{ij}}=\min\{\theta_{ij}\pi-\theta_{ij}\}/2\in(0,\pi/4]$ and (ii) comes from the trigonometrical identity
$$\tan\widetilde{\theta_{ij}}+\cot2\widetilde{\theta_{ij}}=\tan\widetilde{\theta_{ij}}
+\frac{1-\tan^2\widetilde{\theta_{ij}}}{2\tan\widetilde{\theta_{ij}}}
=\frac{1+\tan^2\widetilde{\theta_{ij}}}{2\tan\widetilde{\theta_{ij}}}=\frac{1}{\sin2\widetilde{\theta_{ij}}}=\csc2\widetilde{\theta_{ij}}.$$
Reminding (\ref{eth_ij_1}), we have therefore
\begin{align}\nonumber
\eth_{ij}&=\mathbb P(|\boldsymbol w_r(0)^\top\boldsymbol x_i|\leqslant R\quad\text{and}\quad|\boldsymbol w_r(0)^\top\boldsymbol x_j|\leqslant R)\\\nonumber
&=1-e^{-R^2/2}+\frac{1}{2\pi}\int_R^{+\infty}\left(\int_{\aleph(\theta_{ij})}d\alpha\right)\rho e^{-\rho^2/2}d\rho\\\nonumber
&=\frac{R^2}{2}+\mathcal O\left(R^4\right)+\frac{4\csc2\widetilde{\theta_{ij}}-\pi}{2\pi}R^2+\mathcal O\left(R^4\right)\\\nonumber
&=\frac{2\csc2\widetilde{\theta_{ij}}}{\pi}R^2+\mathcal O\left(R^4\right).
\end{align}
The final expression of $\eth_{ij}$ is surprisingly simple, although the intermediate analysis and calculations are very complicated. The value of $\eth_{ij}$ decreases with increasing $\theta_{ij}$, which matches our intuition and in essence is the consequence of high-dimensional geoemtry of the input data. Additionally, as can be seen in the sequel, the asymptotic relation $\eth_{ij}=\mathcal O(R^2)$ is very crucial for upper bounding the off-diagonal entry  $\langle\widetilde{\boldsymbol\psi_i}(k)-\widetilde{\boldsymbol\psi_i}(0),\widetilde{\boldsymbol\psi_j}(k)-\widetilde{\boldsymbol\psi_j}(0)\rangle$ of matrix $\boldsymbol N$ and further upper bounding $\|\boldsymbol\Psi(k)-\boldsymbol\Psi(0)\|_2$ efficiently.\\
Returning to (\ref{inner_prod_variation_Psi_4}), we arrive at with probability exceeding $1-\delta$,
\begin{align}\nonumber
&\langle\widetilde{\boldsymbol\psi_i}(k)-\widetilde{\boldsymbol\psi_i}(0),\widetilde{\boldsymbol\psi_j}(k)-\widetilde{\boldsymbol\psi_j}(0)\rangle\\\nonumber
\leqslant& m\eth_{ij}+\frac{2}{3}\log\frac{1}{\delta}+\sqrt{2\eth_{ij}(1-\eth_{ij})m\log\frac{1}{\delta}}\\\nonumber
=&\frac{2\csc2\widetilde{\theta_{ij}}}{\pi}R^2m+\frac{2}{3}\log\frac{1}{\delta}+\sqrt{\left[\frac{4\csc2\widetilde{\theta_{ij}}}{\pi}R^2+\mathcal O\left(R^4\right)\right]m\log\frac{1}{\delta}}+\mathcal O\left(R^4m\right)\\\nonumber
=&\frac{2\csc2\widetilde{\theta_{ij}}}{\pi}R^2m+\frac{2}{3}\log\frac{1}{\delta}+2R\sqrt{\frac{\csc2\widetilde{\theta_{ij}}}{\pi}m\log\frac{1}{\delta}}\sqrt{1+\mathcal O\left(R^2\right)}+\mathcal O\left(R^4m\right)\\\nonumber
\stackrel{\text{(i)}}=&\frac{2\csc2\widetilde{\theta_{ij}}}{\pi}R^2m+2R\sqrt{\frac{\csc2\widetilde{\theta_{ij}}}{\pi}m\log(1/\delta)}+\frac{2}{3}\log\frac{1}{\delta}+\mathcal O\left(R^3\sqrt{m\log(1/\delta)}+R^4m\right)\\\nonumber
\stackrel{\text{(ii)}}\leqslant&\frac{2\csc\theta^*}{\pi}R^2m+2R\sqrt{\frac{\csc\theta^*}{\pi}m\log(1/\delta)}+\frac{2}{3}\log\frac{1}{\delta}+\mathcal O\left(R^3\sqrt{m\log(1/\delta)}+R^4m\right),
\end{align}
where (i) depends on the fact that
$$
\sqrt{1+\mathcal O(f)}=1+\frac{1}{2}\mathcal O(f)+\mathcal O(f^2)=1+\mathcal O(f),\text{ if }f\to0,
$$ whereas (ii) results from $\pi/4\geqslant\widetilde{\theta_{ij}}\geqslant\theta^*/2>0$.\\
\\{\textcolor{blue}{{\bf Step 3: Invoking Ger\v{s}gorin and Weyl to Upper Bound $\|\boldsymbol\Psi(k)-\boldsymbol\Psi(0)\|_2^2$}}\\
Thus, taking note of the symmetry of $\langle\widetilde{\boldsymbol\psi_i}(k)-\widetilde{\boldsymbol\psi_i}(0),\widetilde{\boldsymbol\psi_j}(k)-\widetilde{\boldsymbol\psi_j}(0)\rangle$ and applying a union bound over $j\in[n]$ and $j\not=i$, we have with probability exceeding $1-2(n-1)\delta$,
\begin{align}\nonumber
&\sum_{1\leqslant j\not=i\leqslant n}|\boldsymbol N_{ij}|\\\nonumber
\leqslant&(n-1)\left[\frac{2\csc\theta^*}{\pi}R^2m+2R\sqrt{\frac{\csc\theta^*}{\pi}m\log(1/\delta)}+\frac{2}{3}\log\frac{1}{\delta}+\mathcal O\left(R^3\sqrt{m\log(1/\delta)}+R^4m\right)\right].
\end{align}
With the help of {\textcolor{red}{Ger\v{s}gorin disc theorem} (\cite{gersgorin_1931}) for symmetric matrix $\boldsymbol N$, there exists $i\in[n]$, with probability exceeding $1-2(n-1)\delta$,
\begin{align}\nonumber
\|\boldsymbol N\|_2&\leqslant\boldsymbol N_{ii}+\sum_{1\leqslant j\not=i\leqslant n}|\boldsymbol N_{ij}|\\\nonumber
&\leqslant(n-1)\left[\frac{2\csc\theta^*}{\pi}R^2m+2R\sqrt{\frac{\csc\theta^*}{\pi}m\log(1/\delta)}+\frac{2}{3}\log\frac{1}{\delta}+\mathcal O\left(R^3\sqrt{m\log(1/\delta)}+R^4m\right)\right].
\end{align}
As a consequence, with probability exceeding $1-n\delta-2(n-1)\delta=1-(3n-2)\delta$,
\begin{align}\nonumber
&\|\boldsymbol\Psi(k)-\boldsymbol\Psi(0)\|_2^2\leqslant
\max_{1\leqslant i\leqslant n}\|\widetilde{\boldsymbol\psi_i}(k)-\widetilde{\boldsymbol\psi_i}(0)\|_2^2+\|\boldsymbol N\|_2\\\nonumber
\leqslant& \wp(R)m+\frac{2}{3}\log\frac{1}{\delta}+\sqrt{2\wp(R)[1-\wp(R)]m\log\frac{1}{\delta}}\\\nonumber
&+(n-1)\left[\frac{2\csc\theta^*}{\pi}R^2m+2R\sqrt{\frac{\csc\theta^*}{\pi}m\log(1/\delta)}+\frac{2}{3}\log\frac{1}{\delta}+\mathcal O\left(R^3\sqrt{m\log(1/\delta)}+R^4m\right)\right]\\\nonumber
\stackrel{\text{(ii)}}<&\frac{2\csc\theta^*}{\pi}R^2mn+Rm+\frac{2n}{3}\log(1/\delta)+\mathcal O\left((\sqrt{R}n+1)\sqrt{Rm\log(1/\delta)}+R^4mn\right),
\end{align}
where (i) is due to $0<\wp(R)<R<1/2$ (see (\ref{wp_R_bound})).
\subsection{Piecing Together a Whole Picture: Global Convergence of GD}\label{subsection_pieceing_together}
We now return to the Taylor expansion of $\mathcal L(\boldsymbol W(k+1))$ with Peano's remainder in (\ref{loss_function_Taylor})
\begin{align}\nonumber
&\mathcal L(\boldsymbol W(k+1))\\\nonumber
=&\mathcal L(\boldsymbol W(k))-\eta\|\nabla \mathcal L(\boldsymbol W(k))\|_{\text{F}}^2\\\nonumber
&+\frac{\eta^2}{2}\text{vec}(\nabla \mathcal L(\boldsymbol W(k)))^\top\nabla^2\mathcal L(\boldsymbol W(k))\text{vec}(\nabla \mathcal L(\boldsymbol W(k)))+o(\eta^2\|\nabla \mathcal L(\boldsymbol W(k))\|_{\text{F}}^2).
\end{align}
Invoking Lemma \ref{lemma_gradient_lower_bound}, Lemma \ref{lemma_Hessian_upper_bound} and Lemma \ref{lemma_Phi_fact_1} yields
\begin{align}\nonumber
&\mathcal L(\boldsymbol W(k+1))\leqslant\mathcal L(\boldsymbol W(k))-\eta\cdot\frac{2}{m}\sigma_{\min}^2(\boldsymbol\Psi(k))\lambda_{\min}(\boldsymbol X^\top\boldsymbol X)\mathcal L(\boldsymbol W(k))\\\nonumber
+&\frac{\eta^2}{m^2}\left[\|\boldsymbol\Psi\|_{\text{F}}^2+\sum_{1\leqslant r\not=s\leqslant m}\sqrt{\|\boldsymbol\psi_r(k)\|_2\boldsymbol\psi_s(k)\|_2\langle\boldsymbol\psi_r(k),\boldsymbol\psi_s(k)\rangle}
\right]\sqrt{\sum_{1\leqslant i,j\leqslant n}\langle \boldsymbol x_i,\boldsymbol x_j\rangle^4}\mathcal L(\boldsymbol W(k))+Rem\\\nonumber
<&\bigg\{1-\eta\cdot\frac{2}{m}\sigma_{\min}^2(\boldsymbol\Psi(k))\lambda_{\min}(\boldsymbol X^\top\boldsymbol X)\\\nonumber
&+\frac{\eta^2}{m^2}\left[\|\boldsymbol\Psi(k)\|_{\text{F}}^2+ 
m\|\boldsymbol\Psi(k)\|_{\text{F}}\left[\sum_{i=1}^n\|\widetilde{\boldsymbol\psi_i}(k)\|_2^4+\sum_{1\leqslant i\not=j\leqslant n}\langle\widetilde{\boldsymbol\psi_i}(k),\widetilde{\boldsymbol\psi_j}(k)\rangle^2-\frac{\|\boldsymbol\Psi(k)\|_{\text{F}}^4}{m}\right]^{1/4}
\right]\\\label{loss_function_Taylor_2}
&\cdot\sqrt{\sum_{1\leqslant i,j\leqslant n}\langle \boldsymbol x_i,\boldsymbol x_j\rangle^4}\bigg\}\cdot\mathcal L(\boldsymbol W(k))+Rem.
\end{align}
In a similar manner in Lemma \ref{lemma_gradient_lower_bound}, we can control Peano remainder $o(1)\eta^2\|\nabla \mathcal L(\boldsymbol W(k))\|_{\text{F}}^2$ via upper bounding $\|\nabla \mathcal L(\boldsymbol W(k))\|_{\text{F}}^2$ as
\begin{align}\nonumber
&m\|\nabla \mathcal L(\boldsymbol W)\|_{\text{F}}^2
=\text{tr}(\boldsymbol D_{\boldsymbol r}^\top\boldsymbol X^\top\boldsymbol X\boldsymbol D_{\boldsymbol r}\boldsymbol\Psi\boldsymbol\Psi^\top)
\stackrel{\text{(i)}}{\leqslant}\lambda_{\max}(\boldsymbol\Psi\boldsymbol\Psi^\top)\text{tr}(\boldsymbol D_{\boldsymbol r}^\top\boldsymbol X^\top\boldsymbol X\boldsymbol D_{\boldsymbol r})\\\nonumber
\stackrel{\text{(ii)}}{=}&\lambda_{\max}(\boldsymbol\Psi\boldsymbol\Psi^\top)\text{tr}(\boldsymbol X^\top\boldsymbol X\boldsymbol D_{\boldsymbol r}\boldsymbol D_{\boldsymbol r}^\top)
=\lambda_{\max}(\boldsymbol\Psi\boldsymbol\Psi^\top)\text{tr}(\boldsymbol X^\top\boldsymbol X\boldsymbol D_{\boldsymbol r}^2)
\stackrel{\text{(iii)}}{\leqslant}\lambda_{\max}(\boldsymbol\Psi\boldsymbol\Psi^\top)\lambda_{\max}(\boldsymbol X^\top\boldsymbol X)\text{tr}(\boldsymbol D_{\boldsymbol r}^2)\\\nonumber
=&2\sigma_{\max}^2(\boldsymbol\Psi)\lambda_{\max}(\boldsymbol X^\top\boldsymbol X)\mathcal L(\boldsymbol W),
\end{align}
where (i) and (iii) come from the {\textcolor{red}{trace inequality} $\text{tr}(\boldsymbol A\boldsymbol B)\leqslant\lambda_{\max}(\boldsymbol A)\text{tr}(\boldsymbol B)$ if $\boldsymbol A$ and $\boldsymbol B$ are positive semidefinite matrices. (ii) is due to the famous {\textcolor{red}{trace identity} $\text{tr}(\boldsymbol A\boldsymbol B)=\text{tr}(\boldsymbol B\boldsymbol A)$. We also use the simple facts that and the obvious facts that $\text{tr}(\boldsymbol D_{\boldsymbol r}^2)=2\mathcal L(\boldsymbol W)$.
\begin{align}\nonumber
&Rem=o(\eta^2\|\nabla \mathcal L(\boldsymbol W(k))\|_{\text{F}}^2)=\alpha\eta^2\|\nabla \mathcal L(\boldsymbol W(k))\|_{\text{F}}^2\leqslant|\alpha|\eta^2\|\nabla \mathcal L(\boldsymbol W(k))\|_{\text{F}}^2\\\nonumber
\leqslant&2|\alpha|\eta^2\sigma_{\max}^2(\boldsymbol\Psi(k))\lambda_{\max}(\boldsymbol X^\top\boldsymbol X)\mathcal L(\boldsymbol W(k))/m.
\end{align}
Obviously, $0<|\alpha|<1/2$ and thus Peano's remainder $$Rem<\eta^2\sigma_{\max}^2(\boldsymbol\Psi(k))\lambda_{\max}(\boldsymbol X^\top\boldsymbol X)\mathcal L(\boldsymbol W(k))/m.$$
Therefore, we get
\begin{align}\nonumber
&\mathcal L(\boldsymbol W(k+1))\\\nonumber
<&\bigg\{1-\eta\cdot\frac{2}{m}\sigma_{\min}^2(\boldsymbol\Psi(k))\lambda_{\min}(\boldsymbol X^\top\boldsymbol X)\\\nonumber
&+\frac{\eta^2}{m^2}\left[\|\boldsymbol\Psi(k)\|_{\text{F}}^2+ 
m\|\boldsymbol\Psi(k)\|_{\text{F}}\left[\sum_{i=1}^n\|\widetilde{\boldsymbol\psi_i}(k)\|_2^4+\sum_{1\leqslant i\not=j\leqslant n}\langle\widetilde{\boldsymbol\psi_i}(k),\widetilde{\boldsymbol\psi_j}(k)\rangle^2-\frac{\|\boldsymbol\Psi(k)\|_{\text{F}}^4}{m}\right]^{1/4}
\right]\\\label{loss_function_Taylor_2}
&\cdot\sqrt{\sum_{1\leqslant i,j\leqslant n}\langle \boldsymbol x_i,\boldsymbol x_j\rangle^4}+\frac{\eta^2}{m}\sigma_{\max}^2(\boldsymbol\Psi(k))\lambda_{\max}(\boldsymbol X^\top\boldsymbol X)\bigg\}\cdot\mathcal L(\boldsymbol W(k)).
\end{align}
\\To analyze (\ref{loss_function_Taylor_2}) further, it suffices to control quantities of the activation matrix $\boldsymbol\Psi(k)$: $\sigma_{\min}^2(\boldsymbol\Psi(k)),\sigma_{\max}^2(\boldsymbol\Psi(k)),\|\boldsymbol\Psi(k)\|_{\text{F}}^2, \sum_{i=1}^n\|\widetilde{\boldsymbol\psi_i}(k)\|_2^4$ and $\sum_{1\leqslant i\not=j\leqslant n}\langle\widetilde{\boldsymbol\psi_i}(k),\widetilde{\boldsymbol\psi_j}(k)\rangle^2.$
\\\\
{\textcolor{blue}{\bf Step 1: Bounding $\sigma_{\min}^2(\boldsymbol\Psi(k))$}}\\
In view of the importance of $\sigma_{\min}^2(\boldsymbol\Psi(k))$}}, we state the result as the following lemma.
\begin{lemma}\label{lemma_Psi_k_lower_bound}
(\textcolor{blue}{The Smallest Singular Value of Activation Matrix}) In the over-parameterized setting $n\leqslant m$, for any weight vector $\boldsymbol w_r\in\mathbb R^d$ $(r\in[m])$ and any $k\geqslant1$, $\|\boldsymbol w_r(k)-\boldsymbol w_r(0)\|_2\leqslant R:=\sqrt{(\pi\lambda^*\sin\theta^*)/[(2+\varepsilon)n]}$ for any $\varepsilon>0$, if $\lambda_{\text{min}}(\boldsymbol X^\top\boldsymbol X)=\Omega(\log^{-2\hbar}n)$ for certain $\hbar>0$ and $m=\Omega(n^2\log^{1+\varpi}(n/\delta))$ for any $\varpi>4\hbar$, then for any $\delta\in(0,1)$, with probability exceeding $1-(mn-m+4n)\delta$ over the random initialization, one has: 
\begin{align}\nonumber
\sigma_{\min}^2(\boldsymbol\Psi(k))>&\left(1-\sqrt{\frac{2}{2+\varepsilon}}\right)^2\lambda^*m+\left(\sqrt{\frac{\pi}{2+\varepsilon}}-\sqrt{\frac{1}{2\pi}}+o(1)\right)m\sqrt{\frac{\lambda^*\sin\theta^*}{n}}\\\nonumber
&+\mathcal O\left(n\sqrt{m\log(2n/\delta)}+n\log(2n/\delta)/\sqrt{\delta}\right),
\end{align}
where $\lambda^*=\lambda_{\text{min}}(\mho)$ and the matrix $\mho$ is defined in (\ref{matrix_mho_definition}) in Section \ref{section_problem} and $\theta^*$ are defined in (\ref{theta_star_definition}) in Section \ref{section_problem}, respectively.
\end{lemma}
{\bf Remark}: Roughly speaking, Lemma \ref{lemma_Psi_k_lower_bound} states that if $m$ is sufficiently large and each weight vector stays in a close neighbhood of the initial vector, then the smallest singular value of activation matrix $\boldsymbol\Psi(k)$ for given input data is lower bounded by a sufficiently large number which is approximately proportional to $m$. The real meaning of this lemma is the effect of over-parameterization for controlling the spectrum of random activation matrix $\boldsymbol\Psi(k)$. More explicitely, although the smallest singular value of activation matrix $\boldsymbol\Psi(k)$ has very complex bound expression, many terms in the bound can be efficiently controlled under the over-parameterization condition of $m=\Omega(n^2\log^{1+\varpi}(n/\delta))$. Therefore, it is the over-parameterization condition $m=\Omega(n^2\log^{1+\varpi}(n/\delta))$ that ensures the smallest singular value of activation matrix $\boldsymbol\Psi(k)$ remains large enough in the whole GD iterations, which is crucial to guarantee the global convergence of GD. 
\\In addition, $R$ is a rather small number in $(0,1/2)$. It should be emphasized that our choice of $R=\sqrt{\pi\lambda^*\sin\theta^*/(2+\varepsilon)n}$ for any $\varepsilon>0$ is much larger than previous choices of $R$ in the literature, which is often $\Theta(\lambda_0/n)$. It is obvious that larger choice of $R$ makes the theory be more applicable in practice. Meanwhile, it is the choice of larger $R$ than previous choices that enables the milder over-parameterization condition.
\\
\textbf{Proof}:
Regarding $\sigma_{\min}^2(\boldsymbol\Psi(k))$, applying \textcolor{red}{Weyl's inequality} (\cite{weyl_1912}) and a union bound, followed from Lemmas \ref{lemma_Psi0} and \ref{lemma_Psik_Psi0}, we conclude that: for any $k\in\mathbb Z^{+}$, with probability exceeding $1-(mn-m+n+2)\delta-(3n-2)\delta=1-(mn-m+4n)\delta$ over the random initialization,
\begin{align}\nonumber
&\sigma_{\min}(\boldsymbol\Psi(k))\\\nonumber
\geqslant&\sigma_{\min}(\boldsymbol\Psi(0))-\|\boldsymbol\Psi(k)-\boldsymbol\Psi(0)\|_2\\\nonumber
>&\sqrt{\lambda^*m-n\sqrt{\left[2-\frac{\theta^*(\theta^*+3\pi)}{2\pi^2}\right]m\log(2n/\delta)}+\mathcal O\left(\sqrt{m\log(1/\delta)}+n\log(2n/\delta)/\sqrt{\delta}\right)}\\\label{Step1_Psi_k_lower_bound}
-&\sqrt{\frac{2\csc\theta^*}{\pi}R^2mn+Rm+\frac{2n}{3}\log(1/\delta)+\mathcal O\left((\sqrt{R}n+1)\sqrt{Rm\log(1/\delta)}+R^4mn\right)}.
\end{align}
In learning over-parameterized high dimensional networks where $m\geqslant n$ and $n$ is sufficiently large. As a consequence, if $R=\sqrt{\pi\lambda^*\sin\theta^*/(2+\varepsilon)n}$ for any $\varepsilon>0$, then the right hand side of (\ref{Step1_Psi_k_lower_bound}) is positive for sufficiently large $n$ and $m$, more explicitly, as $n$ tends to $+\infty$:
\begin{align}\label{infinitely_small_1}
n\sqrt{\left[2-\frac{\theta^*(\theta^*+3\pi)}{2\pi^2}\right]m\log(2n/\delta)}+\mathcal O\left(\sqrt{m\log(1/\delta)}+n\log(2n/\delta)/\sqrt{\delta}\right)=o(\lambda^*m);\\\label{infinitely_small_2}
Rm+\frac{2n}{3}\log(1/\delta)+\mathcal O\left((\sqrt{R}n+1)\sqrt{Rm\log(1/\delta)}+R^4mn\right)=o(2\csc\theta^*R^2mn/\pi).
\end{align}
In other words, $\lambda^*m$ and $2\csc\theta^*R^2mn/\pi$ are the dominant terms, with the remaining terms being negligible compared to them.\\
We first demonstrate the asymptotic behavior (\ref{infinitely_small_1}) as $n$ tends to $+\infty$,
\begin{align}\nonumber
&n\sqrt{\left[2-\frac{\theta^*(\theta^*+3\pi)}{2\pi^2}\right]m\log(2n/\delta)}(\lambda^*m)^{-1}\stackrel{\text{(i)}}{<}\frac{4\sqrt{2}n}{\lambda_{\text{min}}(\boldsymbol X^\top\boldsymbol X)}\sqrt{\frac{\log(2n/\delta)}{m}}\\\label{infinitely_small_1_1}
\stackrel{\text{(ii)}}\lesssim&\sqrt{\frac{\log^{4\hbar}n\log(2n/\delta)}{\log^{1+\varpi}(n/\delta)}}=\sqrt{\frac{\log^{4\hbar}n(\log(n/\delta)+\log2)}{\log^{1+\varpi}(n/\delta)}}=\sqrt{\frac{\log^{4\hbar}n}{\log^{\varpi}(n/\delta)}+\frac{\log2\log^{\hbar}n}{\log^{1+\varpi}(n/\delta)}}=o(1),
\end{align}
where (i) follows from $\theta^*>0$ and $\lambda^*\geqslant\lambda_{\text{min}}(\boldsymbol X^\top\boldsymbol X)/4$ from (\ref{lambda_star_lower_bound}); (ii) uses the condition $\lambda_{\text{min}}(\boldsymbol X^\top\boldsymbol X)=\Omega(\log^{-2\hbar}n)$ for certain $\hbar>0$ and $m=\Omega(n^2\log^{1+\varpi}(n/\delta))$ for any $\varpi>4\hbar$. On the other hand, as $n$ tends to $+\infty$,
\begin{align}\nonumber
&\mathcal O\left(\sqrt{m\log(1/\delta)}+n\log(2n/\delta)/\sqrt{\delta}\right)(\lambda^*m)^{-1}
=\mathcal O\left(\frac{1}{\lambda^*}\sqrt{\frac{\log(1/\delta)}{m}}+\frac{n\log(2n/\delta)/\sqrt{\delta}}{\lambda^*m}\right)\\\nonumber
\stackrel{\text{(i)}}{\lesssim}&\log^{2\hbar}n\left(\sqrt{\frac{\log(1/\delta)}{n^2\log^{1+\varpi}(n/\delta)}}
+\frac{4n\log(2n/\delta)/\sqrt{\delta}}{n^2\log^{1+\varpi}(n/\delta)}\right)\\\label{infinitely_small_1_2}
=&
\frac{\log^{2\hbar}n}{n\sqrt{\log^{\varpi}(n/\delta)}}
+\frac{\log^{2\hbar}n(\log(n/\delta)+\log2)}{n\log^{1+\varpi}(n/\delta)\sqrt{\delta}}=o(1),
\end{align}
where (i) comes from $\theta^*>0$, $\lambda^*\geqslant\lambda_{\text{min}}(\boldsymbol X^\top\boldsymbol X)/4$ from (\ref{lambda_star_lower_bound}) and the condition $\lambda_{\text{min}}(\boldsymbol X^\top\boldsymbol X)=\Omega(\log^{-2\hbar}n)$ for certain $\hbar>0$ and $m=\Omega(n^2\log^{1+\varpi}(n/\delta))$ for any $\varpi>4\hbar$. Combinig (\ref{infinitely_small_1_1}) and (\ref{infinitely_small_1_2}) leads to (\ref{infinitely_small_1}).\\
Turning to (\ref{infinitely_small_2}), in a similar manner, we have as $n$ tends to $+\infty$,
\begin{align}\label{infinitely_small_2_1}
&\frac{Rm}{2\pi^{-1}\csc\theta^*R^2mn}=\frac{\pi\sin\theta^*}{2Rn}\stackrel{\text{(i)}}=
\frac{\pi\sin\theta^*}{2n}\sqrt{\frac{(2+\varepsilon)n}{\pi\lambda^*\sin\theta^*}}=\frac{1}{2}\sqrt{\frac{(2+\varepsilon)\pi\sin\theta^*}{\lambda^*n}}\\\nonumber
\leqslant&\sqrt{\frac{(2+\varepsilon)\pi\sin\theta^*}{\lambda_{\text{min}}(\boldsymbol X^\top\boldsymbol X)n}}\lesssim\sqrt{\frac{(2+\varepsilon)\pi\sin\theta^*\log^{2\hbar}n}{n}}=o(1),
\end{align}
where (i) uses the condition $R=\sqrt{\pi\lambda^*\sin\theta^*/(2+\varepsilon)n}$ for any $\varepsilon>0$.
\begin{align}\label{infinitely_small_2_2}
&\frac{2n\log(1/\delta)/3}{2\pi^{-1}\csc\theta^*R^2mn}\stackrel{\text{(i)}}\lesssim\frac{\pi\sin\theta^*n\log(1/\delta)}{3[\pi\lambda^*\sin\theta^*][(2+\varepsilon)n]^{-1}n^2\log^{1+\varpi}(n/\delta)\cdot n}=\frac{(2+\varepsilon)\log(1/\delta)}{3\lambda^*n\log^{1+\varpi}(n/\delta)}=o(1),
\end{align}
where (i) comes from $R=\sqrt{\pi\lambda^*\sin\theta^*/(2+\varepsilon)n}$ for any $\varepsilon>0$, $\lambda^*\geqslant\lambda_{\text{min}}(\boldsymbol X^\top\boldsymbol X)/4$ from (\ref{lambda_star_lower_bound}) and the condition $\lambda_{\text{min}}(\boldsymbol X^\top\boldsymbol X)=\Omega(\log^{-2\hbar}n)$ for certain $\hbar>0$ and $m=\Omega(n^2\log^{1+\varpi}(n/\delta))$ for any $\varpi>4\hbar$. Moreover, as $n$ tends to $+\infty$,
\begin{align}\nonumber
&\frac{(\sqrt{R}n+1)\sqrt{Rm\log(1/\delta)}+R^4mn}{2\pi^{-1}\csc\theta^*R^2mn}=\frac{\pi\sin\theta^*}{2}\left[\frac{\sqrt{\log(1/\delta)}}{Rm^{1/2}}+\frac{\sqrt{\log(1/\delta)}}{R^{3/2}m^{1/2}n}+R^2\right]\\\nonumber
\stackrel{\text{(i)}}\lesssim&\frac{\pi\sin\theta^*}{2}\bigg\{\frac{\sqrt{\log(1/\delta)}}{\sqrt{\pi\lambda^*\sin\theta^*/(2+\varepsilon)n}\sqrt{n^2\log^{1+\varpi}(n/\delta)}}\\\nonumber
&+\frac{\sqrt{\log(1/\delta)}}{[\pi\lambda^*\sin\theta^*/(2+\varepsilon)n]^{3/4}[n^2\log^{1+\varpi}(n/\delta)]^{1/2}n}+\frac{\pi\lambda^*\sin\theta^*}{(2+\varepsilon)n}\bigg\}\\\nonumber
=&\frac{\pi\sin\theta^*}{2}\bigg\{\frac{\sqrt{(2+\varepsilon)\log(1/\delta)}}{\sqrt{\pi\lambda^*\sin\theta^*\cdot n\log^{1+\varpi}(n/\delta)}}+\left[\frac{2+\varepsilon}{\pi\lambda^*\sin\theta^*}\right]^{3/4}\sqrt{\frac{\log(1/\delta)}{\log^{1+\varpi}(n/\delta)}}\cdot\frac{1}{n^{5/4}}+\frac{\pi\lambda^*\sin\theta^*}{(2+\varepsilon)n}\bigg\}\\\nonumber
=&o(1).
\end{align}
where (i) comes from the conditions $R=\sqrt{\pi\lambda^*\sin\theta^*/(2+\varepsilon)n}$ for any $\varepsilon>0$, $\lambda^*\geqslant\lambda_{\text{min}}(\boldsymbol X^\top\boldsymbol X)/4$ from (\ref{lambda_star_lower_bound}) and the condition $\lambda_{\text{min}}(\boldsymbol X^\top\boldsymbol X)=\Omega(\log^{-2\hbar}n)$ for certain $\hbar>0$ and $m=\Omega(n^2\log^{1+\varpi}(n/\delta))$ for any $\varpi>4\hbar$.\\
Together with (\ref{infinitely_small_2_1}) and (\ref{infinitely_small_2_2}), this enables us to establish (\ref{infinitely_small_2}).
\\With the aid of (\ref{infinitely_small_1}) and (\ref{infinitely_small_2}), we conclude that if $R=\sqrt{\pi\lambda^*\sin\theta^*/(2+\varepsilon)n}$ for any $\varepsilon>0$, then the right hand side of (\ref{Step1_Psi_k_lower_bound}) is positive for sufficiently large $n$ and $m$. Therefore, with probability exceeding $1-(mn-m+4n)\delta$, we can derive
\begin{align}\nonumber
&\sigma_{\min}^2(\boldsymbol\Psi(k))\\\nonumber
>&\bigg\{\sqrt{\lambda^*m-n\sqrt{\left[2-\frac{\theta^*(\theta^*+3\pi)}{2\pi^2}\right]m\log(2n/\delta)}+\mathcal O\left(\sqrt{m\log(1/\delta)}+n\log(2n/\delta)/\sqrt{\delta}\right)}\\\nonumber
-&\sqrt{\frac{2\csc\theta^*}{\pi}R^2mn+Rm+\frac{2n}{3}\log(1/\delta)+\mathcal O\left((\sqrt{R}n+1)\sqrt{Rm\log(1/\delta)}+R^4mn\right)}\bigg\}^2\\\nonumber
=&\lambda^*m-n\sqrt{\left[2-\frac{\theta^*(\theta^*+3\pi)}{2\pi^2}\right]m\log(2n/\delta)}+\mathcal O\left(\sqrt{m\log(1/\delta)}+n\log(2n/\delta)/\sqrt{\delta}\right)
\\\nonumber
&+\frac{2\csc\theta^*}{\pi}R^2mn+Rm+\frac{2n}{3}\log(1/\delta)+\mathcal O\left((\sqrt{R}n+1)\sqrt{Rm\log(1/\delta)}+R^4mn\right)\\\nonumber
&-2\sqrt{\lambda^*m-n\sqrt{\left[2-\frac{\theta^*(\theta^*+3\pi)}{2\pi^2}\right]m\log(2n/\delta)}+\mathcal O\left(\sqrt{m\log(1/\delta)}+n\log(2n/\delta)/\sqrt{\delta}\right)}\\\nonumber
&\cdot\sqrt{\frac{2\csc\theta^*}{\pi}R^2mn+Rm+\frac{2n}{3}\log(1/\delta)+\mathcal O\left((\sqrt{R}n+1)\sqrt{Rm\log(1/\delta)}+R^4mn\right)}\\\nonumber
=&\left[\lambda^*+\frac{2\csc\theta^*}{\pi}R^2n\right]m-n\sqrt{\left[2-\frac{\theta^*(\theta^*+3\pi)}{2\pi^2}\right]m\log(2n/\delta)}+Rm-2\sqrt{\textcolor{red}{\clubsuit}}\\\nonumber
&+\mathcal O\left(Rn\sqrt{m\log(1/\delta)}+R^4mn+\sqrt{m\log(1/\delta)}
+n\log(2n/\delta)/\sqrt{\delta}+n\log(1/\delta)+\sqrt{Rm\log(1/\delta)}\right)\\\nonumber
\stackrel{\text{(i)}}{=}&\frac{4+\varepsilon}{2+\varepsilon}\lambda^*m-n\sqrt{\left[2-\frac{\theta^*(\theta^*+3\pi)}{2\pi^2}\right]m\log(2n/\delta)}+Rm-2\sqrt{\textcolor{red}{\clubsuit}}\\\label{step1_estimate}
&+\mathcal O\left(Rn\sqrt{m\log(1/\delta)}+R^4mn+n\log(2n/\delta)/\sqrt{\delta}\right),
\end{align}
where (i) comes from $R=\sqrt{(\pi\lambda^*\sin\theta^*)/[(2+\varepsilon)n]}$ for any $\varepsilon>0$ and $m=\Omega(n^2\log^{1+\varpi}(n/\delta))$ for any $\varpi>4\hbar>0$; whereas the asymptotic relation in (\ref{step1_estimate}) is due to the following three asymptotic relations:
\begin{align}\nonumber
&\sqrt{m\log(1/\delta)}=o(\sqrt{[\pi\lambda^*\sin\theta^*n/(2+\varepsilon)]m\log(1/\delta)})
\\\nonumber
\Longrightarrow&\sqrt{m\log(1/\delta)}=o(Rn\sqrt{m\log(1/\delta)}),\\\nonumber
&n\log(1/\delta)=o(n\log(2n/\delta)/\sqrt{\delta}),\\\nonumber &\sqrt{Rm\log(1/\delta)}=\sqrt[4]{(\pi\lambda^*\sin\theta^*)/[(2+\varepsilon)n]}\sqrt{m\log(1/\delta)}\\\nonumber
=&o(\sqrt{\pi\lambda^*\sin\theta^*n/(2+\varepsilon)}\sqrt{m\log(1/\delta)})\\\nonumber
\Longrightarrow&\sqrt{Rm\log(1/\delta)}=o(Rn\sqrt{m\log(1/\delta)}).
\end{align}
To proceed, we can evaluate $\textcolor{red}{\clubsuit}$ as
\begin{align}\nonumber
\textcolor{red}{\clubsuit}=&\left[\lambda^*m-n\sqrt{\left[2-\frac{\theta^*(\theta^*+3\pi)}{2\pi^2}\right]m\log(2n/\delta)}+\mathcal O\left(\sqrt{m\log(1/\delta)}+n\log(2n/\delta)/\sqrt{\delta}\right)\right]\\\nonumber
&\cdot\left[\frac{2\csc\theta^*}{\pi}R^2mn+Rm+\frac{2n}{3}\log(1/\delta)+\mathcal O\left((\sqrt{R}n+1)\sqrt{Rm\log(1/\delta)}+R^4mn\right)\right]\\\nonumber
\stackrel{\text{(ii)}}{<}&\left[\lambda^*m-\frac{3}{4}n\sqrt{2m\log(2n/\delta)}+\mathcal O\left(\sqrt{m\log(1/\delta)}+n\log(2n/\delta)/\sqrt{\delta}\right)\right]\\\nonumber
\cdot&\left[\frac{2\csc\theta^*}{\pi}R^2mn+Rm+\mathcal O\left(Rn\sqrt{m\log(1/\delta)}+R^4mn\right)\right].
\end{align}
Here (ii) results from $\theta^*\leqslant\pi/2$, $\sqrt{Rm\log(1/\delta)}=o(Rn\sqrt{m\log(1/\delta)})$ and $$R^4mn\gtrsim\frac{(\pi\lambda^*\sin\theta^*)^2}{(2+\varepsilon)^2n^2} n^2\log^{1+\varpi}(n/\delta)n=\frac{(\pi\lambda^*\sin\theta^*)^2}{(2+\varepsilon)^2} n\log^{1+\varpi}(n/\delta)\gtrsim2n\log(1/\delta)/3.$$
Furthermore, there exists a constant $c_1$ for sufficiently large $n$ such that
\begin{align}\nonumber
&\lim_{n\to+\infty}\frac{\sqrt{m\log(1/\delta)}}{n\sqrt{2m\log(2n/\delta)}}
=\lim_{n\to+\infty}\frac{\sqrt{\log(1/\delta)}}{n\sqrt{2\log(2n/\delta)}}=0,\\\nonumber
&\lim_{n\to+\infty}\frac{n\log(2n/\delta)/\sqrt{\delta}}{n\sqrt{2m\log(2n/\delta)}}=\lim_{n\to+\infty}\sqrt{\frac{\log(2n/\delta)}{2m\delta}}\leqslant\lim_{n\to+\infty}\sqrt{\frac{\log(2n/\delta)}{2c_1n^2\log^{1+\varpi}(n/\delta)\delta}}=0,\\\nonumber
&\lim_{n\to+\infty}\frac{Rn\sqrt{m\log(1/\delta)}}{Rm}=\lim_{n\to+\infty}n\sqrt{\frac{\log(1/\delta)}{m}}\leqslant\lim_{n\to+\infty}\sqrt{\frac{\log(1/\delta)}{c_1\log^{1+\varpi}(n/\delta)\delta}}=0,
\\\nonumber
&\lim_{n\to+\infty}\frac{R^4mn}{Rm}=\lim_{n\to+\infty}\frac{1}{Rn}=\lim_{n\to+\infty}\frac{1}{n}\sqrt{\frac{(2+\varepsilon)n}{\pi\lambda^*\sin\theta^*}}=0.
\end{align} 
giving rise to
\begin{align}\nonumber
&\sqrt{m\log(1/\delta)}=o(n\sqrt{2m\log(2n/\delta)}),n\log(2n/\delta)/\sqrt{\delta}=o(n\sqrt{2m\log(2n/\delta)}),
\\\nonumber
&Rn\sqrt{m\log(1/\delta)}=o(Rm), R^4mn=o(Rm).
\end{align}
This further implies that
\begin{align}\nonumber
\textcolor{red}{\clubsuit}<&\left[\lambda^*m-\frac{3}{4}n\sqrt{2m\log(2n/\delta)}+o\left(n\sqrt{2m\log(2n/\delta)}\right)\right]
\left[\frac{2\csc\theta^*}{\pi}R^2mn+Rm+o\left(Rm\right)\right]\\\nonumber
=&\left[\lambda^*m+\mathcal O\left(n\sqrt{m\log(2n/\delta)}\right)\right]
\left[\frac{2\csc\theta^*}{\pi}R^2mn+Rm+o\left(Rm\right)\right]\\\nonumber
=&\frac{2\lambda^*\csc\theta^*}{\pi}R^2m^2n+\lambda^*Rm^2+o(Rm^2)\\\nonumber
&+\mathcal O\left(R^2mn^2\sqrt{m\log(2n/\delta)}\right)+\mathcal O\left(Rmn\sqrt{m\log(2n/\delta)}\right)+o(Rmn\sqrt{m\log(2n/\delta)})\\\nonumber
=&\frac{2\lambda^*\csc\theta^*}{\pi}R^2m^2n+\lambda^*Rm^2+\mathcal O\left(R^2mn^2\sqrt{m\log(2n/\delta)}+Rmn\sqrt{m\log(2n/\delta)}\right)+o(Rm^2)\\\label{step1_estimate_7}
\stackrel{\text{(i)}}=&\frac{2\lambda^*\csc\theta^*}{\pi}R^2m^2n+\lambda^*Rm^2+\mathcal O\left(R^2mn^2\sqrt{m\log(2n/\delta)}\right)+o(Rm^2).
\end{align}
Here (i) results from
\begin{align}
&\lim_{n\to+\infty}\frac{Rmn\sqrt{m\log(2n/\delta)}}{R^2mn^2\sqrt{m\log(2n/\delta)}}=\lim_{n\to+\infty}\frac{1}{Rn}=\lim_{n\to+\infty}\frac{1}{n}\sqrt{\frac{(2+\varepsilon)n}{\pi\lambda^*\sin\theta^*}}=0.
\end{align}
Putting (\ref{step1_estimate_7}) in (\ref{step1_estimate}), we arrive at
\begin{align}\nonumber
&\sigma_{\min}^2(\boldsymbol\Psi(k))\\\nonumber
>&\frac{4+\varepsilon}{2+\varepsilon}\lambda^*m-n\sqrt{\left[2-\frac{\theta^*(\theta^*+3\pi)}{2\pi^2}\right]m\log(2n/\delta)}+Rm\\\nonumber
&-2\sqrt{\frac{2\lambda^*\csc\theta^*}{\pi}R^2m^2n+\lambda^*Rm^2+\mathcal O\left(R^2mn^2\sqrt{m\log(2n/\delta)}\right)+o(Rm^2)}\\\nonumber
&+\mathcal O\left(Rn\sqrt{m\log(1/\delta)}+R^4mn+n\log(2n/\delta)/\sqrt{\delta}\right)\\\nonumber
=&\frac{4+\varepsilon}{2+\varepsilon}\lambda^*m-n\sqrt{\left[2-\frac{\theta^*(\theta^*+3\pi)}{2\pi^2}\right]m\log(2n/\delta)}+Rm\\\nonumber
&-2\sqrt{\frac{2\lambda^*\csc\theta^*}{\pi}R^2m^2n\left[1+\frac{\pi\sin\theta^*}{2Rn}+\mathcal O\left(n\sqrt{\frac{\log(2n/\delta)}{m}}\right)+o\left(\frac{1}{Rn}\right)\right]}\\\nonumber
&+\mathcal O\left(Rn\sqrt{m\log(1/\delta)}+R^4mn+n\log(2n/\delta)/\sqrt{\delta}\right)\\\nonumber
\stackrel{\text{(i)}}\geqslant&\frac{4+\varepsilon}{2+\varepsilon}\lambda^*m-n\sqrt{\left[2-\frac{\theta^*(\theta^*+3\pi)}{2\pi^2}\right]m\log(2n/\delta)}+Rm\\\nonumber
&-2Rm\sqrt{\frac{2\lambda^*\csc\theta^*}{\pi}n}\left[1+\frac{\pi\sin\theta^*}{4Rn}+\mathcal O\left(n\sqrt{\frac{\log(2n/\delta)}{m}}\right)+o\left(\frac{1}{Rn}\right)\right]\\\nonumber
&+\mathcal O\left(Rn\sqrt{m\log(1/\delta)}+R^4mn+n\log(2n/\delta)/\sqrt{\delta}\right)\\\nonumber
=&\frac{4+\varepsilon}{2+\varepsilon}\lambda^*m-2Rm\sqrt{\frac{2\lambda^*\csc\theta^*}{\pi}n}-n\sqrt{\left[2-\frac{\theta^*(\theta^*+3\pi)}{2\pi^2}\right]m\log(2n/\delta)}-m\sqrt{\frac{\lambda^*\sin\theta^*}{2\pi n}}+Rm\\\nonumber
&+\mathcal O\left(Rn^{3/2}\sqrt{m\log(2n/\delta)}\right)+o(m/\sqrt{n})+\mathcal O\left(Rn\sqrt{m\log(1/\delta)}+R^4mn+n\log(2n/\delta)/\sqrt{\delta}\right)\\\nonumber
\stackrel{\text{(ii)}}=&\left(\frac{4+\varepsilon}{2+\varepsilon}-2\sqrt{\frac{2}{2+\varepsilon}}\right)\lambda^*m-n\sqrt{\left[2-\frac{\theta^*(\theta^*+3\pi)}{2\pi^2}\right]m\log(2n/\delta)}-m\sqrt{\frac{\lambda^*\sin\theta^*}{2\pi n}}+\sqrt{\frac{\pi\lambda^*\sin\theta^*}{(2+\varepsilon)n}}m\\\nonumber
&+\mathcal O\left(Rn^{3/2}\sqrt{m\log(2n/\delta)}+Rn\sqrt{m\log(1/\delta)}+R^4mn+n\log(2n/\delta)/\sqrt{\delta}\right)+o(m/\sqrt{n}).
\\\nonumber
\stackrel{\text{(iii)}}=&\left(1-\sqrt{\frac{2}{2+\varepsilon}}\right)^2\lambda^*m+\left(\sqrt{\frac{\pi}{2+\varepsilon}}-\sqrt{\frac{1}{2\pi}}+o(1)\right)m\sqrt{\frac{\lambda^*\sin\theta^*}{n}}\\
&+\mathcal O\left(n\sqrt{m\log(2n/\delta)}+n\log(2n/\delta)/\sqrt{\delta}\right).
\end{align}
Here, (i) makes use of {\textcolor{red}{Bernoulli's inequality}: $$(1+x)^\alpha\leqslant1+\alpha x\quad\text{for } x\geqslant-1\text{ and }0<\alpha<1.$$
and there exists a constant $c_1$ for sufficiently large $n$ such that
\begin{align}\nonumber
&\lim_{n\to+\infty}\frac{1}{Rn}
=\lim_{n\to+\infty}\frac{1}{\sqrt{\pi\lambda^*\sin\theta^*/(2+\varepsilon)n}\cdot n}=0;\\\nonumber
&\lim_{n\to+\infty}n\sqrt{\frac{\log(2n/\delta)}{m}}\leqslant\lim_{n\to+\infty}n\sqrt{\frac{\log(2n/\delta)}{c_1n^2\log^{1+\varpi}(n/\delta)}}=
\lim_{n\to+\infty}\sqrt{\frac{\log(2n/\delta)}{c_1\log^{1+\varpi}(n/\delta)}}=0,
\end{align}
yielding that when $n$ is sufficiently large,
$$\frac{\pi\sin\theta^*}{2Rn}+\mathcal O\left(n\sqrt{\frac{\log(2n/\delta)}{m}}\right)+o\left(\frac{1}{Rn}\right)>-1.$$
(ii) uses $R=\sqrt{\pi\lambda^*\sin\theta^*/(2+\varepsilon)n}$ for any $\varepsilon>0$ and (iii) depends on the following asymptotic relations
\begin{align}\nonumber
&Rn\sqrt{m\log(2n/\delta)}=o(Rn^{3/2}\sqrt{m\log(2n/\delta)}), \quad R^4mn=o(m/\sqrt{n});\\\nonumber
&Rn^{3/2}\sqrt{m\log(2n/\delta)}=\Theta(n\sqrt{m\log(2n/\delta)}).
\end{align}
Therefore, we complete the proof of Lemma \ref{lemma_Psi_k_lower_bound}.
\\\\
\textcolor{blue}{{\bf Step 2: Bounding $\sigma_{\max}^2(\boldsymbol\Psi(k))$}}
\begin{lemma}\label{lemma_Psi_k_upper_bound}
(\textcolor{blue}{The Largest Singular Value of Activation Matrix}) For any weight vector $\boldsymbol w_r\in\mathbb R^d$ $(r\in[m])$ and any $k\geqslant1$, $\|\boldsymbol w_r(k)-\boldsymbol w_r(0)\|_2\leqslant R:=\sqrt{(\pi\lambda^*\sin\theta^*)/[(2+\varepsilon)n]}$ for any $\varepsilon>0$, if $m=\Omega(n^2\log^{1+\varpi}(n/\delta))$ for any $\varpi>0$, then for any $\delta\in(0,1)$, with probability exceeding $1-(4n-2)\delta$ over the random initialization, one has: 
\begin{align}\nonumber
\sigma_{\max}^2(\boldsymbol\Psi(k))<&\frac{mn}{2}+2m\sqrt{\lambda^*n}+2\lambda^*m+\mathcal O\left(n\sqrt{m\log(1/\delta)}\right),
\end{align}
where $\lambda^*=\lambda_{\text{min}}(\mho)$ and the matrix $\mho$ is defined in (\ref{matrix_mho_definition}) in Section \ref{section_problem}.
\end{lemma}
{\bf Remark}: Similar to Lemma \ref{lemma_Psi_k_lower_bound}, Lemma \ref{lemma_Psi_k_upper_bound} reveals the effect of $m,n$ and $\lambda^*$ on controlling the largest singular value of random activation matrix $\boldsymbol\Psi(k)$.\\
\textbf{Proof}:
Applying \textcolor{red}{Weyl's inequality}(\cite{weyl_1912}) and a union bound, followed from Lemmas \ref{lemma_Psi0} and \ref{lemma_Psik_Psi0}, we conclude that: for any $k\in\mathbb Z^{+}$, with probability exceeding $1-n\delta-(3n-2)\delta=1-(4n-2)\delta$ over the random initialization,
\begin{align}\nonumber
&\sigma_{\max}(\boldsymbol\Psi(k))\\\nonumber
\leqslant&\sigma_{\max}(\boldsymbol\Psi(0))+\|\boldsymbol\Psi(k)-\boldsymbol\Psi(0)\|_2\leqslant\|\boldsymbol\Psi(0))\|_{\text{F}}+\|\boldsymbol\Psi(k)-\boldsymbol\Psi(0)\|_2\\\nonumber
\leqslant&\sqrt{\frac{mn}{2}+n\sqrt{\frac{m}{2}\log\frac{1}{\delta}}}\\\label{Step2_Psi_k_upper_bound}
&+\sqrt{\frac{2\csc\theta^*}{\pi}R^2mn+Rm+\frac{2n}{3}\log(1/\delta)+\mathcal O\left((\sqrt{R}n+1)\sqrt{Rm\log(1/\delta)}+R^4mn\right)}.
\end{align}
We know from Step 1 of deriving the lower bound of $\sigma_{\min}(\boldsymbol\Psi(k))$ that as $n$ tends to $+\infty$:
\begin{align}\nonumber
Rm+\frac{2n}{3}\log(1/\delta)+\mathcal O\left((\sqrt{R}n+1)\sqrt{Rm\log(1/\delta)}+R^4mn\right)=o(2\csc\theta^*R^2mn/\pi).
\end{align}
Hence $\sigma_{\max}(\boldsymbol\Psi(k))$ can be upper bounded as
\begin{align}\nonumber
&\sigma_{\max}(\boldsymbol\Psi(k))\\\nonumber
\leqslant&\sqrt{\frac{mn}{2}+n\sqrt{\frac{m}{2}\log\frac{1}{\delta}}}+\sqrt{(1+o(1))\frac{2\csc\theta^*}{\pi}R^2mn}\\\nonumber
=&\sqrt{\frac{mn}{2}}\left[1+\sqrt{\frac{2\log(1/\delta)}{m}}\right]^{1/2}+\sqrt{(1+o(1))\frac{2}{2+\varepsilon}\lambda^*m}\\\nonumber
\stackrel{\text{(i)}}<&\sqrt{\frac{mn}{2}}\left[1+\mathcal O\left(\sqrt{\frac{2\log(1/\delta)}{m}}\right)\right]+\sqrt{2\lambda^*m}\\\nonumber
=&\sqrt{mn/2}+\sqrt{2\lambda^*m}+\mathcal O\left(\sqrt{n\log(1/\delta)}\right).
\end{align}
where (i) comes from that
$$
\sqrt{1+\alpha}=1+\frac{\alpha}{2}+\mathcal O(\alpha^2)=1+\mathcal O(\alpha),\text{ if }\alpha\to0.
$$
Henceforth, it follows that
\begin{align}\nonumber
&\sigma_{\max}^2(\boldsymbol\Psi(k))\\\nonumber
<&\left[\sqrt{mn/2}+\sqrt{2\lambda^*m}+\mathcal O\left(\sqrt{n\log(1/\delta)}\right)\right]^2\\\nonumber
=&\frac{mn}{2}\left[1+2\sqrt{\frac{\lambda^*}{n}}+\mathcal O\left(\sqrt{\frac{\log(1/\delta)}{m}}\right)\right]^2\\\nonumber
=&\frac{mn}{2}\left\{1+4\sqrt{\frac{\lambda^*}{n}}+\mathcal O\left(\sqrt{\frac{\log(1/\delta)}{m}}\right)+\left[2\sqrt{\frac{\lambda^*}{n}}+\mathcal O\left(\sqrt{\frac{\log(1/\delta)}{m}}\right)\right]^2\right\}\\\nonumber
=&\frac{mn}{2}\left\{1+4\sqrt{\frac{\lambda^*}{n}}+\mathcal O\left(\sqrt{\frac{\log(1/\delta)}{m}}\right)+\frac{4\lambda^*}{n}\left[1+\mathcal O\left(\sqrt{\frac{n\log(1/\delta)}{m}}\right)\right]^2\right\}\\\nonumber
=&\frac{mn}{2}\left\{1+4\sqrt{\frac{\lambda^*}{n}}+\mathcal O\left(\sqrt{\frac{\log(1/\delta)}{m}}\right)+\frac{4\lambda^*}{n}\left[1+\mathcal O\left(\sqrt{\frac{n\log(1/\delta)}{m}}\right)+\mathcal O\left(\frac{n\log(1/\delta)}{m}\right)\right]\right\}\\\nonumber
=&\frac{mn}{2}\left\{1+4\sqrt{\frac{\lambda^*}{n}}+\frac{4\lambda^*}{n}+\mathcal O\left(\sqrt{\frac{\log(1/\delta)}{m}}\right)+\mathcal O\left(\sqrt{\frac{\log(1/\delta)}{mn}}\right)+\mathcal O\left(\frac{\log(1/\delta)}{m}\right)\right\}\\\nonumber
=&\frac{mn}{2}\left\{1+4\sqrt{\frac{\lambda^*}{n}}+\frac{4\lambda^*}{n}+\mathcal O\left(\sqrt{\frac{\log(1/\delta)}{m}}\right)\right\}\\\nonumber
=&\frac{mn}{2}+2m\sqrt{\lambda^*n}+2\lambda^*m+\mathcal O\left(n\sqrt{m\log(1/\delta)}\right).
\end{align}
This completes the proof of Lemma \ref{lemma_Psi_k_upper_bound}.
\\\\
\textcolor{blue}{{\bf Step 3: Bounding $\|\boldsymbol\Psi(k)\|_{\text{F}}^2$}}\\
We now move on to bounding $\|\boldsymbol\Psi(k)\|_{\text{F}}^2$ via bounds in Lemmas \ref{lemma_Psi0} and \ref{lemma_Psik_Psi0}. Using a union bound, we have that with probability exceeding $1-2n\delta$ over the random initialization,
\begin{align}\nonumber
&\|\boldsymbol\Psi(k)\|_{\text{F}}^2\\\nonumber
=&\|\boldsymbol\Psi(0)+[\boldsymbol\Psi(k)-\boldsymbol\Psi(0)]\|_{\text{F}}^2\\\nonumber
=&\|\boldsymbol\Psi(0)\|_{\text{F}}^2
+2\text{tr}\left[\boldsymbol\Psi(0)^\top(\boldsymbol\Psi(k)-\boldsymbol\Psi(0))\right]
+\|\boldsymbol\Psi(k)-\boldsymbol\Psi(0)\|_{\text{F}}^2\\\nonumber
\leqslant&\|\boldsymbol\Psi(0)\|_{\text{F}}^2+\|\boldsymbol\Psi(k)-\boldsymbol\Psi(0)\|_{\text{F}}^2+2\|\boldsymbol\Psi(0)\|_{\text{F}}\|\boldsymbol\Psi(k)-\boldsymbol\Psi(0)\|_{\text{F}}\\\nonumber
\leqslant&\frac{mn}{2}+n\sqrt{\frac{m}{2}\log\frac{1}{\delta}}
+\wp(R)mn+\frac{2}{3}n\log\frac{1}{\delta}+n\sqrt{2\wp(R)[1-\wp(R)]m\log\frac{1}{\delta}}\\\nonumber
&+2\sqrt{\left[\frac{mn}{2}+n\sqrt{\frac{m}{2}\log\frac{1}{\delta}}\right]
\left[\wp(R)mn+\frac{2}{3}n\log\frac{1}{\delta}+n\sqrt{2\wp(R)[1-\wp(R)]m\log\frac{1}{\delta}}\right] }\\\label{step3_upper_estimate}
\stackrel{\text{(i)}}{<}&\left(\frac{1}{2}+R\right)mn+n\sqrt{\frac{m}{2}\log\frac{1}{\delta}}+n\sqrt{2Rm\log\frac{1}{\delta}}+\frac{2}{3}n\log\frac{1}{\delta}+2\sqrt{\textcolor{red}{\clubsuit}},
\end{align}
where (i) are due to $\wp(R)<R$ (see (\ref{wp_R_bound})) and $2\wp(R)[1-\wp(R)]<2\wp(R)<2R$.\\
Turning to \textcolor{red}{$\clubsuit$}, we have
\begin{align}\nonumber
&\textcolor{red}{\clubsuit}\\\nonumber
=&\left[\frac{mn}{2}+n\sqrt{\frac{m}{2}\log\frac{1}{\delta}}\right]
\left[\wp(R)mn+\frac{2}{3}n\log\frac{1}{\delta}+n\sqrt{2\wp(R)[1-\wp(R)]m\log\frac{1}{\delta}}\right]\\\nonumber
<&\left[\frac{mn}{2}+n\sqrt{\frac{m}{2}\log\frac{1}{\delta}}\right]
\left[Rmn+n\sqrt{2Rm\log\frac{1}{\delta}}+\frac{2}{3}n\log\frac{1}{\delta}\right]\\\label{est_1221_1}
=&Rm^2n^2\underbrace{\left[\frac{1}{2}+\sqrt{\frac{\log(1/\delta)}{2m}}\right]
\left[1+\sqrt{\frac{2\log(1/\delta)}{Rm}}+\frac{2\log(1/\delta)}{3Rm}\right]}_{\textcolor{blue}{\spadesuit}},
\end{align}
where
\begin{align}\nonumber
&\textcolor{blue}{\spadesuit}\\\nonumber
=&\frac{1}{2}+\sqrt{\frac{\log(1/\delta)}{2Rm}}
+\sqrt{\frac{\log(1/\delta)}{2m}}+\frac{\log(1/\delta)}{3Rm}+\frac{\log(1/\delta)}{\sqrt{R}m}+\frac{\sqrt{2}\log^{3/2}(1/\delta)}{3Rm^{3/2}}\\\nonumber
=&\frac{1}{2}\left[1+\sqrt{\frac{2\log(1/\delta)}{Rm}}
+\sqrt{\frac{2\log(1/\delta)}{m}}+\frac{2\log(1/\delta)}{3Rm}+\frac{2\log(1/\delta)}{\sqrt{R}m}+\frac{2\sqrt{2}\log^{3/2}(1/\delta)}{3Rm^{3/2}}\right]\\\nonumber
\stackrel{\text{(i)}}=&\frac{1}{2}\left[1+\sqrt{\frac{2\log(1/\delta)}{Rm}}
+\sqrt{\frac{2\log(1/\delta)}{m}}+\mathcal O\left(\frac{\log(1/\delta)}{Rm}\right)\right],
\end{align}
where (i) uses $R=\sqrt{\pi\lambda^*\sin\theta^*/(2+\varepsilon)n}$ for any $\varepsilon>0$. We therefore have
\begin{align}\nonumber
2\sqrt{\textcolor{red}{\clubsuit}}=&\sqrt{2R}mn\left[1+\sqrt{\frac{\log(1/\delta)}{2Rm}}
+\sqrt{\frac{\log(1/\delta)}{2m}}+\mathcal O\left(\frac{\log(1/\delta)}{Rm}\right)\right]^{1/2}\\\nonumber
\stackrel{\text{(i)}}=&\sqrt{2R}mn\bigg\{1+\sqrt{\frac{2\log(1/\delta)}{Rm}}
+\sqrt{\frac{2\log(1/\delta)}{m}}+\mathcal O\left(\frac{\log(1/\delta)}{Rm}\right)\\\nonumber
&+\left(\sqrt{\frac{2\log(1/\delta)}{Rm}}
+\sqrt{\frac{2\log(1/\delta)}{m}}+\mathcal O\left(\frac{\log(1/\delta)}{Rm}\right)\right)^2\bigg\}\\\nonumber
=&\sqrt{2R}mn\left[1+\sqrt{\frac{2\log(1/\delta)}{Rm}}
+\sqrt{\frac{2\log(1/\delta)}{m}}+\mathcal O\left(\frac{\log(1/\delta)}{Rm}\right)\right]\\\label{est_1221_2}
=&\sqrt{2R}mn+2n\sqrt{m\log(1/\delta)}+2n\sqrt{Rm\log(1/\delta)}+\mathcal O\left(\frac{n\log(1/\delta)}{R}\right).
\end{align}
where (i) comes from that
$$
\sqrt{1+\alpha}=1+\frac{\alpha}{2}+\mathcal O(\alpha^2),\text{ if }\alpha\to0.
$$
On substituting (\ref{est_1221_2}) into (\ref{step3_upper_estimate}), one obtains that with probability exceeding $1-2n\delta$,
\begin{align}\nonumber
&\|\boldsymbol\Psi(k)\|_{\text{F}}^2\\\nonumber
<&\left(\frac{1}{2}+R\right)mn+n\sqrt{\frac{m}{2}\log\frac{1}{\delta}}+n\sqrt{2Rm\log\frac{1}{\delta}}+\frac{2}{3}n\log\frac{1}{\delta}\\\nonumber
&+\sqrt{2R}mn+2n\sqrt{m\log(1/\delta)}+2n\sqrt{Rm\log(1/\delta)}+\mathcal O\left(\frac{n\log(1/\delta)}{R}\right)\\\nonumber
=&\left[\frac{1}{\sqrt{2}}+\sqrt{R}\right]^2mn+\left[2+\frac{1}{\sqrt{2}}\right]n\sqrt{m\log(1/\delta)}\\\label{step3_upper_estimate_2}
&+(2+\sqrt{2})n\sqrt{Rm\log(1/\delta)}+\mathcal O\left(n\log(1/\delta)/R\right).
\end{align}
As for the lower bound of $\|\boldsymbol\Psi(k)\|_{\text{F}}^2$, applying Lemmas \ref{lemma_Psi0} and \ref{lemma_Psik_Psi0} and using a union bound, with probability exceeding $1-4n\delta$,
\begin{align}\nonumber
&\|\boldsymbol\Psi(k)\|_{\text{F}}^2\\\nonumber
=&\|\boldsymbol\Psi(0)+[\boldsymbol\Psi(k)-\boldsymbol\Psi(0)]\|_{\text{F}}^2\\\nonumber
=&\|\boldsymbol\Psi(0)\|_{\text{F}}^2
+2\text{tr}\left[\boldsymbol\Psi(0)^\top(\boldsymbol\Psi(k)-\boldsymbol\Psi(0))\right]
+\|\boldsymbol\Psi(k)-\boldsymbol\Psi(0)\|_{\text{F}}^2\\\nonumber
\geqslant&\|\boldsymbol\Psi(0)\|_{\text{F}}^2+\|\boldsymbol\Psi(k)-\boldsymbol\Psi(0)\|_{\text{F}}^2-2\|\boldsymbol\Psi(0)\|_{\text{F}}\|\boldsymbol\Psi(k)-\boldsymbol\Psi(0)\|_{\text{F}}\\\nonumber
\geqslant&\frac{mn}{2}-n\sqrt{\frac{m}{2}\log\frac{1}{\delta}}
+\wp(R)mn-\frac{2}{3}n\log\frac{1}{\delta}-n\sqrt{2\wp(R)[1-\wp(R)]m\log\frac{1}{\delta}}\\\nonumber
&-2\sqrt{\left[\frac{mn}{2}+n\sqrt{\frac{m}{2}\log\frac{1}{\delta}}\right]
\left[\wp(R)mn+\frac{2}{3}n\log\frac{1}{\delta}+n\sqrt{2\wp(R)[1-\wp(R)]m\log\frac{1}{\delta}}\right] }\\\label{step3_lower_estimate_1}
\stackrel{\text{(i)}}{>}&\left(\frac{1}{2}+R\right)mn-n\sqrt{\frac{m}{2}\log\frac{1}{\delta}}-n\sqrt{\frac{7}{10}Rm\log\frac{1}{\delta}}-\frac{2}{3}n\log\frac{1}{\delta}-2\sqrt{\textcolor{red}{\clubsuit}},
\end{align}
where (i) holds since $$2\wp(R)[1-\wp(R)]>2(7R/10)[1-\wp(R)]>(7R/5)[1-R]>(7R/5)\cdot1/2=7R/10$$ which is due to $0<7R/10<\wp(R)<R<1/2<5/7$ in (\ref{wp_R_bound}).\\
Substituting (\ref{est_1221_2}) into (\ref{step3_lower_estimate_1}), one obtains that with probability exceeding $1-4n\delta$,
\begin{align}\nonumber
&\|\boldsymbol\Psi(k)\|_{\text{F}}^2\\\nonumber
>&\left(\frac{1}{2}+R\right)mn-n\sqrt{\frac{m}{2}\log\frac{1}{\delta}}-n\sqrt{\frac{7}{10}Rm\log\frac{1}{\delta}}-\frac{2}{3}n\log\frac{1}{\delta}\\\nonumber
&-\sqrt{2R}mn-2n\sqrt{m\log(1/\delta)}-2n\sqrt{Rm\log(1/\delta)}+\mathcal O\left(\frac{n\log(1/\delta)}{R}\right)\\\nonumber
=&\left[\frac{1}{\sqrt{2}}-\sqrt{R}\right]^2mn-\left[2+\frac{1}{\sqrt{2}}\right]n\sqrt{m\log(1/\delta)}\\\label{step3_lower_estimate_2}
&-(2+\sqrt{7/10})n\sqrt{Rm\log(1/\delta)}+\mathcal O\left(n\log(1/\delta)/R\right).
\end{align}
\\
\textcolor{blue}{{\bf Step 4: Bounding $\sum_{i=1}^n\|\widetilde{\boldsymbol\psi_i}(k)\|_2^4$}}\\
With regards to $\|\widetilde{\boldsymbol\psi_i}(k)\|_2^4$, applying Lemmas \ref{lemma_Psi0} and \ref{lemma_Psik_Psi0} with similar computations in Step 2 and a union bound gives, with probability exceeding $1-2\delta$ over the random initialization,
\begin{align}\nonumber
&\|\widetilde{\boldsymbol\psi_i}(k)\|_2^2\\\nonumber
=&\|\widetilde{\boldsymbol\psi_i}(0)+[\widetilde{\boldsymbol\psi_i}(k)-\widetilde{\boldsymbol\psi_i}(0)]\|_2^2\\\nonumber
=&\|\widetilde{\boldsymbol\psi_i}(0)\|_2^2
+2\langle\widetilde{\boldsymbol\psi_i}(0), \widetilde{\boldsymbol\psi_i}(k)-\widetilde{\boldsymbol\psi_i}(0)\rangle
+\|\widetilde{\boldsymbol\psi_i}(k)-\widetilde{\boldsymbol\psi_i}(0)\|_2^2\\\nonumber
\leqslant&\|\widetilde{\boldsymbol\psi_i}(0)\|_2^2
+\|\widetilde{\boldsymbol\psi_i}(k)-\widetilde{\boldsymbol\psi_i}(0)\|_2^2
+2\|\widetilde{\boldsymbol\psi_i}(0)\|_2\|\widetilde{\boldsymbol\psi_i}(k)-\widetilde{\boldsymbol\psi_i}(0)\|_2\\\nonumber
\leqslant&\frac{m}{2}+\sqrt{\frac{m}{2}\log\frac{1}{\delta}}+\wp(R)m+\frac{2}{3}\log\frac{1}{\delta}+\sqrt{2\wp(R)[1-\wp(R)]m\log\frac{1}{\delta}}\\\nonumber
&+2\sqrt{\left[\frac{m}{2}+\sqrt{\frac{m}{2}\log\frac{1}{\delta}}\right]
\left[\wp(R)m+\frac{2}{3}\log\frac{1}{\delta}+\sqrt{2\wp(R)[1-\wp(R)]m\log\frac{1}{\delta}}\right]  }\\\nonumber
<&\frac{m}{2}+Rm+\sqrt{\frac{m}{2}\log\frac{1}{\delta}}+\sqrt{2Rm\log\frac{1}{\delta}}+\frac{2}{3}\log\frac{1}{\delta}\\\nonumber
&+\sqrt{2R}m+2\sqrt{m\log(1/\delta)}+2\sqrt{Rm\log(1/\delta)}+\mathcal O\left(\frac{\log(1/\delta)}{R}\right)\\\nonumber
=&\left[\frac{1}{\sqrt{2}}+\sqrt{R}\right]^2m+\left[2+\frac{1}{\sqrt{2}}\right]\sqrt{m\log(1/\delta)}\\\nonumber
&+(2+\sqrt{2})\sqrt{Rm\log(1/\delta)}+\mathcal O\left(\log(1/\delta)/R\right),
\end{align}
thus indicating that by taking a union bound, with probability exceeding $1-2n\delta$,
\begin{align}\nonumber
&\sum_{i=1}^n\|\widetilde{\boldsymbol\psi_i}(k)\|_2^4\\\nonumber
<&\sum_{i=1}^n\left\{\left[\frac{1}{\sqrt{2}}+\sqrt{R}\right]^2m+\left[2+\frac{1}{\sqrt{2}}\right]\sqrt{m\log(1/\delta)}+(2+\sqrt{2})\sqrt{Rm\log(1/\delta)}+\mathcal O\left(\log(1/\delta)/R\right)\right\}^2\\\nonumber
=&\sum_{i=1}^n\left\{\left[\frac{1}{\sqrt{2}}+\sqrt{R}\right]^2m\left[1+\frac{4+\sqrt{2}}{(1+\sqrt{2R})^2}\sqrt{\frac{\log(1/\delta)}{m}}+\frac{4+2\sqrt{2}}{(1+\sqrt{2R})^2}\sqrt{\frac{R\log(1/\delta)}{m}}+\mathcal O\left(\frac{\log(1/\delta)}{Rm}\right)\right]\right\}^2\\\nonumber
\stackrel{\text{(i)}}=&\sum_{i=1}^n\left[\frac{1}{\sqrt{2}}+\sqrt{R}\right]^4m^2\bigg\{1+\frac{8+2\sqrt{2}}{(1+\sqrt{2R})^2}\sqrt{\frac{\log(1/\delta)}{m}}+\frac{8+4\sqrt{2}}{(1+\sqrt{2R})^2}\sqrt{\frac{R\log(1/\delta)}{m}}+\mathcal O\left(\frac{\log(1/\delta)}{Rm}\right)\\\nonumber
&+\mathcal O\left(\frac{4+\sqrt{2}}{(1+\sqrt{2R})^2}\sqrt{\frac{\log(1/\delta)}{m}}+\frac{4+2\sqrt{2}}{(1+\sqrt{2R})^2}\sqrt{\frac{R\log(1/\delta)}{m}}+\mathcal O\left(\frac{\log(1/\delta)}{Rm}\right) \right)^2\bigg\}\\\nonumber
=&\sum_{i=1}^n\left[\frac{1}{\sqrt{2}}+\sqrt{R}\right]^4m^2\left[1+\frac{8+2\sqrt{2}}{(1+\sqrt{2R})^2}\sqrt{\frac{\log(1/\delta)}{m}}+\frac{8+4\sqrt{2}}{(1+\sqrt{2R})^2}\sqrt{\frac{R\log(1/\delta)}{m}}+\mathcal O\left(\frac{\log(1/\delta)}{Rm}\right)\right]\\\nonumber
=&\left[\frac{1}{\sqrt{2}}+\sqrt{R}\right]^4m^2n+\frac{(4+\sqrt{2})(1+\sqrt{2R})^2}{2}m^{3/2}n\sqrt{\log(1/\delta)}\\\label{step3_estimate}
&+(2+\sqrt{2})(1+\sqrt{2R})^2m^{3/2}n\sqrt{R\log(1/\delta)}+\mathcal O\left(mn\log(1/\delta)/R\right).
\end{align}
Here (i) comes from that
$$
\sqrt{1+\alpha}=1+\frac{\alpha}{2}+\mathcal O(\alpha^2),\text{ if }\alpha\to0.
$$
{\textcolor{blue}{{\bf Step 5: Bounding $\sum_{1\leqslant i\not=j\leqslant n}\langle\widetilde{\boldsymbol\psi_i}(k),\widetilde{\boldsymbol\psi_j}(k)\rangle^2$}}\\
We see from step 3 that with probability exceeding $1-2\delta$ over the random initialization,
\begin{align}\nonumber
\|\widetilde{\boldsymbol\psi_i}(k)\|_2^2
<\left[\frac{1}{\sqrt{2}}+\sqrt{R}\right]^2m+\left[2+\frac{1}{\sqrt{2}}\right]\sqrt{m\log(1/\delta)}+(2+\sqrt{2})\sqrt{Rm\log(1/\delta)}+\mathcal O\left(\log(1/\delta)/R\right).
\end{align}
Hence, applying a union bound yields that with probability exceeding $1-4\delta$ results in
\begin{align}\nonumber
&\langle\widetilde{\boldsymbol\psi_i}(k),\widetilde{\boldsymbol\psi_j}(k)\rangle^2\leqslant\|\widetilde{\boldsymbol\psi_i}(k)\|_2^2\|\widetilde{\boldsymbol\psi_j}(k)\|_2^2\\\nonumber
<&\left\{\left[\frac{1}{\sqrt{2}}+\sqrt{R}\right]^2m+\left[2+\frac{1}{\sqrt{2}}\right]\sqrt{m\log(1/\delta)}+(2+\sqrt{2})\sqrt{Rm\log(1/\delta)}+\mathcal O\left(\log(1/\delta)/R\right)\right\}^2\\\nonumber
=&\left[\frac{1}{\sqrt{2}}+\sqrt{R}\right]^4m^2+\frac{(4+\sqrt{2})(1+\sqrt{2R})^2}{2}m^{3/2}\sqrt{\log(1/\delta)}\\\nonumber
&+(2+\sqrt{2})(1+\sqrt{2R})^2m^{3/2}\sqrt{R\log(1/\delta)}+\mathcal O\left(m\log(1/\delta)/R\right).
\end{align}
Finally, taking a union bound again, with probability exceeding $1-(n(n-1)/2)\cdot4\delta=1-2n(n-1)\delta$, we obtain
\begin{align}\nonumber
\sum_{1\leqslant i\not=j\leqslant n}\langle\widetilde{\boldsymbol\psi_i}(k),\widetilde{\boldsymbol\psi_j}(k)\rangle^2<&\left[\frac{1}{\sqrt{2}}+\sqrt{R}\right]^4m^2n^2+\frac{(4+\sqrt{2})(1+\sqrt{2R})^2}{2}m^{3/2}n^2\sqrt{\log(1/\delta)}\\\label{step4_estimate}
&+(2+\sqrt{2})(1+\sqrt{2R})^2m^{3/2}n^2\sqrt{R\log(1/\delta)}+\mathcal O\left(mn^2\log(1/\delta)/R\right).
\end{align}
{\textcolor{blue}{{\bf Step 6: Bounding the Main Ingredient of Directional Curvature via Step 3 to 5}}\\
We turn to bound the main ingredient of directional curvature in (\ref{loss_function_Taylor_2}) using the above bounds obtained from step 3 to step 5. Firstly, by (\ref{step3_estimate}) and (\ref{step4_estimate}). Since if the event (\ref{step3_estimate}) happens, then the event (\ref{step4_estimate}) also happens, thus we conclude that with probability exceeding $1-2n\delta$ over the random initialization,
\begin{align}\nonumber
&\sum_{i=1}^n\|\widetilde{\boldsymbol\psi_i}(k)\|_2^4+\sum_{1\leqslant i\not=j\leqslant n}\langle\widetilde{\boldsymbol\psi_i}(k),\widetilde{\boldsymbol\psi_j}(k)\rangle^2\\\nonumber
<&\left[\frac{1}{\sqrt{2}}+\sqrt{R}\right]^4m^2n+\frac{(4+\sqrt{2})(1+\sqrt{2R})^2}{2}m^{3/2}n\sqrt{\log(1/\delta)}\\\nonumber
&+(2+\sqrt{2})(1+\sqrt{2R})^2m^{3/2}n\sqrt{R\log(1/\delta)}+\mathcal O\left(mn\log(1/\delta)/R\right)\\\nonumber
&+\left[\frac{1}{\sqrt{2}}+\sqrt{R}\right]^4m^2n^2+\frac{(4+\sqrt{2})(1+\sqrt{2R})^2}{2}m^{3/2}n^2\sqrt{\log(1/\delta)}\\\nonumber
&+(2+\sqrt{2})(1+\sqrt{2R})^2m^{3/2}n^2\sqrt{R\log(1/\delta)}+\mathcal O\left(mn^2\log(1/\delta)/R\right)\\\nonumber
\stackrel{\text{(i)}}=&\left[\frac{1}{\sqrt{2}}+\sqrt{R}\right]^4m^2n+\frac{(4+\sqrt{2})(1+\sqrt{2R})^2}{2}m^{3/2}n\sqrt{\log(1/\delta)}\\\nonumber
&+(2+\sqrt{2})(1+\sqrt{2R})^2m^{3/2}n\sqrt{R\log(1/\delta)}+o\left(m^{3/2}n\sqrt{R\log(1/\delta)}\right)\\\nonumber
&+\left[\frac{1}{\sqrt{2}}+\sqrt{R}\right]^4m^2n^2+\frac{(4+\sqrt{2})(1+\sqrt{2R})^2}{2}m^{3/2}n^2\sqrt{\log(1/\delta)}\\\nonumber
&+(2+\sqrt{2})(1+\sqrt{2R})^2m^{3/2}n^2\sqrt{R\log(1/\delta)}+o\left(m^{3/2}n^2\sqrt{R\log(1/\delta)}\right)
\\\nonumber
=&\left[\frac{1}{\sqrt{2}}+\sqrt{R}\right]^4m^2n^2
+\left[\frac{1}{\sqrt{2}}+\sqrt{R}\right]^4m^2n+\frac{(4+\sqrt{2})(1+\sqrt{2R})^2}{2}m^{3/2}n^2\sqrt{\log(1/\delta)}\\\nonumber
&+\frac{(4+\sqrt{2})(1+\sqrt{2R})^2}{2}m^{3/2}n\sqrt{\log(1/\delta)}+\mathcal O\left(m^{3/2}n\sqrt{R\log(1/\delta)}\right)+\mathcal O\left(m^{3/2}n^2\sqrt{R\log(1/\delta)}\right)\\\nonumber
\stackrel{\text{(ii)}}=&\left[\frac{1}{\sqrt{2}}+\sqrt{R}\right]^4m^2n^2
+\left[\frac{1}{\sqrt{2}}+\sqrt{R}\right]^4m^2n+\frac{(4+\sqrt{2})(1+\sqrt{2R})^2}{2}m^{3/2}n^2\sqrt{\log(1/\delta)}\\\label{Step5_estimate_1}
&+\mathcal O\left(m^{3/2}n^{7/4}\sqrt{\log(1/\delta)}\right),
\end{align}
where (i) uses $R=\sqrt{\pi\lambda^*\sin\theta^*/(2+\varepsilon)n}$ for any $\varepsilon>0$ and $m=\Omega(n^2\log^{1+\varpi}(n/\delta))$ for any $\varpi>0$ which leads to
\begin{align}\nonumber
&\lim_{n\to+\infty}\frac{mn\log(1/\delta)/R}{m^{3/2}n\sqrt{R\log(1/\delta)}}
=\lim_{n\to+\infty}\frac{\sqrt{\log(1/\delta)}}{R^{3/2}m^{1/2}}\\\nonumber
\leqslant&\lim_{n\to+\infty}\left[\frac{(2+\varepsilon)n}{\pi\lambda^*\sin\theta^*}\right]^{3/4}\sqrt{\frac{\log(1/\delta)}{c_1n^2\log^{1+\varpi}(n/\delta)}}\\\nonumber
\stackrel{\text{(i)}}=&\lim_{n\to+\infty}\frac{1}{n^{1/4}}\left[\frac{(2+\varepsilon)}{\pi\lambda^*\sin\theta^*}\right]^{3/4}\sqrt{\frac{\log(1/\delta)}{c_1\log^{1+\varpi}(n/\delta)}}\\\nonumber
\leqslant&\lim_{n\to+\infty}\frac{c}{n^{1/4}}\left[\frac{(2+\varepsilon)}{\pi\sin\theta^*}\right]^{3/4}\sqrt{\frac{\log^{3\hbar}n\log(1/\delta)}{c_1\log^{1+\varpi}(n/\delta)}}=0\quad(c\text{ is a contant}), 
\end{align}
whereas (i) results from
\begin{align}\nonumber
m^{3/2}n^2\sqrt{R\log(1/\delta)}\asymp m^{3/2}n^{7/4}\sqrt{\log(1/\delta)}
\end{align}
and the last line $\lambda^*\geqslant\lambda_{\text{min}}(\boldsymbol X^\top\boldsymbol X)/4$ from (\ref{lambda_star_lower_bound}) and the condition $\lambda_{\text{min}}(\boldsymbol X^\top\boldsymbol X)=\Omega(\log^{-2\hbar}n)$ for certain $\hbar>0$.
Moreover, one can apply (\ref{step3_lower_estimate_2}) to demonstrate that with probability exceeding $1-4n\delta$,
\begin{align}\nonumber
&\frac{1}{m}\|\boldsymbol\Psi(k)\|_{\text{F}}^4\\\nonumber
>&\frac{1}{m}\bigg\{  \left[\frac{1}{\sqrt{2}}-\sqrt{R}\right]^2mn-\left[2+\frac{1}{\sqrt{2}}\right]n\sqrt{m\log(1/\delta)}\\\nonumber
&-(2+\sqrt{7/10})n\sqrt{Rm\log(1/\delta)}+\mathcal O\left(n\log(1/\delta)/R\right)  \bigg\}^2\\\nonumber
=&\frac{1}{m}\left\{\left[\frac{1}{\sqrt{2}}-\sqrt{R}\right]^4m^2n^2\left[1  -\frac{4+\sqrt{2}}{(1-\sqrt{2R})^2}\sqrt{\frac{\log(1/\delta)}{m}}-\frac{4+2\sqrt{7/10}}{(1-\sqrt{2R})^2}\sqrt{\frac{R\log(1/\delta)}{m}}+\mathcal O\left(\frac{\log(1/\delta)}{Rm}\right)\right]^2\right\}\\\nonumber
\stackrel{\text{(i)}}=&\left[\frac{1}{\sqrt{2}}-\sqrt{R}\right]^4mn^2\bigg\{1-\frac{4+\sqrt{2}}{(1-\sqrt{2R})^2}\sqrt{\frac{\log(1/\delta)}{m}}-\frac{4+2\sqrt{7/10}}{(1-\sqrt{2R})^2}\sqrt{\frac{R\log(1/\delta)}{m}}+\mathcal O\left(\frac{\log(1/\delta)}{Rm}\right)\\\nonumber
&+\mathcal O\left(\frac{4+\sqrt{2}}{(1-\sqrt{2R})^2}\sqrt{\frac{\log(1/\delta)}{m}}+\frac{4+2\sqrt{7/10}}{(1-\sqrt{2R})^2}\sqrt{\frac{R\log(1/\delta)}{m}}+\mathcal O\left(\frac{\log(1/\delta)}{Rm}\right)\right)^2\bigg\}\\\nonumber
=&\left[\frac{1}{\sqrt{2}}-\sqrt{R}\right]^4mn^2\left[1-\frac{4+\sqrt{2}}{(1-\sqrt{2R})^2}\sqrt{\frac{\log(1/\delta)}{m}}-\frac{4+2\sqrt{7/10}}{(1-\sqrt{2R})^2}\sqrt{\frac{R\log(1/\delta)}{m}}+\mathcal O\left(\frac{\log(1/\delta)}{Rm}\right)\right]\\\nonumber
=&\left[\frac{1}{\sqrt{2}}-\sqrt{R}\right]^4mn^2-\frac{(4+\sqrt{2})(1-\sqrt{2R})^2}{4}n^2\sqrt{m\log(1/\delta)}\\\label{Step5_estimate_2}
&-\frac{(4+2\sqrt{7/10})(1-\sqrt{2R})^2}{4}n^2\sqrt{Rm\log(1/\delta)}+\mathcal O\left(n^{5/2}\log(1/\delta)\right).
\end{align}
Here (i) comes from that
$$
\sqrt{1+\alpha}=1+\frac{\alpha}{2}+\mathcal O(\alpha^2),\text{ if }\alpha\to0,
$$
whereas the last line uses $R=\sqrt{\pi\lambda^*\sin\theta^*/(2+\varepsilon)n}$ for any $\varepsilon>0$ which means $R\asymp n^{-1/2}$.\\
Since if the event (\ref{Step5_estimate_2}) happens, then the event (\ref{Step5_estimate_1}) also happens, thus we conclude that with probability exceeding $1-4n\delta$,
\begin{align}\nonumber
&\sum_{i=1}^n\|\widetilde{\boldsymbol\psi_i}(k)\|_2^4+\sum_{1\leqslant i\not=j\leqslant n}\langle\widetilde{\boldsymbol\psi_i}(k),\widetilde{\boldsymbol\psi_j}(k)\rangle^2-\frac{1}{m}\|\boldsymbol\Psi(k)\|_{\text{F}}^4
\\\nonumber
<&\left[\frac{1}{\sqrt{2}}+\sqrt{R}\right]^4m^2n^2
+\left[\frac{1}{\sqrt{2}}+\sqrt{R}\right]^4m^2n\\\nonumber
&+\frac{(4+\sqrt{2})(1+\sqrt{2R})^2}{2}m^{3/2}n^2\sqrt{\log(1/\delta)}\\\nonumber
&+\mathcal O\left(m^{3/2}n^{7/4}\sqrt{\log(1/\delta)}\right)-\left[\frac{1}{\sqrt{2}}-\sqrt{R}\right]^4mn^2\\\nonumber
&+\frac{(4+\sqrt{2})(1-\sqrt{2R})^2}{4}n^2\sqrt{m\log(1/\delta)}\\\nonumber
&+\frac{(4+2\sqrt{7/10})(1-\sqrt{2R})^2}{4}n^2\sqrt{Rm\log(1/\delta)}+\mathcal O\left(n^{5/2}\log(1/\delta)\right)\\\nonumber
=&\left[\frac{1}{\sqrt{2}}+\sqrt{R}\right]^4m^2n^2
+\left[\frac{1}{\sqrt{2}}+\sqrt{R}\right]^4m^2n+\mathcal O\left(m^{3/2}n^2\sqrt{\log(1/\delta)}\right).
\end{align}
where the last line results from
\begin{align}\nonumber
&\lim_{n\to+\infty}\left\{m^{3/2}n^{7/4}\sqrt{\log(1/\delta)}+mn^2\right\}(m^{3/2}n^2\sqrt{\log(1/\delta)})^{-1}\\\nonumber
=&\lim_{n\to+\infty}\left\{m^{3/2}n^{7/4}\sqrt{\log(1/\delta)}\right\}(m^{3/2}n^2\sqrt{\log(1/\delta)})^{-1}\\\nonumber
&+\lim_{n\to+\infty}\left\{mn^2\right\}(m^{3/2}n^2\sqrt{\log(1/\delta)})^{-1}\\\nonumber
=&\lim_{n\to+\infty}n^{-1/4}+\lim_{n\to+\infty}(m\log(1/\delta))^{-1/2}=0
\end{align}
and
\begin{align}\nonumber
&\lim_{n\to+\infty}\left\{n^2\sqrt{m\log(1/\delta)}\right\}(m^{3/2}n^2\sqrt{\log(1/\delta)})^{-1}\\\nonumber
=&\lim_{n\to+\infty}m^{-1}\leqslant
\lim_{n\to+\infty}(n^2\log^{1+\varpi}(n/\delta))^{-1}=0,\\\nonumber
&n^2\sqrt{Rm\log(1/\delta)}\asymp n^{7/4}\sqrt{m\log(1/\delta)}
\\\nonumber
=&o(m^{3/2}n^2\sqrt{\log(1/\delta)}),\\\nonumber
&\frac{n^{5/2}\log(1/\delta)}{m^{3/2}n^2\sqrt{\log(1/\delta)}}=\frac{n^{1/2}\sqrt{\log(1/\delta)}}{m^{3/2}}\\\nonumber
\lesssim&\frac{n^{1/2}\sqrt{\log(1/\delta)}}{(n^2\log^{1+\varpi}(n/\delta))^{3/2}}=\frac{\sqrt{\log(1/\delta)}}{n^{5/2}\log^{3(1+\varpi)/2}(n/\delta)}=o(1).
\end{align}
We therefore arrive at the following estimate with probability exceeding $1-4n\delta$,
\begin{align}\nonumber
&\left[\sum_{i=1}^n\|\widetilde{\boldsymbol\psi_i}(k)\|_2^4+\sum_{1\leqslant i\not=j\leqslant n}\langle\widetilde{\boldsymbol\psi_i}(k),\widetilde{\boldsymbol\psi_j}(k)\rangle^2-\frac{1}{m}\|\boldsymbol\Psi(k)\|_{\text{F}}^4\right]^{1/4}
\\\nonumber
<&\left[  \left[\frac{1}{\sqrt{2}}+\sqrt{R}\right]^4m^2n^2
+\left[\frac{1}{\sqrt{2}}+\sqrt{R}\right]^4m^2n+\mathcal O\left(m^{3/2}n^2\sqrt{\log(1/\delta)}\right)  \right]^{1/4}\\\nonumber
=&\left\{ \left[\frac{1}{\sqrt{2}}+\sqrt{R}\right]^4m^2n^2 \left[1+\frac{1}{n}+\mathcal O\left(\sqrt{\log(1/\delta)/m}\right)  \right]\right\}^{1/4}
\\\nonumber
\stackrel{\text{(i)}}=&\left[\frac{1}{\sqrt{2}}+\sqrt{R}\right]\sqrt{mn}\left[1+\frac{1}{4n}+\mathcal O\left(\sqrt{\log(1/\delta)/m}\right)+\left( \frac{1}{n}+\mathcal O\left(\sqrt{\log(1/\delta)/m}\right) \right)^2  \right]\\\nonumber
=&\left[\frac{1}{\sqrt{2}}+\sqrt{R}\right]\sqrt{mn}\left[1+\frac{1}{4n}+\mathcal O\left(\sqrt{\log(1/\delta)/m}\right)+\frac{1}{n^2}\left[1+\mathcal O\left(n\sqrt{\log(1/\delta)/m}\right)\right]^2 \right]\\\nonumber
=&\left[\frac{1}{\sqrt{2}}+\sqrt{R}\right]\sqrt{mn}\left[1+\frac{1}{4n}+\mathcal O\left(\sqrt{\log(1/\delta)/m}\right)+\frac{1}{n^2}+\mathcal O\left(\frac{\sqrt{\log(1/\delta)/m}}{n}\right) \right]\\\nonumber
=&\left[\frac{1}{\sqrt{2}}+\sqrt{R}\right]\sqrt{mn}\left[1+\frac{1}{4n}+\frac{1}{n^2}+\mathcal O\left(\sqrt{\log(1/\delta)/m}\right) \right]\\\label{step5_upper_estimate_1}
=&\left[\frac{1}{\sqrt{2}}+\sqrt{R}\right]\sqrt{mn}+\frac{1+\sqrt{2R}}{4\sqrt{2}}\sqrt{m/n}+\frac{1+\sqrt{2R}}{\sqrt{2}}\sqrt{m/n^3}+\mathcal O\left(\sqrt{n\log(1/\delta)}\right),
\end{align}
where (i) makes use of the \textcolor{blue}{binomial series}
$$
\sqrt[4]{1+f}=1+{1/4 \choose 1}f+{1/4 \choose 2}f^2+{1/4 \choose 3}f^3+\cdots=1+\frac{f}{4}+\mathcal O(f^2),\text{ as }f\to0.
$$
As a result, combining (\ref{step3_upper_estimate_2}) and (\ref{step5_upper_estimate_1}), since if the event (\ref{step5_upper_estimate_1}) happens, then the event (\ref{step3_upper_estimate_2}) also happens, we arrive at that with probability exceeding $1-4n\delta$,
\begin{align}\nonumber
&\|\boldsymbol\Psi(k)\|_{\text{F}}^2+ 
m\|\boldsymbol\Psi(k)\|_{\text{F}}\left[\sum_{i=1}^n\|\widetilde{\boldsymbol\psi_i}(k)\|_2^4+\sum_{1\leqslant i\not=j\leqslant n}\langle\widetilde{\boldsymbol\psi_i}(k),\widetilde{\boldsymbol\psi_j}(k)\rangle^2-\frac{\|\boldsymbol\Psi(k)\|_{\text{F}}^4}{m}\right]^{1/4}\\\nonumber
<&\left[\frac{1}{\sqrt{2}}+\sqrt{R}\right]^2mn+\left[2+\frac{1}{\sqrt{2}}\right]n\sqrt{m\log(1/\delta)}+(2+\sqrt{2})n\sqrt{Rm\log(1/\delta)}+\mathcal O\left(n\log(1/\delta)/R\right)\\\nonumber
&+m\sqrt{\left[\frac{1}{\sqrt{2}}+\sqrt{R}\right]^2mn+\left[2+\frac{1}{\sqrt{2}}\right]n\sqrt{m\log(1/\delta)}+(2+\sqrt{2})n\sqrt{Rm\log(1/\delta)}+\mathcal O\left(n\log(1/\delta)/R\right)}\\\nonumber
&\cdot \left\{\left[\frac{1}{\sqrt{2}}+\sqrt{R}\right]\sqrt{mn}+\frac{1+\sqrt{2R}}{4\sqrt{2}}\sqrt{m/n}+\frac{1+\sqrt{2R}}{\sqrt{2}}\sqrt{m/n^3}+\mathcal O\left(\sqrt{n\log(1/\delta)}\right)\right\},
\end{align}
\textcolor{red}{Having a cup of tea}, we continue to estimate
\begin{align}\nonumber
&\sqrt{\left[\frac{1}{\sqrt{2}}+\sqrt{R}\right]^2mn+\left[2+\frac{1}{\sqrt{2}}\right]n\sqrt{m\log(1/\delta)}+(2+\sqrt{2})n\sqrt{Rm\log(1/\delta)}+\mathcal O\left(n\log(1/\delta)/R\right)}\\\nonumber
&\cdot \left\{\left[\frac{1}{\sqrt{2}}+\sqrt{R}\right]\sqrt{mn}+\frac{1+\sqrt{2R}}{4\sqrt{2}}\sqrt{m/n}+\frac{1+\sqrt{2R}}{\sqrt{2}}\sqrt{m/n^3}+\mathcal O\left(\sqrt{n\log(1/\delta)}\right)\right\}\\\nonumber
=&\sqrt{ \left[\frac{1}{\sqrt{2}}+\sqrt{R}\right]^2mn \left[1+ \frac{4+\sqrt{2}}{(1+\sqrt{2R})^2}\sqrt{\frac{\log(1/\delta)}{m}}+\frac{4+2\sqrt{2}}{(1+\sqrt{2R})^2}\sqrt{\frac{R\log(1/\delta)}{m}}+\mathcal O\left(\frac{\log(1/\delta)}{Rm}\right)\right]}\\\nonumber
&\cdot\left[\frac{1}{\sqrt{2}}+\sqrt{R}\right]\sqrt{mn}\left[1+\frac{1}{4n}+\frac{1}{n^2}+\mathcal O\left(\sqrt{\frac{\log(1/\delta)}{m}}\right) \right]\\\nonumber
\stackrel{\text{(i)}}=&\left[\frac{1}{\sqrt{2}}+\sqrt{R}\right]^2mn\left[1+\frac{4+\sqrt{2}}{(1+\sqrt{2R})^2}\sqrt{\frac{\log(1/\delta)}{m}}+\mathcal O\left(  \sqrt{\frac{R\log(1/\delta)}{m}}\right)\right]\\\nonumber
& \left[1+\frac{1}{4n}+\frac{1}{n^2}+\mathcal O\left(\sqrt{\frac{\log(1/\delta)}{m}}\right) \right]\\\nonumber
=&\left[\frac{1}{\sqrt{2}}+\sqrt{R}\right]^2mn\bigg\{1+\frac{1}{4n}+\frac{4+\sqrt{2}}{(1+\sqrt{2R})^2}\sqrt{\frac{\log(1/\delta)}{m}}+\frac{1}{n^2}+\frac{4+\sqrt{2}}{4(1+\sqrt{2R})^2n}\sqrt{\frac{\log(1/\delta)}{m}}\\\nonumber
&+\frac{4+\sqrt{2}}{(1+\sqrt{2R})^2n^2}\sqrt{\frac{\log(1/\delta)}{m}}+\mathcal O\left(\sqrt{\frac{R\log(1/\delta)}{m}}+\sqrt{\frac{\log(1/\delta)}{m}}\right)\bigg\}\\\nonumber
=&\left[\frac{1}{\sqrt{2}}+\sqrt{R}\right]^2mn\bigg\{1+\frac{1}{4n}+\frac{1}{n^2}+\mathcal O\left(\sqrt{\frac{\log(1/\delta)}{m}}\right)\bigg\}\\\nonumber
=&\left[\frac{1}{\sqrt{2}}+\sqrt{R}\right]^2mn+\frac{(1+\sqrt{2R})^2}{8}m+\frac{(1+\sqrt{2R})^2}{2}\frac{m}{n}+\mathcal O\left(n\sqrt{m\log(1/\delta)}\right).
\end{align}
(i) stems from $R=\sqrt{\pi\lambda^*\sin\theta^*/(2+\varepsilon)n}$ for any $\varepsilon>0$ and $m=\Omega(n^2\log^{1+\varpi}(n/\delta))$ for any $\varpi>0$, which give rise to
\begin{align}\nonumber
&\lim_{n\to+\infty}\frac{\log(1/\delta)}{Rm}\left(\sqrt{\frac{R\log(1/\delta)}{m}}\right)^{-1}=\lim_{n\to+\infty}\frac{\sqrt{\log(1/\delta)}}{R^{3/2}m^{1/2}}\\\nonumber
\leqslant&\lim_{n\to+\infty}\frac{\sqrt{\log(1/\delta)}}{[\sqrt{\pi\lambda^*\sin\theta^*/(2+\varepsilon)n}]^{3/2}[c_1n^2\log^{1+\varpi}(n/\delta)]^{1/2}}\\\nonumber
=&
\lim_{n\to+\infty}\frac{\sqrt{\log(1/\delta)}(2+\varpi)^{3/4}}{c_1^{1/2}(\pi\lambda^*\sin\theta^*)^{3/4}n^{1/4}\log^{(1+\varpi)/2}(n/\delta)}
=0\\\nonumber
\leqslant&
\lim_{n\to+\infty}\frac{c_2(\log^{3\hbar/2}n)\sqrt{\log(1/\delta)}(2+\varpi)^{3/4}}{c_1^{1/2}(\pi\sin\theta^*)^{3/4}n^{1/4}\log^{(1+\varpi)/2}(n/\delta)}
=0\quad(\text{$c_1,c_2$ are constants}).
\end{align}
Therefore, we conclude that with probability exceeding $1-4n\delta$,
\begin{align}\nonumber
&\|\boldsymbol\Psi(k)\|_{\text{F}}^2+ 
m\|\boldsymbol\Psi(k)\|_{\text{F}}\left[\sum_{i=1}^n\|\widetilde{\boldsymbol\psi_i}(k)\|_2^4+\sum_{1\leqslant i\not=j\leqslant n}\langle\widetilde{\boldsymbol\psi_i}(k),\widetilde{\boldsymbol\psi_j}(k)\rangle^2-\frac{\|\boldsymbol\Psi(k)\|_{\text{F}}^4}{m}\right]^{1/4}\\\nonumber
<&\left[\frac{1}{\sqrt{2}}+\sqrt{R}\right]^2mn+\left[2+\frac{1}{\sqrt{2}}\right]n\sqrt{m\log(1/\delta)}+(2+\sqrt{2})n\sqrt{Rm\log(1/\delta)}+\mathcal O\left(n\log(1/\delta)/R\right)\\\nonumber
&+m\left\{\left[\frac{1}{\sqrt{2}}+\sqrt{R}\right]^2mn+\frac{(1+\sqrt{2R})^2}{8}m+\frac{(1+\sqrt{2R})^2}{2}\frac{m}{n}+\mathcal O\left(n\sqrt{m\log(1/\delta)}\right)\right\}\\\label{step5_estimate}
\stackrel{\text{(i)}}=&\left(\frac{1}{\sqrt{2}}+\sqrt{R}\right)^2m^2n+\frac{(1+\sqrt{2R})^2}{8}m^2+\frac{(1+\sqrt{2R})^2m^2}{2n}+\mathcal O\left(m^{3/2}n\sqrt{\log(1/\delta)}\right).
\end{align}
Here (i) is due to the fact
\begin{align}\nonumber
\lim_{n\to+\infty}\frac{mn}{m^2/n}=\lim_{n\to+\infty}\frac{n^2}{m}\leqslant\lim_{n\to+\infty}\frac{n^2}{c_1n^2\log^{1+\varpi}(n/\delta)}=0.
\end{align}
{\textcolor{blue}{{\bf Step 7: Bounding $\sum_{1\leqslant i,j\leqslant n}\langle \boldsymbol x_i,\boldsymbol x_j\rangle^4$}}\\
Recall the anglular paramters defined in Section \ref{section_problem}:
\begin{align}\nonumber
\theta_{\text{min}}&=\min_{1\leqslant i\not=j\leqslant n}\theta_{ij}=\min_{1\leqslant i\not=j\leqslant n}\angle(\boldsymbol x_i,\boldsymbol x_j)=\min_{1\leqslant i\not=j\leqslant n}\{\arccos\langle\boldsymbol x_i, \boldsymbol x_j\rangle\}\in(0,2\pi/3],\\\nonumber
\theta_{\text{max}}&=\max_{1\leqslant i\not=j\leqslant n}\theta_{ij}=\max_{1\leqslant i\not=j\leqslant n}\angle(\boldsymbol x_i,\boldsymbol x_j)=\max_{1\leqslant i\not=j\leqslant n}\{\arccos\langle\boldsymbol x_i, \boldsymbol x_j\rangle\}\in(0,\pi),\\\nonumber
\theta^*&=\min\{\theta_{\text{min}},\pi-\theta_{\text{max}}\}\in(0,\pi/2].
\end{align}
Thus, $\max_{1\leqslant i\not=j\leqslant n}|\langle \boldsymbol x_i,\boldsymbol x_j\rangle|=\cos\theta^*$ and this allows us to derive
\begin{align}\nonumber
&\sum_{1\leqslant i,j\leqslant n}\langle \boldsymbol x_i,\boldsymbol x_j\rangle^4
=\sum_{i=1}^n\langle \boldsymbol x_i,\boldsymbol x_i\rangle^4
+\sum_{1\leqslant i\not=j\leqslant n}\langle \boldsymbol x_i,\boldsymbol x_j\rangle^4
\leqslant n+\max_{1\leqslant i\not=j\leqslant n}\langle \boldsymbol x_i,\boldsymbol x_j\rangle^2
\sum_{1\leqslant i\not=j\leqslant n}\langle \boldsymbol x_i,\boldsymbol x_j\rangle^2\\\nonumber
=&n+\cos^2\theta^*(\|\boldsymbol X^\top\boldsymbol X\|_{\text{F}}^2-n)
=n\sin^2\theta^*+\|\boldsymbol X^\top\boldsymbol X\|_{\text{F}}^2\cos^2\theta^*
\leqslant n\sin^2\theta^*+n\sigma_{\max}^2(\boldsymbol X^\top\boldsymbol X)\cos^2\theta^*\\\nonumber
\stackrel{\text{(i)}}=&n(\sin^2\theta^*+\lambda_{\max}^2(\boldsymbol X^\top\boldsymbol X)\cos^2\theta^*)=n[\lambda_{\max}^2(\boldsymbol X^\top\boldsymbol X)+(1-\lambda_{\max}^2(\boldsymbol X^\top\boldsymbol X))\sin^2\theta^*]\stackrel{\text{(ii)}}\leqslant n\lambda_{\max}^2(\boldsymbol X^\top\boldsymbol X),
\end{align}
where (i) results from that $\boldsymbol X^\top\boldsymbol X$ is a real symmetric matrix; whereas (ii) is due to the fact that $\lambda_{\max}(\boldsymbol X^\top\boldsymbol X)\geqslant1$ which is the consequence of {\textcolor{red}{Ger\v{s}gorin disc theorem} (\cite{gersgorin_1931}), making use of the fact that all the diagonal entries of matrix $\boldsymbol X^\top\boldsymbol X$ are one.\\\\
{\textcolor{blue}{{\bf Step 8: Back to Loss Function Iteration: Combining Bounds in Step 1 to 7}}\\
Putting the lower bound of the smallest singular value of activation matrix in Lemma \ref{lemma_Psi_k_lower_bound}, the upper bound of the largest singular value of activation matrix in Lemma \ref{lemma_Psi_k_upper_bound} and the main term of Hessian upper bound (\ref{step5_estimate}) together into (\ref{loss_function_Taylor_2}). Notice that if the events in Lemma \ref{lemma_Psi_k_lower_bound} and the main term of Hessian upper bound (\ref{step5_estimate}) happen, then the event Lemma \ref{lemma_Psi_k_upper_bound} also happens. Hence, applying a union bound, we conclude that in the over-parameterized setting $m\geqslant n$, if $n>70$ which obviously holds in practice, then with probability exceeding $1-(mn-m+4n)\delta-4n\delta>1-1.1mn\delta$ over the random initialization,
\begin{align}\nonumber
&\mathcal L(\boldsymbol W(k+1))\\\nonumber
<&\Big\{1-\eta\cdot\frac{2}{m}\sigma_{\min}^2(\boldsymbol\Psi(k))\lambda_{\min}(\boldsymbol X^\top\boldsymbol X)\\\nonumber
&+\eta^2\cdot\frac{1}{m^2}\left[\|\boldsymbol\Psi(k)\|_{\text{F}}^2+ 
m\|\boldsymbol\Psi(k)\|_{\text{F}}\left[\sum_{i=1}^n\|\widetilde{\boldsymbol\psi_i}(k)\|_2^4+\sum_{1\leqslant i\not=j\leqslant n}\langle\widetilde{\boldsymbol\psi_i}(k),\widetilde{\boldsymbol\psi_j}(k)\rangle^2-\frac{\|\boldsymbol\Psi(k)\|_{\text{F}}^4}{m}\right]^{1/4}
\right]\\\nonumber
&\cdot\sqrt{\sum_{1\leqslant i,j\leqslant n}\langle \boldsymbol x_i,\boldsymbol x_j\rangle^4}+\frac{\eta^2}{m}\sigma_{\max}^2(\boldsymbol\Psi(k))\lambda_{\max}(\boldsymbol X^\top\boldsymbol X)\Big\}\cdot\mathcal L(\boldsymbol W(k))\\\nonumber
<&\bigg\{1-\frac{2\eta}{m}\bigg\{\left(1-\sqrt{\frac{2}{2+\varepsilon}}\right)^2\lambda^*m+\left(\sqrt{\frac{\pi}{2+\varepsilon}}-\sqrt{\frac{1}{2\pi}}+o(1)\right)m\sqrt{\frac{\lambda^*\sin\theta^*}{n}}\\\nonumber
&+\mathcal O\left(n\sqrt{m\log(2n/\delta)}+n\log(2n/\delta)/\sqrt{\delta}\right)\bigg\}\lambda_{\min}(\boldsymbol X^\top\boldsymbol X)\\\nonumber
&
+\frac{\eta^2}{m^2}\bigg\{  \left(\frac{1}{\sqrt{2}}+\sqrt{R}\right)^2m^2n+\frac{(1+\sqrt{2R})^2}{8}m^2\\\nonumber
&+\frac{(1+\sqrt{2R})^2m^2}{2n}+\mathcal O\left(m^{3/2}n\sqrt{\log(1/\delta)}\right)  \bigg\}
\sqrt{n}\lambda_{\max}(\boldsymbol X^\top\boldsymbol X)+\frac{\eta^2}{m}\sigma_{\max}^2(\boldsymbol\Psi(k))\lambda_{\max}(\boldsymbol X^\top\boldsymbol X)\bigg\}\mathcal L(\boldsymbol W(k))\\\nonumber
<&\bigg\{1-\eta\bigg\{ 2\left(1-\sqrt{\frac{2}{2+\varepsilon}}\right)^2\lambda^*+2\left(\sqrt{\frac{\pi}{2+\varepsilon}}-\sqrt{\frac{1}{2\pi}}+o(1)\right)\sqrt{\frac{\lambda^*\sin\theta^*}{n}}\\\nonumber  &+\mathcal O\left(n\sqrt{\log(2n/\delta)/m}+n\log(2n/\delta)/(m\sqrt{\delta})\right)\bigg\}\lambda_{\min}(\boldsymbol X^\top\boldsymbol X)\\\nonumber
&
+\eta^2\bigg\{  \left(\frac{1}{\sqrt{2}}+\sqrt{R}\right)^2n^{3/2}+\frac{(1+\sqrt{2R})^2}{8}n^{1/2}+\frac{(1+\sqrt{2R})^2}{2}n^{-1/2}+\mathcal O\left(n^{3/2}\sqrt{\log(1/\delta)/m}\right)  \bigg\}\\\nonumber
&\lambda_{\max}(\boldsymbol X^\top\boldsymbol X)+\frac{\eta^2}{m}\left[\frac{mn}{2}+2m\sqrt{\lambda^*n}+2\lambda^*m+\mathcal O\left(n\sqrt{m\log(1/\delta)}\right)\right]\lambda_{\max}(\boldsymbol X^\top\boldsymbol X)\bigg\}\mathcal L(\boldsymbol W(k))\\\nonumber
=&\bigg\{1-\eta\bigg\{ 2\left(1-\sqrt{\frac{2}{2+\varepsilon}}\right)^2\lambda^*+2\left(\sqrt{\frac{\pi}{2+\varepsilon}}-\sqrt{\frac{1}{2\pi}}+o(1)\right)\sqrt{\frac{\lambda^*\sin\theta^*}{n}}\\\nonumber  &+\mathcal O\left(n\sqrt{\log(2n/\delta)/m}+n\log(2n/\delta)/(m\sqrt{\delta})\right)\bigg\}\lambda_{\min}(\boldsymbol X^\top\boldsymbol X)\\\nonumber
&
+\eta^2\bigg\{  \left(\frac{1}{\sqrt{2}}+\sqrt{R}\right)^2n^{3/2}+\frac{n}{2}+\left(\frac{(1+\sqrt{2R})^2}{8}+2\sqrt{\lambda^*}\right)n^{1/2}+2\lambda^*+\frac{(1+\sqrt{2R})^2}{2}n^{-1/2}\\\nonumber
&+\mathcal O\left(n^{3/2}\sqrt{\log(1/\delta)/m}\right)  \bigg\}
\lambda_{\max}(\boldsymbol X^\top\boldsymbol X)\bigg\}\mathcal L(\boldsymbol W(k)).
\end{align}
Further, it is evident that if $m=\Omega(n^2\log^{1+\varpi}(n/\delta))$ for any $\varpi>0$, then as $n\to+\infty$,
\begin{align}\nonumber
&n\sqrt{\frac{\log(2n/\delta)}{m}}\lesssim n\sqrt{\frac{\log(2n/\delta)}{n^2\log^{1+\varpi}(n/\delta)}}=\sqrt{\frac{\log(2n/\delta)}{\log^{1+\varpi}(n/\delta)}}=o(1);\\\nonumber
&n^{3/2}\sqrt{\frac{\log(1/\delta)}{m}}\lesssim n^{3/2}\sqrt{\frac{\log(1/\delta)}{n^2\log^{1+\varpi}(n/\delta)}}
=\sqrt{\frac{n\log(1/\delta)}{\log^{1+\varpi}(n/\delta)}}=o(n^{1/2}).
\end{align}
Let $\delta'=1.1mn\delta$ and if $m=\Omega(n^3\log^{1+\varpi'}(n/\delta))$ for any $\varpi'>1$, then as $n\to+\infty$,
\begin{align}\nonumber
&\frac{n\log(2n/\delta)}{m\sqrt{\delta}}=\frac{n\log(2n/\delta)}{m}\sqrt{\frac{1.1mn}{\delta'}}=\frac{\sqrt{1.1}n^{3/2}\log(2n/\delta)}{\sqrt{m\delta'}}\lesssim\frac{n^{3/2}\log(2n/\delta)}{\sqrt{n^3\log^{1+\varpi'}(n/\delta)\delta'}}=o(\delta'^{-1/2}).
\end{align}
Since $\delta'$ is lower bounded from zero in real applications, for example, it is enough for $\delta'>10^{-5}$ in real applications, which implies $n\log(2n/\delta)/(m\sqrt{\delta})=o(1)$ in practice.\\
We consider  very high dimensional data, i.e., $d\geqslant n$, thus $\lambda_{\text{min}}(\boldsymbol X^\top\boldsymbol X)=\sigma_{\text{min}}^2(\boldsymbol X)$ and $\lambda_{\text{max}}(\boldsymbol X^\top\boldsymbol X)=\|\boldsymbol X\|_2^2$. Taking limit of $\varepsilon\to0$ and $R<1/2$ in (i) (see (\ref{wp_R_bound})) and rewritting $\delta'$ as $\delta$ and $\varpi'$ as $\varpi$, we conclude that with if the width of neural network $m=\Omega(n^3\log^{1+\varpi}(n/\delta))$ for any $\varpi>1$, then as $n$ tends to infinity, for $k\in\mathbb Z^+$ and any $\delta\in(0,1)$, with probability exceeding $1-\delta$ over the random initialization,
\begin{align}\nonumber
&\mathcal L(\boldsymbol W(k+1))<\bigg\{1-\eta\left[2\lambda^*+o(1)\right]\sigma_{\text{min}}^2(\boldsymbol X)
+\eta^2\left[2n^{3/2}+n/2+\mathcal O(n^{1/2})\right]
\|\boldsymbol X\|_2^2\bigg\}\mathcal L(\boldsymbol W(k)),
\end{align}
where $\lambda^*=\lambda_{\text{min}}(\mho)$ and $\mho$ is defined in (\ref{matrix_mho_definition}) in Section \ref{section_problem}. \textcolor{blue}{Theorem \ref{theorem_GD_lossfunction_update} is therefore proved}.
\\
Let $\kappa=\|\boldsymbol X\|_2/\sigma_{\min}(\boldsymbol X)$ denote the condition number of data matrix $\boldsymbol X$ and
set $\eta=\lambda^*/(2\kappa^2 n^{3/2})$, straightforward calculations with $\lambda^*\geqslant\sigma_{\min}^2(\boldsymbol X)/4\gtrsim\Omega(\log^{-2\hbar}n)$ for certain $\hbar>0$ show that with probability exceeding $1-\delta$ over the random initialization,
\begin{align}\nonumber
&\mathcal L(\boldsymbol W(k+1))\\\nonumber
<&\left(1-\frac{(\lambda^*)^2\sigma_{\min}^4(\boldsymbol X)}{n^{3/2}\|\boldsymbol X\|_2^2}+o\left(\frac{\lambda^*\sigma_{\min}^4(\boldsymbol X)}{n^{3/2}\|\boldsymbol X\|_2^2}\right)+\frac{(\lambda^*)^2\sigma_{\min}^4(\boldsymbol X)}{2n^{3/2}\|\boldsymbol X\|_2^2}+\mathcal O(n^{-5/2})\right)\mathcal L(\boldsymbol W(k))\\\nonumber
=&\left[1-\eta(1+o(1))\lambda^*\sigma_{\min}^2(\boldsymbol X)\right]\mathcal L(\boldsymbol W(k)).
\end{align}
Therefore, as $n$ tends to infinity, the final convergence rate of GD is characterized as
\begin{align}\label{convergence_linear_rate}
&\mathcal L(\boldsymbol W(k))<\left[1-\eta(1+o(1))\lambda^*\sigma_{\min}^2(\boldsymbol X)\right]^k\mathcal L(\boldsymbol W(0))\quad(k\in\mathbb Z^+).
\end{align}
By means of $\eta=\lambda^*\sigma_{\text{min}}^2(\boldsymbol X)/(2n^{3/2}\|\boldsymbol X\|_2^2)$, we obtain another explicit form as $n\to+\infty$,
\begin{align}\label{convergence_linear_rate_explicit_form}
\mathcal L(\boldsymbol W(k))&<\left(1-\frac{(1+o(1))(\lambda^*)^2\sigma_{\min}^4(\boldsymbol X)}{2n^{3/2}\|\boldsymbol X\|_2^2}\right)^k\mathcal L(\boldsymbol W(0))\quad(k\in\mathbb Z^+).
\end{align}
Henceforth, we obtain another explicit form that for sufficiently large number of samples $n$,
\begin{align}\label{convergence_linear_rate_2}
\mathcal L(\boldsymbol W(k))<\left(1-\frac{3(\lambda^*)^2\sigma_{\min}^4(\boldsymbol X)}{8n^{3/2}\|\boldsymbol X\|_2^2}\right)^k\mathcal L(\boldsymbol W(0))\quad(k\in\mathbb Z^+).
\end{align}
Finally, we require that the actural moving distance of weight vector from initialization via applying Corollary 4.1 in (\cite{du_iclr}), which states if (\ref{convergence_linear_rate}) holds then
\begin{align}\label{R_requirement}
\|\boldsymbol w_r(k+1)-\boldsymbol w_r(0)\|_2\leqslant\frac{2\sqrt{2n\mathcal L(\boldsymbol W(0))}}{(1+o(1))\sqrt{m}\lambda^*\sigma_{\min}^2(\boldsymbol X)}<R,
\end{align}
which means
\begin{align}\nonumber
m\gtrsim\frac{n\mathcal L(\boldsymbol W(0))}{(\lambda^*)^2\sigma_{\min}^4(\boldsymbol X)R^2} .
\end{align}
Plugging in the choice of $R=\sqrt{(\pi\lambda^*\sin\theta^*)/[(2+\varepsilon)n]}$ for any $\varepsilon>0$. Further, by Claim 3.10 in (\cite{song_2020}), we see that for any $\delta\in(0,1)$, with probability exceeding $1-\delta$,
\begin{align}\label{LW_0_upper_bound}
\mathcal L(\boldsymbol W(0))=\mathcal O(n\log(m/\delta)\log^2(n/\delta)).
\end{align}
Therefore, it suffices to choose
\begin{align}\nonumber
m\gtrsim\frac{n^2\log(m/\delta)\log^2(n/\delta)}{(\lambda^*)^2\sigma_{\min}^4(\boldsymbol X)R^2}=\frac{(2+\varepsilon)n^3\log(m/\delta)\log^2(n/\delta)}{\pi(\lambda^*)^3\sin\theta^*\sigma_{\min}^4(\boldsymbol X)}
\end{align}
to satisfy the condition (\ref{R_requirement}). Notice that this requirement of $m$ obviously satisfies the condition of $m=\Omega(n^3\log^{1+\varpi}(n/\delta))$ for any $\varpi>1$.
\\Finally, by a union bound of two events (\ref{convergence_linear_rate}) and (\ref{LW_0_upper_bound}) hold simultaneously, we infer that, if $m=\Omega([n^3\log(m/\delta)\log^2(n/\delta)]/[(\lambda^*)^3\sin\theta^*\sigma_{\min}^4(\boldsymbol X)])$ and $\eta=\lambda^*/(2\kappa^2 n^{3/2})$,
then gradient descent is able to find the global minimum from a random initialization at the convergence linear rate in (\ref{convergence_linear_rate}) with probability exceeding $1-2\delta$. Rewriting $2\delta$ as $\delta$, we get the over-parameterization size $m=\Omega([n^3\log(2m/\delta)\log^2(2n/\delta)]/[(\lambda^*)^3\sin\theta^*\sigma_{\min}^4(\boldsymbol X)])=\Omega([n^3\log(m/\delta)\log^2(n/\delta)]/[(\lambda^*)^3\sin\theta^*\sigma_{\min}^4(\boldsymbol X)])$. With the aid of $\log(n/\delta)=\Theta(\log(m/\delta))$, we arrive at the desired over-parameterization size 
\begin{align}\label{final_overpara_result_1}
\textcolor{red}{m=\Omega\left(\frac{n^3\log^3(n/\delta)}{(\lambda^*)^3\sin\theta^*\sigma_{\min}^4(\boldsymbol X)}\right)}.
\end{align}

Furthermore, by virture of (\ref{convergence_linear_rate_2}) and (\ref{LW_0_upper_bound}), we have that for sufficiently large $n$, there exists a positive constant $C$ such that
\begin{align}\nonumber
\mathcal L(\boldsymbol W(k))<C\left(1-\frac{3(\lambda^*)^2\sigma_{\min}^4(\boldsymbol X)}{8n^{3/2}\|\boldsymbol X\|_2^2}\right)^kn\log(m/\delta)\log^2(n/\delta).
\end{align}
Therefore, $\mathcal L(\boldsymbol W(k))$ will be upper bounded as $\epsilon$ after the number of iterations
\begin{align}\label{iteration_bound_1}
T=\mathcal O\left(\frac{\log(Cn\log(m/\delta)\log^2(n/\delta)/\epsilon)}{\log\left(\{1-[3(\lambda^*)^2\sigma_{\min}^4(\boldsymbol X)]/[8n^{3/2}\|\boldsymbol X\|_2^2]\}^{-1}\right)}\right).
\end{align}
By means of $(1-x)^{-1}=\sum_{\nu=0}^{+\infty}x^{\nu}=1+x+\mathcal O(x^2)$ as $x\to0$ and Taylor expansion of logarithmic function $\log(1+x)=x+\mathcal O(x^2)$ as $x\to0$, with $0<\lambda^*\leqslant1/2$ and $0<\sigma_{\min}^4(\boldsymbol X)/\|\boldsymbol X\|_2^2<1$, we arrive at as $n\to+\infty$,
\begin{align}\nonumber
&\log\left(\{1-[3(\lambda^*)^2\sigma_{\min}^4(\boldsymbol X)]/[8n^{3/2}\|\boldsymbol X\|_2^2]\}^{-1}\right)\\\label{iteration_bound_2}
=&\log\left(1+\frac{3(\lambda^*)^2\sigma_{\min}^4(\boldsymbol X)}{8n^{3/2}\|\boldsymbol X\|_2^2}+\mathcal O(n^{-3})  \right)=\frac{3(\lambda^*)^2\sigma_{\min}^4(\boldsymbol X)}{8n^{3/2}\|\boldsymbol X\|_2^2}+\mathcal O(n^{-3}).
\end{align}
It is straightforward to show that
\begin{align}\nonumber
\log(Cn\log(m/\delta)\log^2(n/\delta)/\epsilon)=\log(1/\epsilon)+\alpha
=\mathcal O(1)\log(1/\epsilon),
\end{align}
where $\alpha=\mathcal O(\log(n/\delta))$ as $n$ tends to infinity. It is evident that $\alpha/\log(1/\epsilon)=\mathcal O(1)$, leading to $\log(Cn\log(m/\delta)\log^2(n/\delta)/\epsilon)
=\mathcal O(1)\log(1/\epsilon)$. Combining the estimates of (\ref{iteration_bound_1}) and (\ref{iteration_bound_2}), we obtain that after $T=\mathcal O([n^{3/2}\|\boldsymbol X\|_2^2\log(1/\epsilon)]/[(\lambda^*)^2\sigma_{\text{min}}^4(\boldsymbol X)])$ iterations, the empirical loss will be upper bounded as $\epsilon$.\\
Looking back (\ref{final_overpara_result_1}), with the aid of $\lambda^*\geqslant\sigma_{\min}^2(\boldsymbol X)/4\gtrsim\Omega(\log^{-2\hbar}n)$ for certain $\hbar>0$, a piece of paper, a pencil and a cup of tea, I see that
\begin{align}\nonumber
\frac{n^3\log^3(n/\delta)}{(\lambda^*)^3\sin\theta^*\sigma_{\min}^4(\boldsymbol X)}\leqslant\frac{4^3\cdot n^3\log^3(n/\delta)}{\sin\theta^*\cdot\sigma_{\min}^{10}(\boldsymbol X)}
\lesssim\frac{n^3\log^{10\hbar+3}(n/\delta)}{\sin\theta^*}.
\end{align}
Therefore, after a long analysis and calculation, I arrive at a concise characterization of desired overparameterization condition, \textcolor{blue}{revealing the instrinsic and subtle connection between the network width and the geometrical/spectral properties of the input data}:
\begin{align}
\textcolor{red}{m=\Omega\left(\frac{n^3\log^{10\hbar+3}(n/\delta)}{\sin\theta^*}\right).}
\end{align}
This formula No. 100 is the final result and therefrom Theorem \ref{theorem_GD_convergence} is established.
\section{Concluding Remarks, My Conjectures and Reflections}\label{section_conclusion}
In this paper, we develop a convergence theory for over-parameterized two-layer ReLU neural networks learned by gradient descent for very high dimensional dataset, shedding light on demystifying the global convergence puzzle and leading to state-of-the-art statistical guarantees of trainability. We perform a fine-grained analysis of the trajectory of objective function and characterize the convergence rate using geometrical and spectral quantities of training samples. By characterizing the local geometry including the gradient and directional curvature of loss function, we find the pivotal role of random activation matrix in optimization. This allows us to dissect random activation matrix in depth and reveal how the angular information in high-dimensional data space with over-parameterization determine the global convergence. The key novelty of our analysis is to make full use of angular information in dataset, providing more elaborate characterization of activation matrices and dynamics of loss function in the asymptotic expansion manner. This idea allows us to uncover how training works at a fundamental level and obtain \textcolor{blue}{\it a geometric understanding of global convergence mechanism: high-dimensional data geometry (angular distribution) $\mapsto$ spectrums of overparameterized activation matrices $\mapsto$ favorable geometrical properties of empirical loss function $\mapsto$ global convergence phenomenon}. Along the way, we improve the over-parameterization size and learning rate of gradient descent under very mild assumptions, bridging the significent gap between pratice and theory in prior work. Moreover, our theory demonstrates the particular advantage of learning overparameterized nonlinear nets for very high dimensional dataset. More explicitly, our results imply that higher data dimension may give rise to milder over-parameterization size, faster global convergence rate (iteration complexity) with more practical learning rate (step size).

Moving forward, the findings of this paper further may open the door to a new research direction in learning very high dimensional dataset. The methodology of this paper may help future theoretical investigations of analyzing other overparameterized neural networks learning tasks from three aspects: loss functions (e.g. cross-entropy loss), optimization algorithms (e.g. stochastic gradient descent) and network architectures (e.g. deep non-linear networks). Intutively, we can expect milder minimal over-parameterization size for provable learning nets via stochastic gradient descent, which is more challenging than that of gradient desent. It seems very challenging to obtain tighter overparameterization condition without additional assumptions, if possible. Additionally, our results imply that the data dimension $d$ can decrease the required neuron number $m$. 
The exploration in this paper implies that there is every indication that the role of data dimension $d$ in training has been underestimated so far. To further understand the global convergence phenomenon, we should consider the lower bound of $md$ rather than $m$. 

\textcolor{blue}{\it In the end, I raise three conjectures concerning ReLU networks:\\
(1) The minimal over-parameterization condition for provable learning two-layer ReLU networks via gradient descent using only non-degenerate training data is $md=\widetilde\Omega(n^2/\sin\theta^*)$.\\ 
(2) Among all the multi-layer ReLU networks, the minimal over-parameterization condition for provable learning via gradient descent using only non-degenerate training data is $md=\widetilde\Omega(n^{3/2}/\sin\theta^*)$.\\
(3) Phase transition phenomenon will occur at the minimal overparameterization condition for learning overparameterized ReLU networks, which in essence corresponds to the spontaneous magnetization phenomenon of a two-dimensional Ising model in statistical physics}.

Over the years, AI has changed enormously tastes form perceptons, expert systems, artificial neural networks, Bayesian networks, SVM, Adaboot, graphical models to deep learning. In 1969, Minsky and Papert published a mathematical critical paper on neural networks (perceptons), doubting the  ability of neural networks for learning complex tasks. Neural networks have experienced renaissance for several times. Although the theorectical understanding of deep learning is still lacking, recent empirical results have hinted us the instrinsic limitation of deep learning. I think it is wise to keep an open and critical attitude of deep learning, and I believe ideas and techniques of analyzing angular distributions in high-dimensional space will be of importance and shed new light on future AI development.

Albert Einstein said that \textcolor{red}{the basic laws of nature are always closely related to the most sophisticated and delicate mathematics}. In the great discoveries of Issac Newton, Albert Einstein , Paul Dirac to Chen-Ning Yang, elegant mathematics always played a pivotal role. Our exploration makes us further believe that it is the inherent and elegant mathematics in modern networks that makes it be so powerful in practice. Albert Einstein also said that \textcolor{red}{no problems can be solved from the same level of consciousness that created it}. From the mathematical viewpoint, the structures of modern networks such as multi-layer ReLU networks are still very fundamental, without much essential differences than perceptions discovered many years old. However, there are still many myterious things of mathematical nature of these fundamental AI models, which may be concealed in a rather complex and ingenious mechanism. The explorations in this paper makes me have reason to believe that \textcolor{blue}{\it the ultimate cause of the global convergence puzzle and hidden machinery probably origin in the wonderful connection between random parameter matrix and intrinsic data geometry in high dimension, which can be and only be discovered by critical attitude of mind and much deeper mathematical analysis than existing theories, which is just what I long for and I strongly believe the final answer will be an extremely concise and beautiful mathematical form, just like Paul Dirac's great discovery in 1928}.
\section{Commentary: The Genesis of This Research}
After receiving a Ph.D degree in Chinese Academy of Sciences, completing postdoctoral reseach in Tsinghua University and publishing two theoretical papers as the first author on machine learning problems in AISTATS 2014 (one of only three persons publishing at least two papers as  the first author in AISTATS 2014), I became a chief mathematics competition coach, a director of mathematics in an international school and a chief math teacher for teaching top students in top high schools in Liaoning Province. I taught smart students for participating National Mathematics Olympics, international math contests, Independent Entrance Examinations and National College Entrance Math Examination. Teaching lovely boys and girls how to solve fascinating and challenging math problems is full of fun. I like to get along with my students and am proud of them. Many of my students won gold medals in CMO (Chinese Mathematical Olympiad) and entered national mathematics training team for IMO (International Mathematical Olympiad), got high scores in AIME (American Invitational Mathematics Examination) and full scores in National College Entrance Math Examination. In the past few years, more than one hundred boys and girls of my students entered Tsinghua University and Peking University, with one provincial champion in National College Entrance Examination. In addition, dozens of my students entered top American universities including Yale, Stanford, UC Berkeley and so on. Apart from teaching math, I spends my spare time climbing, swimming and reading math books especially of classical analysis and geometry for fun. 

Very occasionally, one day in November 2021, as I waited for a train at a railway station in Shenyang, I happened to know the lackness of rigorous mathematical theory of deep learning. For the sake of curiousity, I kept thinking about the global convergence puzzle. One day, when I was running at Zhongshan park in Shenyang, I suddenly recognized that the keypoint is to dissect random activation matrix since it connects the dynamics of loss function and angular distribution in high-dimensional space in a natural manner. Furthermore, all of the most crucial expectations concerning activation patterns and random events can be exactly evaluated in terms of angle parameters from angular distribution. Once these ideas were there, it was thus led to an entertaining calculation. On March 6, 2022, all the pieces fitted together. The final expression is surprisingly simple, although the intermediate analysis and calculation steps are very complicated.

\addcontentsline{toc}{chapter}{References}
\bibliography{my}

\begin{thebibliography}{134}
\providecommand{\natexlab}[1]{#1}
\providecommand{\url}[1]{\texttt{#1}}
\expandafter\ifx\csname urlstyle\endcsname\relax
  \providecommand{\doi}[1]{doi: #1}\else
  \providecommand{\doi}{doi: \begingroup \urlstyle{rm}\Url}\fi

\bibitem[Ahlswede and Winter(2002)]{ahlswede_2002}
R.~Ahlswede and A.~Winter.
\newblock Strong converse for identification via quantum channels.
\newblock \emph{ITIT}, 48\penalty0 (3):\penalty0 569--579, 2002.

\bibitem[Aitken(1944)]{aitken_1944}
A.~C. Aitken.
\newblock \emph{Determinants and Matrices, Third Edition}.
\newblock University Mathematical Texts, 1. Edinburgh and London, 1944.

\bibitem[Allen-Zhu et~al.(2018)Allen-Zhu, Li, and Liang]{zhu_2018a}
Z.~Allen-Zhu, Y.~Li, and Y.~Liang.
\newblock Learning and generalization in overparameterized neural networks,
  going beyond two layers.
\newblock \emph{arXiv preprint arXiv:1811.04918}, 2018.

\bibitem[Allen-Zhu et~al.(2019)Allen-Zhu, Li, and Song]{zhu_2018b}
Z.~Allen-Zhu, Y.~Li, and Z.~Song.
\newblock A convergence theory for deep learning via overparameterization.
\newblock In \emph{International Conference on Machine Learning}, 2019.

\bibitem[Ara\'{u}jo et~al.(2019)Ara\'{u}jo, Oliveira, and
  Yukimura]{araujo_2019}
D.~Ara\'{u}jo, R.~I. Oliveira, and D.~Yukimura.
\newblock A mean-field limit for certain deep neural networks.
\newblock \emph{arXiv preprint arXiv:1906.00193}, 2019.

\bibitem[Arora et~al.(2019{\natexlab{a}})Arora, Cohen, Golowich, and
  Hu]{arora_2019a}
S.~Arora, N.~Cohen, N.~Golowich, and W.~Hu.
\newblock A convergence analysis of gradient descnet for deep linear neural
  networks.
\newblock In \emph{International Conference on Learning Representations},
  2019{\natexlab{a}}.

\bibitem[Arora et~al.(2019{\natexlab{b}})Arora, Du, Hu, Li, Salakhutdinov, and
  Wang]{arora_net_2019}
S.~Arora, S.~S. Du, W.~Hu, Z.~Li, R.~Salakhutdinov, and R.~Wang.
\newblock On exact computation with an infinitely wide neural net.
\newblock \emph{arXiv preprint arXiv:1904.11955}, 2019{\natexlab{b}}.

\bibitem[Arora et~al.(2019{\natexlab{c}})Arora, Du, Hu, Zhi, and
  Wang]{arora_2019}
S.~Arora, S.~S. Du, W.~Hu, Z.~Zhi, and R.~Wang.
\newblock Fine-grained analysis of optimization and generalization for
  overparameterized two-layer neural networks.
\newblock \emph{arXiv preprint arXiv:1901.08584}, 2019{\natexlab{c}}.

\bibitem[Bartlett et~al.(2018)Bartlett, Helmbold, and Long]{bartlett_2018}
P.~L. Bartlett, D.~Helmbold, and P.~Long.
\newblock Gradient descent with identity initialization efficiently learns
  positive definite linear transformations by deep residual networks.
\newblock In \emph{International Conference on Machine Learning}, pages
  521--530, 2018.

\bibitem[Belkin et~al.(2018{\natexlab{a}})Belkin, Ma, and Mandal]{belkin_2018a}
M.~Belkin, S.~Ma, and S.~Mandal.
\newblock To understand deep learning we need to understand kernel learning.
\newblock \emph{arXiv preprint arXiv:1802.01396}, 2018{\natexlab{a}}.

\bibitem[Belkin et~al.(2018{\natexlab{b}})Belkin, Rakhlin, and
  Tsybakov]{belkin_2018b}
M.~Belkin, A.~Rakhlin, and A.~B. Tsybakov.
\newblock Does data interpolation contradict statistical optimality?
\newblock \emph{arXiv preprint arXiv:1806.09471}, 2018{\natexlab{b}}.

\bibitem[Bellman(1960)]{bellman_1960}
R.~Bellman.
\newblock \emph{Introduction to Matrix Analysis}.
\newblock McGraw-Hill New York, 1960.

\bibitem[Billingsley(1979)]{billingsley_1979}
P.~Billingsley.
\newblock \emph{Probability and Measure, First Edition}.
\newblock Wiley, Chichester, 1979.

\bibitem[Billingsley(1999)]{billingsley_1999}
P.~Billingsley.
\newblock \emph{Convergence of Probability Measures, Second Edition}.
\newblock John Wiley and Sons, New York, 1999.

\bibitem[Blum and Rivest(1989)]{blum_1989}
A.~Blum and R.~L. Rivest.
\newblock Training a 3-node neural network is np-complete.
\newblock In \emph{Advances in Neural Information Processing Systems}, pages
  491--501, 1989.

\bibitem[Boob and Lan(2017)]{boob_2017}
D.~Boob and G.~Lan.
\newblock Theoretical properties of the global optimizer of two layer neural
  network.
\newblock \emph{arXiv preprint arXiv:1710.11241}, 2017.

\bibitem[Brutzkus and Globerson(2017)]{brutzkus_2017}
A.~Brutzkus and A.~Globerson.
\newblock Globally optimal gradient descent for a convnet with gaussian inputs.
\newblock \emph{arXiv preprint arXiv:1702.07966}, 2017.

\bibitem[Cand\`{e}s et~al.(2015)Cand\`{e}s, Li, and
  Soltanolkotabi]{candes_2018}
E.~J. Cand\`{e}s, X.~Li, and M.~Soltanolkotabi.
\newblock Phase retrieval via wirtinger flow: Theory and algorithms.
\newblock \emph{IEEE Transactions on Information Theory}, 61\penalty0
  (4):\penalty0 1985--2007, 2015.

\bibitem[Cao and Gu(2019{\natexlab{a}})]{cao_2019}
Y.~Cao and Q.~Gu.
\newblock A generalization theory of gradient descent for learning over-
  parameterized deep relu networks.
\newblock \emph{arXiv preprint arXiv:1902.01384}, 2019{\natexlab{a}}.

\bibitem[Cao and Gu(2019{\natexlab{b}})]{cao_2019b}
Y.~Cao and Q.~Gu.
\newblock Generalization bounds of stochastic gradient descent for wide and
  deep neural networks.
\newblock In \emph{Advances in Neural Information Processing Systems},
  2019{\natexlab{b}}.

\bibitem[Cauchy(1821)]{cauchy_1821}
A.~L. Cauchy.
\newblock \emph{Cours d'analyse de l'\'{e}cole polytechnique. Part I. Analyse
  Alg\'{e}brique}.
\newblock Paris, 1821.

\bibitem[Chen et~al.(2021)Chen, Chi, Fan, and Ma]{y_chen_2021}
Y.~Chen, Y.~Chi, J.~Fan, and C.~Ma.
\newblock Spectral methods for data science: a statistical perspective.
\newblock \emph{Foundations and Trends in Machine Learning}, 2021.

\bibitem[Chen et~al.(2020)Chen, Cao, and ang T.~Zhang]{chen_2020}
Z.~Chen, Y.~Cao, and Q.~Gu ang T.~Zhang.
\newblock Mean-field analysis of two-layer neural networks: non-asymptotic
  rates and generalization bounds.
\newblock \emph{arXiv preprint arXiv:2002.04026}, 2020.

\bibitem[Chi et~al.(2019)Chi, Lu, and Chen]{chi_2019}
Y.~Chi, Y.~M. Lu, and Y.~Chen.
\newblock Nonconvex optimization meets low-rank matrix factorization: An
  overview.
\newblock \emph{IEEE Transactions on Signal Processing}, 67\penalty0
  (20):\penalty0 5239--5269, 2019.

\bibitem[Chizat and Bach(2018{\natexlab{a}})]{chizat_2018a}
L.~Chizat and F.~Bach.
\newblock On the global convergence of gradient descent for over-parameterized
  models using optimal transport.
\newblock In \emph{Advances in Neural Information Processing Systems},
  2018{\natexlab{a}}.

\bibitem[Chizat and Bach(2018{\natexlab{b}})]{chizat_2018b}
L.~Chizat and F.~Bach.
\newblock A note on lazy training in supervised differentiable programming.
\newblock \emph{arXiv preprint arXiv:1812.07956}, 2018{\natexlab{b}}.

\bibitem[Chung(1968)]{chung_1968}
K.~L. Chung.
\newblock \emph{A Course of Probability}.
\newblock New York: Harcout, Brace and Wold, 1968.

\bibitem[Cram\'{e}r(1946)]{cramer_1946}
H.~Cram\'{e}r.
\newblock \emph{Mathematical Methods of Statistics}.
\newblock Princeton University, 1946.

\bibitem[Daniely(2017)]{daniely_2017}
A.~Daniely.
\newblock Sgd learns the conjugage kernel class of the network.
\newblock In \emph{Advances in Neural Information Processing Systems}, pages
  2422--2430, 2017.

\bibitem[Daniely et~al.(2016)Daniely, Frostig, and Singer]{daniely_2016}
A.~Daniely, R.~Frostig, and Y.~Singer.
\newblock Toward deeper understanding of neural networks: The power of
  initialization and a dual view on expressivity.
\newblock In \emph{Advances in Neural Information Processing Systems}, 2016.

\bibitem[Das and Golikov(2019)]{das_2019}
B.~Das and E.~A. Golikov.
\newblock Quadratic number of nodes is sufficient to learn a dataset via
  gradient descent.
\newblock \emph{arXiv preprint arXiv:1911.05402}, 2019.

\bibitem[Devlin et~al.(2018)Devlin, Chang, Lee, and Toutanova]{devlin_2018}
J.~Devlin, M.~W. Chang, K.~Lee, and K.~Toutanova.
\newblock Bert: Pre-training of deep bidirectional transformers for language
  understanding.
\newblock \emph{arXiv preprint arXiv:1810.04805}, 2018.

\bibitem[Doob(1953)]{doob_1953}
J.~L. Doob.
\newblock \emph{Stochastic Processes}.
\newblock John Wiley and Sons, New York, 1953.

\bibitem[Du and Lee(2018)]{du_2018}
S.~S. Du and J.~D. Lee.
\newblock On the power of over-parametrization in neural networks with
  quadratic activation.
\newblock \emph{arXiv preprint arXiv:1803.01206}, 2018.

\bibitem[Du et~al.(2017{\natexlab{a}})Du, Lee, and Tian]{du_2017a}
S.~S. Du, J.~D. Lee, and Y.~Tian.
\newblock When is a convolutional filter easy to learn?
\newblock \emph{arXiv preprint arXiv:1709.06129}, 2017{\natexlab{a}}.

\bibitem[Du et~al.(2017{\natexlab{b}})Du, Lee, Tian, Poczos, and
  Singh]{du_2017b}
S.~S. Du, J.~D. Lee, Y.~Tian, B.~Poczos, and A.~Singh.
\newblock Gradient descent learns one-hidden-layer cnn: Don't be afraid of
  spurious local minima.
\newblock \emph{arXiv preprint arXiv:1712.00779}, 2017{\natexlab{b}}.

\bibitem[Du et~al.(2019{\natexlab{a}})Du, Lee, Li, Wang, and Zhai]{du_2018a}
S.~S. Du, J.~D. Lee, H.~Li, L.~Wang, and X.~Zhai.
\newblock Gradient descent finds glolal minima of deep neural networks.
\newblock In \emph{International Conference on Machine Learning},
  2019{\natexlab{a}}.

\bibitem[Du et~al.(2019{\natexlab{b}})Du, Zhai, Poczos, and Singh]{du_iclr}
S.~S. Du, X.~Zhai, B.~Poczos, and A.~Singh.
\newblock Gradient descent provably optimizes over-parameterized neural
  networks.
\newblock \emph{In ICLR. arXiv preprint arXiv:1810.02054}, 2019{\natexlab{b}}.

\bibitem[Euler(1748)]{euler_1748}
L.~Euler.
\newblock \emph{Introductio in Analysin Infinitorum}.
\newblock Lausanne, 1748.

\bibitem[Euler(1755)]{euler_1755}
L.~Euler.
\newblock \emph{Institutiones Calculi Differentialis Cum Ejus Usu in Analysi
  Infinitorum ac Doctrina Serierum}.
\newblock Berlin, 1755.

\bibitem[Fang et~al.(2020)Fang, Lee, Yang, and Zhang]{fang_2020}
C.~Fang, J.~D. Lee, P.~Yang, and T.~Zhang.
\newblock Modeling from features: a mean-field framework for over-parameterized
  deep neural networks.
\newblock \emph{arXiv preprint arXiv:2007.01452}, 2020.

\bibitem[Feller(1957)]{feller_1957}
W.~Feller.
\newblock \emph{An Introduction to Probability Theory and Its Applications}.
\newblock John Wiley and Sons, New York, 1957.

\bibitem[Fisher(1941)]{fisher_1941}
R.~A. Fisher.
\newblock \emph{Statistical Methods for Research Workers}.
\newblock Edinburgh and London, 1941.

\bibitem[Foucart and Rauhut(2013)]{foucart_2013}
S.~Foucart and H.~Rauhut.
\newblock \emph{A Mathematical Introduction to Compressive Sensing}.
\newblock Springer New York, 2013.

\bibitem[Gantmacher(1966)]{gantmacher_1966}
F.~R. Gantmacher.
\newblock \emph{The Theory of Matrices (In Russian)}.
\newblock Science Press, 1966.

\bibitem[Gao et~al.(2021)Gao, Liu, Liu, Rajan, and Gao]{gao_2021}
T.~Gao, H.~Liu, J.~Liu, H.~Rajan, and H.~Gao.
\newblock A global convergence theory for deep relu implicit network via
  over-parameterization.
\newblock In \emph{International Conference on Learning Representations}, 2021.

\bibitem[Garcia and Horn(2017)]{garcia_2017}
S.~R. Garcia and R.~Horn.
\newblock \emph{A Second Course in Linear Algebra}.
\newblock Cambridge University Press, 2017.

\bibitem[Ge et~al.(2018)Ge, Lee, and Ma]{ge_2018}
R.~Ge, J.~D. Lee, and T.~Ma.
\newblock Learning one-hidden-layer neural networks with landscape design.
\newblock In \emph{International Conference on Learning Representations}, 2018.

\bibitem[Ger\v{s}gorin(1931)]{gersgorin_1931}
S.~Ger\v{s}gorin.
\newblock \"{U}ber die abgrenzung der eigenwerte einer matrix.
\newblock pages 749--754. Izv. Akad. Nauk. S. S. S. R., 1931.

\bibitem[Golub and Loan(2012)]{golub_2012}
G.~H. Golub and C.~F.~V. Loan.
\newblock \emph{Matrix Computations}, volume~3.
\newblock Johns Hopkins University Press, 2012.

\bibitem[Goursat(1902)]{goursat_1902}
E.~Goursat.
\newblock \emph{Cours d'Analyse math\'{e}mat\'{i}que}.
\newblock Paris, 1902.

\bibitem[Hardy(1908)]{hardy_1908}
G.~H. Hardy.
\newblock \emph{A Course of Pure Mathematics , First Edition}.
\newblock Cambridge University Press, 1908.

\bibitem[He et~al.(2016)He, Zhang, Ren, and Sun]{he_2016}
K.~He, X.~Zhang, S.~Ren, and J.~Sun.
\newblock Deep residual learning for image recognition.
\newblock In \emph{IEEE confernce on computer vision and pattern recognition},
  pages 770--778, 2016.

\bibitem[Hinton et~al.(2012)Hinton, Deng, Yu, Dahl, Mohamed, Jaitly, Senior,
  Vanhouche, Nguyen, and et~al]{hinton_2012}
G.~Hinton, L.~Deng, D.~Yu, G.~E. Dahl, A.~R. Mohamed, N.~Jaitly, A.~Senior,
  V.~Vanhouche, P.~Nguyen, and T.~Sainath et~al.
\newblock Deep neural networks for acoustic modeling in speech recognition: The
  shared views of four research groups.
\newblock \emph{IEEE Signal Processing Magazine}, 29\penalty0 (6):\penalty0
  82--97, 2012.

\bibitem[Horn and Johnson(2013)]{horn_2013}
R.~A. Horn and C.~R. Johnson.
\newblock \emph{Matrix Analysis}.
\newblock Cambridge University Press, 2013.

\bibitem[Hu et~al.(2020)Hu, Xiao, and Pennington]{hu_2020}
W.~Hu, L.~Xiao, and J.~Pennington.
\newblock Provable benefit of orthogonal initialization in optimizing deep
  linear networks.
\newblock In \emph{International Conference on Learning Representations}, 2020.

\bibitem[Hua(1984)]{hua_1984}
Luo-Geng Hua.
\newblock \emph{An Introduction to Advanced Mathematics (In Chinese)}.
\newblock Science Press, 1984.

\bibitem[Huang et~al.(2020)Huang, Wang, Tao, and Zhao]{huang_2020}
K.~Huang, Y.~Wang, M.~Tao, and T.~Zhao.
\newblock Why do deep residual networks generalize better than deep feedforward
  networks? -- a neural tangent kernel perspective.
\newblock \emph{arXiv preprint arXiv:2002.06262}, 2020.

\bibitem[Jacot et~al.(2018)Jacot, Gabriel, and Hongler]{jacot_2018}
A.~Jacot, F.~Gabriel, and C.~Hongler.
\newblock Neural tangent kernel: Convergence and generalization in neural
  networks.
\newblock \emph{arXiv preprint arXiv:1806.07572}, 2018.

\bibitem[Ji and Telgarsky(2019)]{ji_2019}
Z.~Ji and M.~J. Telgarsky.
\newblock Polylogarithmic width suffices for gradient descent to achieve
  arbitrarily small test error with shallow relu networks.
\newblock \emph{arXiv preprint arXiv:1909.12292}, 2019.

\bibitem[Jordan(1894)]{jordan_1894}
M.~C. Jordan.
\newblock \emph{Cours d'Analyse}.
\newblock Paris, 1894.

\bibitem[Kawaguchi(2021)]{kawaguchi_2021}
K.~Kawaguchi.
\newblock On the theory of implicit deep learning: Global convergence with
  implicit layers.
\newblock \emph{arXiv preprint arXiv:2102.07346}, 2021.

\bibitem[Kawaguchi and Huang(2020)]{kawaguchi_2020}
K.~Kawaguchi and J.~Huang.
\newblock Gradient descent finds global minima for generalizable deep neural
  networks of practical sizes.
\newblock \emph{arXiv preprint arXiv:1908.02419}, 2020.

\bibitem[Lang(1993)]{lang_1993}
S.~Lang.
\newblock \emph{Real and Functional Analysis}, volume~10.
\newblock Springer-Verlag New York, 1993.

\bibitem[Ledoux(2001)]{ledoux_2001}
M.~Ledoux.
\newblock \emph{The Concentration of Measure Phenomenon}.
\newblock Number 89 in Mathematical Surveys and Monographs. American
  Mathematical Society, 2001.

\bibitem[Ledoux and Talagrand(1991)]{ledoux_1991}
M.~Ledoux and M.~Talagrand.
\newblock \emph{Probability in Banach Spaces}.
\newblock Springer-Verlag Berlin, 1991.

\bibitem[Lee et~al.(2019)Lee, Xiao, Schoenholz, Bahri, Novak, Sohl-Dickstein,
  and Pennington]{lee_2019}
J.~Lee, L.~Xiao, S.~Schoenholz, Y.~Bahri, R.~Novak, J.~Sohl-Dickstein, and
  J.~Pennington.
\newblock Wide neural networks of any depth evolve as linear models under
  gradient descent.
\newblock In \emph{Advances in Neural Information Processing Systems}, 2019.

\bibitem[Li and Liang(2018)]{li_2018}
Y.~Li and Y.~Liang.
\newblock Learning overparameterized neural networks via stochastic gradient
  descent on structred data.
\newblock In \emph{Advances in Neural Information Processing Systems}, pages
  8157--8166, 2018.

\bibitem[Li and Yuan(2017)]{li_2017}
Y.~Li and Y.~Yuan.
\newblock Convergence analysis of two-layer neural networks with relu
  activation.
\newblock \emph{arXiv preprint arXiv:1705.09886}, 2017.

\bibitem[Li et~al.(2019)Li, Wang, Yu, Du, Hu, Salakhutdinov, and
  Arora]{li_2019}
Z.~Li, R.~Wang, D.~Yu, S.~S. Du, W.~Hu, R.~Salakhutdinov, and S.~Arora.
\newblock Enhanced convolutional neural tangent kernels.
\newblock \emph{arXiv preprint arXiv:1911.00809}, 2019.

\bibitem[Liang and Rakhlin(2018)]{liang_2018_kernel}
T.~Y. Liang and A.~Rakhlin.
\newblock Just interpolate: Kernel "ridgeless" regression can generalize.
\newblock \emph{arXiv preprint arXiv:1808.00387}, 2018.

\bibitem[Liao and Kyrillidis(2021)]{liao_2021}
F.~Liao and A.~Kyrillidis.
\newblock On the convergence of shallow neural network training with randomly
  masked neurons.
\newblock \emph{arXiv preprint arXiv:2112.02668}, 2021.

\bibitem[Lin and Bai(2010)]{lin_2010}
Z.~Lin and Z.~Bai.
\newblock \emph{Probability Inequalities}.
\newblock Science Press, 2010.

\bibitem[Liu et~al.(2020)Liu, Zhu, and Belkin]{liu_2020}
C.~Liu, L.~Zhu, and M.~Belkin.
\newblock On the linearity of large non-linear models: when and why the tangent
  kernel is constant.
\newblock In \emph{Advances in Neural Information Processing Systems},
  volume~33, 2020.

\bibitem[Liu et~al.(2021)Liu, Zhu, and Belkin]{liu_2021}
C.~Liu, L.~Zhu, and M.~Belkin.
\newblock Loss landscapes and optimization in over-parameterized non-linear
  systems and neural networks.
\newblock \emph{arXiv preprint arXiv:2003.00307}, 2021.

\bibitem[Livni et~al.(2014)Livni, Shwartz, and Shamir]{livni_2014}
R.~Livni, S.~S. Shwartz, and O.~Shamir.
\newblock On the computational efficiency of training neural network.
\newblock In 855-863, editor, \emph{Advances in Neural Information Processing
  Systems}, 2014.

\bibitem[Lojasiewicz(1963)]{lojasiewicz_1963}
S.~Lojasiewicz.
\newblock A topological property of real analytic subsets.
\newblock \emph{Coll. du CNRS, Les \'{e}quations aux d\'{e}riv\'{e}es
  partiells}, 117\penalty0 (87-89), 1963.

\bibitem[Lu et~al.(2020)Lu, Ma, Lu, Lu, and Ying]{lu_2020}
Y.~Lu, C.~Ma, Y.~Lu, J.~Lu, and L.~Ying.
\newblock A mean field analysis of deep resnet and beyond: Towards provably
  optimization via overparameterization from depth.
\newblock In \emph{International Conference on Machine Learning}, pages
  6426--6436, 2020.

\bibitem[Ma et~al.(2018)Ma, Wang, Chi, and Chen]{ma_2018_1}
C.~Ma, K.~Wang, Y.~Chi, and Y.~Chen.
\newblock Implicit regularization in nonconvex statistical estimation: Gradient
  descent converges linearly for phase retrieval and matrix completion.
\newblock In \emph{International Conference on Machine Learning}, pages
  3345--3354, 2018.

\bibitem[Mannelli et~al.(2020)Mannelli, Vanden-Eijnden, and
  Zdeborov\'{a}]{mannelli_2020}
S.~S. Mannelli, E.~Vanden-Eijnden, and L.~Zdeborov\'{a}.
\newblock Optimization and generalization of shallow neural networks with
  quadratic activation functions.
\newblock In \emph{Advances in Neural Information Processing Systems}, 2020.

\bibitem[Mei and Montanari(2019)]{mei_2019}
S.~Mei and A.~Montanari.
\newblock The generalization error of random features regression: Precise
  asymptotics and double descent curve.
\newblock \emph{arXiv preprint arXiv:1908.05355}, 2019.

\bibitem[Mei et~al.(2018)Mei, Montanari, and Nguyen]{mei_2018}
S.~Mei, A.~Montanari, and P.~M. Nguyen.
\newblock A mean field view of the landscape of two-layer neural networks.
\newblock 115, 2018.

\bibitem[Mianjy and Arora(2020)]{mianjy_2020}
P.~Mianjy and R.~Arora.
\newblock On convergence and generalization of dropout training.
\newblock 2020.

\bibitem[Moivre(1733)]{moivre_1733}
A.~De. Moivre.
\newblock \emph{Miscellanea Analytica, Second suppl.}
\newblock 1733.

\bibitem[Nguyen and Pham(2020)]{ngugen_2020}
P.~M. Nguyen and H.~T. Pham.
\newblock A rigorous framework for the mean field limit of multilayer neural
  networks.
\newblock \emph{arXiv preprint arXiv:2001.11443}, 2020.

\bibitem[Nguyen(2021)]{nguyen_2021}
Q.~Nguyen.
\newblock On the proof of global convergence of gradient descent for deep relu
  networks with linear widths.
\newblock In \emph{International Conference on Machine Learning}, 2021.

\bibitem[Nguyen and Hein(2017)]{nguyen_2017}
Q.~Nguyen and M.~Hein.
\newblock The loss surface of deep and wide neural networks.
\newblock In \emph{International Conference on Machine Learning}, pages
  2603--2612, 2017.

\bibitem[Nguyen and Mondelli(2020)]{nguyen_2020}
Q.~Nguyen and M.~Mondelli.
\newblock Global convergence of deep networks with one wide layer followed by
  pyramidal topology.
\newblock \emph{arXiv preprint arXiv:2002.07867}, 2020.

\bibitem[Nitanda and Suzuki(2017)]{nitanda_2017}
A.~Nitanda and T.~Suzuki.
\newblock Stochastic partice gradient descent for infinite ensembles.
\newblock \emph{arXiv preprint arXiv:1712.05438}, 2017.

\bibitem[Noy et~al.(2021)Noy, Xu, Afalo, Manor, and Jin]{noy_2021}
A.~Noy, Y.~Xu, Y.~Afalo, L.~Z. Manor, and R.~Jin.
\newblock A convergence theory towards practical over-parameterized deep neural
  networks.
\newblock \emph{arXiv preprint arXiv:2101.04243}, 2021.

\bibitem[Oymak and Soltanolkotabi(2020)]{oymak_2020}
S.~Oymak and M.~Soltanolkotabi.
\newblock Towars moderate overparameterization: global convergence guarantees
  for training shallow neural networks.
\newblock \emph{IEEE Journal on Selected Areas in Information Theory},
  1\penalty0 (1):\penalty0 84--105, 2020.

\bibitem[Parthasarathy(1967)]{parthasarathy_1967}
K.~R. Parthasarathy.
\newblock \emph{Probability Measures on Metric Spaces}.
\newblock Academic Press, New York, 1967.

\bibitem[Poincar\'{e}(1912)]{poincare_1912}
H.~Poincar\'{e}.
\newblock \emph{Calcul des Probabilit\'{e}s, Second Edition}.
\newblock Paris, 1912.

\bibitem[Polyak(1963)]{polyak_1963}
B.~T. Polyak.
\newblock Gradient methods for minimizing functionals.
\newblock \emph{Zhurnal Vychislitel'noi Matematiki i Matematicheskoi Fiziki},
  3.4:\penalty0 643--653, 1963.

\bibitem[Ross(1976)]{ross_1976}
S.~Ross.
\newblock \emph{A First Course in Probability}.
\newblock Macmillan, New York, 1976.

\bibitem[Ross(1983)]{ross_1983}
S.~Ross.
\newblock \emph{Stochastic Processes}.
\newblock Wiley, New York, 1983.

\bibitem[Rotskoff and Vanden-Eijnden(2018)]{rotskoff_2018}
G.~M. Rotskoff and E.~Vanden-Eijnden.
\newblock Trainability and accuracy of neural networks: An interacting particle
  system approach.
\newblock \emph{arXiv preprint arXiv:1805.00915}, 2018.

\bibitem[Rudelson(1999)]{rudelson_1999}
M.~Rudelson.
\newblock Random vectors in the isotropi position.
\newblock \emph{Journal of Functional Analysis}, 164\penalty0 (1):\penalty0
  60--72, 1999.

\bibitem[Rudin(1964)]{rudin_1964}
W.~Rudin.
\newblock \emph{Principles of Mathematical Analysis}, volume~3.
\newblock McGraw-Hill New York, 1964.

\bibitem[Safran and Shamir(2016)]{safran_2016}
I.~Safran and O.~Shamir.
\newblock On the quality of the initial basin in overspecified neural networks.
\newblock In \emph{International Conference on Machine Learning}, pages
  774--782, 2016.

\bibitem[Safran and Shamir(2017)]{safran_2017}
I.~Safran and O.~Shamir.
\newblock Spurious local minima are common in two-layer relu neural networks.
\newblock \emph{arXiv preprint arXiv:1712.08968}, 2017.

\bibitem[Schur(1911)]{schur_1911}
J.~Schur.
\newblock Bemerkungen zur theorie der beschr\"{a}nkten bilinearformen mit
  unendlich vielen ver\"{a}nderlichen.
\newblock In \emph{Journal f\"{u}r die reine and angewandte Mathematik (Crelles
  Journal)}, volume 140, pages 1--28, 1911.

\bibitem[Silver et~al.(2017)Silver, Schrittwieser, Simonyan, Antonoglou, Huang,
  Guez, Hubert, Baker, Lai, and A.~Bolton]{silver_2017}
D.~Silver, J.~Schrittwieser, K.~Simonyan, I.~Antonoglou, A.~Huang, A.~Guez,
  T.~Hubert, L.~Baker, M.~Lai, and et~al. A.~Bolton.
\newblock Mastering the game of go without human knowledge.
\newblock \emph{Nature}, 550\penalty0 (7676):\penalty0 354, 2017.

\bibitem[Sirignano and Spiliopoulos(2020)]{sirignano_2020}
J.~Sirignano and K.~Spiliopoulos.
\newblock Mean field analysis of neural networks: A central limit theorem.
\newblock \emph{Stochastic Processes and their Applications}, 130\penalty0
  (3):\penalty0 1820--1852, 2020.

\bibitem[Soltanolkotabi(2017)]{soltanolkotabi_2017}
M.~Soltanolkotabi.
\newblock Learning relus via gradient descent.
\newblock \emph{arXiv preprint arXiv:1705.04591}, 2017.

\bibitem[Soltanolkotabi et~al.(2018)Soltanolkotabi, Javanmard, and
  Lee]{soltanolkotabi_2018}
M.~Soltanolkotabi, A.~Javanmard, and J.~D. Lee.
\newblock Theoretical insights into the optimization landscape of
  over-parameterized shallow neural networks.
\newblock \emph{IEEE Transactions on Information Theory}, 2018.

\bibitem[Song and Yang(2020)]{song_2020}
Z.~Song and X.~Yang.
\newblock Over-parametrization for learning and generalization in two-layer
  neural network.
\newblock \emph{arXiv preprint arXiv:1906.03593}, 2020.

\bibitem[Stewart and Sun(1990)]{stewart_1990}
G.~W. Stewart and J.~G. Sun.
\newblock \emph{Matrix Perturbation Theory}.
\newblock Academic Press, 1990.

\bibitem[Su and Yang(2019)]{su_2019}
L.~Su and P.~Yang.
\newblock On learning over-parameterized neural networks: A functional
  approximation perspective.
\newblock 2019.

\bibitem[Sun(1987)]{sun_1987}
J.~G. Sun.
\newblock \emph{Matrix Perturbation Analysis (in Chinese)}, volume~6.
\newblock Science Press, 1987.

\bibitem[Tao(2012)]{tao_2012}
T.~Tao.
\newblock \emph{Topics in Random Matrix Theory}, volume 132.
\newblock AMS Graduate Studies in Mathematics, 2012.

\bibitem[Tian(2017)]{tian_2017}
Y.~Tian.
\newblock An analytical formula of population gradient for two-layered relu
  network and its applications in convergence and critical point analysis.
\newblock In \emph{International Conference on Machine Learning}, pages
  3404--3413, 2017.

\bibitem[Titchmarsh(1932)]{titchmarsh_1932}
E.~C. Titchmarsh.
\newblock \emph{The Theory of Funcions, First Edition}.
\newblock Oxford University Press, 1932.

\bibitem[Tropp(2015)]{tropp_2015}
J.~A. Tropp.
\newblock An introduction to matrix concentration inequalities.
\newblock \emph{Foundations and Trends in Machine Learning}, 8\penalty0
  (1-2):\penalty0 1--230, 2015.

\bibitem[Tsuchida et~al.(2017)Tsuchida, Roosta-Khorasani, and
  Gallagher]{tsuchida_2017}
R.~Tsuchida, F.~Roosta-Khorasani, and M.~Gallagher.
\newblock Invariance of weight distributions in rectified mlps.
\newblock \emph{arXiv preprint arXiv:1711.09090}, 2017.

\bibitem[Varga(2004)]{varga_2004}
R.~S. Varga.
\newblock \emph{Ger\v{s}gorin And His Circles}, volume Springer Series in
  Computational Mathematics.
\newblock Springer-Verlag Berlin, 2004.

\bibitem[Venturi et~al.(2018)Venturi, Bandeira, and Bruna]{venturi_2018}
L.~Venturi, A.~Bandeira, and J.~Bruna.
\newblock Neural networks with finite intrinsic dimension have no spurious
  valleys.
\newblock \emph{arXiv preprint arXiv:1802.06384}, 2018.

\bibitem[Vershynin(2018)]{vershy_2018}
R.~Vershynin.
\newblock \emph{High-Dimensional Probability: An Introduction with Applications
  in Data Science}.
\newblock Cambridge University Press, 2018.

\bibitem[Voulodimos et~al.(2018)Voulodimos, Doulamis, Doulamis, and
  Protopapadakis]{voulodimos_2018}
A.~Voulodimos, N.~Doulamis, A.~Doulamis, and E.~Protopapadakis.
\newblock Deep learning for computer vision: A brief review.
\newblock \emph{Computational Intelligence and Neuroscience}, 2018.

\bibitem[Wang et~al.(2006)Wang, Wu, and Jiang]{wang_2006_book}
S.~G. Wang, M.~X. Wu, and Z.~Z. Jiang.
\newblock \emph{Matrix Inequalities (In Chinese)}.
\newblock Science Press, 2006.

\bibitem[Weyl(1912)]{weyl_1912}
H.~Weyl.
\newblock Das asymptotische verteilungsgesetz der eigenwerte linearer
  partieller differentialgleichungen (mit einer anwendung auf die theorie der
  hohlraumstrahlung).
\newblock \emph{Mathematische Annalen}, 71\penalty0 (4):\penalty0 441--479,
  1912.

\bibitem[Whittaker and Watson(1902)]{whittaker_1902}
E.~T. Whittaker and G.~N. Watson.
\newblock \emph{A Course of Modern Analysis , First Edition}.
\newblock Cambridge University Press, 1902.

\bibitem[Wu et~al.(2019{\natexlab{a}})Wu, Wang, and Ma]{wu_2019}
L.~Wu, Q.~Wang, and C.~Ma.
\newblock Global convergence of gradient descent for deep linear residual
  networks.
\newblock In \emph{Advances in Neural Information Processing Systems},
  2019{\natexlab{a}}.

\bibitem[Wu et~al.(2019{\natexlab{b}})Wu, Du, and Ward]{wu_xiaoxia_2019}
X.~Wu, S.~S. Du, and R.~Ward.
\newblock Global convergence of adaptive gradient methods for an
  over-parameterized neural network.
\newblock \emph{arXiv preprint arXiv:1902.07111}, 2019{\natexlab{b}}.

\bibitem[Wu et~al.(2016)Wu, Schuster, Chen, Le, Norouzi, Macherey, Krikun, Cao,
  Gao, and et~al.]{wu_2016}
Y.~Wu, M.~Schuster, Z.~Chen, Q.~V. Le, M.~Norouzi, W.~Macherey, M.~Krikun,
  Y.~Cao, Q.~Gao, and K.~Macherey et~al.
\newblock Google's neural machine translation system: Bridging the gap between
  human and machine translation.
\newblock \emph{arXiv preprint arXiv:1609.08144}, 2016.

\bibitem[Xie et~al.(2017)Xie, Liang, and Song]{xie_2017}
B.~Xie, Y.~Liang, and L.~Song.
\newblock Diverse neural network learns true target functions.
\newblock In \emph{Artifical Intelligence and Statistics}, pages 1216--1224,
  2017.

\bibitem[Xu et~al.(2015)Xu, Wang, Chen, and Li]{xu_2015}
B.~Xu, N.~Wang, T.~Q. Chen, and M.~Li.
\newblock Empirical evaluation of rectified activations in convolutional
  network.
\newblock \emph{arXiv preprint arXiv:1505.00853}, 2015.

\bibitem[Zhang et~al.(2017)Zhang, Bengio, Hardt, Recht, and
  Vinyals]{zhang_chi_2017}
C.~Zhang, S.~Bengio, M.~Hardt, B.~Recht, and O.~Vinyals.
\newblock Understanding deep learning requires rethinking generalization.
\newblock In \emph{International Conference on Learning Representations}, 2017.

\bibitem[Zhang et~al.(2019)Zhang, Yu, Wang, and Gu]{zhang_2019}
X.~Zhang, Y.~Yu, L.~Wang, and Q.~Gu.
\newblock Learning one-hidden-layer relu networks via gradient descent.
\newblock In PMLR, editor, \emph{International Conference on Artificial
  Intelligence and Statistics}, pages 1524--1534, 2019.

\bibitem[Zhang et~al.(2020)Zhang, Plevrakis, Du, Li, Song, and
  Sanjeev]{zhang_yi_2020}
Y.~Zhang, O.~Plevrakis, S.~S. Du, X.~Li, Z.~Song, and Sanjeev.
\newblock Over-parameterized adversarial training: An analysis overcoming the
  curse of dimensionality.
\newblock \emph{arXiv preprint arXiv:2002.06668}, 2020.

\bibitem[Zhong et~al.(2017)Zhong, Song, and Dhillon]{zhong_song_2017}
K.~Zhong, Z.~Song, and I.~S. Dhillon.
\newblock Learning non-overlapping convolutional neural networks with multiple
  kernels.
\newblock \emph{arXiv preprint arXiv:1711.03440}, 2017.

\bibitem[Zhou and Liang(2017)]{zhou_2017}
Y.~Zhou and Y.~Liang.
\newblock Critical points of neural networks: analytical forms and landscape
  properties.
\newblock \emph{arXiv preprint arXiv:1710.11205}, 2017.

\bibitem[Zou and Gu(2019)]{zou_2019}
D.~Zou and Q.~Gu.
\newblock An improved analysis of training over-parameterized deep neural
  networks.
\newblock In \emph{Advances in Neural Information Processing Systems}, 2019.

\bibitem[Zou et~al.(2018)Zou, Cao, Zhou, and Gu]{zou_2018}
D.~Zou, Y.~Cao, D.~Zhou, and Q.~Gu.
\newblock Stochastic gradient descent optimized over-parameterized deep relu
  networks.
\newblock \emph{arXiv preprint arXiv:1811.08888}, 2018.

\end{thebibliography}


\appendix\label{section_appendix}

\end{document}